\documentclass[10pt,twocolumn,letterpaper]{article}

\usepackage[pagenumbers]{cvpr} 

\usepackage{microtype}
\usepackage{graphicx}
\usepackage{amsmath}
\usepackage{amssymb}
\usepackage{algorithm}
\usepackage[accsupp]{axessibility} 

\usepackage{mathtools}
\usepackage{bm}
\usepackage{booktabs}
\usepackage{multirow}
\usepackage{pifont}
\usepackage{tabularx}
\usepackage{color}
\usepackage{xcolor}
\usepackage{subcaption}
\usepackage[noend]{algpseudocode}
\usepackage[pagebackref,breaklinks,colorlinks,citecolor=cvprblue]{hyperref}
\usepackage[capitalize]{cleveref}

\Crefname{section}{Section}{Sections}
\Crefname{table}{Table}{Tables}
\crefname{section}{Sec.}{Secs.}
\crefname{table}{Tab.}{Tabs.}

\definecolor{cvprblue}{rgb}{0.21,0.49,0.74}
\newcommand*{\ourmodel}{UniDepth\@\xspace}

\newcommand\normx[1]{\left\Vert#1\right\Vert}

\definecolor{_blue}{RGB}{0, 90, 181}
\definecolor{_red}{RGB}{220, 50, 32}
\definecolor{_green}{RGB}{26, 255, 26}

\def\paperID{5061}
\def\confName{CVPR}
\def\confYear{2024}

\title{UniDepth: Universal Monocular Metric Depth Estimation}
\begin{document}
\author{
Luigi Piccinelli\textsuperscript{1} \quad Yung-Hsu Yang\textsuperscript{1} \quad Christos Sakaridis\textsuperscript{1} \quad \\[0.1cm]
Mattia Segu\textsuperscript{1} \quad Siyuan Li\textsuperscript{1} \quad Luc Van Gool\textsuperscript{1,2} \quad Fisher Yu\textsuperscript{1}\\[0.4cm]
$^1$ETH Z\"urich \quad $^2$INSAIT}


\maketitle
\begin{abstract}
Accurate monocular metric depth estimation (MMDE) is crucial to solving downstream tasks in 3D perception and modeling.
However, the remarkable accuracy of recent MMDE methods is confined to their training domains.
These methods fail to generalize to unseen domains even in the presence of moderate domain gaps, which hinders their practical applicability.
We propose a new model, \ourmodel, capable of reconstructing metric 3D scenes from solely single images across domains.
Departing from the existing MMDE methods, \ourmodel directly predicts metric 3D points from the input image at inference time without any additional information, striving for a universal and flexible MMDE solution.
In particular, \ourmodel implements a self-promptable camera module predicting dense camera representation to condition depth features.
Our model exploits a pseudo-spherical output representation, which disentangles camera and depth representations.
In addition, we propose a geometric invariance loss that promotes the invariance of camera-prompted depth features.
Thorough evaluations on ten datasets in a zero-shot regime consistently demonstrate the superior performance of \ourmodel, even when compared with methods directly trained on the testing domains.
Code and models are available at: \href{https://github.com/lpiccinelli-eth/unidepth}{github.com/lpiccinelli-eth/unidepth}.
\end{abstract}    
\vspace{-10pt}
\section{Introduction}
\label{sec:introduction}

\begin{figure}[ht]
    \centering
    \footnotesize
    \includegraphics[width=1.0\linewidth]{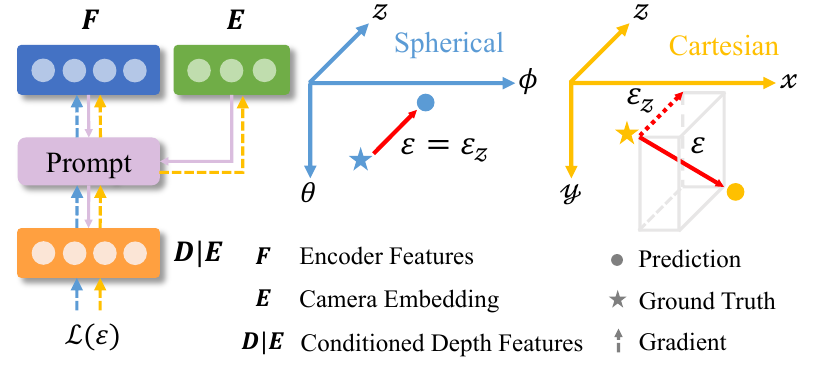}
    \vspace{-14pt}
    \caption{
    We introduce \ourmodel, a novel approach that directly predicts 3D points in a scene with only one image as input. 
    \ourmodel incorporates a camera self-prompting mechanism and leverages a pseudo-spherical 3D output space defined by azimuth and elevation angles, and depth ($\theta$, $\phi$, $z$).
    This design effectively separates camera and depth optimization by avoiding gradient flowing to the camera module due to depth-related error ($\varepsilon_z$).}
    \label{fig:teaser}
    \vspace{-10pt}
\end{figure}

The precise pixel-wise depth estimation is crucial to understanding the geometric scene structure, with applications in 3D modeling~\cite{deng2022nerf}, robotics~\cite{Zhou2019, dong2022depth4robotics}, and autonomous vehicles~\cite{wang2019depth4vehicles, park2021dd3d}.
However, delivering reliable metric scaled depth outputs is necessary to perform 3D reconstruction effectively, thus motivating the challenging and inherently ill-posed task of Monocular Metric Depth Estimation (MMDE).

While existing MMDE methods~\cite{Eigen2014, Fu2018Dorn, Bhat2020adabins, Ranftl2021dpt, Patil2022p3depth, Yuan2022newcrf, piccinelli2023idisc} have demonstrated remarkable accuracy across different benchmarks, they require training and testing on datasets with similar camera intrinsics and scene scales.
Moreover, the training datasets typically have a limited size and contain little diversity in scenes and cameras.
These characteristics result in poor generalization to real-world inference scenarios~\cite{Wang2020traingermany}, where images are captured in uncontrolled, arbitrarily structured environments and cameras with arbitrary intrinsics.

Only a few methods~\cite{yin2023metric3d, guizilini2023zerodepth} have addressed the challenging task of generalizable MMDE.
However, these methods assume controlled setups at test time, including camera intrinsics. 
While this assumption simplifies the task, it has two notable drawbacks.
Firstly, it does not address the full application spectrum, \eg in-the-wild video processing and crowd-sourced image analysis.
Secondly, the inherent camera parameter noise is directly injected into the model, leading to large inaccuracies in the high-noise case.

In this work, we address the more demanding task of generalizable MMDE \emph{without} any reliance on additional external information, such as camera parameters, thus defining the universal MMDE task.
Our approach, named \ourmodel, is the first that attempts to solve this challenging task without restrictions on scene composition and setup and distinguishes itself through its general and adaptable nature. 
Unlike existing methods, \ourmodel delivers metric 3D predictions for any scene \emph{solely} from a single image, waiving the need for extra information about scene or camera.
Furthermore, \ourmodel flexibly allows for the incorporation of additional camera information at test time.

Our design introduces a camera module that outputs a non-parametric, \ie dense camera representation, serving as the prompt to the depth module. 
However, relying only on this single additional module clearly results in challenges related to training stability and scale ambiguity.
We propose an effective pseudo-spherical representation of the output space to disentangle the camera and depth dimensions of this space.
This representation employs azimuth and elevation angle components for the camera and a radial component for the depth, forming a perfect orthogonal space between the camera plane and the depth axis.
Moreover, the camera components are embedded through Laplace spherical harmonic encoding.
Figure~\ref{fig:teaser} depicts our camera self-prompting mechanism and the output space.
Additionally, we introduce a geometric invariance loss to enhance the robustness of depth estimation. 
The underlying idea is that the camera-conditioned depth features from two views of the same image should exhibit reciprocal consistency.
In particular, we sample two geometric augmentations, creating a pair of different views for each training image, thus simulating different apparent cameras for the original scene.

Our overall contribution is the first universal MMDE method, \ourmodel, that predicts a point in metric 3D space for each pixel without \emph{any} input other than a single image.
In particular, first, we design a promptable camera module, an architectural component that learns a dense camera representation and allows for non-parametric camera conditioning.
Second, we propose a pseudo-spherical representation of the output space, thus solving the intertwined nature of camera and depth prediction.
In addition, we introduce a geometric invariance loss to disentangle the camera information from the underlying 3D geometry of the scene.
Moreover, we extensively test \ourmodel and re-evaluate seven MMDE State-of-the-Art (SotA) methods on ten different datasets in a fair and comparable zero-shot setup to lay the ground for the generalized MMDE task.
Owing to its design, \ourmodel consistently sets the new state of the art even compared with non-zero-shot methods, ranking first in the competitive official KITTI Depth Prediction Benchmark.

\section{Related Work}
\label{sec:relwork}

\begin{figure*}[t]
    \centering
    \footnotesize
    \includegraphics[width=0.98\linewidth]{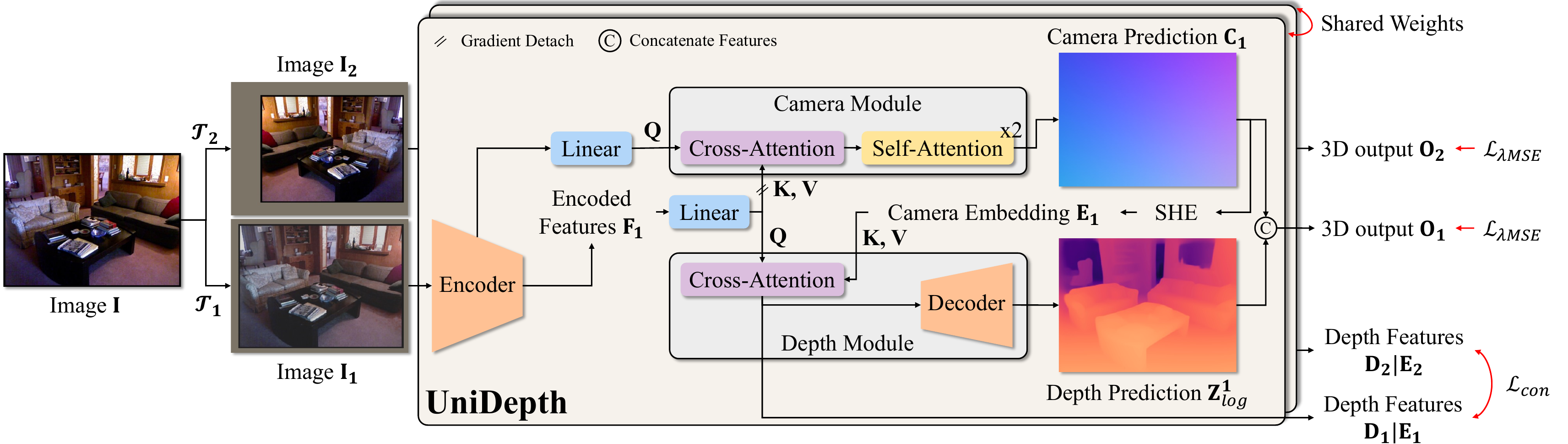}
    \vspace{-8pt}
    \caption{\textbf{Model Architecture.} \ourmodel utilizes solely the input image to generate the 3D output ($\mathbf{O}$). It bootstraps dense camera prediction ($\mathbf{C}$) from the Camera Module, injecting prior knowledge on scene scale into the Depth Module via a cross-attention layer. The camera representation corresponds to azimuth and elevation angles. The geometric invariance loss ($\mathcal{L}_{\mathrm{con}}$) enforces consistency between depth features tensors conditioned on the camera from different geometric augmentations ($\mathcal{T}_1$, $\mathcal{T}_2$). Stop-gradient is applied to the encoded feature ($\mathbf{F}$) flowing to the Camera Module to prevent the camera gradient from dominating the depth gradient in the encoder. The depth output ($\mathbf{Z}_{\log}$) is obtained through three self-attention blocks interleaved with learnable $2\mathrm{x}$ upsampling. The final output is the concatenation of the camera and depth tensors ($\mathbf{C} || \mathbf{Z}_{\log}$), creating two independent optimization spaces for $\mathcal{L}_{\lambda MSE}$.}
    \label{fig:results:overview}
    \vspace{-15pt}
\end{figure*}

\noindent{}\textbf{Metric and Scale-agnostic Depth Estimation.}
It is crucial to distinguish Monocular Metric Depth Estimation (MMDE) from scale-agnostic, namely up-to-a-scale, monocular depth estimation.
MMDE SotA approaches typically confine training and testing to the same domain.
However, challenges arise, such as overfitting to the training scenario leading to considerable performance drops in the presence of minor domain gaps,
often overlooked in benchmarks like NYU-Depthv2~\cite{silberman2012nyu} (NYU) and KITTI~\cite{Geiger2012kitti}.
On the other hand, scale-agnostic depth methods, including MiDaS~\cite{ranftl2020midas}, OmniData~\cite{eftekhar2021omnidata}, and LeReS~\cite{yin2021leres}, show robust generalization by training on extensive datasets.
Their limitation lies in the absence of a metric output, hindering practical usage in downstream applications.

\noindent{}\textbf{Monocular Metric Depth Estimation.}
The introduction of end-to-end trainable neural networks in MMDE, pioneered by \cite{Eigen2014}, marked a significant milestone, also introducing the optimization process through the Scale-Invariant log loss ($\mathrm{SI}_{\log}$).
Subsequent developments witnessed the emergence of advanced networks, ranging from convolution-based architectures~\cite{Fu2018Dorn, Laina2016, Liu2015, Patil2022p3depth} to transformer-based approaches~\cite{Yang2021, Bhat2020adabins, Yuan2022newcrf, piccinelli2023idisc}.
Despite impressive achievements on established benchmarks, MMDE models face challenges in zero-shot scenarios, revealing the need for robust generalization
against domain shifts in appearance and geometry.

\noindent{}\textbf{General Monocular Metric Depth Estimation.}
Recent efforts focus on developing MMDE models~\cite{bhat2023zoedepth, guizilini2023zerodepth, yin2023metric3d} for general depth prediction across diverse domains.
These models often leverage camera awareness, either by directly incorporating external camera parameters into computations~\cite{facil2019camconvs, guizilini2023zerodepth} or by normalizing the shape or output depth based on intrinsic properties, as seen in~\cite{Lee2019bts, Lopez2020mapillary, yin2023metric3d}.

However, these generalizable MMDE methods often adopt specific strategies to enhance performance, \eg geometric pretraining~\cite{bhat2023zoedepth} or dataset-specific prior like reshaping~\cite{yin2023metric3d}.
In addition, these methods assume access to noiseless camera intrinsics both at training and test time, also limiting their applicability to pinhole camera models.
Additionally, SotA methods depend on a predefined backprojection operation, blurring the distinction between learning depth and the 3D scene.
In contrast, our approach aims to overcome these limitations, presenting a more demanding perspective, \eg universal MMDE.
Universal MMDE involves directly predicting the 3D scene from the input image \emph{without any} additional information other than the latter.
Notably, we do not require any additional prior information at test time, such as access to camera information.

\section{\ourmodel}
\label{sec:method}


MMDE SotA methods typically assume access to the camera intrinsics, thus blurring the line between pure depth estimation and actual 3D estimation.
In contrast, \ourmodel aims to create a universal MMDE model deployable in diverse scenarios without relying on any other external information, such as camera intrinsic, thus leading to 3D space estimation by design.
However, attempting to directly predict 3D points from a single image without a proper internal representation neglects geometric prior knowledge, \ie perspective geometry, burdening the learning process with re-learning laws of perspective projection from data. 

\cref{ssec:method:spherical} introduces a pseudo-spherical representation of the output space to inherently disentangle camera rays' angles from depth.
In addition, our preliminary studies indicate that depth prediction clearly benefits from prior information on the acquisition sensor, leading to the introduction of a self-prompting camera operation in \cref{ssec:method:camera_module}.
Further disentanglement at the level of internal depth features is achieved through a geometric invariance loss, outlined in \cref{ssec:method:consistency}. 
This loss ensures depth features remain invariant when conditioned on the bootstrapped camera predictions, promoting robust camera-aware depth predictions.
The overall architecture and the resulting optimization induced by the combination of design choices are detailed in \cref{ssec:method:design}.

\subsection{3D Representation}
\label{ssec:method:spherical}

The general purpose nature of our MMDE method requires inferring both depth and camera intrinsics to make 3D predictions based only on imagery observations. 
We design the 3D output space presenting a natural disentanglement of the two sub-tasks, namely depth estimation and camera calibration.
In particular, we exploit the pseudo-spherical representation where the basis is defined by azimuth, elevation, and log-depth, \ie ($\theta$,$\phi$,$z_{\log}$), in contrast to the Cartesian representation ($x$,$y$,$z$).
The strength of the proposed pseudo-spherical representation lies in the decoupling of camera ($\theta$,$\phi$) and depth ($z_{\log}$) components, ensuring their orthogonality by design, in contrast to the entanglement present in Cartesian representation.

It is worth highlighting that in this output space, the non-parametric dense representation of the camera is mathematically represented as a tensor $\mathbf{C} \in \mathbb{R}^{H \times W \times 2}$, where $H$ and $W$ are the height and width of the input image and the last dimension corresponds to azimuth and elevation values.
While in the typical Cartesian space, the backprojection involves the multiplication of homogeneous camera rays and depth, the backprojection operation in the proposed representation space accounts for the concatenation of camera and depth representations.
The pencil of rays are defined as $(\mathbf{r}_1, \mathbf{r}_2, \mathbf{r}_3) = \mathbf{K}^{-1} [\mathbf{u}, \mathbf{v}, \mathbf{1}]^T$, where $\mathbf{K}$ is the calibration matrix, $\mathbf{u}$ and $\mathbf{v}$ are pixel positions in pixel coordinates, and $\mathbf{1}$ is a vector of ones. Therefore, the homogeneous camera rays $(\mathbf{r}_x, \mathbf{r}_y)$ correspond to $(\frac{\mathbf{r}_1}{\mathbf{r}_3}, \frac{\mathbf{r}_2}{\mathbf{r}_3})$.

Moreover, the angular dense representation can be embedded via the Laplace Spherical Harmonic Encoding (SHE).
The camera embedding tensor is defined as $\mathbf{E} = \mathrm{SHE}(\mathbf{C}), \mathbf{E} \in \mathbb{R}^{H \times W \times d}$, where $d$ is the number of harmonics chosen.
$\mathrm{SHE}(\cdot)$ computes the set of spherical harmonics, \ie, $\{\mathcal{Y}\}_{l,m}$ with degree $l$ and order $m$, and concatenating along the channel dimension, with $\mathbf{Y}^l_m$ as
\begin{equation}
    \label{eqn:sht}
    Y^l_m(\theta, \phi) = \alpha^l_m \mathcal{P}^l_m(\cos \theta)e^{im\phi},
\end{equation}
where $\mathcal{P}^l_m$ is the associated Legendre polynomial of degree $l$ and order $m$, and $\alpha^l_m$ is a normalizing constant.
In particular, the spherical harmonics on the unit sphere form an orthogonal basis of the spherical manifold and preserve inner products.
The total number of harmonics utilized is $81$, resulting from capping the degree $l$ to 8.
SHE is utilized as a mathematic sounder choice compared to, \eg the Fourier Transform, to produce the camera embeddings.

\subsection{Self-Promptable Camera}
\label{ssec:method:camera_module}

The camera module plays a crucial role in the final 3D predictions since its angular dense output accounts for two dimensions of the output space, namely azimuth and elevation.
Most importantly, these embeddings prompt the depth module to ensure a bootstrapped prior knowledge of the input scene's global depth scale.
The prompting is fundamental to avoid mode collapse in the scene scale and to alleviate the depth module from the burden of predicting depth from scratch as the scale is already modeled by camera output.

Nonetheless, the internal representation of the camera module is based on a pinhole parameterization, namely via focal length ($f_x$, $f_y$) and principal point ($c_x$, $c_y$).
The four tokens conceptually corresponding to the intrinsics are then projected to scalar values, \ie, $\Delta f_x$, $\Delta f_y$, $\Delta c_x$, $\Delta c_y$.
However, they do not directly represent the camera parameters, but the multiplicative residuals to a pinhole camera initialization, namely $\frac{H}{2}$ for y-components and $\frac{W}{2}$ for x-components, leading to $f_x = \frac{\Delta f_x W}{2}$, $f_y = \frac{\Delta f_y H}{2}$, $c_x = \frac{\Delta c_x W}{2}$, $c_y = \frac{\Delta c_y H}{2}$, leading to invariance towards input image sizes.

Subsequently, a backprojection operation based on the intrinsic parameters is applied to every pixel coordinate to produce the corresponding rays.
The rays are normalized and thus represent vectors on a unit sphere.
The critical step involves extracting azimuth and elevation from the backprojected rays, effectively creating a ``dense'' angular camera representation.
This dense representation undergoes SHE to produce the embeddings $\mathbf{E}$.
The embedded representations are then seamlessly passed to the depth module as a prompt, where they play a vital role as a conditioning factor.
The conditioning is enforced via a cross-attention layer between the initialized feature of Depth Module $\mathbf{D} \in \mathbb{R}^{h \times w \times C}$ and the camera embeddings $\mathbf{E}$ where $(h,w)=(H/16, W/16)$. The camera-prompted depth features $\mathbf{D|E} \in \mathbb{R}^{h \times w \times C}$ are defined as 
\begin{equation}
    \mathbf{D|E} = \mathrm{MLP}(\mathrm{CA}(\mathbf{D}, \mathbf{E})),
\end{equation}
where $\mathrm{CA}$ is a cross-attention block and $\mathrm{MLP}$ is a MultiLayer Perceptron with one $4C$-channel hidden layer.

\begin{figure}[t]
    \centering
    \footnotesize
    \includegraphics[width=1\columnwidth]{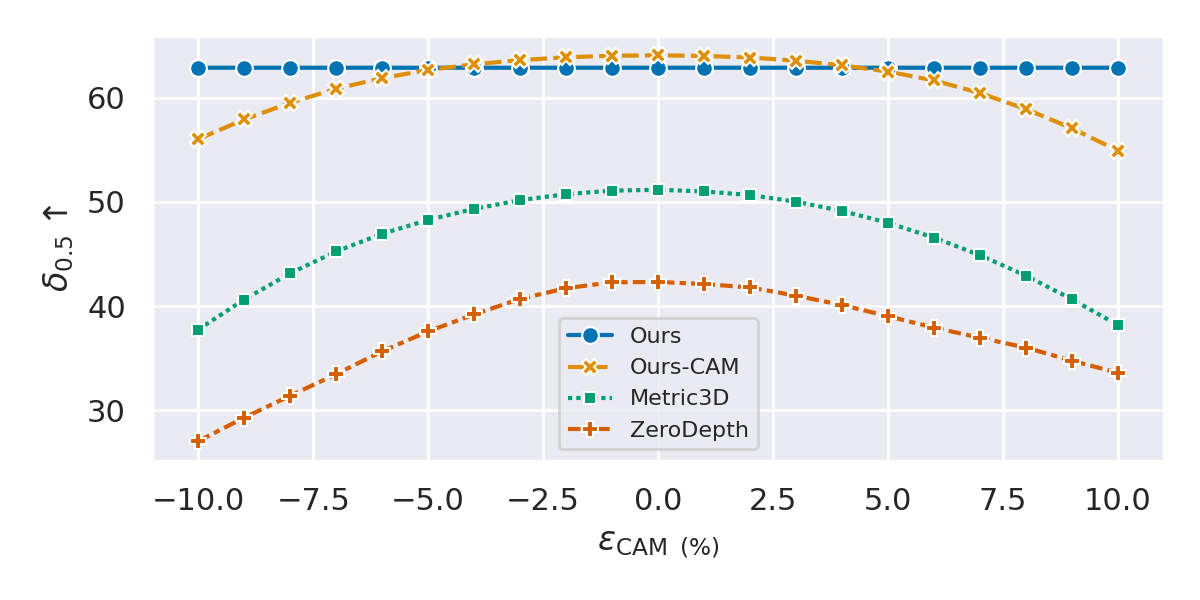} \\
    \vspace{-14pt}
    \caption{\textbf{Impact of noise in camera intrinsics.}
    The amount of relative distortion ($\varepsilon_{\mathrm{CAM} (\%)}$) of the intrinsics is shown on the x-axis, while $\delta_{0.5}$ performance on OOD test sets on the y-axis.
    Relying on external input inherently leads to being subject to its noise.
    \ourmodel functions in dual regimes, with and without external intrinsic.
    In situations of unknown intrinsics or high noise, \ourmodel exhibits total robustness by bootstrapping camera prediction (\textit{Ours}).
    In contrast, with low-noise intrinsics, we leverage it for enhanced peak performance (\textit{Ours-CAM}).}
    \label{fig:results:noise_intrinsics}
    \vspace{-14pt}
\end{figure}

\Cref{fig:results:noise_intrinsics} illustrates one of the main benefits of our camera module.
In particular, in high-noise intrinsics or camera-agnostic scenarios, \ourmodel can bootstrap the camera prediction, thus displaying total noise insensitivity.
However, we can substitute the camera module output to improve 3D reconstruction peak performance if any external dense camera representation is provided.
This adaptability enhances the model's versatility, allowing it to operate seamlessly in diverse setups.
Moreover, \Cref{fig:results:noise_intrinsics} suggests that training with noisy self-prompts enhances the robustness of \ourmodel to noisier external intrinsics if given at test time.

\subsection{Geometric Invariance Loss}
\label{ssec:method:consistency}

The spatial locations from the same scene captured by different cameras should correspond when the depth module is conditioned on the specific camera.
To this end, we propose a geometric invariance loss to enforce the consistency of camera-prompted depth features of the same scene from different acquisition sensors.
In particular, consistency is enforced on features extracted from identical 3D locations.

For each image, we perform $N$ distinct geometrical augmentations, denoted as $\{\mathcal{T}_i\}_{i=1}^N$, with $N=2$ in our experiments. 
This operation involves involves sampling a rescaling factor $r \sim 2^{\mathcal{U}_{[-1, 1]}}$ and a relative translation on the $x$-axis $t \sim \mathcal{U}_{[-0.1, 0.1]}$, then cropping it to the network's input shape.
This is analogous to sampling a pair of images from the same scene and extrinsic parameters but captured by different cameras. 
Let $\mathbf{C}_i$ and $\mathbf{D_i|E_i}$ describe the predicted camera representation and camera-prompted depth features, respectively, corresponding to augmentation $\mathcal{T}_i$. 
It is evident that the camera representations differ when two diverse geometric augmentations are applied, i.e., $\mathbf{C}_i \neq \mathbf{C}_j$ if $\mathcal{T}_i \neq \mathcal{T}_j$. 
Therefore, the geometric invariance loss can be expressed as
\begin{equation}
\begin{split}
        &\mathcal{L}_{\mathrm{con}}(\mathbf{D}_1|\mathbf{E}_1, \mathbf{D}_2|\mathbf{E}_2) =\\
        &\normx{\mathcal{T}_2 \circ \mathcal{T}^{-1}_1 \circ (\mathbf{D}_1|\mathbf{E}_1) - \mathrm{sg}(\mathbf{D}_2|\mathbf{E}_2)}_1,
\end{split}
\label{eqn:method:selfconst}
\end{equation}
where $\mathbf{D}_i|\mathbf{E}_i$ represents the depth feature after being conditioned by camera prompt $\mathbf{E}_i$, as outlined in \cref{ssec:method:camera_module}, and $\mathrm{sg}(\cdot)$ corresponds to the stop-gradient detach operation needed to exploit $\mathbf{D}_2|\mathbf{E}_2$ as pseudo ground-truth (GT).
The bi-directional loss can be computed as: $\frac{1}{2} (\mathcal{L}_{\mathrm{con}}(\mathbf{D}_1|\mathbf{E}_1, \mathbf{D}_2|\mathbf{E}_2) + \mathcal{L}_{\mathrm{con}}(\mathbf{D}_2|\mathbf{E}_2, \mathbf{D}_1|\mathbf{E}_1))$.
It is necessary to apply the geometric invariance loss \emph{after} the features are conditioned on the viewing information, \ie, camera. Otherwise, the loss would enforce consistency across features that inherently carry distinct camera information.

\subsection{Network Design}
\label{ssec:method:design}

\noindent{}\textbf{Architecture.} Our network, described in \cref{fig:results:overview}, comprises an Encoder Backbone, a Camera Module, and a Depth Module.
The encoder can be either convolutional or ViT-based~\cite{Dosovitskiy2020VIT}, producing features at different ``scales'', \ie $\mathbf{F} \in \mathbb{R}^{h \times w \times C \times B}$, where $(h,w) = (\frac{H}{16}, \frac{W}{16})$ and $B=4$.

The Camera Module parameters are initialized class tokens for ViT-style or pooled feature maps for convolutional-style backbones.
The encoded features from the Encoder Backbone are passed to the Camera Module as a stack of detached tokens, the encoder class tokens are utilized as camera parameters initialization.
The features are processed to obtain the final dense representation $\mathbf{C}$ as detailed in \cref{ssec:method:camera_module}, and further embedded to $\mathbf{E}$ via $\mathrm{SHE}(\cdot)$ outlined in \cref{ssec:method:spherical}.
Note that the stop-gradient operation is necessary because of the low variety of effective cameras compared to the image diversity.
In fact, the Camera Module component easily overfits and clearly dominates the overall backbone gradient.

The Depth Module is fed with the encoder features to condition the initial latent features $\mathbf{L} \in \mathbb{R}^{h \times w \times C}$ via one cross-attention layer to obtain the initial depth features, $\mathbf{D}$.
The latent feature tensor $\mathbf{L}$ is obtained as the average of the features $\mathbf{F}$ along the $B$ dimension.
Furthermore, the depth features are conditioned on the camera prompts $\mathbf{E}$ to obtain $\mathbf{D|E}$ as described in \cref{ssec:method:camera_module}.
The camera-prompted depth features are further processed via self-attention layers where the positional encoding utilized is $\mathbf{E}$ and upsampled to produce a multi-scale output.
The log-depth prediction $\mathbf{Z}_{\log} \in \mathbb{R}^{H \times W \times 1}$ corresponds to the mean of the interpolated intermediate representations.
The final 3D output $\mathbf{O} \in \mathbb{R}^{H \times W \times 3}$ is the concatenation of predicted rays and depth, $\mathbf{O} = \mathbf{C} || \mathbf{Z}$, with $\mathbf{Z}$ as element-wise exponentiation of $\mathbf{Z}_{\log}$.

\noindent{}\textbf{Optimization.}  The optimization process is guided by a re-formulation of the Mean Squared Error (MSE) loss in the final 3D output space ($\theta$,$\phi$,$z_{\log}$) from \cref{ssec:method:spherical} as:
\vspace{-4pt}
\begin{equation}
    \vspace{-4pt}
    \begin{split}
        \mathcal{L}_{\lambda\mathrm{MSE}}(\bm{\varepsilon}) = \|\mathbb{V}[\bm{\varepsilon}]\|_1 + \bm{\lambda}^T(\mathbb{E}[\bm{\varepsilon}]\odot\mathbb{E}[\bm{\varepsilon}]),
    \end{split}
    \label{eqn:method:mse}
\end{equation}
where $\bm{\varepsilon} = \hat{\mathbf{o}} - \mathbf{o}^* \in \mathbb{R}^3$, $\hat{\mathbf{o}}=(\hat{\theta},\hat{\phi},\hat{z}_{\log})$ is the predicted 3D output, $\mathbf{o}^*=(\theta^*,\phi^*,z_{\log}^*)$ is the GT 3D value, and $\bm{\lambda} = (\lambda_{\theta},\lambda_{\phi},\lambda_z) \in \mathbb{R}^3$ is a vector of weights for each dimension of the output.
$\mathbb{V}[\bm{\varepsilon}]$ and $\mathbb{E}[\bm{\varepsilon}]$ are computed as the vectors of empirical variances and means for each of the three output dimensions over all pixels, \ie $\{\bm{\varepsilon}^{(i)}\}_{i=1}^{N}$. 
Note that if $\lambda_d=1$ for a dimension $d$, the loss represents the standard MSE loss for that dimension. If $\lambda_d<1$, a scale-invariant loss term is added to that dimension if it is expressed in log space, \eg for the depth dimension $z_{\log}$ and a shift-invariant loss term is added if that output is expressed in linear space. 
In particular, if only the last output dimension is considered, \ie, the one corresponding to depth, and $\lambda_z=0.15$ is utilized, the corresponding loss is the standard $\mathrm{SI}_{\log}$.
In our experiments, we set $\lambda_{\theta}=\lambda_{\phi}=1$ and $\lambda_z=0.15$.
Therefore, the final optimization loss is defined as
\vspace{-5pt}
\begin{equation}
    \vspace{-5pt}
    \mathcal{L} = \mathcal{L}_{\lambda\mathrm{MSE}} + \alpha \mathcal{L}_{\mathrm{con}}, \text{ with } \alpha=0.1.
    \label{eqn:method:loss}
\end{equation}

The loss defined here serves as a motivation for the designed output representation.
Specifically, employing a Cartesian representation and applying the loss directly to the output space would result in backpropagation through ($x$, $y$), and $z_{\log}$ errors.
However, $x$ and $y$ components are derived as $r_x \cdot z$ and $r_y \cdot z$ as detailed in \cref{ssec:method:spherical}.
Consequently, the gradients of camera components, expressed by ($r_x$, $r_y$), and of depth become intertwined, leading to suboptimal optimization as discussed in~\cref{ssec:experiments:ablations}.

\section{Experiments}
\label{sec:experiments}

\begin{table*}[t]
    \centering
    \caption{\textbf{Comparison on zero-shot evaluation.} All methods are tested in a zero-shot setting on eight different datasets without overlap with any of the sets used for training. \ourmodel-\{C, V\}: \ourmodel-\{ConvNext~\cite{liu2022convnext}, ViT~\cite{Dosovitskiy2020VIT}\}. (\dag): DDAD~\cite{Guizilini2020ddad} in training set. (\ddag): predicted intrinsics are utilized for conditioning and backprojecting. Best viewed on a screen and zoomed in.}
    \vspace{-10pt}
    \resizebox{\linewidth}{!}{
    \begin{tabular}{l|ccc|ccc|ccc|ccc|ccc|ccc|ccc|ccc}
        \toprule
        \multirow{2}{*}{\textbf{Method}} & \multicolumn{3}{c|}{NuScenes} & \multicolumn{3}{c|}{DDAD} & \multicolumn{3}{c|}{ETH3D} & \multicolumn{3}{c|}{Diode (Indoor)} & \multicolumn{3}{c|}{SUN-RGBD} & \multicolumn{3}{c|}{VOID} & \multicolumn{3}{c|}{IBims-1} & \multicolumn{3}{c}{HAMMER} \\
         & $\mathrm{\delta}_{1}\uparrow$ & $\mathrm{SI}_{\log}\downarrow$ &  $\mathrm{F}_A\uparrow$ & $\mathrm{\delta}_{1}\uparrow$ & $\mathrm{SI}_{\log}\downarrow$ & $\mathrm{F}_A\uparrow$ & $\mathrm{\delta}_{1}\uparrow$ & $\mathrm{SI}_{\log}\downarrow$ & $\mathrm{F}_A\uparrow$ & $\mathrm{\delta}_{1}\uparrow$ & $\mathrm{SI}_{\log}\downarrow$ & $\mathrm{F}_A\uparrow$ & $\mathrm{\delta}_{1}\uparrow$ & $\mathrm{SI}_{\log}\downarrow$ & $\mathrm{F}_A\uparrow$ &
         $\mathrm{\delta}_{1}\uparrow$ & $\mathrm{SI}_{\log}\downarrow$ & $\mathrm{F}_A\uparrow$ &
         $\mathrm{\delta}_{1}\uparrow$ & $\mathrm{SI}_{\log}\downarrow$ & $\mathrm{F}_A\uparrow$ &
         $\mathrm{\delta}_{1}\uparrow$ & $\mathrm{SI}_{\log}\downarrow$ & $\mathrm{F}_A\uparrow$\\
        \toprule
        BTS~\cite{Lee2019bts} & $33.7$ & $68.0$ & $37.5$ & $43.0$ & $40.8$ & $40.5$ & $26.8$ & $29.9$ & $27.4$ & $19.2$ & $22.8$ & $31.6$ & $76.1$ & $14.6$ & $64.8$ & $47.4$ & $25.8$ & $64.5$ & $53.1$ & $17.5$ & $57.2$ & $3.89$ & $20.9$ & $22.8$\\
        AdaBins~\cite{Bhat2020adabins} & $33.3$ & $61.4$ & $35.2$ & $37.7$ & $44.4$ & $35.6$ & $24.3$ & $28.3$ & $25.2$ & $17.4$ & $21.6$ & $28.7$ & $77.7$ & $13.9$ & $65.4$ & $50.5$ & $23.8$ & $65.0$ & $55.0$ & $15.6$ & $57.8$ & $7.21$ & $21.5$ & $27.7$\\
        NeWCRF~\cite{Yuan2022newcrf} & $44.2$ & $49.4$ & $42.2$ & $45.6$ & $34.9$ & $41.6$ & $35.7$ & $26.1$ & $32.3$ & $20.1$ & $18.5$ & $35.3$ & $75.3$ & $11.9$ & $61.6$ & $53.1$ & $22.3$ & $67.9$ & $53.6$ & $14.7$ & $59.2$ & $1.43$ & $14.9$ & $20.8$\\
        iDisc~\cite{piccinelli2023idisc} & $39.4$ & $37.1$ & $34.5$ & $28.4$ & $32.2$ & $25.8$ & $35.6$ & $27.5$ & $31.4$ & $23.8$ & $15.8$ & $33.4$ & $83.7$ & $12.4$ & $71.0$ & $55.3$ & $20.3$ & $68.6$ & $48.9$ & $13.2$ & $55.4$ & $2.58$ & $14.0$ & $32.6$\\
        ZoeDepth~\cite{bhat2023zoedepth} & $28.3$ & $31.5$ & $26.0$ & $27.2$ & $31.7$ & $21.1$ & $35.0$ & $17.6$ & $26.4$ & $36.9$ & $12.8$ & $40.5$ & $86.7$ & $9.58$ & $75.6$ & $63.4$ & $15.9$ & $72.4$ & $58.0$ & $10.9$ & $59.6$ & $0.72$ & $9.78$ & $21.0$\\
        Metric3D\textsuperscript{\dag}~\cite{yin2023metric3d} & $72.3$ & $29.0$ & $53.9$ & $-$ & $-$ & $-$ & $\underline{45.6}$ & $18.9$ & $\mathbf{35.9}$ & $39.2$ & $11.1$ & $42.1$ & $15.4$ & $13.4$ & $14.4$ & $65.9$ & $16.2$ & $70.4$ & $\mathbf{79.7}$ & $10.1$ & $\mathbf{68.5}$ & $3.40$ & $12.1$ & $29.0$\\
        \midrule
        \ourmodel-C & $\underline{83.3}$ & $\underline{22.9}$ & $62.3$ & $\underline{83.2}$ & $\underline{21.4}$ & $59.3$ & $\mathbf{49.8}$ & $13.2$ & $\underline{33.7}$ & $60.2$ & $9.03$ & $50.0$ & $\underline{94.8}$ & $8.10$ & $\underline{81.4}$ & $86.6$ & $\underline{12.8}$ & $85.1$ & $\mathbf{79.7}$ & $8.92$ & $\underline{66.7}$ & $\mathbf{20.2}$ & $\underline{8.78}$ & $\mathbf{57.1}$\\
        \ourmodel-V & $\mathbf{86.2}$ & $\mathbf{21.7}$ & $\mathbf{64.2}$ & $\mathbf{86.4}$ & $\mathbf{20.3}$ & $\mathbf{61.8}$ & $32.6$ & $\underline{11.6}$ & $24.3$ & $\underline{77.1}$ & $\underline{6.38}$ & $\mathbf{59.4}$ & $\mathbf{96.6}$ & $\mathbf{7.05}$ & $\mathbf{81.9}$ & $\underline{89.4}$ & $\mathbf{10.9}$ & $\underline{85.7}$ & $23.9$ & $\underline{7.22}$ & $37.1$ & $\underline{13.3}$ & $\mathbf{7.41}$ & $\underline{55.9}$\\
        \midrule
        \midrule
        \ourmodel-C\textsuperscript{\ddag} & $\underline{83.3}$ & $\underline{22.9}$ & $60.9$ & $83.1$ & $\underline{21.4}$ & $57.3$ & $22.9$ & $13.1$ & $25.4$ & $60.4$ & $9.01$ & $49.9$ & $92.3$ & $8.27$ & $75.2$ & $86.5$ & $\underline{12.8}$ & $85.0$ & $\underline{79.4}$ & $8.88$ & $64.2$ & $12.7$ & $9.30$ & $54.8$\\
        \ourmodel-V\textsuperscript{\ddag} & $\mathbf{86.2}$ & $\mathbf{21.7}$ & $\underline{63.0}$ & $\mathbf{86.4}$ & $\mathbf{20.3}$ & $\underline{60.4}$ & $17.6$ & $\mathbf{11.4}$ & $21.4$ & $\mathbf{77.4}$ & $\mathbf{6.36}$ & $\underline{58.6}$ & $\underline{94.8}$ & $\underline{7.17}$ & $75.9$ & $\mathbf{90.2}$ & $\mathbf{10.9}$ & $\mathbf{86.2}$ & $17.5$ & $\mathbf{7.20}$ & $36.5$ & $2.56$ & $8.35$ & $53.8$\\
        \bottomrule
    \end{tabular}}
    \label{tab:results:zeroshot}
    \vspace{-12pt}
\end{table*}

\subsection{Experimental Setup}
\label{ssec:experiments:setup}

\noindent{}\textbf{In-domain training datasets.}
The training dataset utilized is the ensemble of Argoverse2~\cite{2021argoverse2}, Waymo~\cite{sun2020waymo}, DrivingStereo~\cite{yang2019drivingstereo},  Cityscapes~\cite{Cordts2016cityscapes}, BDD100K~\cite{yu2020bdd100k}, Mapillary-PSD~\cite{Lopez2020mapillary}, A2D2~\cite{geyer2020a2d2}, ScanNet~\cite{dai2017scannet}, and Taskonomy~\cite{zamir2018taskonomy}.
The resulting dataset amounts roughly to 3M real-world images with different cameras and domains, compared to, \eg Metric3D~\cite{yin2023metric3d} and ZeroDepth~\cite{guizilini2023zerodepth} which exploit 8M and 17M training images, respectively.

\noindent{}\textbf{Zero-shot testing datasets.}
We evaluate the generalizability of the compared models by testing them on ten datasets not seen during training.
More precisely, each method is tested on validation splits from SUN-RGBD~\cite{Song2015sunrgbd} without NYU split, Diode Indoor~\cite{Vasiljevic2019diode}
, IBims-1~\cite{koch2022ibims}, VOID~\cite{wong2020void} HAMMER~\cite{jung2022hammer}, ETH-3D~\cite{schoeps2017eth3d}, nuScenes~\cite{nuscenes}, and DDAD~\cite{Guizilini2020ddad} with split proposed in~\cite{piccinelli2023idisc} and evaluated with official masks.
Also, \ourmodel and the models from~\cite{yin2023metric3d, guizilini2023zerodepth} are zero-shot-tested on NYU-Depth V2~\cite{silberman2012nyu} and KITTI~\cite{Geiger2012kitti}.
In particular, KITTI testing is performed on the corrected Eigen-split test set~\cite{Eigen2014} with the Garg evaluation mask~\cite{Garg2016}, while NYU testing uses the evaluation mask from~\cite{Lee2019bts}.

\noindent{}\textbf{Evaluation Details.}
All methods have been re-evaluated with a fair and consistent pipeline.
In particular, we do not exploit any test-time augmentations.
We use training image shapes for zero-shot testing and evaluate on the same validation splits and masks.
Unfortunately, ZeroDepth lacks full code reproducibility, thus we report results from the original paper only, and for visualization, we utilize their provided code and weights.
When methods do not report the configuration for a specific test dataset, we use the settings of NYU and KITTI for indoor and outdoor testing, respectively.
We utilize common depth estimation evaluation metrics: root mean square error ($\mathrm{RMS}$) and its log variant ($\mathrm{RMS_{log}}$), absolute mean relative error ($\mathrm{A.Rel}$), the percentage of inlier pixels ($\mathrm{\delta}_i$) with threshold $1.25^{i}$, scale-invariant error in log-scale ($\mathrm{SI_{log}}$): $100 \sqrt{\mathrm{Var}(\varepsilon_{\log})}$.
In addition, we report point-cloud-based metrics proposed in~\cite{ornek20222metrics}, namely Chamfer Distance ($\mathrm{CD}$) and F-score ($\mathrm{F_{A}}$), with the latter aggregated as the area under the curve up to $1/20$ of the datasets' maximum depth.
All methods exploit GT intrinsics during evaluation.
Nonetheless, we present results both with and without GT intrinsics for \ourmodel.

\noindent{}\textbf{Implementation Details.}
\ourmodel is implemented in PyTorch~\cite{pytorch} and CUDA~\cite{nickolls2008cuda}.
For training, we use the AdamW~\cite{Loshchilov2017adamw} optimizer ($\beta_1=0.9$, $\beta_2=0.999$) with an initial learning rate of $0.0001$. The learning rate is divided by a factor of 10 for the backbone weights for every experiment and weight decay is set to $0.1$.
As the learning rate scheduler, we exploit Cosine Annealing to one-tenth starting from 30\% of the training.
We run 1M optimization iterations with a batch size of 128, each training dataset is uniformly represented in each batch.
In particular, we sample 64 images and then we sample two different augmented views of the same image for consistency loss.
The augmentations include both geometric and appearance (random brightness, gamma, saturation, hue shift, and grayscale) augmentations.
ViT-L~\cite{Dosovitskiy2020VIT} backbone is initialized with weights from DINO-pre-trained~\cite{caron2021dino} models, and ConvNext-L~\cite{liu2022convnext} is ImageNet~\cite{Deng2010imagenet}-pre-trained.
The required training time amounts to roughly 12 days on 8 NVIDIA A100.
Ablations are conducted with three different seeds and for 100k training iterations, using a randomly sampled subset with a size equal to 20\% of the original training set.

\begin{table}[t]
    \centering
    \caption{\textbf{Comparison on NYU test set.} The first five methods are trained on NYU and tested on it. The last four methods are tested in a zero-shot setting. \ourmodel-\{C, V\}: \ourmodel-\{ConvNext~\cite{liu2022convnext}, ViT~\cite{Dosovitskiy2020VIT}\}. (\dag): MiDaS~\cite{ranftl2020midas} pre-trained.} 
    \vspace{-10pt}
    \resizebox{\columnwidth}{!}{
    \begin{tabular}{l|ccc|ccccc}
        \toprule
        \multirow{2}{*}{\textbf{Method}} & $\mathrm{\delta}_{{0.5}}$ & $\mathrm{\delta}_{1}$ & $\mathrm{F}_A$ & $\mathrm{A.Rel}$ & $\mathrm{RMS}$ & $\mathrm{RMS}_{\log}$ & $\mathrm{CD}$ & $\mathrm{SI}_{\log}$\\
         & \multicolumn{3}{c|}{\textit{Higher is better}} & \multicolumn{5}{c}{\textit{Lower is better}}\\
        \toprule
        BTS~\cite{Lee2019bts} & $66.1$ & $88.5$ & $74.0$ & $10.9$ & $0.391$ & $0.141$ & $0.160$ & $11.5$\\
        AdaBins~\cite{Bhat2020adabins} & $68.1$ & $90.1$ & $74.7$ & $10.3$ & $0.365$ & $0.131$ & $0.156$ & $10.6$\\
        NeWCRF~\cite{Yuan2022newcrf} & $69.6$ & $92.1$ & $75.8$ & $9.56$ & $0.333$ & $0.119$ & $0.147$ & $9.16$\\
        iDisc~\cite{piccinelli2023idisc} & $74.5$ & $93.8$ & $78.2$ & $8.61$ & $0.313$ & $0.110$ & $0.133$ & $8.85$\\
        ZoeDepth\textsuperscript{\dag}~\cite{bhat2023zoedepth} & $78.4$ & $95.2$ & $80.1$ & $7.70$ & $0.278$ & $0.097$ & $0.125$ & $7.19$\\
        ZeroDepth~\cite{guizilini2023zerodepth} & $-$ & $90.1$ & $-$ & $10.0$ & $0.380$ & $-$ & $-$ & $-$\\
        Metric3D~\cite{yin2023metric3d} & $76.3$ & $92.6$ & $77.8$ & $9.38$ & $0.337$ & $0.120$ & $0.146$ & $9.13$\\
        \midrule
        \ourmodel-C & $\underline{85.4}$ & $\underline{97.2}$& $\underline{84.3}$ & $\underline{6.26}$ & $\underline{0.232}$ & $\underline{0.082}$ & $\underline{0.101}$ & $\underline{6.41}$ \\
        \ourmodel-V & $\mathbf{88.6}$ & $\mathbf{98.4}$ & $\mathbf{85.9}$ & $\mathbf{5.78}$ & $\mathbf{0.201}$ & $\mathbf{0.073}$ & $\mathbf{0.092}$ & $\mathbf{5.27}$ \\
        \bottomrule
    \end{tabular}}
    \label{tab:results:nyu}
    \vspace{-14pt}
\end{table}

\subsection{Comparison with the State of the Art}
\label{ssec:experiments:comparison}

Our method consistently outperforms previous SotA methods as shown in \Cref{tab:results:zeroshot}.
We particularly excel in the scale-invariant aspect, represented by $\mathrm{SI}_{\log}$, with an average 34.0\% improvement, and an average 12.3\% improvement for $\mathrm{\delta}_1$ and $\mathrm{F}_A$.
However, \ourmodel could fail to capture the specific scene scales in certain cases, \eg in ETH3D and IBims-1.
This pitfall is demonstrated by the drop in scale-dependent metrics, \eg $\mathrm{F}_A$ drop is 11.8\% and 31.4\%, respectively, although having a clear scale-invariant improvement of 36.9\% and 28.5\%. 
Therefore, we speculate that our method would still greatly benefit from domain-specific fine-tuning.

\begin{table}[t]
    \centering
    \caption{\textbf{Comparison on KITTI Eigen-split test set.} The first five methods are trained on KITTI and tested on it. The last four methods are tested in a zero-shot setting. \ourmodel-\{C, V\}: \ourmodel-\{ConvNext~\cite{liu2022convnext}, ViT~\cite{Dosovitskiy2020VIT}\}. (\dag): MiDaS~\cite{ranftl2020midas} pre-trained. }
    \vspace{-10pt}
    \resizebox{\columnwidth}{!}{
    \begin{tabular}{l|ccc|ccccc}
        \toprule
        \multirow{2}{*}{\textbf{Method}} & $\mathrm{\delta}_{{0.5}}$ & $\mathrm{\delta}_{1}$ & $\mathrm{F}_A$ & $\mathrm{A.Rel}$ & $\mathrm{RMS}$ & $\mathrm{RMS}_{\log}$ & $\mathrm{CD}$ & $\mathrm{SI}_{\log}$\\
         & \multicolumn{3}{c|}{\textit{Higher is better}} & \multicolumn{5}{c}{\textit{Lower is better}}\\
        \toprule
        BTS~\cite{Lee2019bts} & $86.9$ & $96.2$ & $82.0$ & $5.63$ & $2.43$ & $0.089$ & $0.42$ & $8.18$\\
        AdaBins~\cite{Bhat2020adabins} & $86.2$ & $96.3$ & $81.5$ & $5.85$ & $2.38$ & $0.089$ & $0.429$ & $8.10$\\
        NeWCRF~\cite{Yuan2022newcrf} & $88.9$ & $97.5$ & $82.7$ & $5.20$ & $2.07$ & $0.078$ & $0.388$ & $7.00$\\
        iDisc~\cite{piccinelli2023idisc} & $89.2$ & $97.5$ & $83.1$ & $5.09$ & $2.07$ & $0.077$ & $0.380$ & $7.11$\\
        ZoeDepth\textsuperscript{\dag}~\cite{bhat2023zoedepth} & $87.4$ & $96.5$ & $82.1$ & $5.76$ & $2.39$ & $0.089$ & $0.431$ & $7.47$\\
        ZeroDepth~\cite{guizilini2023zerodepth} & $-$ & $89.2$ & $-$ & $10.2$ & $4.38$ & $0.196$ & $-$ & $-$\\
        Metric3D~\cite{yin2023metric3d} & $88.9$ & $97.5$ & $82.9$ & $5.33$ & $2.26$ & $0.081$ & $0.392$ & $7.28$\\
        \midrule
        \ourmodel-C & $\underline{91.1}$ & $\underline{97.9}$ & $\underline{83.9}$ & $\underline{4.69}$ & $\underline{2.00}$ & $\underline{0.072}$ & $\underline{0.371}$ & $\underline{6.71}$\\
        \ourmodel-V & $\mathbf{93.4}$ & $\mathbf{98.6}$ & $\mathbf{85.0}$ & $\mathbf{4.21}$ & $\mathbf{1.75}$ & $\mathbf{0.064}$ & $\mathbf{0.338}$ & $\mathbf{5.84}$\\
        \bottomrule
    \end{tabular}}
    \label{tab:results:kitti}
    \vspace{-14pt}
\end{table}
\begin{figure*}[t]
    \renewcommand{\arraystretch}{2}
    \centering
    \small
    \begin{tabular}{cc|cccc|c}
        \multirow{1}{*}[0.5in]{\rotatebox[origin=c]{90}{KITTI}}
        & \includegraphics[width=0.14\linewidth]{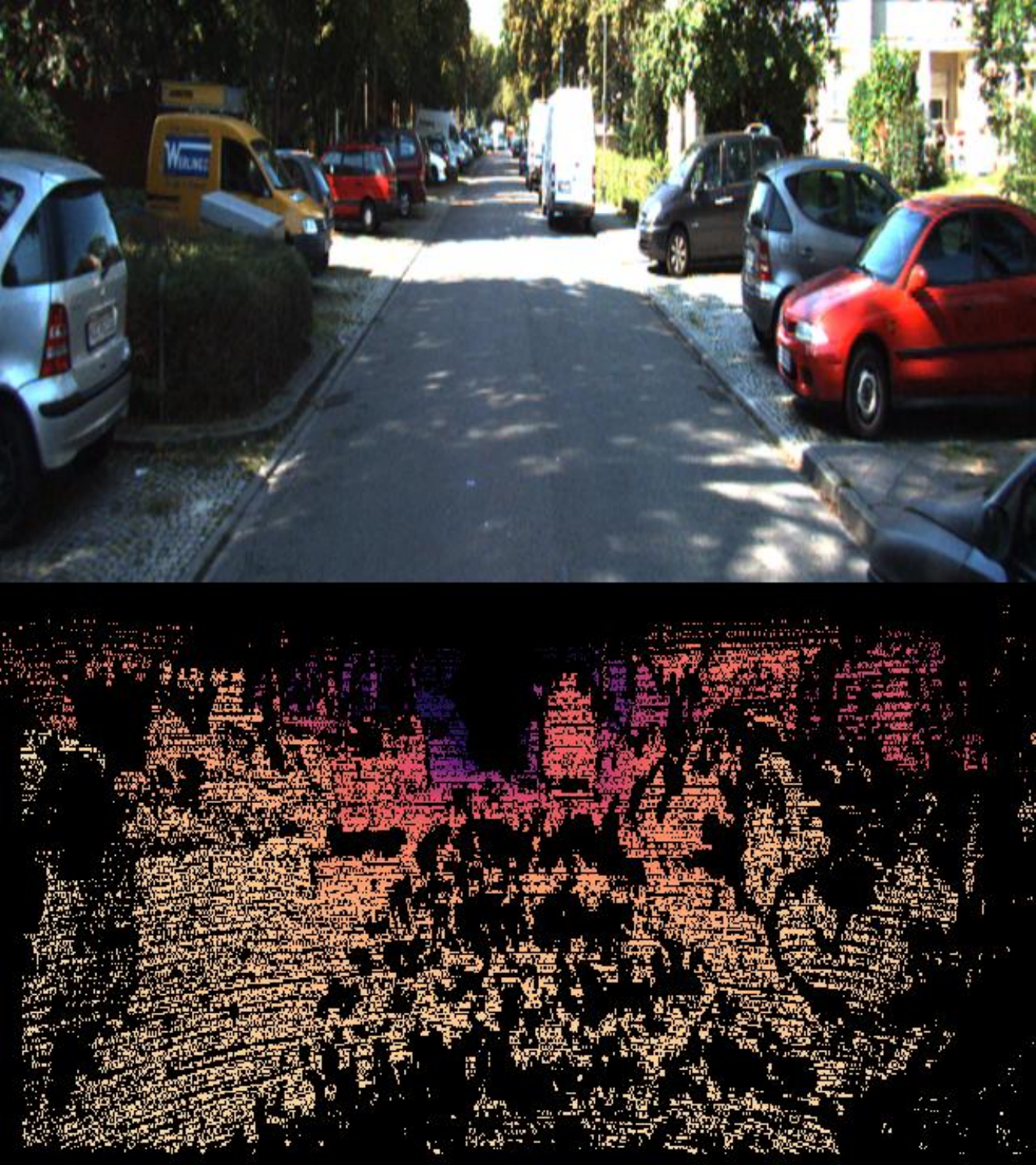}
        & \includegraphics[width=0.14\linewidth]{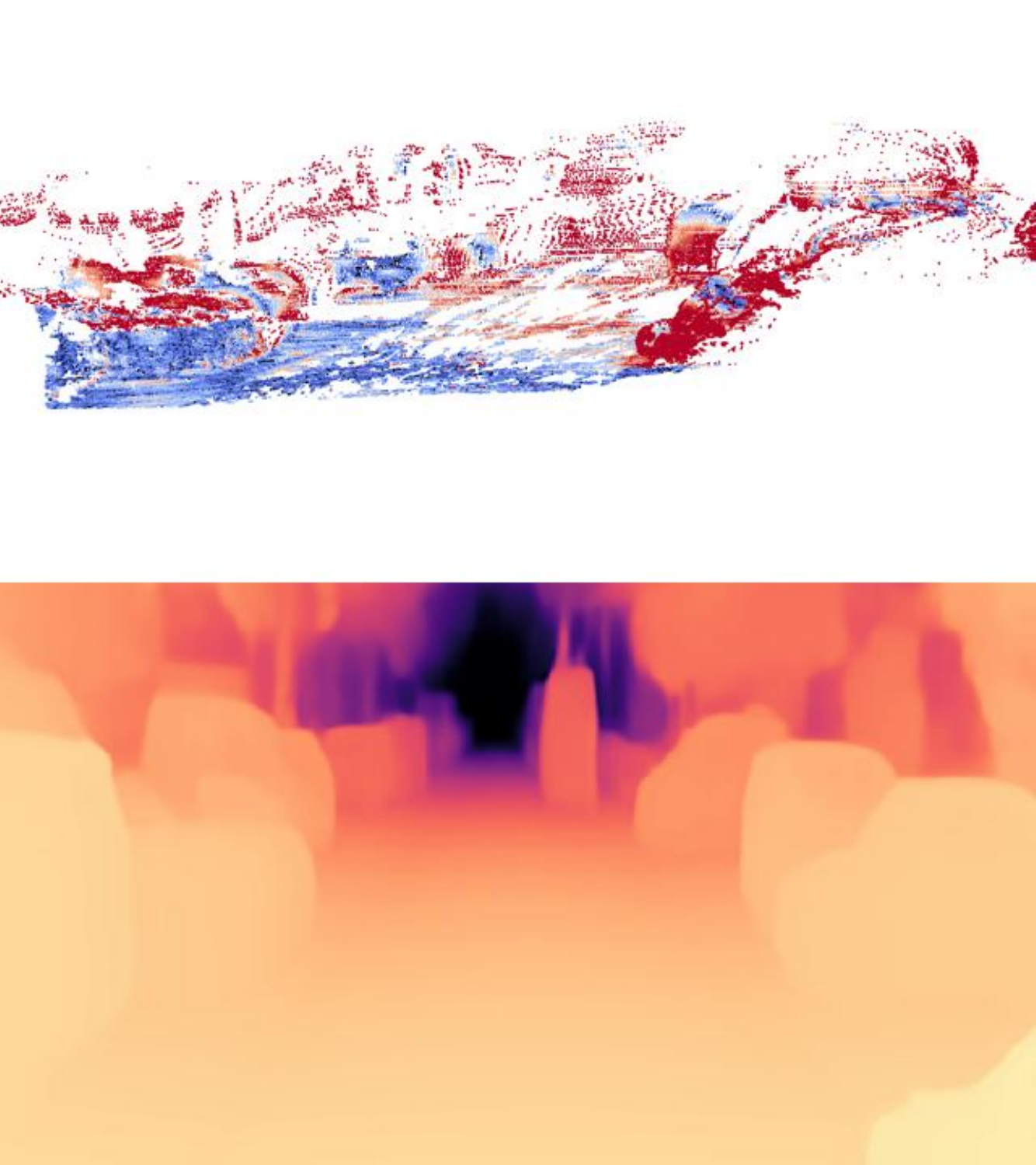}
        & \includegraphics[width=0.14\linewidth]{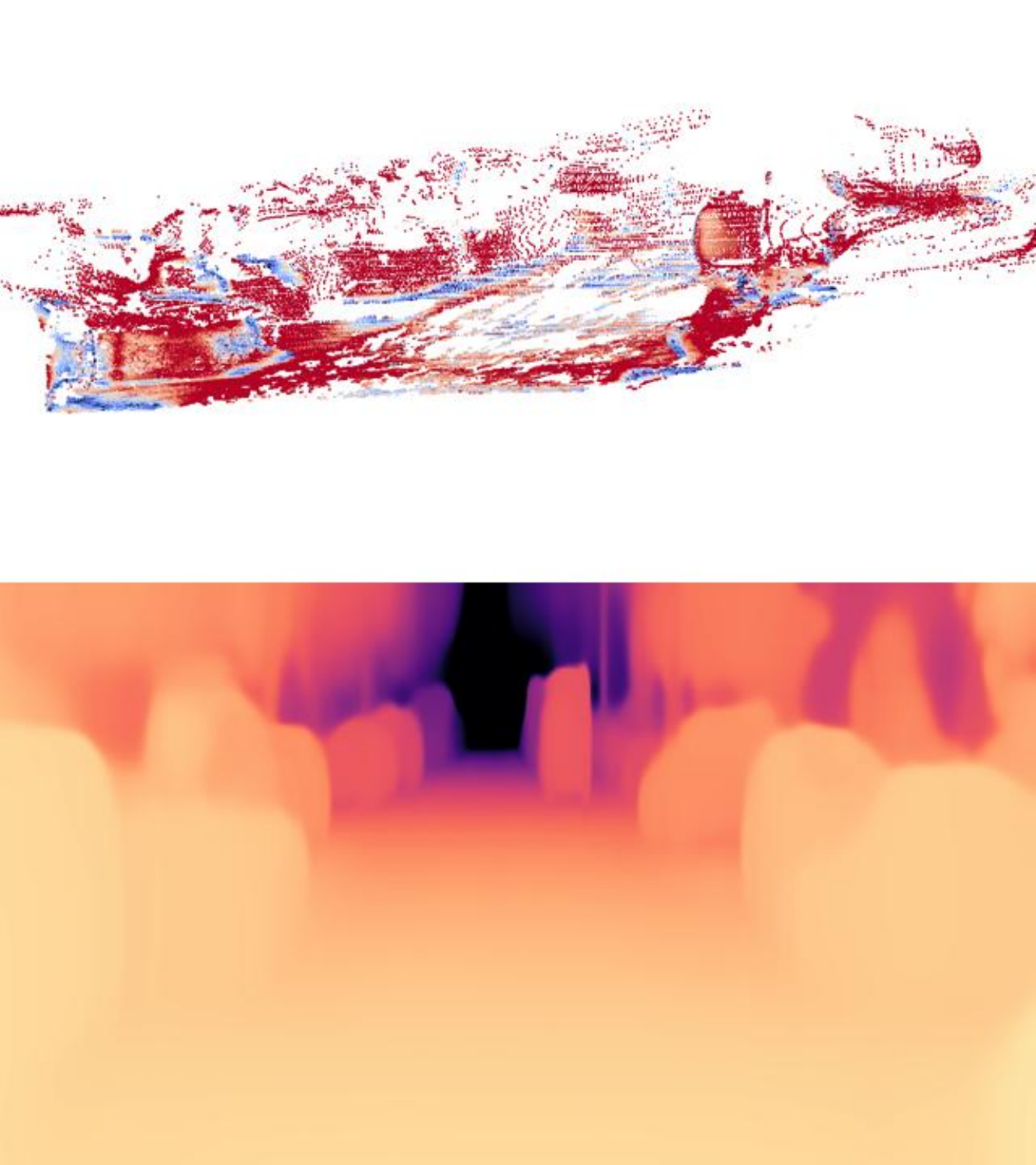}
        & \includegraphics[width=0.14\linewidth]{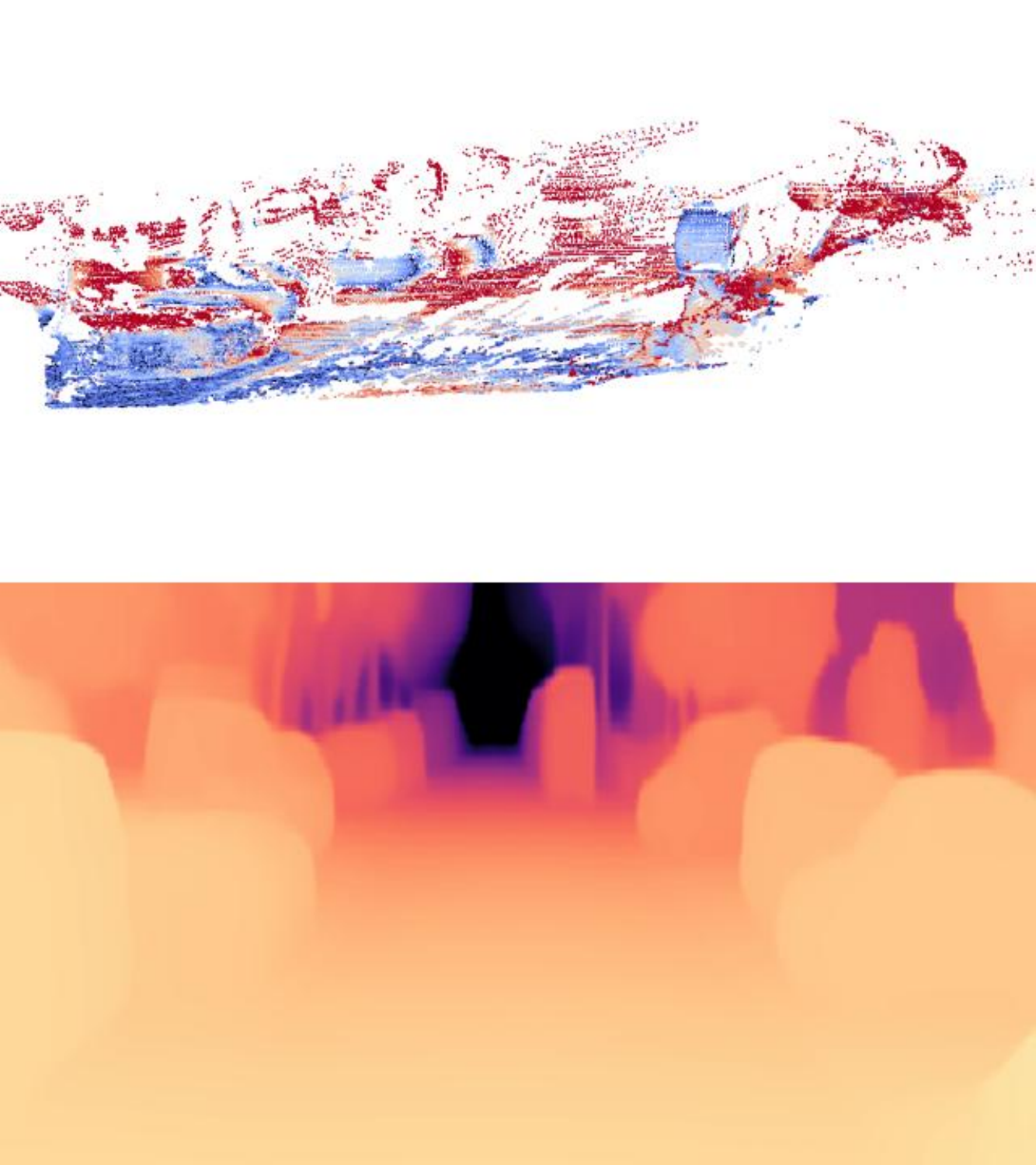}
        & \includegraphics[width=0.14\linewidth]{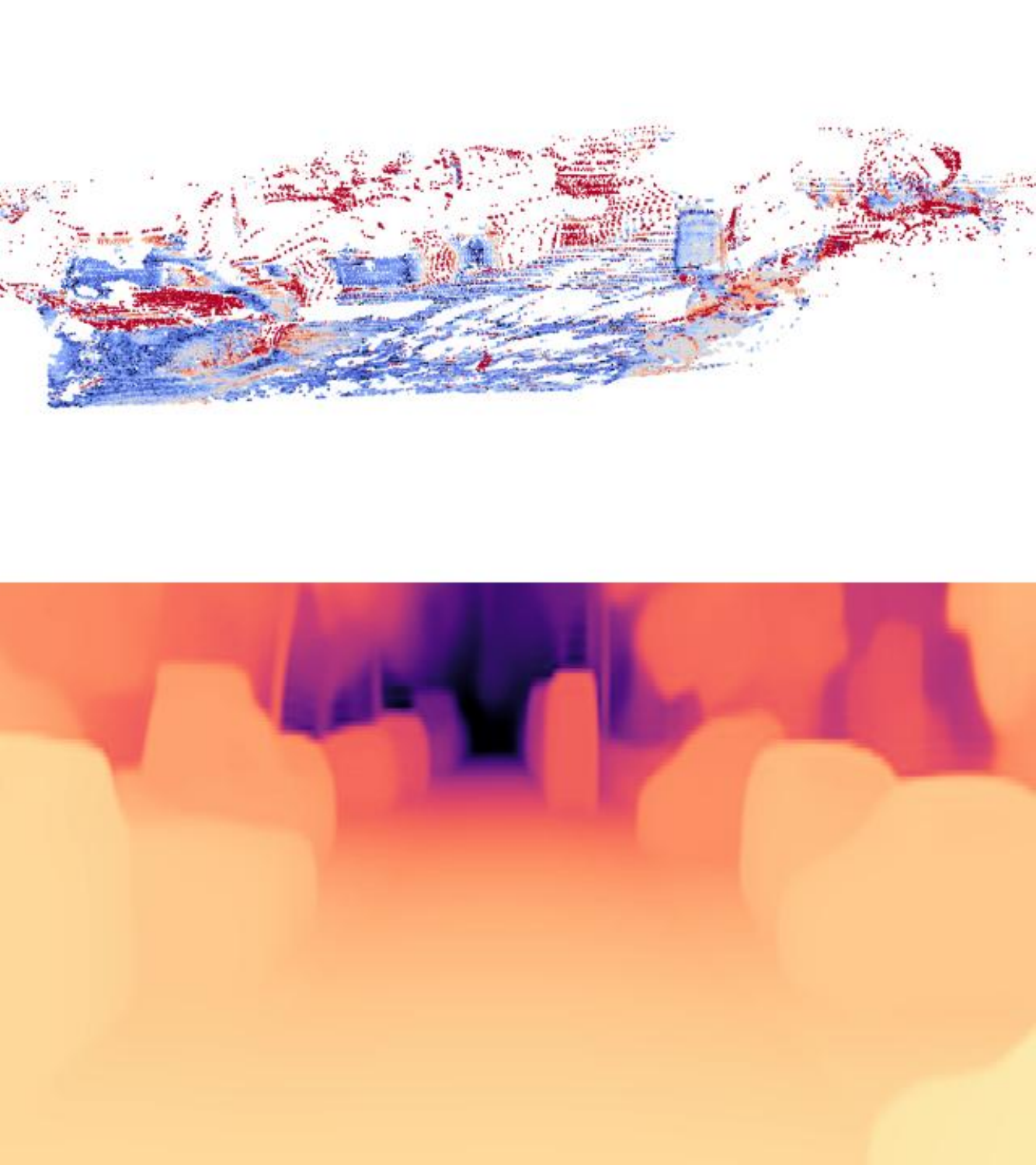}
        & \includegraphics[width=0.075\linewidth]{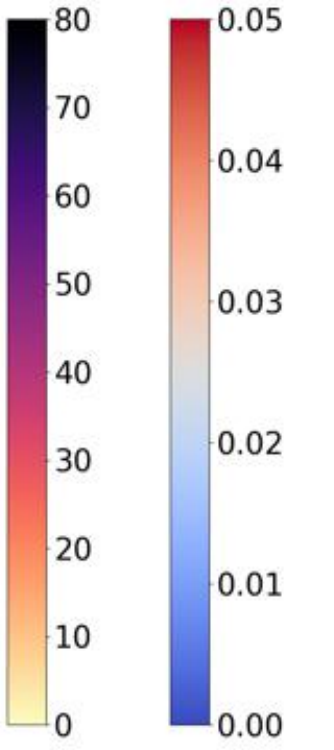} \vspace{-6pt} \\
        \multirow{1}{*}[0.5in]{\rotatebox[origin=c]{90}{NYUv2}}
        & \includegraphics[width=0.14\linewidth]{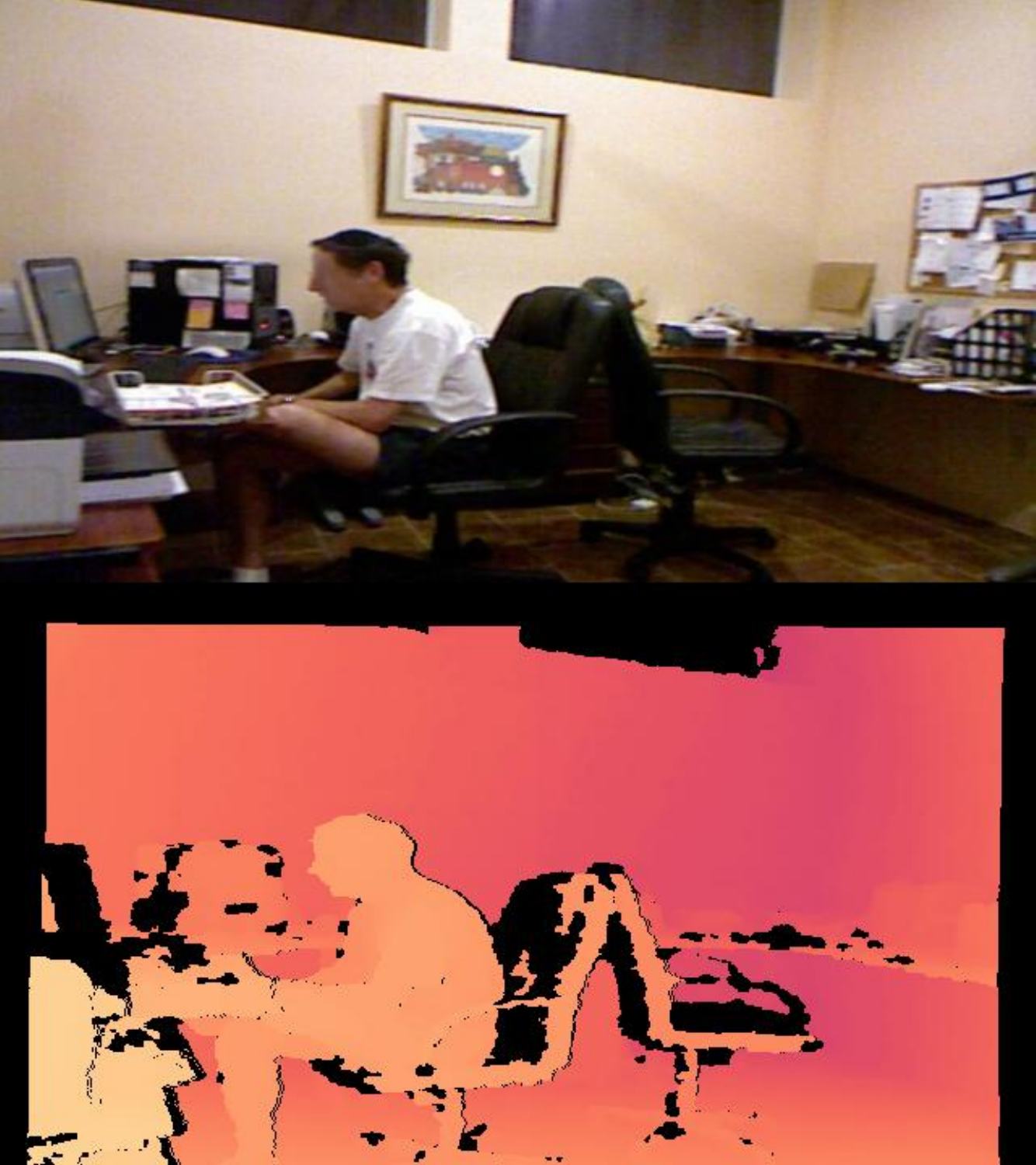}
        & \includegraphics[width=0.14\linewidth]{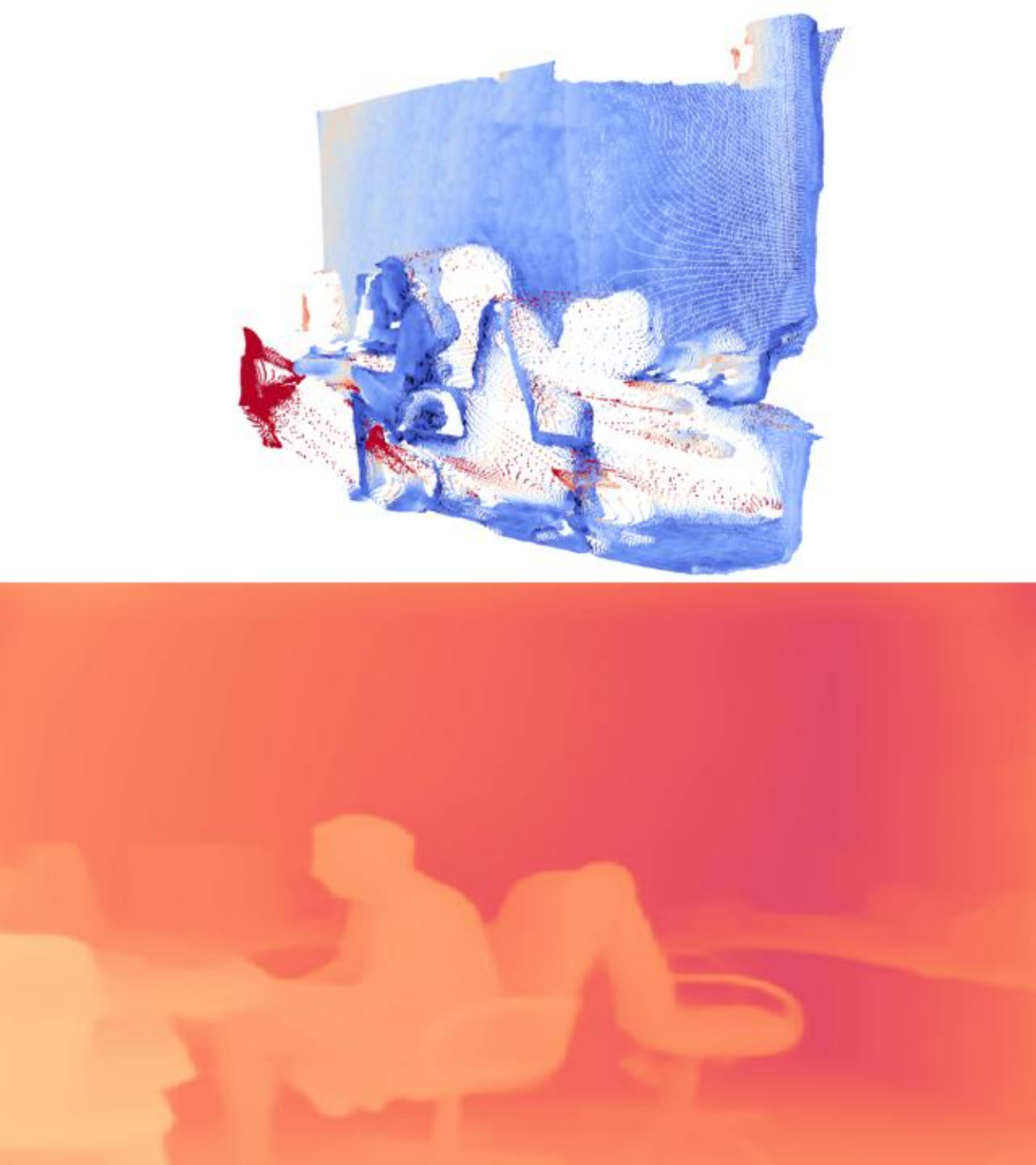}
        & \includegraphics[width=0.14\linewidth]{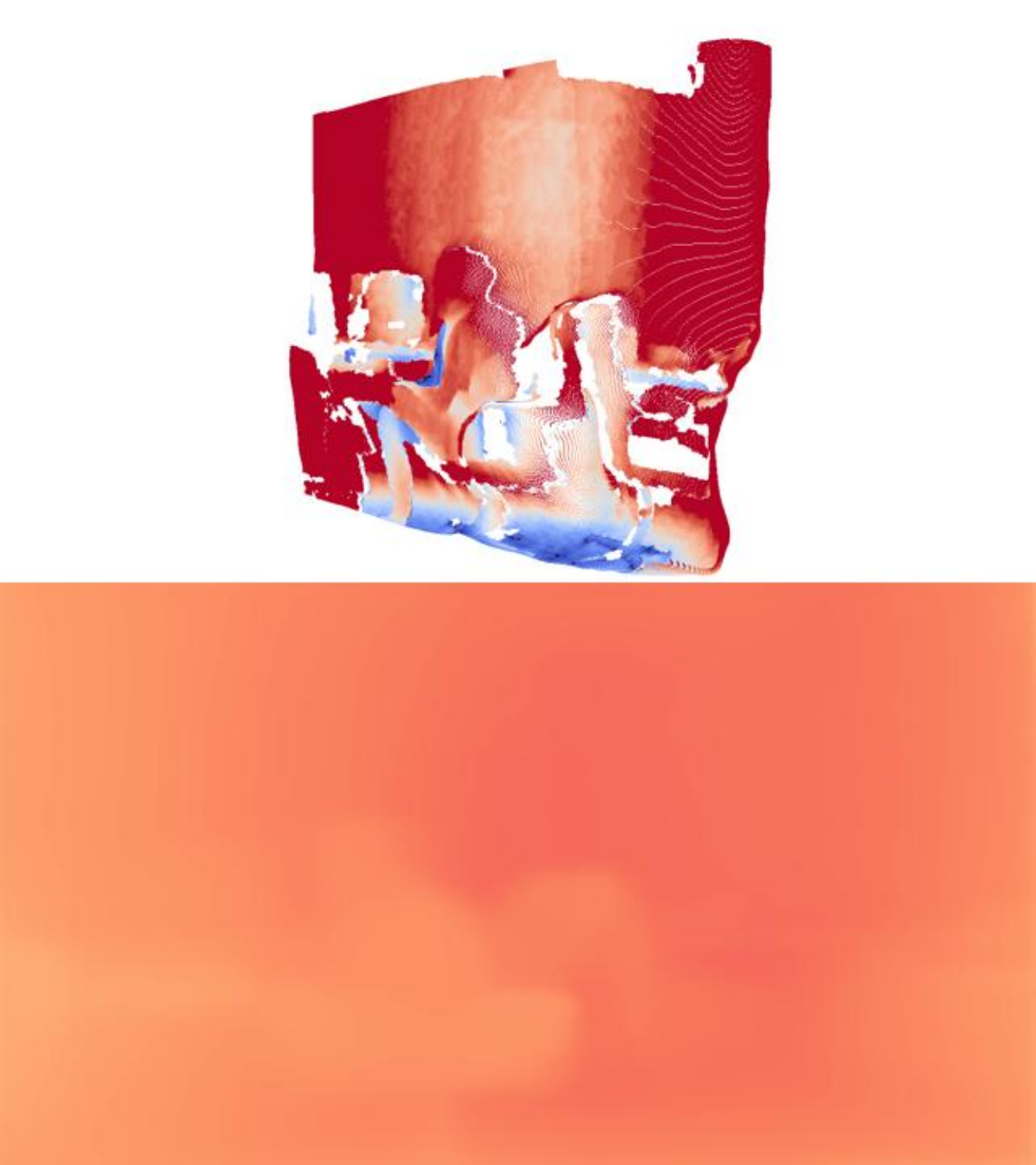}
        & \includegraphics[width=0.14\linewidth]{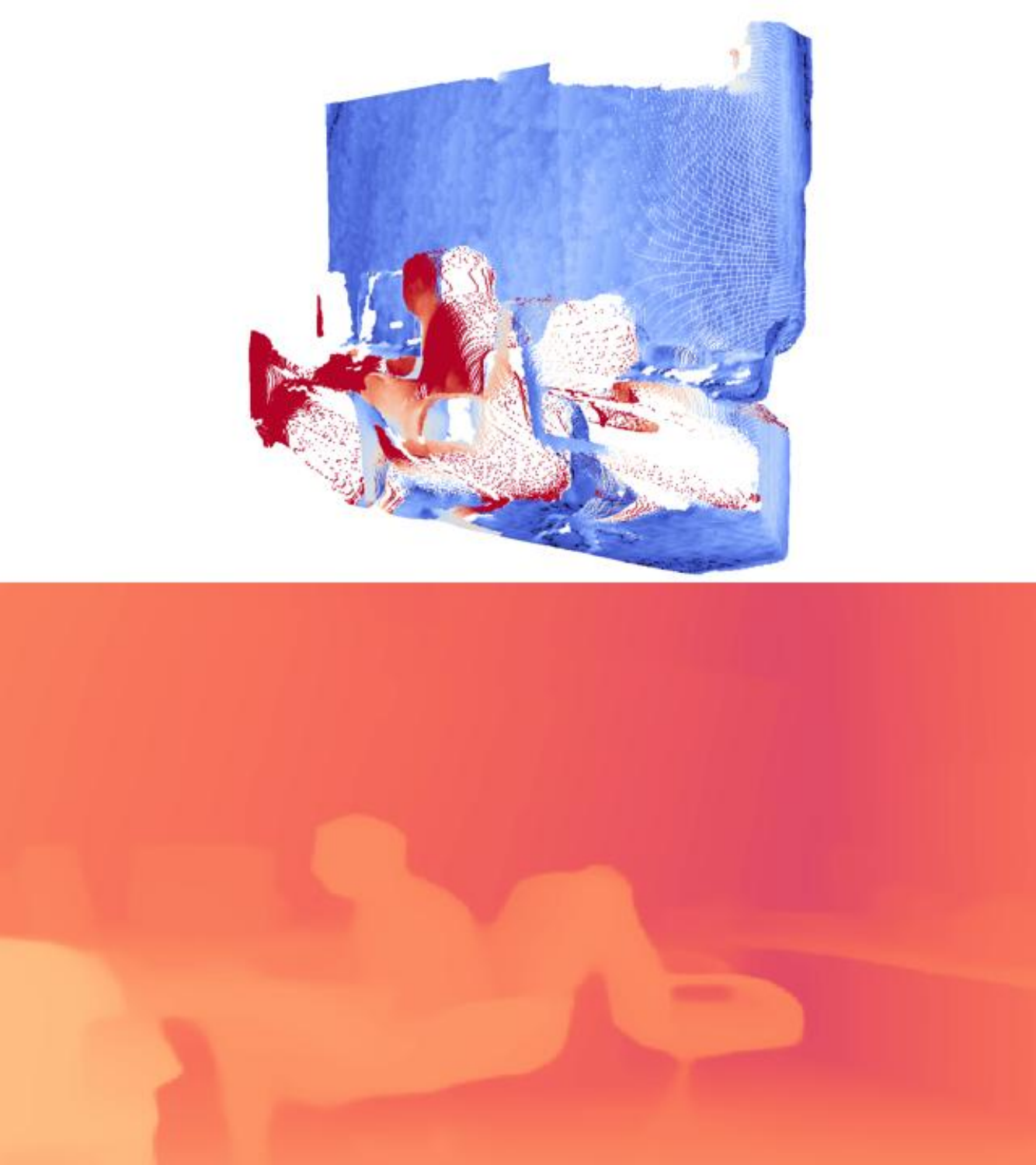}
        & \includegraphics[width=0.14\linewidth]{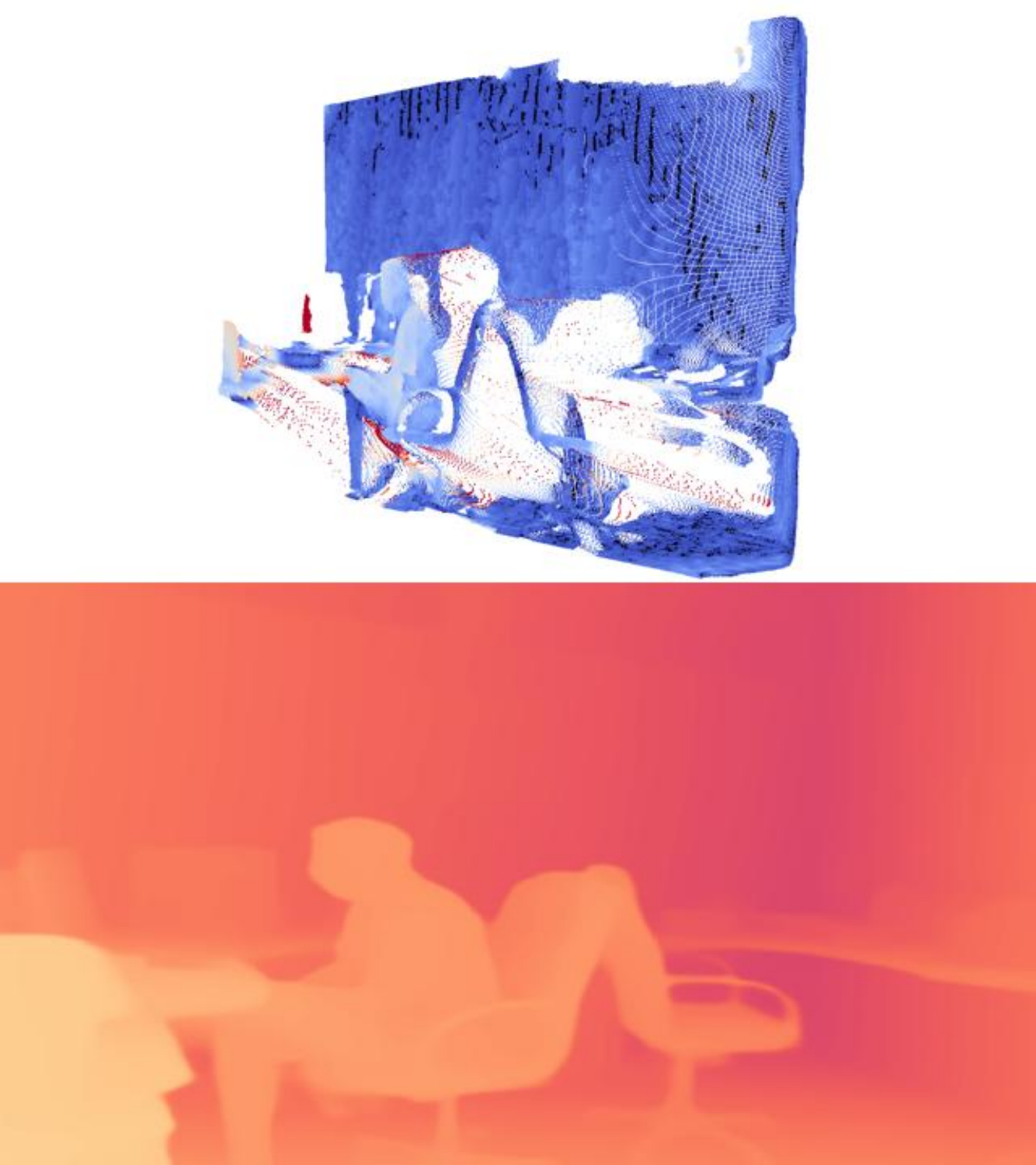}
        & \includegraphics[width=0.075\linewidth]{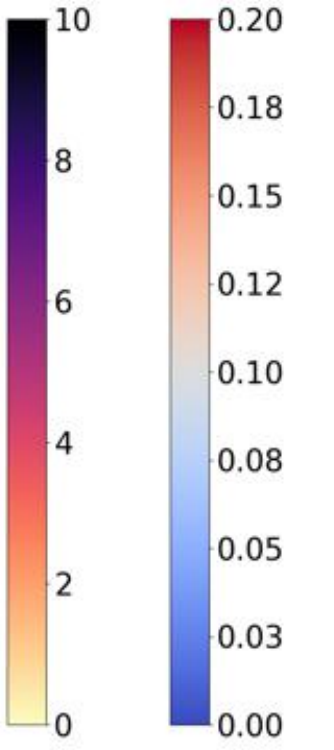} \vspace{-6pt} \\
        \multirow{1}{*}[0.5in]{\rotatebox[origin=c]{90}{Diode}}
        & \includegraphics[width=0.14\linewidth]{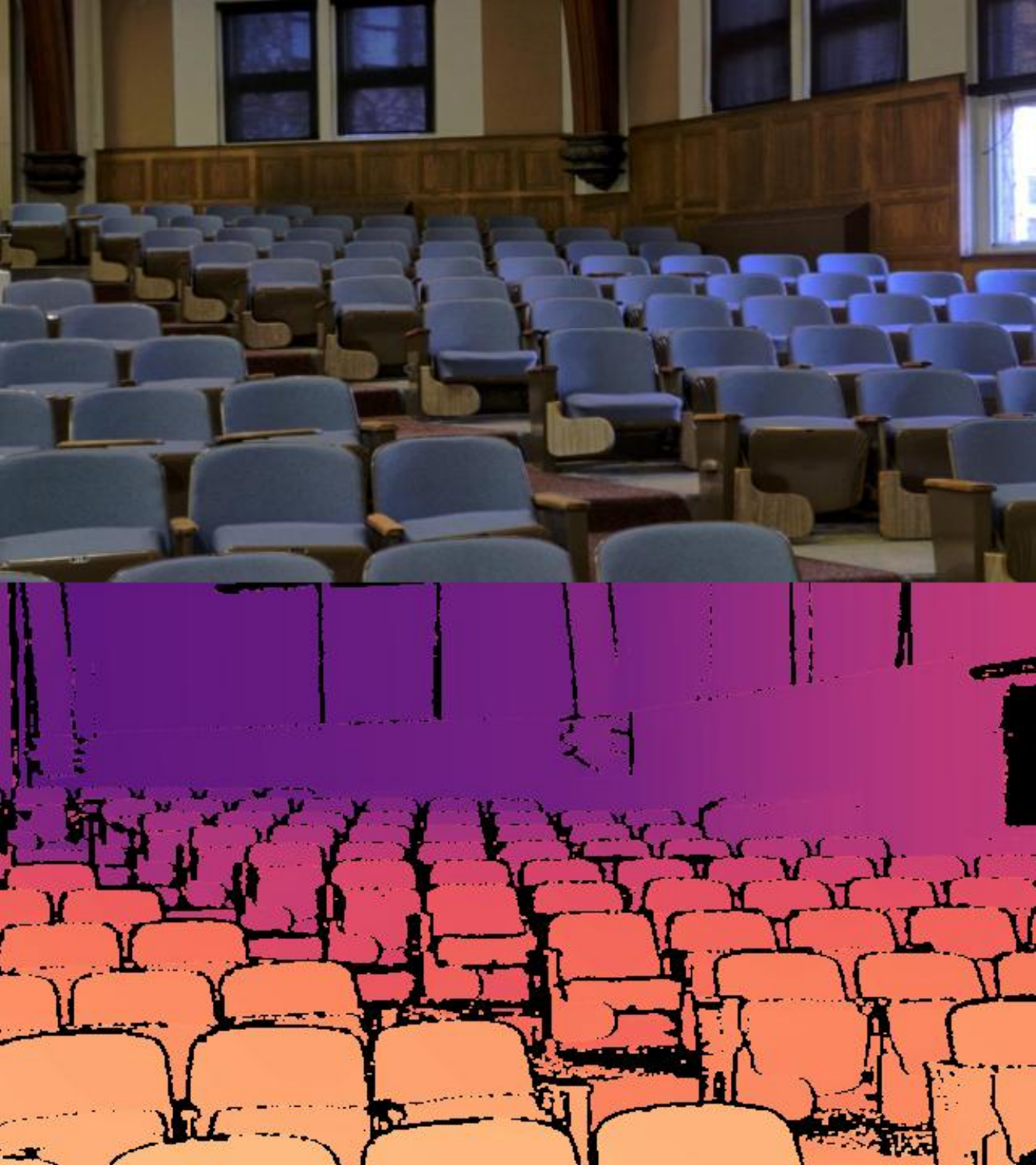}
        & \includegraphics[width=0.14\linewidth]{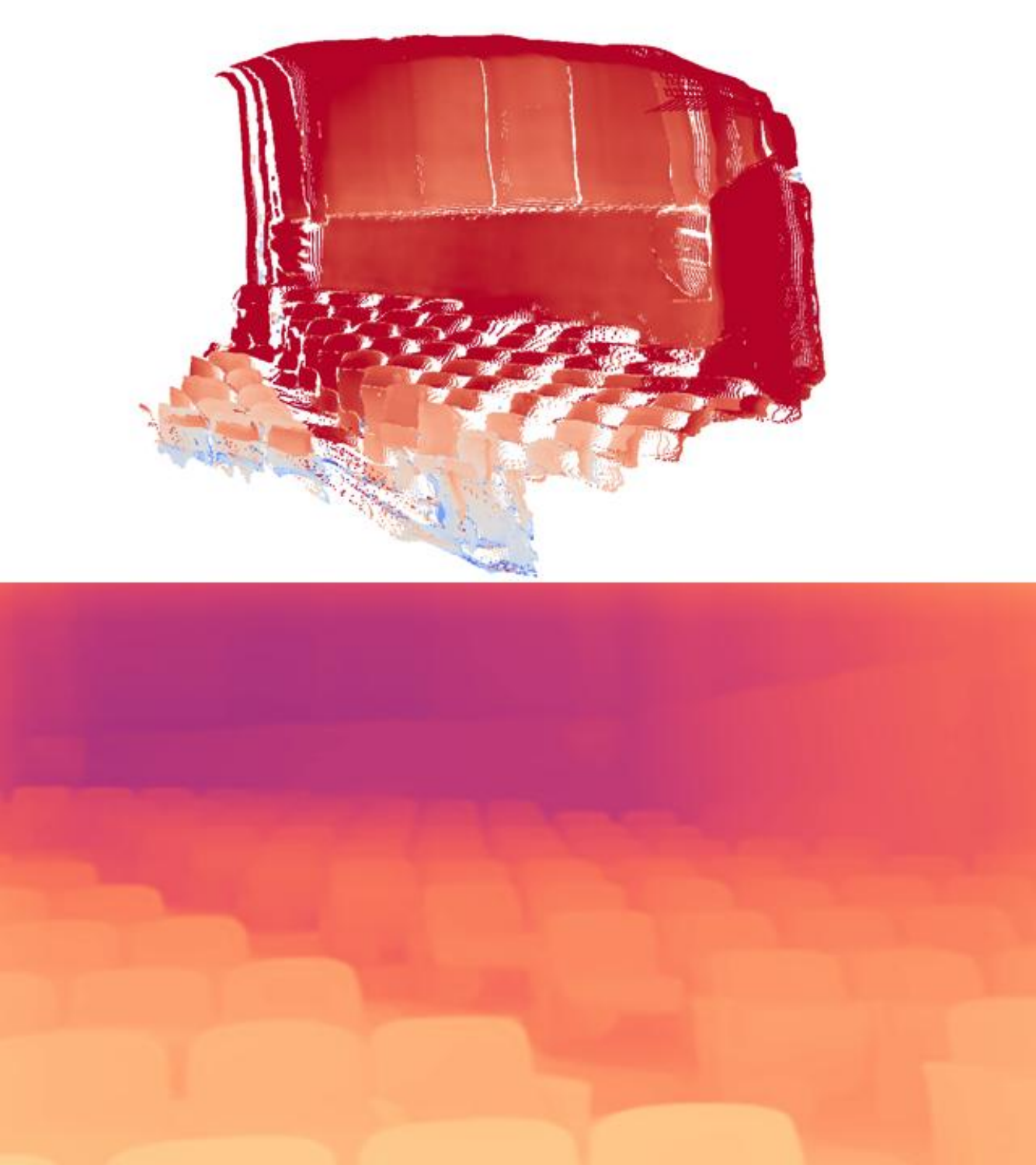}
        & \includegraphics[width=0.14\linewidth]{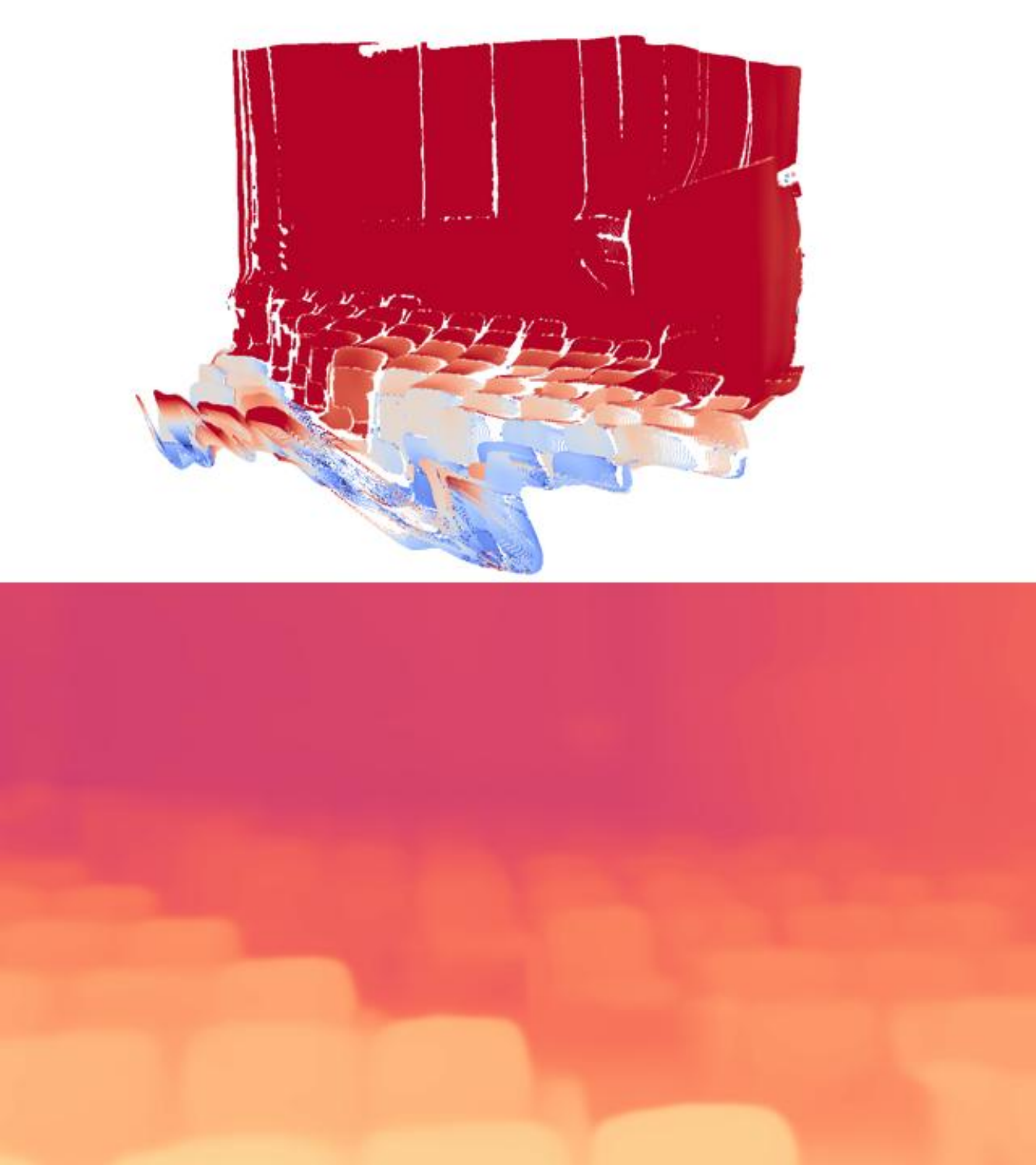}
        & \includegraphics[width=0.14\linewidth]{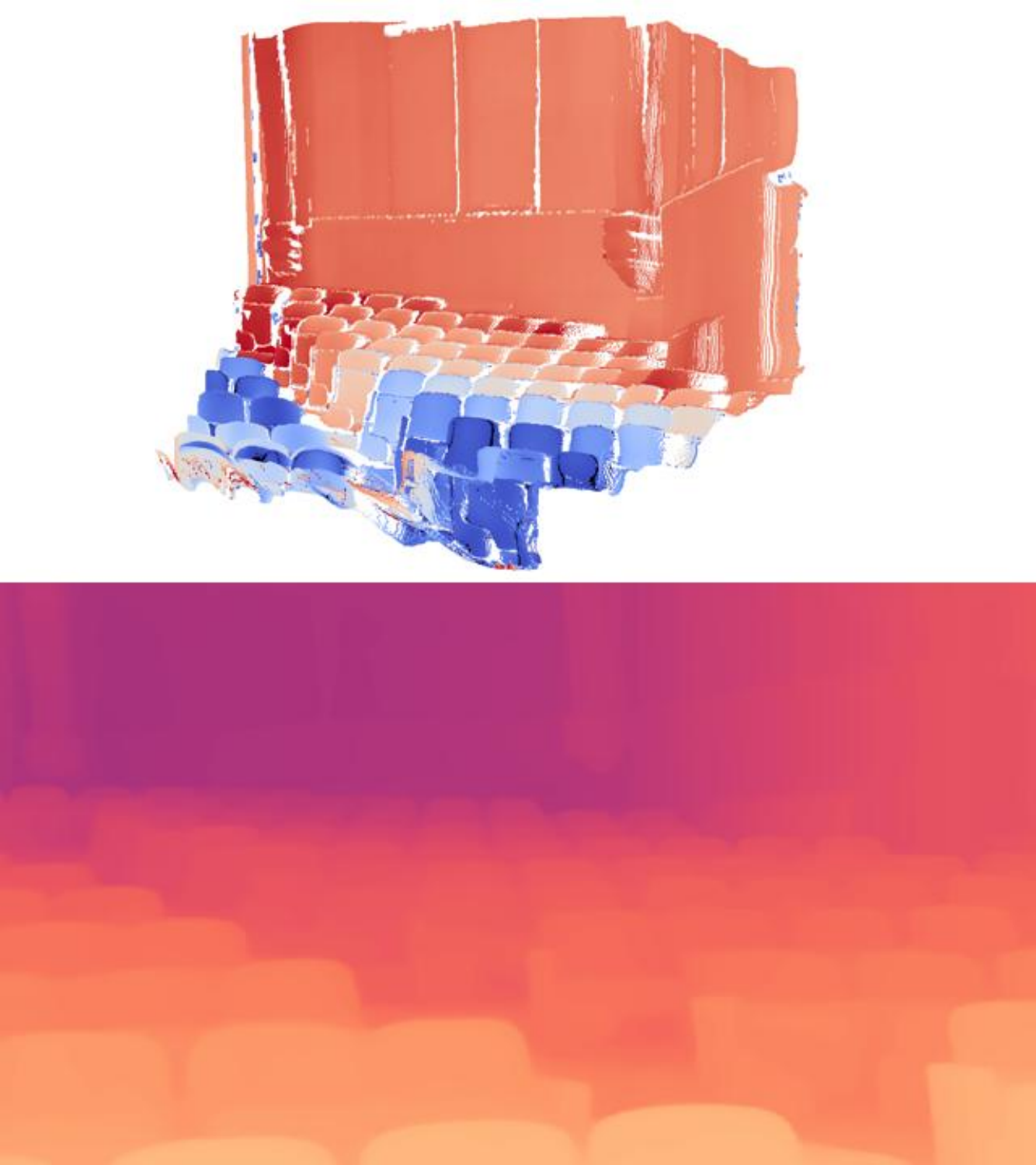}
        & \includegraphics[width=0.14\linewidth]{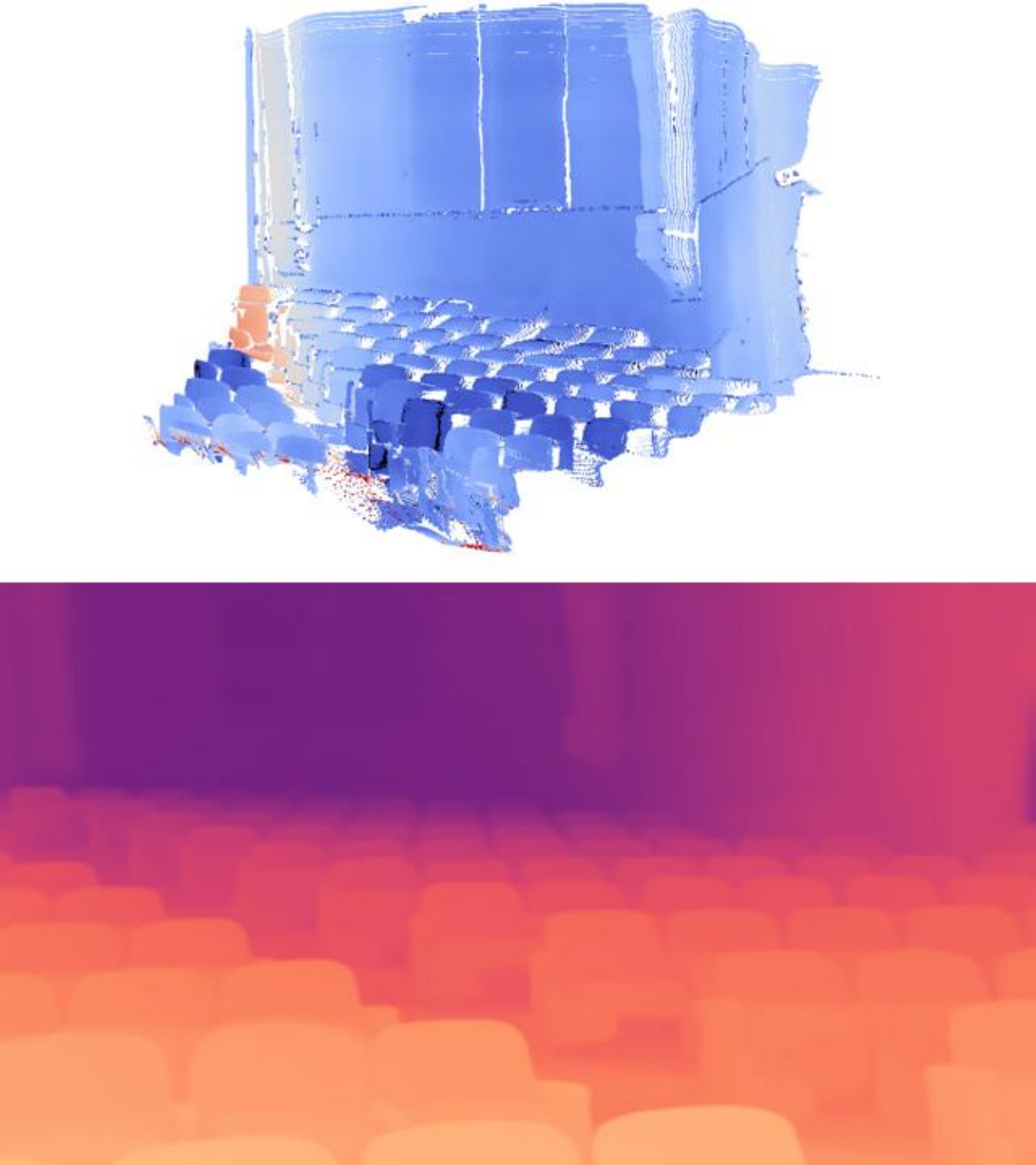}
        & \includegraphics[width=0.075\linewidth]{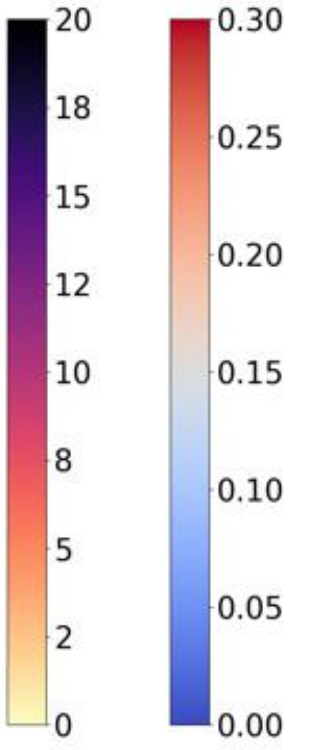} \vspace{-6pt} \\
        & RGB \& GT & ZoeDepth\textsuperscript{\dag}~\cite{bhat2023zoedepth} & ZeroDepth~\cite{guizilini2023zerodepth} & Metric3D~\cite{yin2023metric3d} & \ourmodel & Meters $|$ $\mathrm{A.Rel}$ \\
    \end{tabular}
    \vspace{-8pt}
    \caption{\textbf{Zero-shot qualitative results.} Each pair of consecutive rows corresponds to one test sample. Each odd row shows the input RGB image and the predicted pointcloud color-coded with \textit{coolwarm} based on the absolute relative error. Each even row shows GT depth and the predicted depth. The last column represents the specific colormap ranges for depth and error. (\dag): KITTI and NYU in the training set.}
    \label{fig:results:main_vis}
    \vspace{-12pt}
\end{figure*}

The last two rows in \Cref{tab:results:zeroshot} present \ourmodel in its whole design, namely functioning with solely the input image by self-prompting the predicted dense camera representation, as detailed in~\cref{eqn:method:selfconst}.
Experiments show that not only is the performance preserved for most of the test sets, but \ourmodel with the bootstrapped camera can also outperform models with GT camera, \eg $\mathrm{SI}_{\log}$ in ETH3D and IBims-1.
On the other hand, in cases with particularly out-of-domain camera types, such as ETH3D or HAMMER, bootstrapping camera prediction results in additional noise for scaled depth prediction, thus worsening results for $\mathrm{\delta}_{1}$.

\Cref{tab:results:nyu} and \Cref{tab:results:kitti} display results on the two popular benchmark NYU~\cite{silberman2012nyu} and KITTI~\cite{Geiger2012kitti} Eigen-split.
\ourmodel sets the state of the art in these two benchmarks despite being compared with models trained on the same domain.
Importantly, the KITTI Depth Prediction Benchmark, which provides a perfectly fair evaluation, underscores the excellent zero-shot performance of our method and its robustness compared to the current MMDE SotA methods, as \ourmodel ranks first on this benchmark at the time of submission, with a \emph{15.5\%} improvement in $\mathrm{SI}_{\log}$ over the second-best method.
Performance disparities are not solely attributed to dataset characteristics, as observed in the comparison with Metric3D and ZeroDepth.
Despite being trained on a smaller dataset, \ourmodel outperforms both of these methods.
In particular, \ourmodel improves in $\mathrm{\delta}_1$ over Metric3D and ZeroDepth by 5.8\% and 7.3\%, respectively, on NYU (\Cref{tab:results:nyu}) and by 1.1\% and 9.4\%, respectively, on KITTI (\Cref{tab:results:kitti}).
Moreover, ZoeDepth, which has a capacity similar to our ViT-based approach and is pre-trained on the diverse MiDaS dataset~\cite{ranftl2020midas}, shows limitations in general zero-shot scenarios in \Cref{tab:results:zeroshot}, exhibiting performance comparable to traditional MMDE methods especially on scale-invariant metrics.

\begin{table}[t]
    \centering
    \small
    \caption{\textbf{Comparison with equivalent training setup.} All methods have the same backbone, ConvNext-L~\cite{liu2022convnext} and are tested in a zero-shot regime on KITTI Eigen-split and NYU. iDisc and \ourmodel are retrained on a strict subset of Metric3D for 500k iterations as in~\cite{yin2023metric3d}.}
    \vspace{-10pt}
    \resizebox{\columnwidth}{!}{
    \begin{tabular}{l|ccc|ccc}
        \toprule
        \multirow{2}{*}{\textbf{Method}} & \multicolumn{3}{c|}{KITTI} & \multicolumn{3}{c}{NYU}\\
         & $\mathrm{\delta}_{1}\uparrow$ & $\mathrm{SI}_{\log}\downarrow$ &  $\mathrm{F}_A\uparrow$ & $\mathrm{\delta}_{1}\uparrow$ & $\mathrm{SI}_{\log}\downarrow$ & $\mathrm{F}_A\uparrow$\\
        \toprule
        iDisc~\cite{piccinelli2023idisc} & $93.4$ & $8.36$ & $78.0$ & $92.1$ & $\underline{8.82}$ & $75.0$\\
        Metric3D~\cite{yin2023metric3d} & $\underline{97.5}$ & $\underline{7.28}$ & $\underline{82.9}$ & $\underline{92.6}$ & $9.13$ & $\underline{77.8}$\\
        \midrule
        \ourmodel & $\mathbf{97.9}$ & $\mathbf{6.66}$ & $\mathbf{83.8}$ & $\mathbf{97.1}$ & $\mathbf{6.69}$ & $\mathbf{84.3}$\\
        \bottomrule
    \end{tabular}}
    \label{tab:results:idisc}
    \vspace{-15pt}
\end{table}

For the sake of fair comparison, we provide in \Cref{tab:results:idisc} a comparison between Metric3D, iDisc, and \ourmodel where the latter two are retrained on a \emph{strict} subset of Metric3D's data, namely accounting for one-quarter of the original Metric3D dataset, with same framework detailed in~\cref{ssec:experiments:setup}. 
The results are two-fold: they demonstrate how \ourmodel still surpasses Metric3D with a subsplit of the training set, and how MMDE SotA methods designed for single-domain can not fully exploit the training diversity. Qualitative results in \cref{fig:results:main_vis} emphasize how the method excels in capturing the overall scale and scene complexity in a zero-shot setup.


\subsection{Ablation Study}
\label{ssec:experiments:ablations}

\begin{table*}[t]
    \footnotesize
    \centering
    \caption{\textbf{Ablations of \ourmodel.} \textit{In-Domain} corresponds to the union of the training domain's validation sets, while \textit{Out-of-Domain} involves the union of zero-shot testing sets. \textit{Oracle} is the model with provided GT cameras at training and test time. \textit{Baseline} directly predicts 3D points in Cartesian space, \textit{Baseline++} in pseudo-spherical. \textit{Full} represents the final \ourmodel. All models have the same depth and camera module architecture, if any. $\mathrm{ARel}_C$ is the mean of elementwise absolute relative error for camera intrinsics. (\dag): GT camera intrinsics utilized for backprojection. The backbone used is ConvNext-L~\cite{liu2022convnext}. Medians and median average deviations over three runs are reported.}
    \vspace{-10pt}
    \resizebox{\linewidth}{!}{
    \begin{tabular}{cl|cccc|cccc}
        \toprule
         & \multirow{2}{*}{\textbf{Ablation}} & \multicolumn{4}{c|}{In-Domain} & \multicolumn{4}{c}{Out-of-Domain}\\
         & & $\mathrm{\delta}_{1}\uparrow$ & $\mathrm{SI}_{\log}\downarrow$ & $\mathrm{F}_A\uparrow$ & $\mathrm{ARel}_{C}\downarrow$ & $\mathrm{\delta}_{1}\uparrow$ & $\mathrm{SI}_{\log}\downarrow$ & $\mathrm{F}_A\uparrow$ & $\mathrm{ARel}_{C}\downarrow$\\
        \toprule
    1 & Oracle\textsuperscript{\dag} & $89.06 \scriptstyle \pm 0.03$ & $13.15 \scriptstyle \pm 0.02$ & \textcolor{gray}{$65.45 \scriptstyle \pm 0.13$} & n/a & $68.11 \scriptstyle \pm 0.17$ & $14.78 \scriptstyle \pm 0.01$ & $\textcolor{gray}{57.17 \scriptstyle \pm 0.09}$ & n/a\\
    \midrule
    2 & Full & $88.89 \scriptstyle \pm 0.10$ & $13.13 \scriptstyle \pm 0.01$ & $63.52 \scriptstyle \pm 0.08$ & $2.05 \scriptstyle \pm 0.01$ & $57.06 \scriptstyle \pm 1.48$ & $14.83 \scriptstyle \pm 0.04$ & $49.71 \scriptstyle \pm 0.55$ & $13.54 \scriptstyle \pm 0.85$\\
    3 & -- Camera\textsuperscript{\dag} & $87.42 \scriptstyle \pm 0.04$ & $13.49 \scriptstyle \pm 0.08$ & \textcolor{gray}{$63.78 \scriptstyle \pm 0.02$} & n/a & $48.38 \scriptstyle \pm 0.97$ & $15.55 \scriptstyle \pm 0.15$ & \textcolor{gray}{$45.21 \scriptstyle \pm 0.86$} & n/a\\
    4 & -- Spherical & $61.30 \scriptstyle \pm 1.00$ & $19.36 \scriptstyle \pm 0.09$ & $17.89 \scriptstyle \pm 0.11$ & $48.29 \scriptstyle \pm 4.03$ & $37.09 \scriptstyle \pm 1.37$ & $22.49 \scriptstyle \pm 0.16$ & $21.78 \scriptstyle \pm 0.14$ & $87.51 \scriptstyle \pm 11.1$\\
    5 & -- $\mathcal{L}_{\mathrm{con}}$ & $88.53 \scriptstyle \pm 0.07$ & $13.24 \scriptstyle \pm 0.01$ & $60.89 \scriptstyle \pm 0.15$ & $2.65 \scriptstyle \pm 0.06$ & $52.89 \scriptstyle \pm 0.21$ & $14.85 \scriptstyle \pm 0.01$ & $45.17 \scriptstyle \pm 0.32$ & $14.27 \scriptstyle \pm 0.41$\\
    6 & -- Dense & $87.62 \scriptstyle \pm 0.11$ & $13.41 \scriptstyle \pm 0.05$ & $61.33 \scriptstyle \pm 0.54$ & $1.91 \scriptstyle \pm 0.04$ & $55.65 \scriptstyle \pm 0.18$ & $15.04 \scriptstyle \pm 0.04$ & $43.19 \scriptstyle \pm 0.24$ & $16.61 \scriptstyle \pm 0.41$\\
    7 & -- Detach & $88.16 \scriptstyle \pm 0.12$ & $13.48 \scriptstyle \pm 0.06$ & $64.19 \scriptstyle \pm 0.17$ & $0.93 \scriptstyle \pm 0.02$ & $46.60 \scriptstyle \pm 0.25$ & $15.26 \scriptstyle \pm 0.10$ & $43.85 \scriptstyle \pm 2.01$ & $18.99 \scriptstyle \pm 1.00$\\
    \midrule
    8 & Baseline & $77.36 \scriptstyle \pm 0.22$ & $21.17 \scriptstyle \pm 0.28$ & $16.29 \scriptstyle \pm 0.26$ & n/a & $48.19 \scriptstyle \pm 1.02$ & $23.05 \scriptstyle \pm 0.45$ & $14.29 \scriptstyle \pm 0.36$ & n/a\\
    9 & Baseline++ & $82.41 \scriptstyle \pm 0.13$ & $16.31 \scriptstyle \pm 0.05$ & $41.98 \scriptstyle \pm 0.12$ & n/a & $51.22 \scriptstyle \pm 0.35$ & $18.14 \scriptstyle \pm 0.05$ & $38.27 \scriptstyle \pm 0.02$ & n/a\\
    \bottomrule
    \end{tabular}}
    \label{tab:results:ablations}
    \vspace{-12pt}
\end{table*}

The importance of each component introduced in \ourmodel in \cref{sec:method} is evaluated by ablating the method in \Cref{tab:results:ablations}.
All ablations exploit the predicted camera representation, if not stated otherwise.
The first distinction involves the \textit{Oracle} model, which operates under ideal conditions with known camera information during training and testing, addressing a task similar to~\cite{guizilini2023zerodepth, yin2023metric3d}.
On the other hand, \textit{Baseline} is a straightforward encoder-decoder implementation with a ($x$,$y$,$z$) output, as outlined at the beginning of \cref{sec:method}, while \textit{Baseline++} exploits the proposed pseudo-spherical representation.
Modules' architectures are consistent across experiments.
The \textit{In-Domain} column reflects testing on validation splits of training domains, while \textit{Out-of-Domain} corresponds to zero-shot testing, as detailed in~\cref{ssec:experiments:setup}.
Notably, \textit{In-Domain} results exhibit a higher degree of homogeneity compared to \textit{Out-of-Domain}, which is noisier yet more informative for gauging expected performances in downstream applications and in-the-wild deployment.

\noindent{}\textbf{Architecture.}
The \textit{Oracle} model demonstrates more robust scale-dependent performance during zero-shot testing compared to the \textit{Full} model, highlighting how the proposed task is inherently more demanding.
The \textit{Baseline} model illustrates an approach to the problem without utilizing external information and lacking a proper design for both internal and output space.
This approach yields markedly inferior results for both \textit{In-Domain} and \textit{Out-of-Domain} scenarios in terms of depth and 3D reconstruction metrics.

\noindent{}\textbf{Camera Module.}
In \Cref{tab:results:ablations}, row 3, the benefit of the Camera Module becomes apparent, revealing a substantial disparity in the effect of this module on scale-invariant and scale-dependent metrics for in- and out-of-domain testing.
This disparity stems from the absence of prior knowledge of the model regarding scale, impeding its optimal utilization of the diverse training set.
Concentrating solely on predicting depth, rather than a complete 3D output, proves advantageous in averting convergence issues during training. 
This is evident in comparison with methods predicting 3D, either without reliance on camera information (rows 8 and 9) or influenced by intertwined optimization (row 4), as elucidated in~\cref{sec:method}.
Refraining from relying on the camera also constrains the model's capacity to recover a multimodal distribution for out-of-domain samples. 
The lack of a (bootstrapped) prior prevents the depth module from serving as a corrective mechanism based on an initial scale estimation and imposes an unnecessary computational burden, \ie recovering the depth values from scratch.
This limitation is underscored by the marked variability observed for test sets strongly out-of-distribution, such as KITTI, when comparing the utilization or absence of camera information (rows 2 and 3, respectively).
In particular, \textit{Full} achieves 95.2\% in $\mathrm{\delta}_{1}$ in KITTI, while ``\textit{-- Camera}'' obtains 58.9\% for the same test set, despite a mere 2\% difference between the two versions on nuScenes and DDAD.

\noindent{}\textbf{Optimization and Output Representation.}
All ablations employ the same loss $\mathcal{L}_{\lambda MSE}$, but across different output spaces.
In row 4, a Cartesian output space is used instead of a pseudo-spherical from~\cref{ssec:method:spherical}, which results in substantially inferior performance due to the respective intertwined formulation of camera and depth output spaces. 
The \textit{Baseline} (row 8) also employs a Cartesian representation, but the negative impact of this choice is less pronounced in this model because of the absence of a camera module.
More specifically, the decoder of \textit{Baseline} is not conditioned on inaccurate prior camera and scale information as in row 4.
Moreover, row 9 corresponds to \textit{Baseline} with pseudo-spherical representation.
Comparison between row 8 and row 9 shows that when predicting directly the 3D outputs, the choice of the output representation is still relevant in defining a better internal representation and optimization.
Row 5 demonstrates the positive impact of the geometric invariance loss.
This loss contributes to enhanced in-domain and out-of-domain performance by promoting the invariance of depth features to appearance variations owing to different camera intrinsics.
Furthermore, stopping the gradient from propagating from the Camera Module to the Encoder (row 7), as described in~\cref{ssec:method:design}, proves particularly beneficial in avoiding scale and camera overfitting in zero-shot testing, and stabilizes the training.
The more stable training is obtained by limiting the dominant effect that camera supervision has on the gradient of the Encoder weights compared to depth supervision.

\noindent{}\textbf{Camera Representation.}
In row 6, the model incorporates a sparse camera representation, specifically the pinhole camera model with ($f_x$, $f_y$, $c_x$, $c_y$), leading to sparse camera prompting and scalar supervision; the camera module still predicts the residual components as outlined in~\cref{ssec:method:camera_module}. 
This approach hurts generalization, as evidenced by $\mathrm{ARel}_C$ in the out-of-domain evaluation, despite the slight improvement in in-domain $\mathrm{ARel}_C$. 
We speculate that the four prompts convey less robust information to the depth module than their dense counterpart, resulting in inferior performance for depth metrics compared to \textit{Full} for both in- and out-of-domain.

\section{Conclusion}
\label{sec:conclusion}
In this work, we propose \ourmodel to predict metric 3D points in diverse scenes relying solely on a single input image.
Through meticulous ablation studies, we systematically address the challenges inherent in universal MMDE tasks, underscoring the pivotal contributions of our work.
The designed self-prompting camera allows camera-free test time application and renders the model more robust against camera noise.
The introduced pseudo-spherical output space representation adequately disentangles the camera and depth of the optimization process.
Furthermore, the proposed geometric invariance loss effectively ensures camera-aware depth consistency.
Extensive validations unequivocally exhibit how \ourmodel sets the new state of the art across multiple benchmarks in a zero-shot regime, even surpassing in-domain trained methods.
This attests to the robustness and efficacy of our model and, most importantly, outlines its potential to propel the field of MMDE to new frontiers.

\vfill
\noindent{}\textbf{Acknowledgment.} This work is funded by Toyota Motor Europe via the research project TRACE-Z\"urich.

\newpage

{
    \small
    \bibliographystyle{ieeenat_fullname}
    \bibliography{main}
}

\newpage
\def\thesection{\Alph{section}}
\setcounter{section}{0}
\setcounter{figure}{4}
\setcounter{table}{5}

\twocolumn[{%
 \centering
 \textbf{\Large  Supplementary Material }
 \vspace{20pt}
}]

This supplementary material offers further insights into our work.
In \cref{supp:results} we provide results on the official KITTI benchmark, and standard metric evaluation on KITTI and NYU validation set. 
Moreover, \cref{supp:ablations} includes additional ablations, namely with VIT backbone and a comparison of different pseudo-spherical representations. In addition, differences between convolutional and ViT-based backbones regarding generalization are discussed.
In \cref{supp:dataset}, we describe the datasets used for training and testing and how we propose to amend Diode~\cite{Vasiljevic2019diode} artifacts at boundaries present in ground-truth depth.
We analyze the complexity of \ourmodel and compare it with other methods in \cref{supp:complex}.
Furthermore, we describe in \cref{supp:arch} the network architecture in more detail, necessarily \cref{supp:arch} overlaps with Sec. 3.
Eventually, additional visualizations are provided in \cref{supp:vis}.

\section{Results}
\label{supp:results}

\noindent{}\textbf{KITTI benchmark~\cite{Geiger2012kitti}.} \Cref{tab:supp:kitt_bench} clearly shows the compelling performance of \ourmodel on the official KITTI private test set. Results of the latest published methods are reported. The table is fetched from the official KITTI leaderboard for depth prediction. In particular, \ourmodel ranks first in the KITTI benchmark at the time of submission among all methods, published and not.
\begin{table}[h]
    \centering
    \small
    \caption{\textbf{Results on official KITTI~\cite{Geiger2012kitti} Benchmark.} Comparison of performance of methods trained on KITTI and tested on the official KITTI private test set.}
    \vspace{-10pt}
    \resizebox{\linewidth}{!}{
    \begin{tabular}{l|cccc}
    \toprule
    \multirow{2}{*}{\textbf{Method}} & $\mathrm{SI_{\log}}$ & $\mathrm{Sq.Rel}$ & $\mathrm{A.Rel}$ & $\mathrm{iRMS}$\\
    & \multicolumn{4}{c}{\textit{Lower is better}}\\
    \toprule
    MG~\cite{liu2023mutligaussian} & 9.93 & 1.68 \% & 7.99 \% & 10.63\\
    URCDC-Depth~\cite{shao2023urcdc} & 10.03 & 1.74 \% & 8.24 \% & 10.71\\
    iDisc~\cite{piccinelli2023idisc} & 9.89 & 1.77 \% & 8.11 \% & 10.73\\
    VA-DepthNet~\cite{liu2023vardepth} & 9.84 & 1.66 \% & 7.96 \% & 10.44\\
    IEBins~\cite{shao2023iebins} & 9.63 & 1.60 \% & 7.82 \% & 10.68\\
    NDDepth~\cite{shao2023nddepth} & 9.62 & 1.59 \% & 7.75 \% & 10.62 \\
    \midrule
    \ourmodel & \textbf{8.13} & \textbf{1.09\%} & \textbf{6.54 \%} & \textbf{8.24}\\
    \bottomrule
    \end{tabular}}
    \label{tab:supp:kitt_bench}
\end{table}

\noindent{}\textbf{KITTI Eigen-split and NYUv2-Depth.} For the sake of completeness, we report the ``standard'' metrics results in \Cref{tab:supp:kitti} and \Cref{tab:supp:nyu} on KITTI Eigen-split and NYU validation set, respectively.
It is worth noting that the typical metrics $\delta_2$ and, especially, $\delta_3$ are saturated, thus not informative.
Therefore, we advocate our choice of not reporting them in the main paper and prefer to report $\delta_{0.5}$.
Moreover, we suggest in future works the use of the area under the curve of the $\delta$ metrics as a more informative and comprehensive metric, instead of the values at fixed thresholds, \ie $\{1.25^i\}^3_{i=1}$.

\begin{table}[]
    \centering
    \caption{\textbf{Comparison on KITTI Eigen-split test set.} The first five methods are trained on KITTI and tested on it. The last six methods are tested in a zero-shot setting. \ourmodel-\{C, V\}: \ourmodel-\{ConvNext~\cite{liu2022convnext}, ViT~\cite{Dosovitskiy2020VIT}\}. (\dag): MiDaS~\cite{ranftl2020midas} pre-trained. (\ddag): predicted intrinsics are utilized for conditioning and backprojecting.}
    \vspace{-10pt}
    \resizebox{\columnwidth}{!}{
    \begin{tabular}{l|cccc|ccccc}
        \toprule
        \multirow{2}{*}{\textbf{Method}} & $\mathrm{\delta}_{1}$ & $\mathrm{\delta}_{2}$ & $\mathrm{\delta}_{3}$ & $\mathrm{F}_A$ & $\mathrm{A.Rel}$ & $\mathrm{RMS}$ & $\mathrm{RMS}_{\log}$ & $\mathrm{CD}$ & $\mathrm{SI}_{\log}$\\
         & \multicolumn{4}{c|}{\textit{Higher is better}} & \multicolumn{5}{c}{\textit{Lower is better}}\\
        \toprule
        BTS~\cite{Lee2019bts} & $96.2$ & $99.4$ & $\underline{99.8}$ & $82.0$ & $5.63$ & $2.43$ & $0.089$ & $0.42$ & $8.18$\\
        AdaBins~\cite{Bhat2020adabins} & $96.3$ & $99.5$ & $\underline{99.8}$ & $81.5$ & $5.85$ & $2.38$ & $0.089$ & $0.429$ & $8.10$\\
        NeWCRF~\cite{Yuan2022newcrf} & $97.5$ & $\underline{99.7}$ & $\mathbf{99.9}$ & $82.7$ & $5.20$ & $2.07$ & $0.078$ & $0.388$ & $7.00$\\
        iDisc~\cite{piccinelli2023idisc} & $97.5$ & $\underline{99.7}$ & $\mathbf{99.9}$ & $83.1$ & $5.09$ & $2.07$ & $0.077$ & $0.380$ & $7.11$\\
        ZoeDepth~\cite{bhat2023zoedepth} & $96.5$ & $99.1$ & $99.4$ & $82.1$ & $5.76$ & $2.39$ & $0.089$ & $0.431$ & $7.47$\\
        Metric3D~\cite{yin2023metric3d} & $97.5$ & $99.5$ & $\underline{99.8}$ & $82.9$ & $5.33$ & $2.26$ & $0.081$ & $0.392$ & $7.28$\\
        \midrule
        Ours-C & $\underline{97.8}$ & $\underline{99.7}$ & $\mathbf{99.9}$ & $\underline{83.9}$ & $\underline{4.69}$ & $\underline{2.00}$ & $\underline{0.073}$ & $\underline{0.371}$ & $\underline{6.72}$\\
        Ours-V & $\mathbf{98.6}$ & $\mathbf{99.8}$ & $\mathbf{99.9}$ & $\mathbf{85.0}$ & $\mathbf{4.21}$ & $\mathbf{1.75}$ & $\mathbf{0.064}$ & $\mathbf{0.338}$ & $\mathbf{5.84}$\\
        \midrule
        \midrule
        Ours-C~\textsuperscript{\ddag} & $\mathbf{97.8}$ & $\underline{99.7}$ & $\mathbf{99.9}$ & $80.8$ & $4.77$ & $2.00$ & $0.073$ & $0.427$ & $6.72$\\
        Ours-V~\textsuperscript{\ddag} & $\mathbf{98.6}$ & $\mathbf{99.8}$ & $\mathbf{99.9}$ & $82.7$ & $\mathbf{4.21}$ & $\mathbf{1.75}$ & $\mathbf{0.064}$ & $0.381$ & $\mathbf{5.84}$\\
        \bottomrule
    \end{tabular}}
    \label{tab:supp:kitti}
    \vspace{-5pt}
\end{table}

\begin{table}[]
    \centering
    \caption{\textbf{Comparison on NYU validation set.} The first five methods are trained on NYU and tested on it. The last six methods are tested in a zero-shot setting. \ourmodel-\{C, V\}: \ourmodel-\{ConvNext~\cite{liu2022convnext}, ViT~\cite{Dosovitskiy2020VIT}\}. (\dag): MiDaS~\cite{ranftl2020midas} pre-trained. (\ddag): predicted intrinsics are utilized for conditioning and backprojecting.}
    \vspace{-10pt}
    \resizebox{\columnwidth}{!}{
    \begin{tabular}{l|cccc|ccccc}
        \toprule
        \multirow{2}{*}{\textbf{Method}} & $\mathrm{\delta}_{1}$ & $\mathrm{\delta}_{2}$ & $\mathrm{\delta}_{3}$ & $\mathrm{F}_A$ & $\mathrm{A.Rel}$ & $\mathrm{RMS}$ & $\mathrm{Log}_{10}$ & $\mathrm{CD}$ & $\mathrm{SI}_{\log}$\\
         & \multicolumn{4}{c|}{\textit{Higher is better}} & \multicolumn{5}{c}{\textit{Lower is better}}\\
        \toprule
        BTS~\cite{Lee2019bts} & $88.5$ & $\mathbf{97.8}$ & $99.4$ & $74.0$ & $10.9$ & $0.391$ & $0.046$ & $0.160$ & $11.5$\\
        AdaBins~\cite{Bhat2020adabins} & $90.1$ & $98.3$ & $99.6$ & $74.7$ & $10.3$ & $0.365$ & $0.044$ & $0.156$ & $10.6$\\
        NeWCRF~\cite{Yuan2022newcrf} & $92.1$ & $99.1$ & $\underline{99.8}$ & $75.8$ & $9.56$ & $0.333$ & $0.040$ & $0.147$ & $9.16$\\
        iDisc~\cite{piccinelli2023idisc} & $93.8$ & $99.2$ & $\underline{99.8}$ & $78.2$ & $8.61$ & $0.313$ & $0.037$ & $0.133$ & $8.85$\\
        ZoeDepth~\cite{bhat2023zoedepth} & $95.2$ & $99.5$ & $\underline{99.8}$ & $80.1$ & $7.70$ & $0.278$ & $0.033$ & $0.125$ & $7.19$\\
        Metric3D~\cite{yin2023metric3d} & $92.6$ & $97.9$ & $99.1$ & $77.8$ & $9.38$ & $0.337$ & $0.038$ & $0.146$ & $9.13$\\
        \midrule
        Ours-C & $97.2$ & $\underline{99.6}$ & $\mathbf{99.9}$ & $84.4$ & $6.22$ & $0.231$ & $0.026$ & $0.101$ & $6.39$\\
        Ours-V & $\mathbf{98.4}$ & $\underline{99.7}$ & $\mathbf{99.9}$ & $\mathbf{85.9}$ & $\mathbf{5.78}$ & $\mathbf{0.201}$ & $\mathbf{0.024}$ & $\mathbf{0.092}$ & $\mathbf{5.27}$\\
        \midrule
        \midrule
        Ours-C\textsuperscript{\ddag} & $97.2$ & $\underline{99.6}$ & $\mathbf{99.9}$ & $84.1$ & $6.33$ & $0.232$ & $0.027$ & $0.103$ & $6.40$\\
        Ours-V\textsuperscript{\ddag} & $\underline{98.3}$ & $\underline{99.7}$ & $\mathbf{99.9}$ & $\underline{85.5}$ & $\underline{6.04}$ & $\underline{0.205}$ & $\underline{0.025}$ & $\underline{0.094}$ & $\underline{5.28}$\\
        \bottomrule
    \end{tabular}}
    \label{tab:supp:nyu}
    \vspace{-5pt}
\end{table}

\section{Ablations}
\label{supp:ablations}

\begin{table*}[]
    \footnotesize
    \centering
    \caption{\textbf{Ablations of \ourmodel.} \textit{In-Domain} corresponds to the union of the training domain's validation sets, while \textit{Out-of-Domain} involves the union of zero-shot testing sets. \textit{Oracle} is the model with provided GT cameras at training and test time. \textit{Baseline} directly predicts 3D points in Cartesian space, \textit{Baseline++} in pseudo-spherical. \textit{Full} represents the final \ourmodel. All models have the same depth and camera module architecture, if any. $\mathrm{ARel}_C$ is the mean of elementwise absolute relative error for camera intrinsics. (\dag): GT camera intrinsics utilized for backprojection. The backbone used is ViT-L~\cite{Dosovitskiy2020VIT}. Medians and median average deviations over three runs are reported.}
    \vspace{-10pt}
    \resizebox{\linewidth}{!}{
    \begin{tabular}{cl|cccc|cccc}
        \toprule
         & \multirow{2}{*}{\textbf{Ablation}} & \multicolumn{4}{c|}{In-Domain} & \multicolumn{4}{c}{Out-of-Domain}\\
         & & $\mathrm{\delta}_{1}\uparrow$ & $\mathrm{SI}_{\log}\downarrow$ & $\mathrm{F}_A\uparrow$ & $\mathrm{ARel}_{C}\downarrow$ & $\mathrm{\delta}_{1}\uparrow$ & $\mathrm{SI}_{\log}\downarrow$ & $\mathrm{F}_A\uparrow$ & $\mathrm{ARel}_{C}\downarrow$\\
        \toprule
    1 & Oracle & $91.46 \scriptstyle \pm 0.09$ & $12.12 \scriptstyle \pm 0.02$ & $68.35 \scriptstyle \pm 0.14$ & n/a & $72.17 \scriptstyle \pm 0.44$ & $13.07 \scriptstyle \pm 0.01$ & $59.84 \scriptstyle \pm 0.18$ & n/a\\
    2 & Full & $91.43 \scriptstyle \pm 0.05$ & $12.06 \scriptstyle \pm 0.06$ & $65.44 \scriptstyle \pm 0.84$ & $2.19 \scriptstyle \pm 0.14$ & $64.45 \scriptstyle \pm 0.52$ & $13.0 \scriptstyle \pm 0.02$ & $52.46 \scriptstyle \pm 0.29$ & $12.31 \scriptstyle \pm 0.61$\\
    3 & -- Camera & $89.33 \scriptstyle \pm 0.04$ & $12.54 \scriptstyle \pm 0.04$ & $66.02 \scriptstyle \pm 0.27$ & n/a & $60.67 \scriptstyle \pm 0.22$ & $13.4 \scriptstyle \pm 0.07$ & $52.43 \scriptstyle \pm 0.08$ & n/a\\
    4 & -- $\mathcal{L}_{\mathrm{con}}$ & $90.27 \scriptstyle \pm 0.13$ & $12.21 \scriptstyle \pm 0.01$ & $63.28 \scriptstyle \pm 0.66$ & $1.92 \scriptstyle \pm 0.31$ & $61.98 \scriptstyle \pm 0.41$ & $13.24 \scriptstyle \pm 0.04$ & $50.91 \scriptstyle \pm 0.16$ & $13.11 \scriptstyle \pm 0.36$\\
    5 & -- Spherical & $32.92 \scriptstyle \pm 0.18$ & $18.11 \scriptstyle \pm 0.08$ & $33.62 \scriptstyle \pm 0.07$ & $21.64 \scriptstyle \pm 0.2$ & $48.43 \scriptstyle \pm 1.27$ & $18.53 \scriptstyle \pm 0.35$ & $42.85 \scriptstyle \pm 1.18$ & $17.16 \scriptstyle \pm 0.79$\\
    6 & -- Dense & $90.16 \scriptstyle \pm 0.15$ & $12.23 \scriptstyle \pm 0.01$ & $64.19 \scriptstyle \pm 0.03$ & $1.83 \scriptstyle \pm 0.18$ & $62.44 \scriptstyle \pm 0.19$ & $13.36 \scriptstyle \pm 0.04$ & $49.34 \scriptstyle \pm 0.28$ & $13.68 \scriptstyle \pm 0.61$\\
    7 & -- Detach & $89.93 \scriptstyle \pm 0.02$ & $12.58 \scriptstyle \pm 0.04$ & $66.30 \scriptstyle \pm 0.35$ & $0.94 \scriptstyle \pm 0.03$ & $51.77 \scriptstyle \pm 0.09$ & $13.45 \scriptstyle \pm 0.02$ & $49.91 \scriptstyle \pm 0.01$ & $14.87 \scriptstyle \pm 0.22$\\
    \midrule
    8 & Baseline & $21.26 \scriptstyle \pm 0.23$ & $23.43 \scriptstyle \pm 0.45$ & $29.19 \scriptstyle \pm 0.09$ & n/a & $34.15 \scriptstyle \pm 0.74$ & $20.39 \scriptstyle \pm 0.42$ & $40.14 \scriptstyle \pm 0.52$ & n/a\\
    9 & Baseline++ & $88.84 \scriptstyle \pm 0.11$ & $12.93 \scriptstyle \pm 0.11$ & $42.72 \scriptstyle \pm 0.10$ & n/a & $59.31 \scriptstyle \pm 0.58$ & $14.04 \scriptstyle \pm 0.03$ & $44.12 \scriptstyle \pm 0.10$ & n/a\\
    \bottomrule
    \end{tabular}}
    \vspace{-5pt}
    \label{tab:results:ablations_vit}
\end{table*}

\subsection{Ablations with ViT backbone}
Ablations with ViT backbone are provided in \Cref{tab:results:ablations_vit}.
The trend in \Cref{tab:results:ablations_vit} is consistent with the one outlined for the convolutional backbone.
More specifically, the ablated components contribute similarly between ViT-L~\cite{Dosovitskiy2020VIT} and ConvNext-L~\cite{liu2022convnext} backbones.
However, utilizing a ViT backbone shows a larger variability for out-of-domain results, also showing a stronger effect of the usage of pseudo-spherical representation both for the \textit{Baseline} and \textit{Full}.
The increased susceptibility of the scene's depth scale to domain shift is also related to the backbone comparison in Table 1.
In particular, zero-shot results suggest that the convolutional architecture exhibits superior resilience to scale-related domain shifts, although showing relative disadvantage in handling appearance-related domain shifts.
$\mathrm{SI}_{\log}$ consistently favors ViT over convolutional methods, emphasizing the latter's diminished performance in appearance domain shifts.
However, scale-dependent metrics do not consistently favor ViT, indicating that the constrained receptive field of convolutional methods yields higher robustness to domain shifts associated with scale.

\subsection{Alternative pseudo-spherical representation}
\begin{table*}[]
    \footnotesize
    \centering
    \caption{\textbf{Ablations of specific pseudo-spherical representation.} \textit{In-Domain} corresponds to the union of the training domain's validation sets, while \textit{Out-of-Domain} involves the union of zero-shot testing sets. All models have the same depth and camera module architecture. $\mathrm{ARel}_C$ is the mean of elementwise absolute relative error for camera intrinsics. Medians and median average deviations over three runs are reported.}
    \vspace{-10pt}
    \resizebox{\linewidth}{!}{
    \begin{tabular}{l|c|cccc|cccc}
        \toprule
         \multirow{2}{*}{\textbf{Ablation}} & \multirow{2}{*}{\textbf{Backbone}} & \multicolumn{4}{c|}{In-Domain} & \multicolumn{4}{c}{Out-of-Domain}\\
         & & $\mathrm{\delta}_{1}\uparrow$ & $\mathrm{SI}_{\log}\downarrow$ & $\mathrm{F}_A\uparrow$ & $\mathrm{ARel}_{C}\downarrow$ & $\mathrm{\delta}_{1}\uparrow$ & $\mathrm{SI}_{\log}\downarrow$ & $\mathrm{F}_A\uparrow$ & $\mathrm{ARel}_{C}\downarrow$\\
        \toprule
    \ourmodel & ViT-L~\cite{Dosovitskiy2020VIT}  & $91.43 \scriptstyle \pm 0.05$ & $12.06 \scriptstyle \pm 0.06$ & $65.44 \scriptstyle \pm 0.84$ & $2.19 \scriptstyle \pm 0.14$ & $64.45 \scriptstyle \pm 0.52$ & $13.00 \scriptstyle \pm 0.02$ & $52.46 \scriptstyle \pm 0.29$ & $12.31 \scriptstyle \pm 0.61$\\
    \ourmodel\textsubscript{rays} & ViT-L~\cite{Dosovitskiy2020VIT} & $90.93 \scriptstyle \pm 0.02$ & $12.19 \scriptstyle \pm 0.06$ & $64.70 \scriptstyle \pm 0.05$ & $2.44 \scriptstyle \pm 0.11$ & $65.50 \scriptstyle \pm 0.81$ & $13.03 \scriptstyle \pm 0.01$ & $53.12 \scriptstyle \pm 0.02$ & $11.82 \scriptstyle \pm 0.99$\\
    \midrule
    \ourmodel & ConvNext-L~\cite{liu2022convnext} & $88.89 \scriptstyle \pm 0.10$ & $13.13 \scriptstyle \pm 0.01$ & $63.52 \scriptstyle \pm 0.08$ & $2.05 \scriptstyle \pm 0.01$ & $57.06 \scriptstyle \pm 1.48$ & $14.83 \scriptstyle \pm 0.04$ & $49.71 \scriptstyle \pm 0.55$ & $13.54 \scriptstyle \pm 0.85$\\
    \ourmodel\textsubscript{rays} & ConvNext-L~\cite{liu2022convnext} & $88.55 \scriptstyle \pm 0.31$ & $13.24 \scriptstyle \pm 0.10$ & $62.58 \scriptstyle \pm 1.11$ & $2.74 \scriptstyle \pm 0.13$ & $55.10 \scriptstyle \pm 0.39$ & $14.91 \scriptstyle \pm 0.01$ & $46.38 \scriptstyle \pm 0.61$ & $15.00 \scriptstyle \pm 0.36$\\
    \bottomrule
    \end{tabular}}
    \label{tab:supp:ablations_rays}
\end{table*}
Sec. 3 focuses on describing the pseudo-spherical representation chosen to disentangle the two sub-tasks, namely calibration and depth estimation, and ablations studies confirm the effectiveness of disentangling the sub-tasks.
In particular, \ourmodel exploits an angular pseudo-spherical representation, namely based on azimuth, elevation angle, and log-depth, \ie ($\theta$, $\phi$, $z_{\log}$).
Nevertheless, an alternative solution to disentangle the two different sub-tasks, namely calibration and depth estimation, is to exploit the bearing vector and log-depth.
More specifically, a bearing vector corresponds to the unit-length ray represented by ($r_x$, $r_y$, $r_z$) $\in \mathbb{S}^2$, with $\mathbb{S}^2$ corresponding to the unit-sphere manifold.
The bearing vectors are obtained as the unprojection of image coordinates based on the (pinhole) camera model.
With this design, the output is represented by the tuple ($r_x$, $r_y$, $r_z$, $z_{\log}$) and the loss $\mathcal{L}_{\mathrm{\lambda MSE}}$ is applied seamlessly as depicted in Sec. 3, but with $\lambda_{r_x}=\lambda_{r_y}=\lambda_{r_z}=1$ and $\lambda_z=0.15$.

However, the disentanglement in rays and log-depth can be viewed as an alternative pseudo-spherical representation, in fact, rays and angles share a direct relationship $\theta = \mathrm{arctan}(\frac{r_x}{r_z})$ and $\phi = \mathrm{arccos} (r_y)$.
\Cref{tab:supp:ablations_rays} explores the effectiveness of this alternative representation and compares to the one presented in Sec. 3.
The ablation study reported in \Cref{tab:supp:ablations_rays} highlights how the difference between the two representations is marginal and, in most cases, within the uncertainty range, thus proving their similarity.
The main difference lies in the output space dimensionality. 
In principle, the bearing vectors would span the entire $\mathbb{R}^3$ space.
However, the space is constrained to the unit-sphere manifold by $\mathrm{L}_2$ normalization.

Furthermore, we ablate our camera prompting with respect to CAMConvs~\cite{facil2019camconvs} in \Cref{tab:results:camconvs}
\begin{table}[]
    \footnotesize
    \centering
    \caption{\textbf{Ablate \ourmodel with CAMConvs.} \textit{Full} is complete \ourmodel, as row~2 in Tab.~5. \textit{w/ CAMConvs} represents \ourmodel with CAMConvs~\cite{facil2019camconvs} conditioning instead of our prompting.}
    \vspace{-10pt}
    \resizebox{\linewidth}{!}{
    \begin{tabular}{l|cccc|cccc}
    \toprule
    \multirow{2}{*}{\textbf{Ablation}} & \multicolumn{4}{c|}{In-Domain} & \multicolumn{4}{c}{Out-of-Domain}\\
    & $\mathrm{\delta}_{1}\uparrow$ & $\mathrm{SI}_{\log}\downarrow$ & $\mathrm{F}_A\uparrow$ & $\mathrm{ARel}_{C}\downarrow$ & $\mathrm{\delta}_{1}\uparrow$ & $\mathrm{SI}_{\log}\downarrow$ & $\mathrm{F}_A\uparrow$ & $\mathrm{ARel}_{C}\downarrow$\\
    \midrule
    w/ CAMConvs & $87.81$ & $13.49$ & $60.90$ & $2.55$ & $54.65$ & $15.37$ & $43.09$ & $16.11$\\
    Full & $\mathbf{88.89}$ & $\mathbf{13.13}$ & $\mathbf{63.52}$ & $\mathbf{2.05}$ & $\mathbf{57.06}$ & $\mathbf{14.83}$ & $\mathbf{49.71}$ & $\mathbf{13.54}$\\
    \bottomrule
    \end{tabular}}
    \label{tab:results:camconvs}
\end{table}

\begin{algorithm}[]
    \caption{GT depth boundaries refining.}
    \label{alg:diode}
    \begin{algorithmic}
    \Procedure{BoundaryRefine}{$\mathbf{Z}_{\log}$}
        \State $\mathbf{L} = \mathrm{Laplacian}(\mathbf{Z}_{log}, k=5)$
        \State $\mathbf{M} = \mathbb{I}[\mathbf{L}_{10\%} \leq \mathbf{L} \leq \mathbf{L}_{90\%}]$
        \Comment{Compute Laplacian and threshold at 10-90 percentile}
        \State $\mathbf{M} = (\mathbf{M} \ominus \mathrm{eye}_3) \oplus \mathrm{eye}_3$
        \Comment{Opening with size 3}
        \State $\mathbf{M} = \mathrm{MedianBlur}(\mathbf{M}, k=3)$
        \State $\mathbf{Z} = \exp(\mathbf{Z}_{\log}) \cdot \mathbf{M}$
        \State \textbf{return} $\mathbf{Z}$
    \EndProcedure
    \end{algorithmic}
\end{algorithm}
\begin{table}[t]
    \centering
    \small
    \caption{\textbf{Datasets List.} List of the training and testing datasets: number of images, scene type, and method of acquisition are reported. SfM: Structure-from-Motion. MVS: Multi-View Stereo.} 
    \vspace{-10pt}
    \resizebox{\columnwidth}{!}{
    \begin{tabular}{cl|ccc}
        \toprule
        & \textbf{Dataset} & \textbf{Images} & \textbf{Scene} & \textbf{Acquisition}\\
        \midrule
        \multirow{9}{*}{\rotatebox[origin=c]{90}{\textbf{Training Set}}} & A2D2~\cite{geyer2020a2d2} & 78k & Outdoor & LiDAR\\
        & Argoverse2~\cite{2021argoverse2} & 403k & Outdoor & LiDAR\\
        & BDD100k~\cite{yu2020bdd100k} & 270k & Outdoor & SfM\\
        & CityScapes~\cite{Cordts2016cityscapes} & 24k & Outdoor & MVS\\
        & DrivingStereo~\cite{yang2019drivingstereo} & 63k & Outdoor & MVS\\
        & Mapillary PSD~\cite{Lopez2020mapillary} & 742k & Outdoor & SfM\\
        & ScanNet~\cite{dai2017scannet} & 83k & Indoor & RGB-D\\
        & Taskonomy~\cite{zamir2018taskonomy} & 1940k & Indoor & RGB-D\\
        & Waymo~\cite{sun2020waymo} & 223k & Outdoor & LiDAR\\
        \midrule
        \multirow{9}{*}{\rotatebox[origin=c]{90}{\textbf{Testing Set}}} & DDAD~\cite{Guizilini2020ddad} & 1002 & Outdoor & LiDAR\\
        & Diode~\cite{Vasiljevic2019diode} & 325 & Indoor & LiDAR\\
        & ETH3D~\cite{schoeps2017eth3d} & 454 & Outdoor & RGB-D\\
        & HAMMER~\cite{jung2022hammer} & 496 & Indoor & Mix\\
        & IBims-1~\cite{koch2022ibims} & 100 & Indoor & RGB-D\\
        & KITTI~\cite{Geiger2012kitti} & 652 & Outdoor & LiDAR\\
        & NuScenes~\cite{nuscenes} & 3k & Outdoor & LiDAR\\
        & NYU~\cite{silberman2012nyu} & 654 & Indoor & RGB-D\\
        & SUN-RGBD~\cite{Song2015sunrgbd} & 4.4k & Indoor & RGB-D\\
        & VOID~\cite{wong2020void} & 800 & Indoor & RGB-D\\
        \bottomrule
    \end{tabular}}
    \vspace{-10pt}
    \label{tab:supp:datasets}
\end{table}

\vspace{-5pt}
\section{Datasets}
\label{supp:dataset}

\subsection{Datasets details}
Details of training and testing datasets are presented in 
\Cref{tab:supp:datasets}.
The training datasets are processed in a way that the interval between two consecutive RGB and GT depth frames is not smaller than one second.
We do not apply any post-processing apart from the aforementioned subsampling.
The total amount of training samples accounts for 3'743'000 samples.
SUN-RGBD~\cite{Song2015sunrgbd} validation set involves also NYU~\cite{silberman2012nyu} test set.
Therefore, we removed the samples corresponding to NYU test set to avoid any overlap between test sets.
As per standard practice, KITTI Eigen-split corresponds to the corrected and accumulated GT depth maps with 45 images with inaccurate GT discarded from the original 697 images.

\subsection{Diode Indoor ground-truth correction}
Diode~\cite{Vasiljevic2019diode} ground-truth depth is not perfectly accurate on boundaries, in particular, a simple inspection shows how depth in boundaries presents low values, but greater than zero.
These artifacts present in the GT affect the validation pipeline and results.
Therefore, we design a simple image processing algorithm, outlined in \Cref{alg:diode}, that, first, detects the aforementioned boundary artifacts and, second, masks the depth in the corresponding neighborhoods. 
Thanks to masking those boundaries, the corresponding regions are ignored during validation.

\section{Model Complexity}
\label{supp:complex}
\Cref{tab:supp:resource} displays the parameters and inference complexity of \ourmodel and other SotA methods.
\ourmodel with ViT-L backbone is comparable to ZoeDepth in terms of efficiency and model parameters; however \ourmodel surpasses it in terms of performance as stated in Sec. 4.
Metric3D displays an improved efficiency due to the fully convolutional and relatively low dimensionality designed in the decoder.
It is worth highlighting how ZeroDepth presents a low efficiency although based on ResNet-18, we argue that this is due to the expensive full-resolution cross-attention in the decoder.
The last two rows in \Cref{tab:supp:resource} analyze separately the complexity of the single Camera and Depth Module.
The Camera Module is a lightweight component accounting for 13.4M parameters.
On the other hand, the Depth Module amounts to more than half of the total latency, despite the limited memory consumption.
The Depth Module's high latency is due to the several (6) self-attention layers in the decoder.

\begin{table}[t]
    \centering
    \caption{\textbf{Parameters and efficiency comparison.} Comparison of performance of methods based on latency, throughput, and number of trainable parameters. Tested on RTX3090 GPU, 32-bit precision float, and input image with size (480,~640). The last two rows correspond to the Camera and Depth Moudel evaluated independently. R18: ResNet-18~\cite{He2015}, D161: DenseNet-161~\cite{Huang2016densenet}, EN-B5: EfficientNet-B5-AP~\cite{Xie2019}, CNXT-L: ConvNext-L~\cite{liu2022convnext}.}
    \vspace{-10pt}
    \resizebox{\linewidth}{!}{
    \begin{tabular}{ll|ccc}
    \toprule
    \textbf{Method} & Backbone & Latency (ms) & Throughput (FPS) & Parameters (M)\\
    \toprule
    BTS~\cite{Lee2019bts} & D161 & 28.5 & 35.1 & 47.0\\
    Adains~\cite{Bhat2020adabins} & EN-B5 & 33.2 & 30.1 & 78.3\\
    NewCRF~\cite{Yuan2022newcrf} & SWin-L~\cite{Liu2021swintransf} & 53.1 & 18.8 & 280.0\\
    iDisc~\cite{piccinelli2023idisc} & SWin-L~\cite{Liu2021swintransf} & 81.1 & 12.3 & 209.2\\
    ZoeDepth~\cite{bhat2023zoedepth} & BEiT-L & 144.8 & 6.91 & 345.9\\
    ZeroDepth~\cite{guizilini2023zerodepth} & R18 & 955.6 & 1.05 & 232.6\\
    Metric3D~\cite{yin2023metric3d} & CNXT-L & 40.3 & 24.8 & 203.2\\
    \ourmodel & CNXT-L & 86.6 & 11.5 & 238.9\\
    \ourmodel & ViT-L~\cite{Dosovitskiy2020VIT} & 146.4 & 6.83 & 347.0\\
    \midrule
    Camera Module & - & 5.1 & - & 13.4\\
    Depth Module & - & 49.2 & - & 26.6\\
    \bottomrule
    \end{tabular}}
    \vspace{-10pt}
    \label{tab:supp:resource}
\end{table}

\section{Network Architecture}
\label{supp:arch}
\noindent{}\textbf{Encoder.} We show the effectiveness of our method with different encoders, both convolutional and transformer-based ones, \eg, ConvNext~\cite{liu2022convnext} and ViT~\cite{Dosovitskiy2020VIT}.
However, all of them follow the same structure: the feature maps are extracted at each layer and the features map corresponding to a ``scale'' is obtained as the pixel-wise average. 
For ConvNext, we obtain the class tokens as the average pooled feature maps.
All backbones utilized are originally designed for classification, thus we remove the last 3 layers, \ie, the pooling layer, fully connected layer, and $\mathrm{softmax}$ layer.
The feature maps are flattened, then LayerNorm~\cite{Ba2016layer} (LN) and a linear layer are applied.
The linear layer projects the features to a common channel dimension of 512.
The projected feature maps are interpolated to a common shape, namely $(h,w)=(\frac{H}{16},\frac{W}{16})$, with $H$, $W$ as input height and width, respectively.
Two independent projections are utilized for the features maps, \ie $\mathbf{F} \in \mathbb{R}^{h \times w \times C \times B}$ with $B$ corresponding to the four scales, and $C$ set to 512 as mentioned above, and the class tokens $\in \mathbb{R}^{C \times B}$, the latter fed to the Camera Module only.

\noindent{}\textbf{Camera Module.} The camera parameters are initialized with the four class tokens extracted from the Encoder. 
The flattened and stacked feature maps from the encoder are detached and used as \textit{keys} and \textit{values} in one cross-attention layer, where the \textit{queries} correspond to the four camera parameters.
The output is processed by a MultiLayer Perceptron (MLP) with one hidden layer with dimension of 2048 and non-linear activation Gaussian Error Linear Unit (GELU)~\cite{Hendrycks2016gelu}.
The cross-attention and the MLP present a residual connection.
The four tokens are further processed with two additional self-attention layers, projected to dimension one and then exponentiated.
The camera parameters are obtained as $f_x = \frac{\Delta f_x W}{2}$, $f_y = \frac{\Delta f_y H}{2}$, $c_x = \frac{\Delta c_x W}{2}$, $c_y = \frac{\Delta c_y H}{2}$.
The dense camera representation $\mathbf{C}$ is obtained by backprojecting with the predicted camera parameters: $(\mathbf{r}_x, \mathbf{r}_y, \mathbf{r}_z) = \mathbf{K}^{-1} [\mathbf{u}, \mathbf{v}, \mathbf{1}]^T$ and calculating the azimuth and elevation angles, $\theta$ and $\phi$, as in \cref{supp:ablations}.
The angular representation is embedded through the Laplace Spherical Harmonics Embedding (SHE) leading to 81 channels, resulting in $\mathbf{E} \in \mathbb{R}^{h \times w \times 81}$.

\noindent{}\textbf{Depth Module.} The depth latents are initialized as the average of the features $\mathbf{F}$ along the $B$ dimension.  
Then, the latents are conditioned on the original feature tensor $\mathbf{F}$ via one cross-attention layer where two projections of $\mathbf{F}$ account for \textit{keys} and \textit{values} and $\mathbf{L}$ as \textit{queries}.
In addition, one MLP is applied, seamlessly as in the Camera Module.
Furthermore, the depth features are conditioned on the camera prompts $\mathbf{E}$ with one additional cross-attention layer, where \textit{keys} and \textit{values} are two projections of camera embeddings $\mathbf{E}$, and one MLP as above.
The features are decoded in three consecutive stages.
The first stage applies three self-attention layers with $\mathbf{E}$ as positional encoding.
The features are then processed with one ConvNext~\cite{liu2022convnext} layer, upsampled by a factor of two, and the channels are halved.
The second and third stages are similar, although the second stage presents two self-attention layers and the third only one.
In the second and third stages, MLP's hidden channel dimension is sequentially halved, too, from the initial aforementioned value of 2048.
Each stage's output is projected to a dimension one.
Therefore, the three output maps are interpolated to a common shape, \ie ($\frac{H}{2}$, $\frac{W}{2}$), and pixel-wise averaged.
The final log-depth output $\mathbf{Z}_{\log}$ is obtained by upsampling the obtained tensor to the input shape ($H$, $W$).
The final depth is element-wise exponentiation of $\mathbf{Z}_{\log}$.


\section{Visualization}
\label{supp:vis}
\begin{figure*}[ht]
    \renewcommand{\arraystretch}{2}
    \centering
    \small
    \begin{tabular}{cc|cccc|c}
        \multirow{2}{*}[6pt]{\rotatebox[origin=c]{90}{KITTI}}
        & \includegraphics[width=0.14\linewidth]{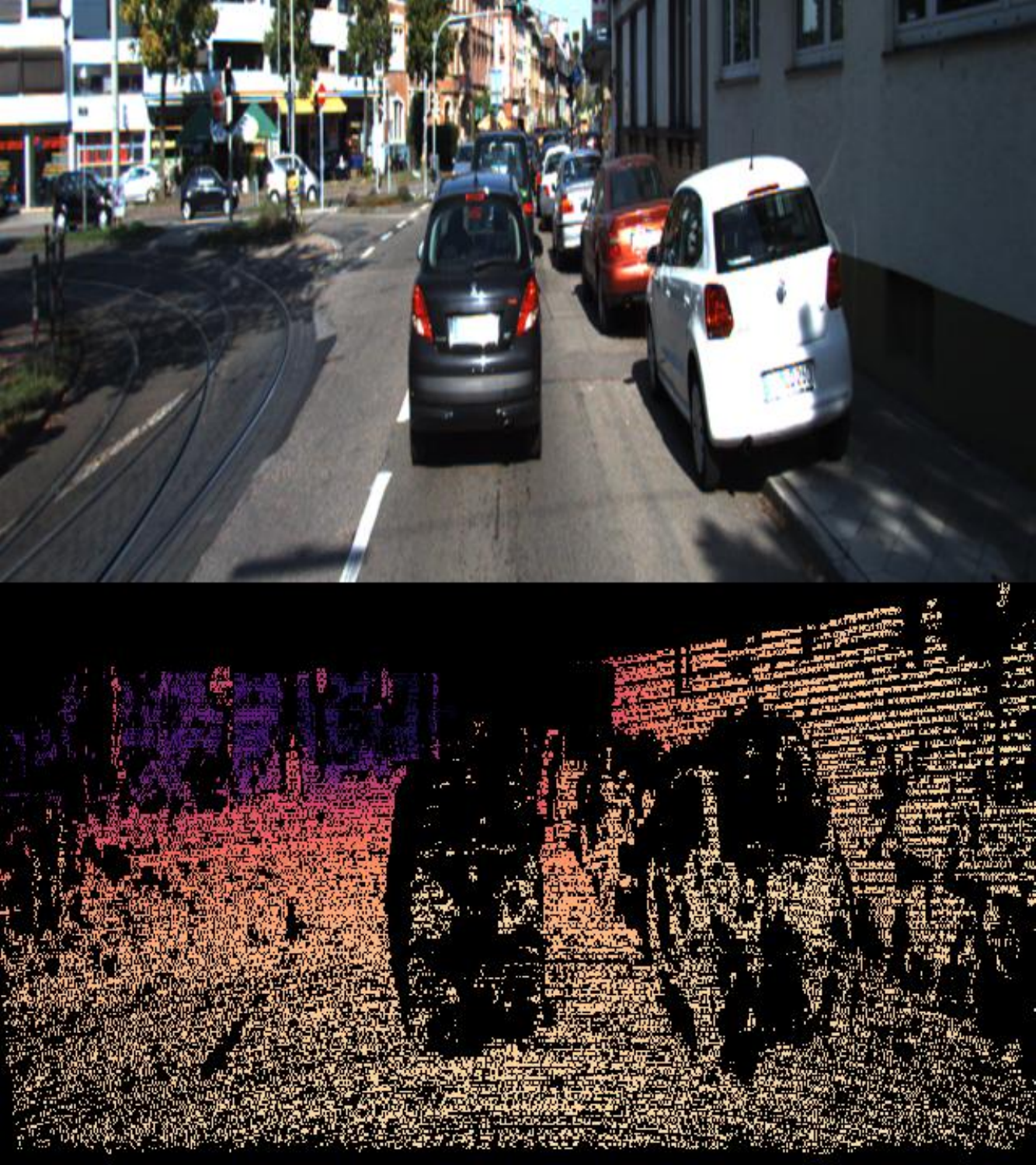}
        & \includegraphics[width=0.14\linewidth]{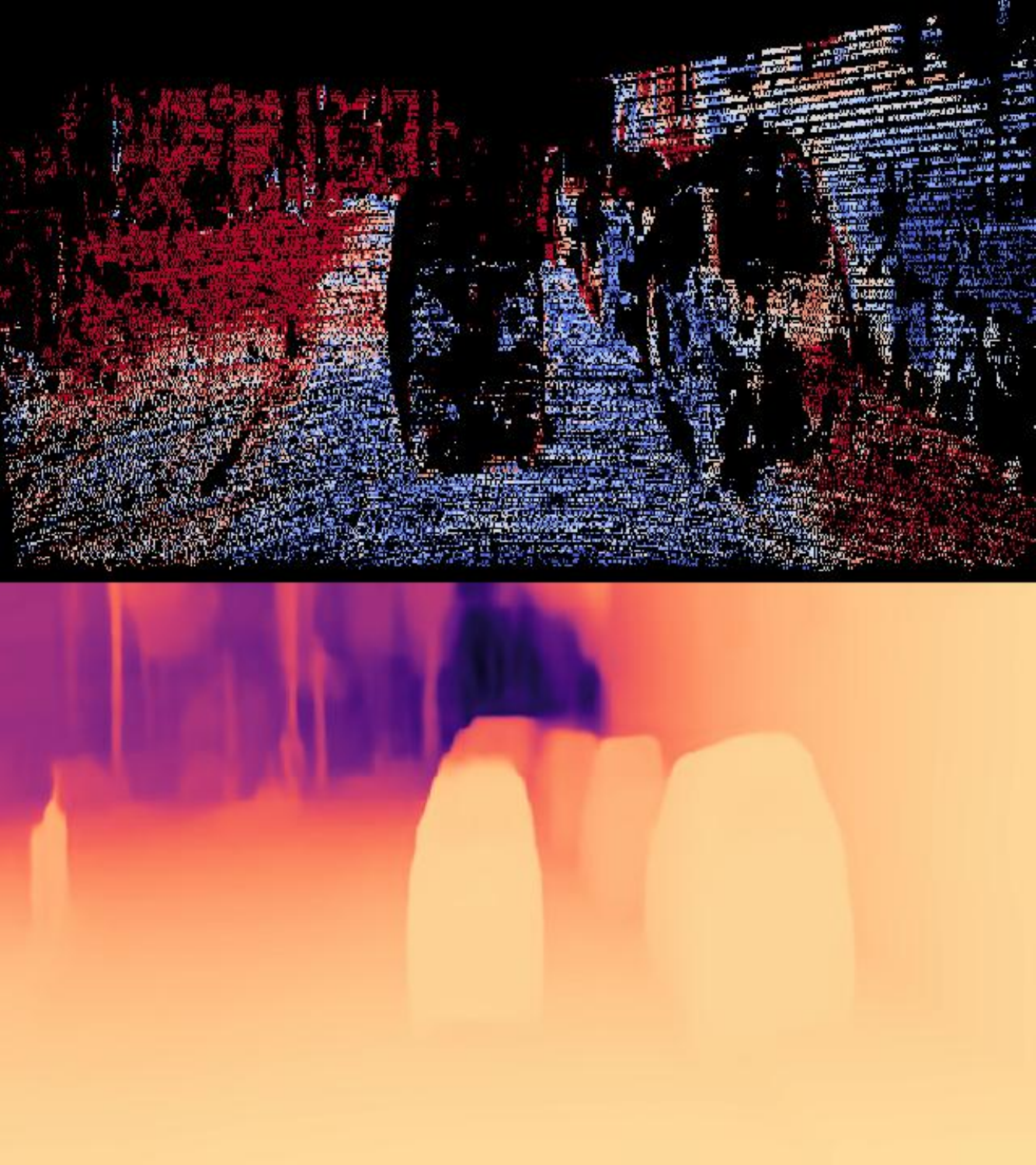}
        & \includegraphics[width=0.14\linewidth]{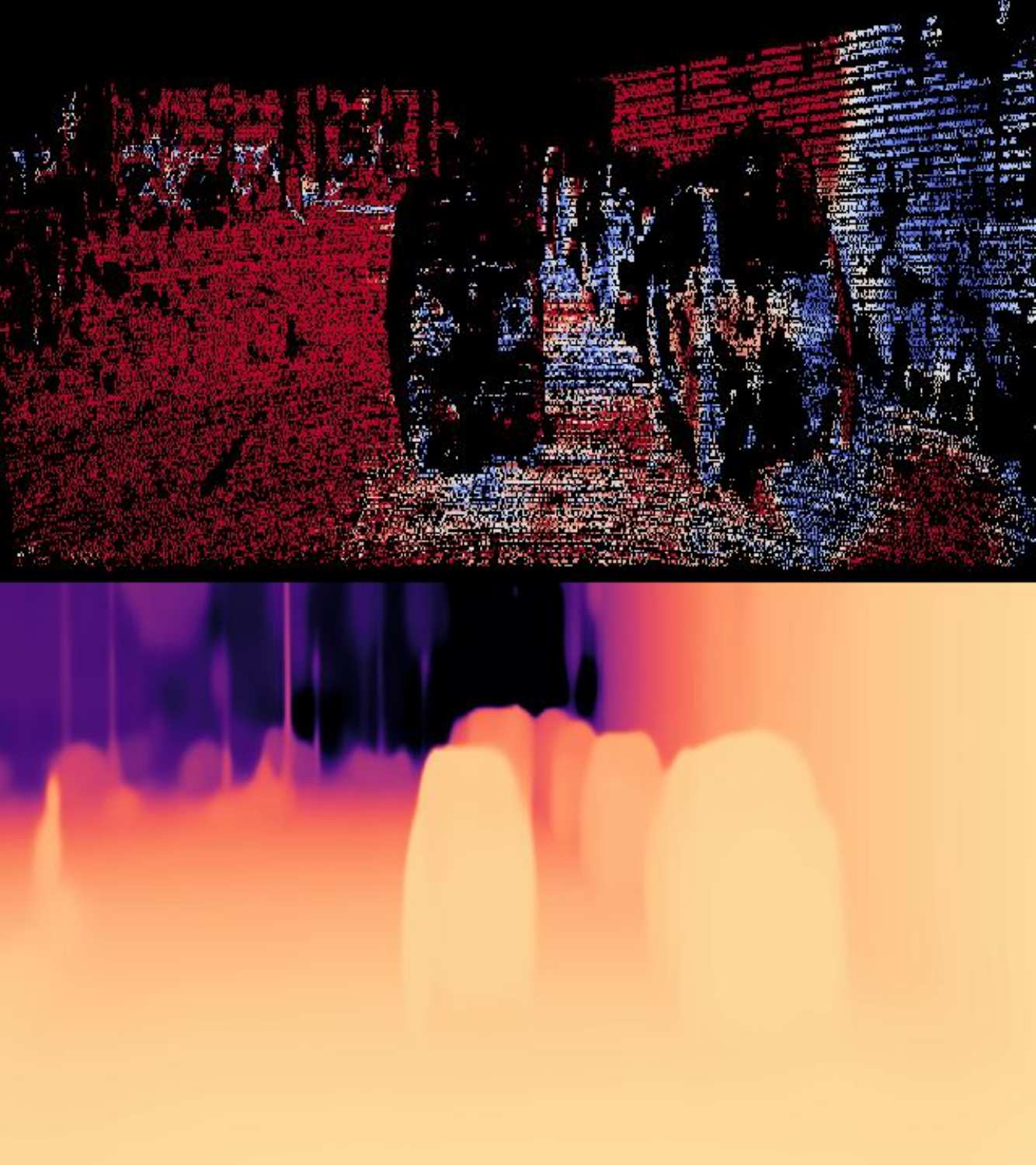}
        & \includegraphics[width=0.14\linewidth]{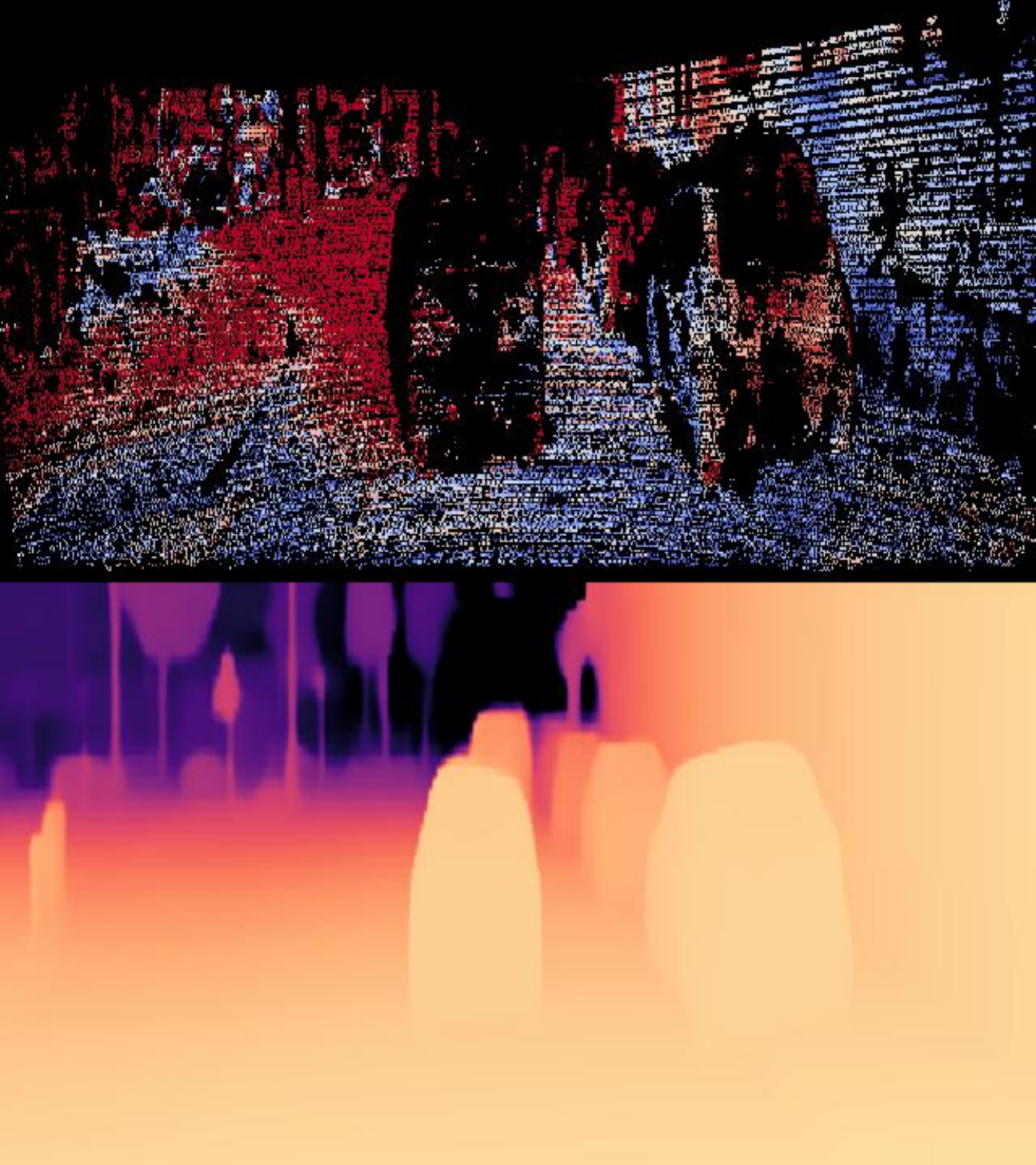}
        & \includegraphics[width=0.14\linewidth]{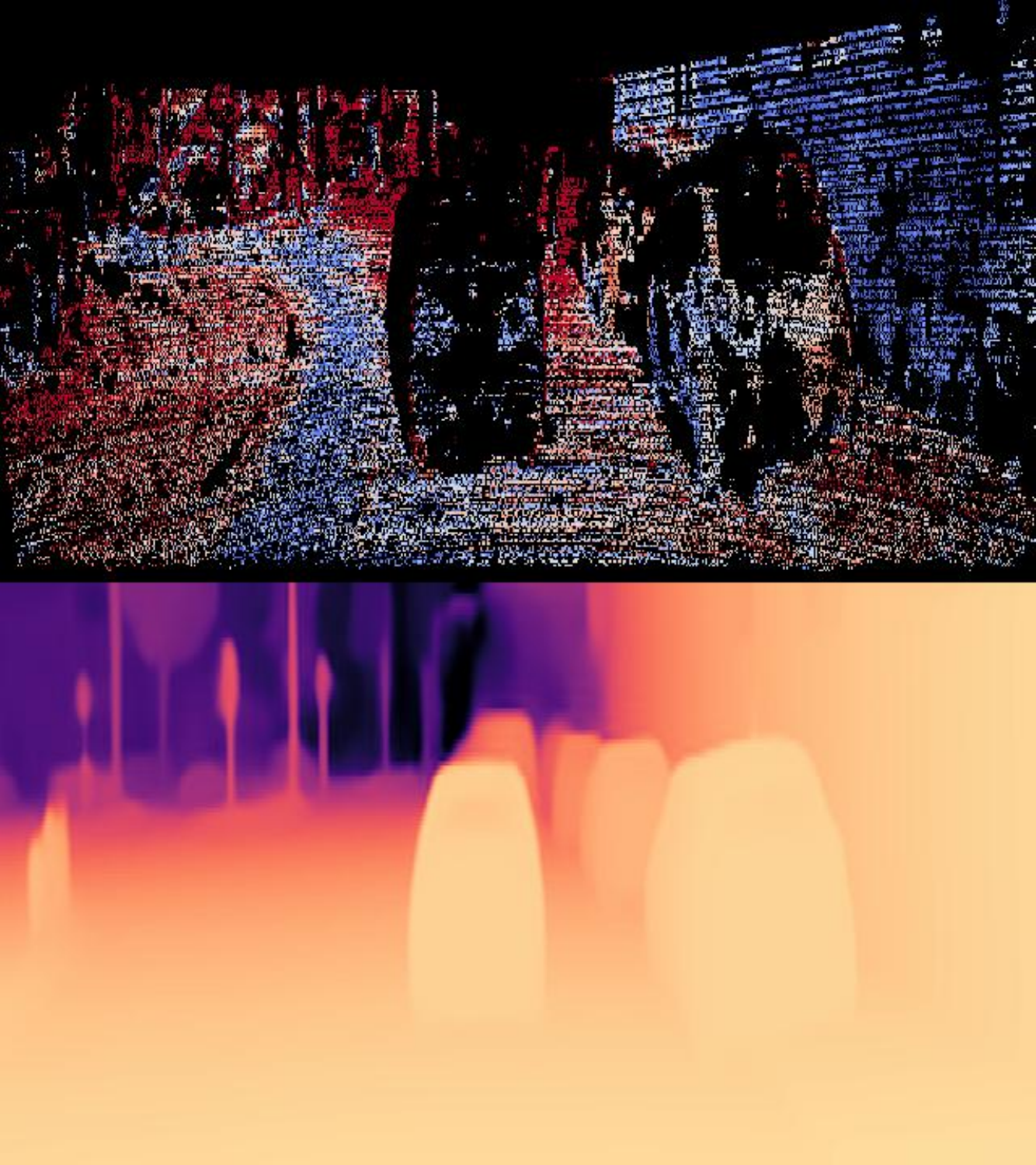}
        & \multirow{2}{*}[25pt]{\includegraphics[width=0.075\linewidth]{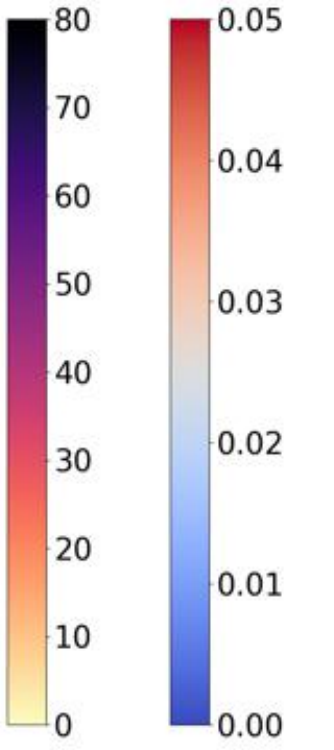}} \vspace{-8pt} \\
        & \includegraphics[width=0.14\linewidth]{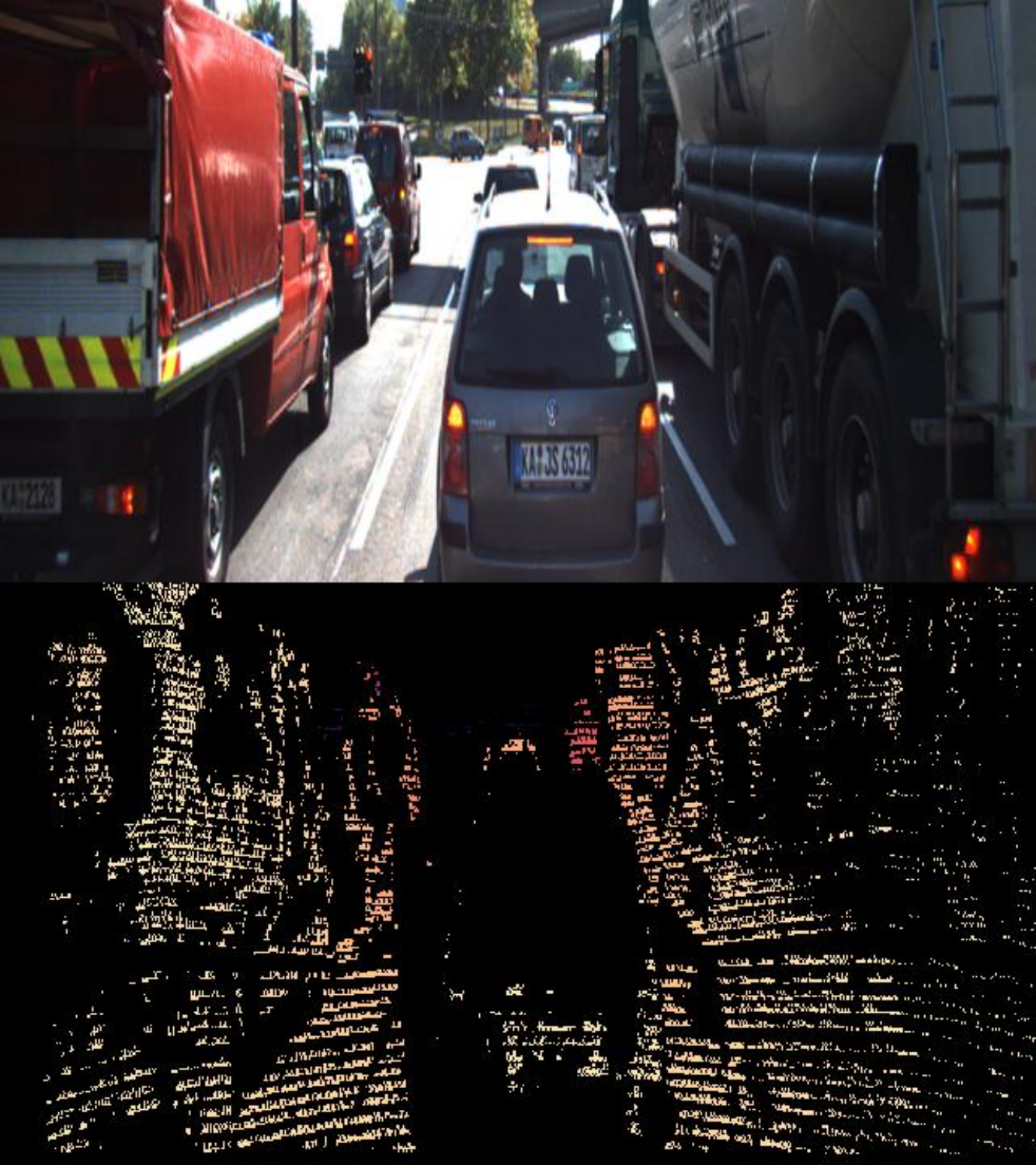}
        & \includegraphics[width=0.14\linewidth]{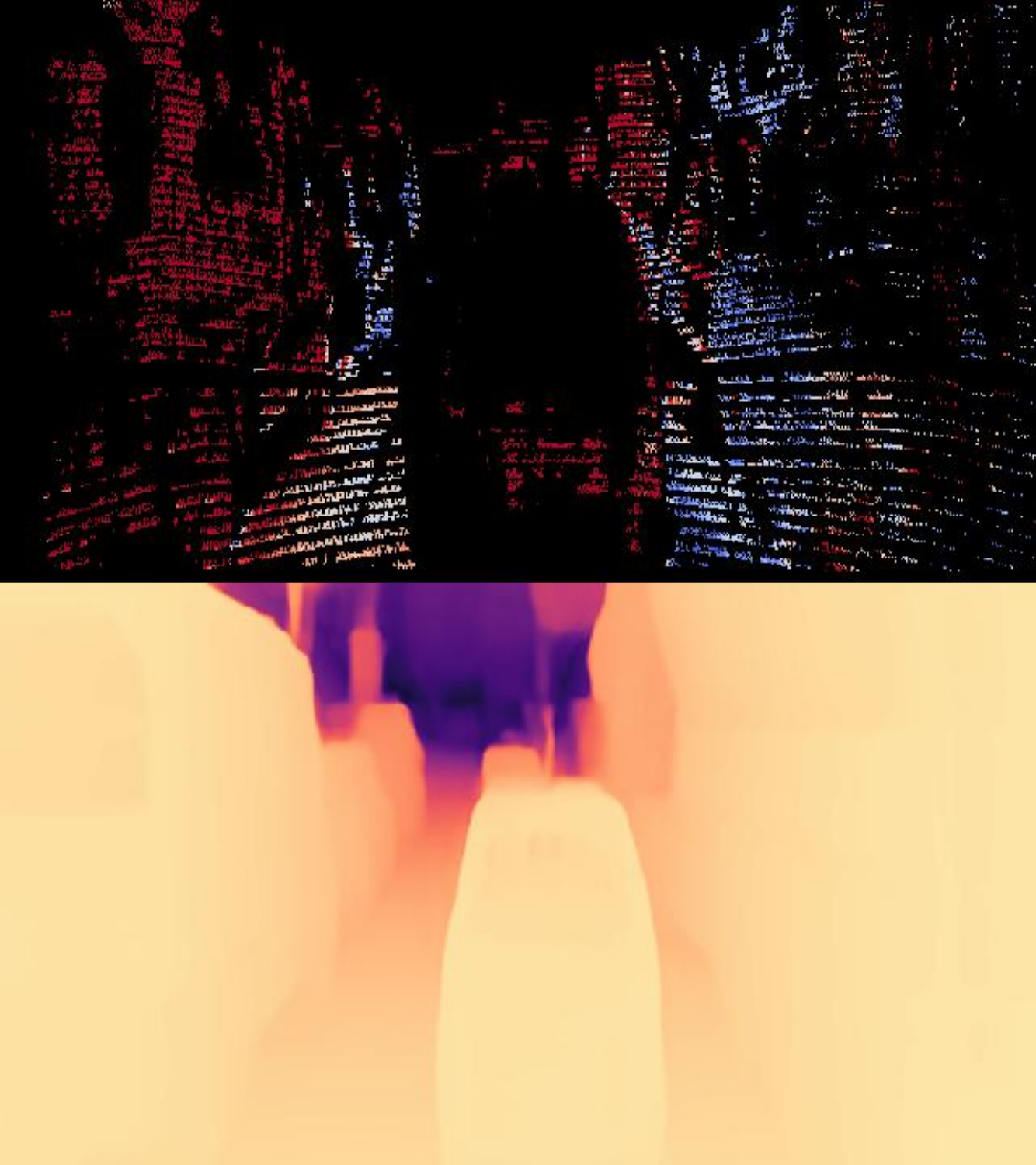}
        & \includegraphics[width=0.14\linewidth]{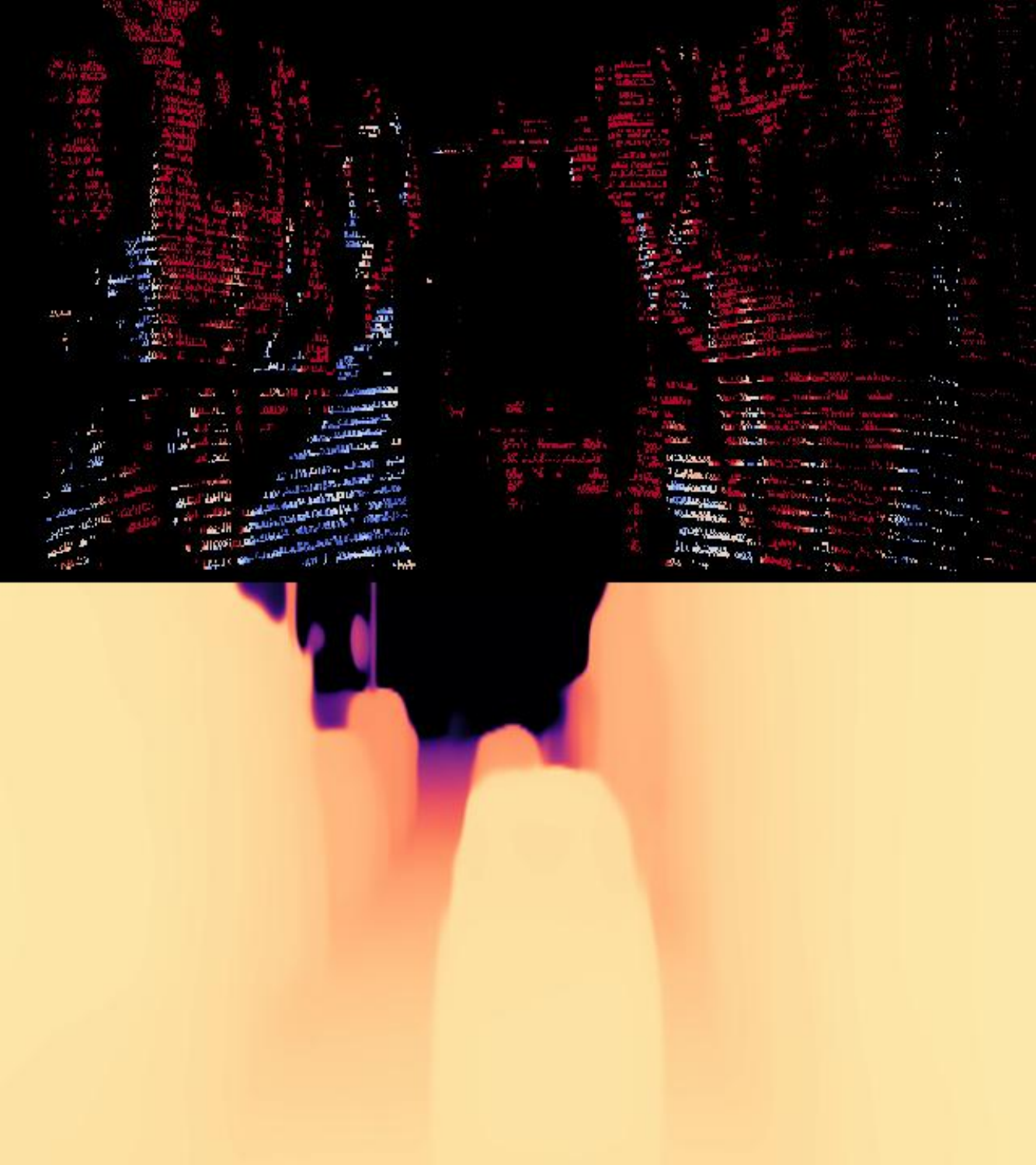}
        & \includegraphics[width=0.14\linewidth]{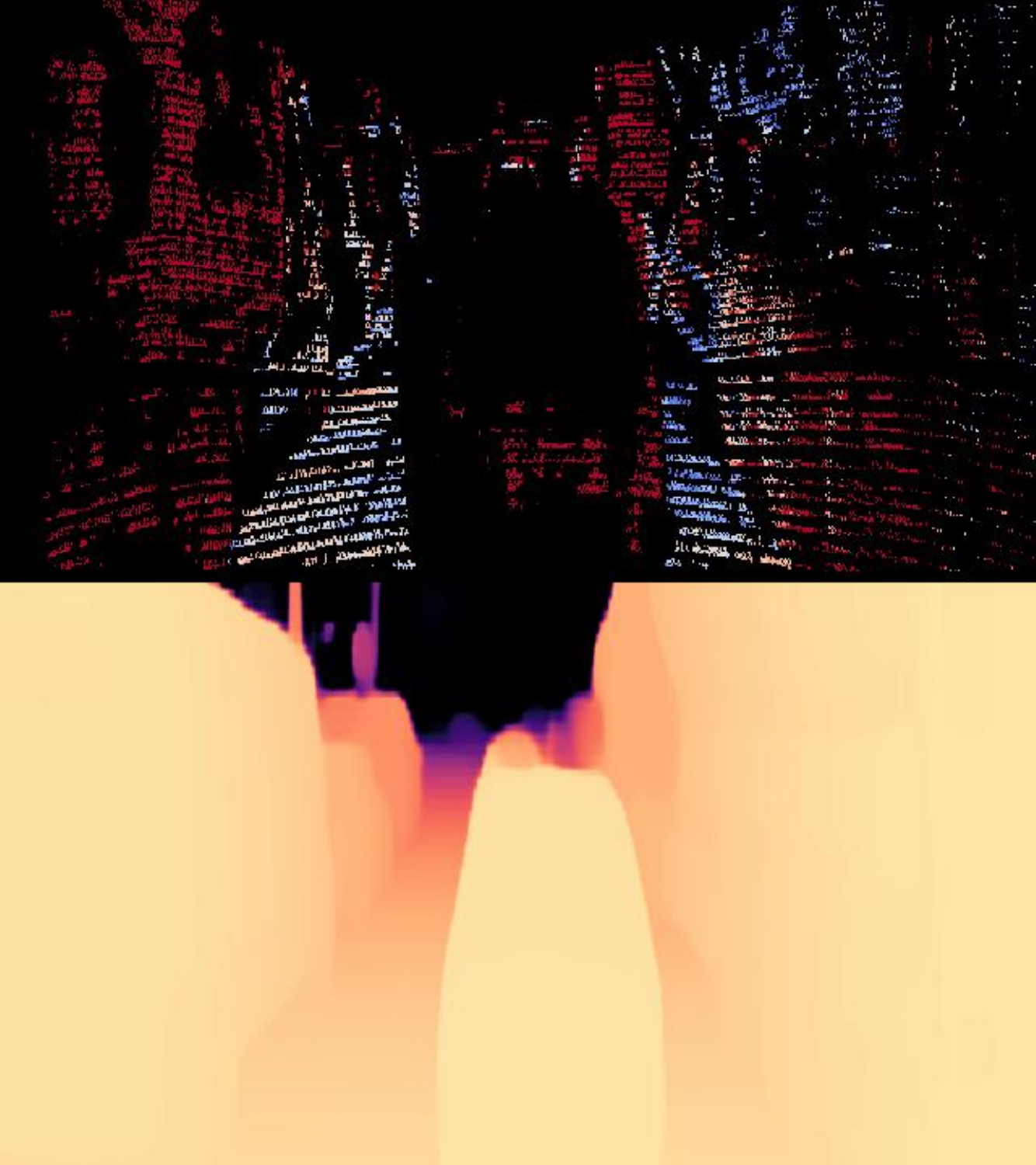}
        & \includegraphics[width=0.14\linewidth]{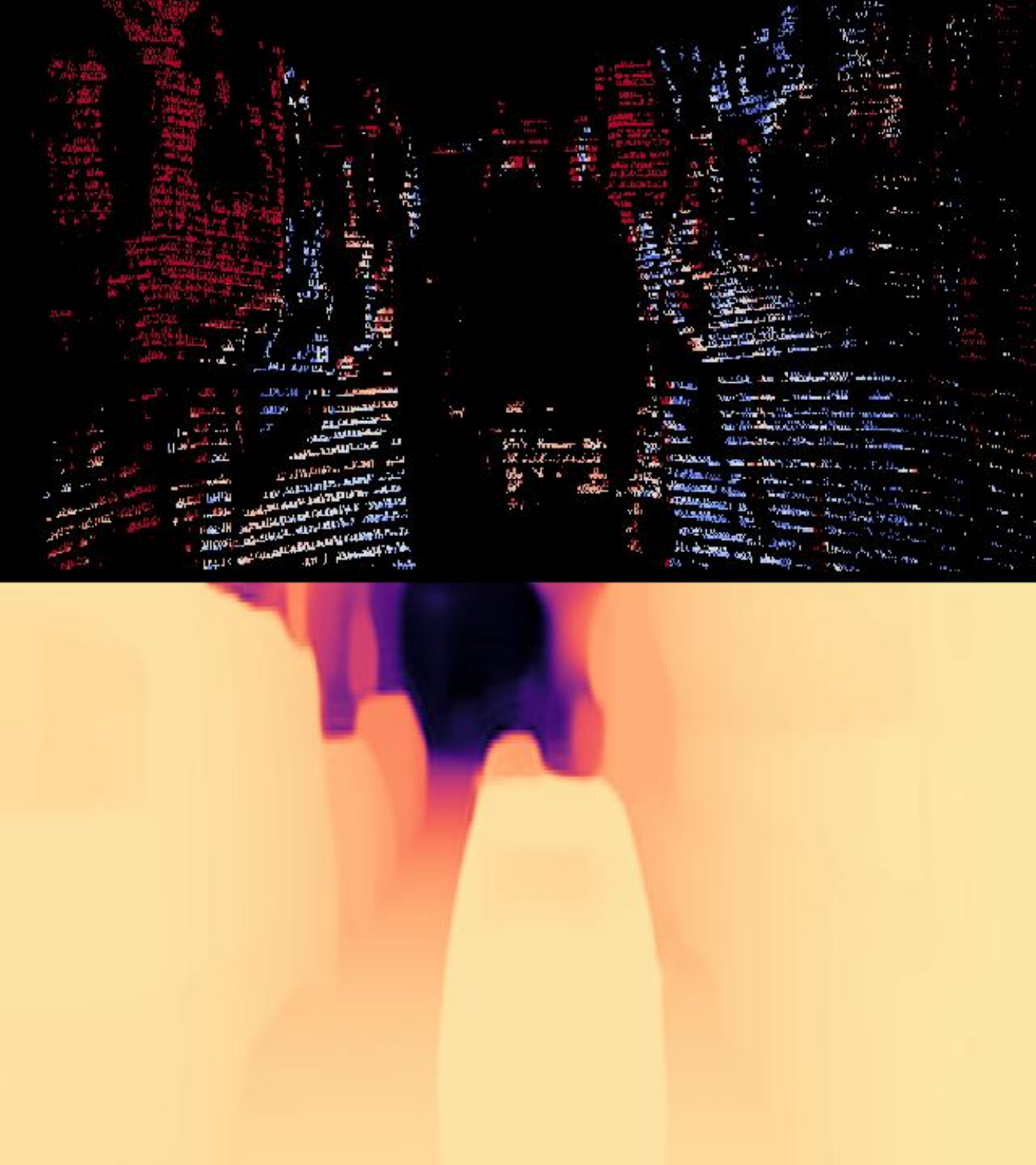}
        & \vspace{-2pt} \\

        \multirow{2}{*}[6pt]{\rotatebox[origin=c]{90}{NYU}}
        & \includegraphics[width=0.14\linewidth]{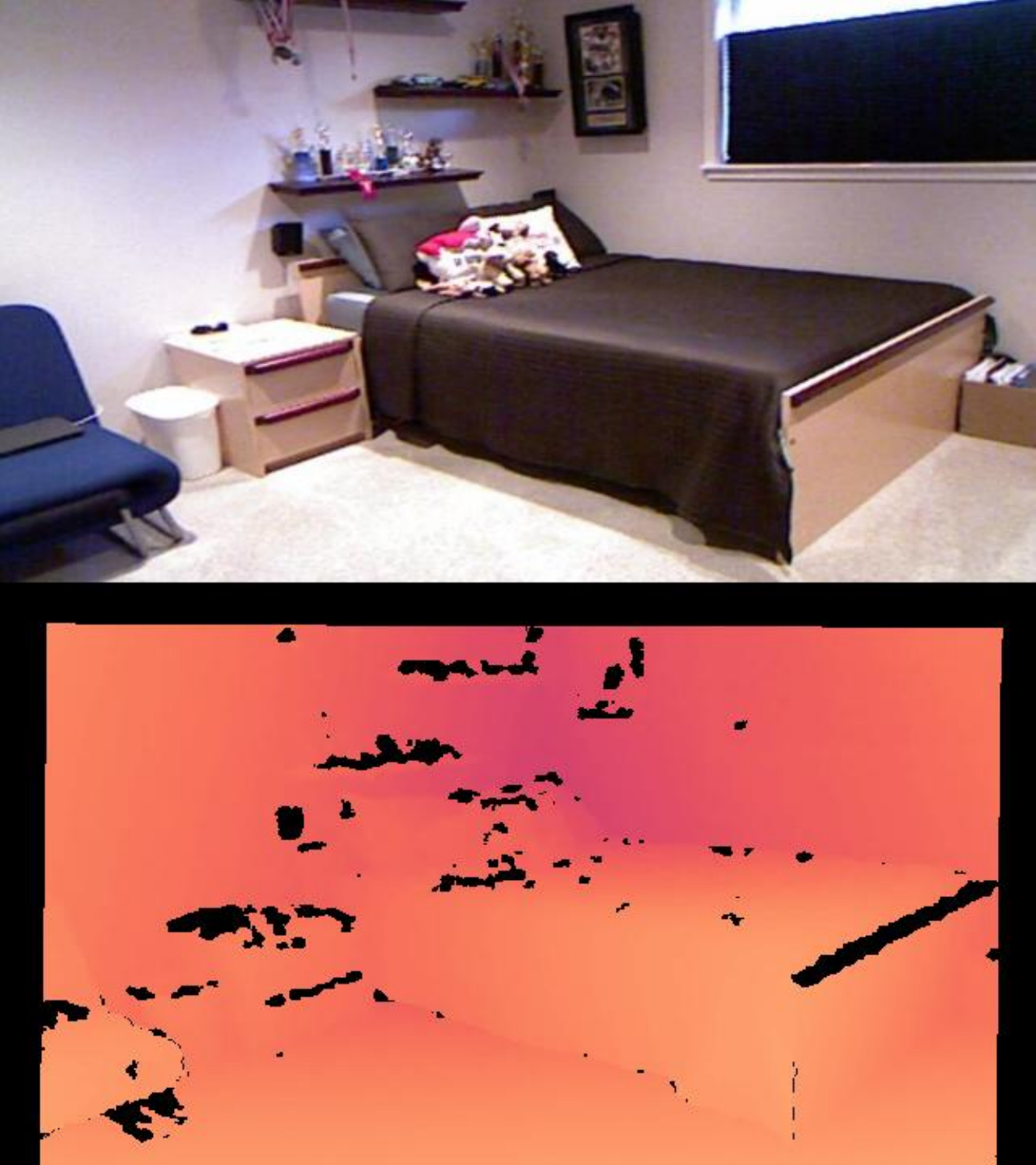}
        & \includegraphics[width=0.14\linewidth]{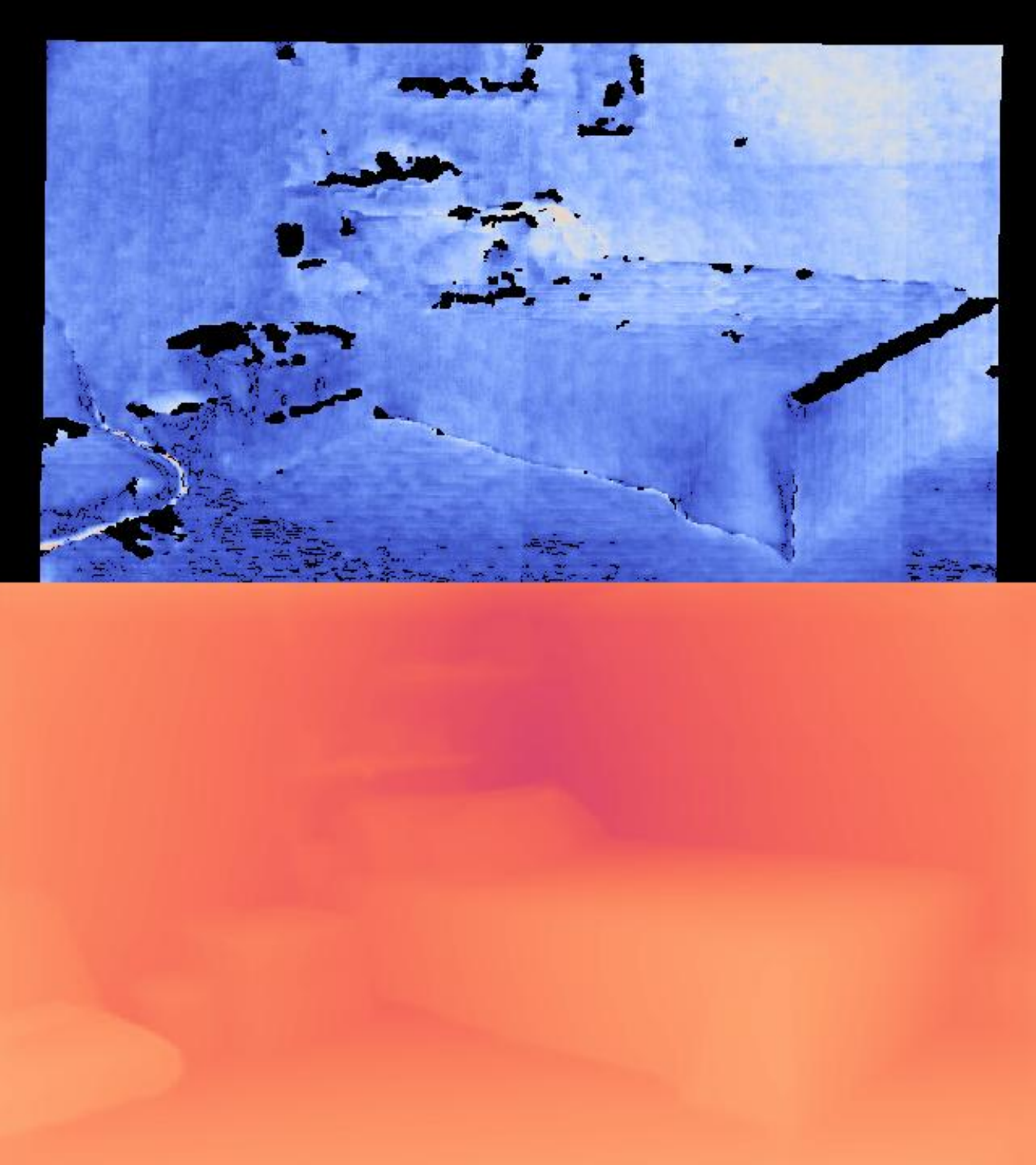}
        & \includegraphics[width=0.14\linewidth]{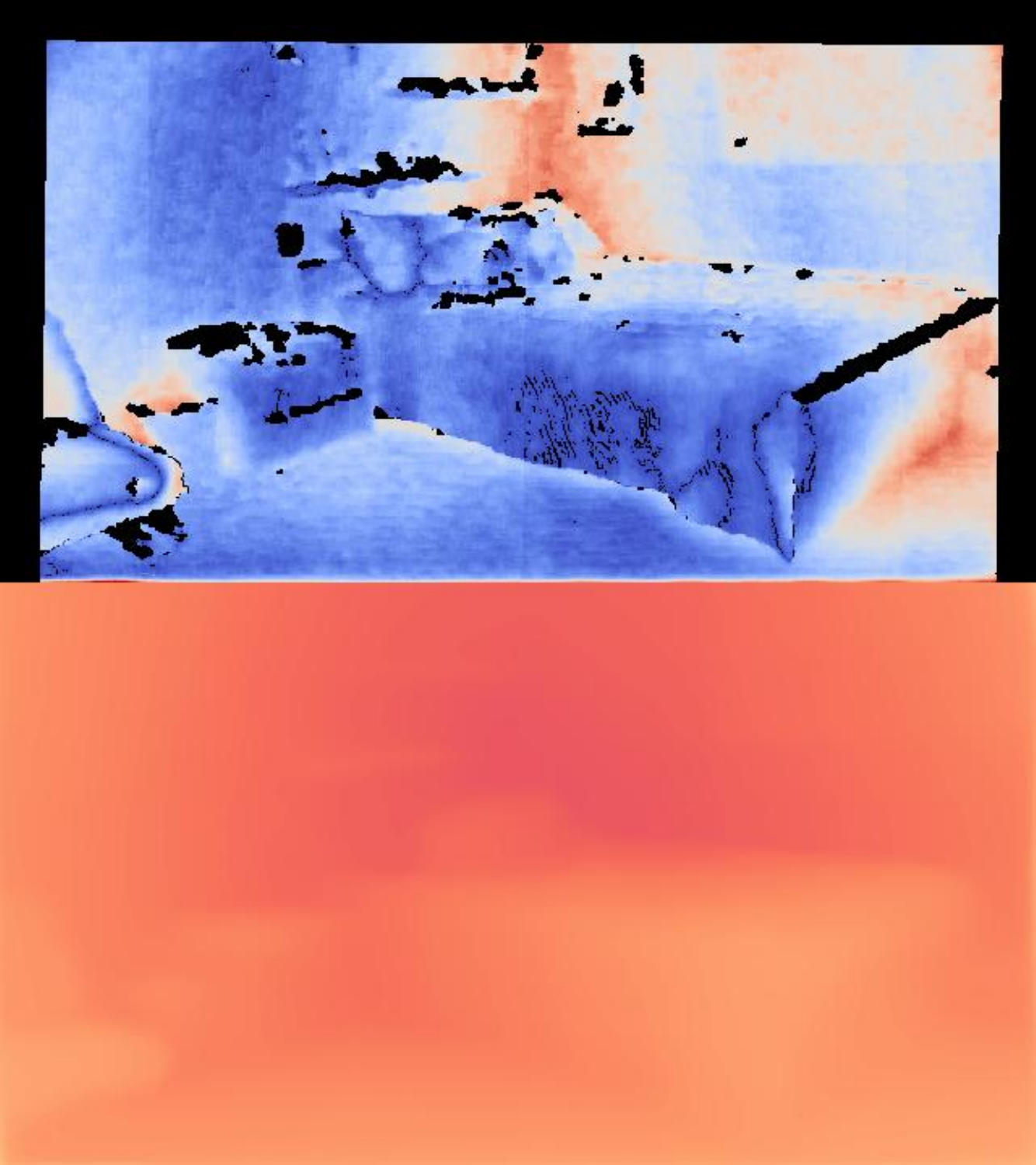}
        & \includegraphics[width=0.14\linewidth]{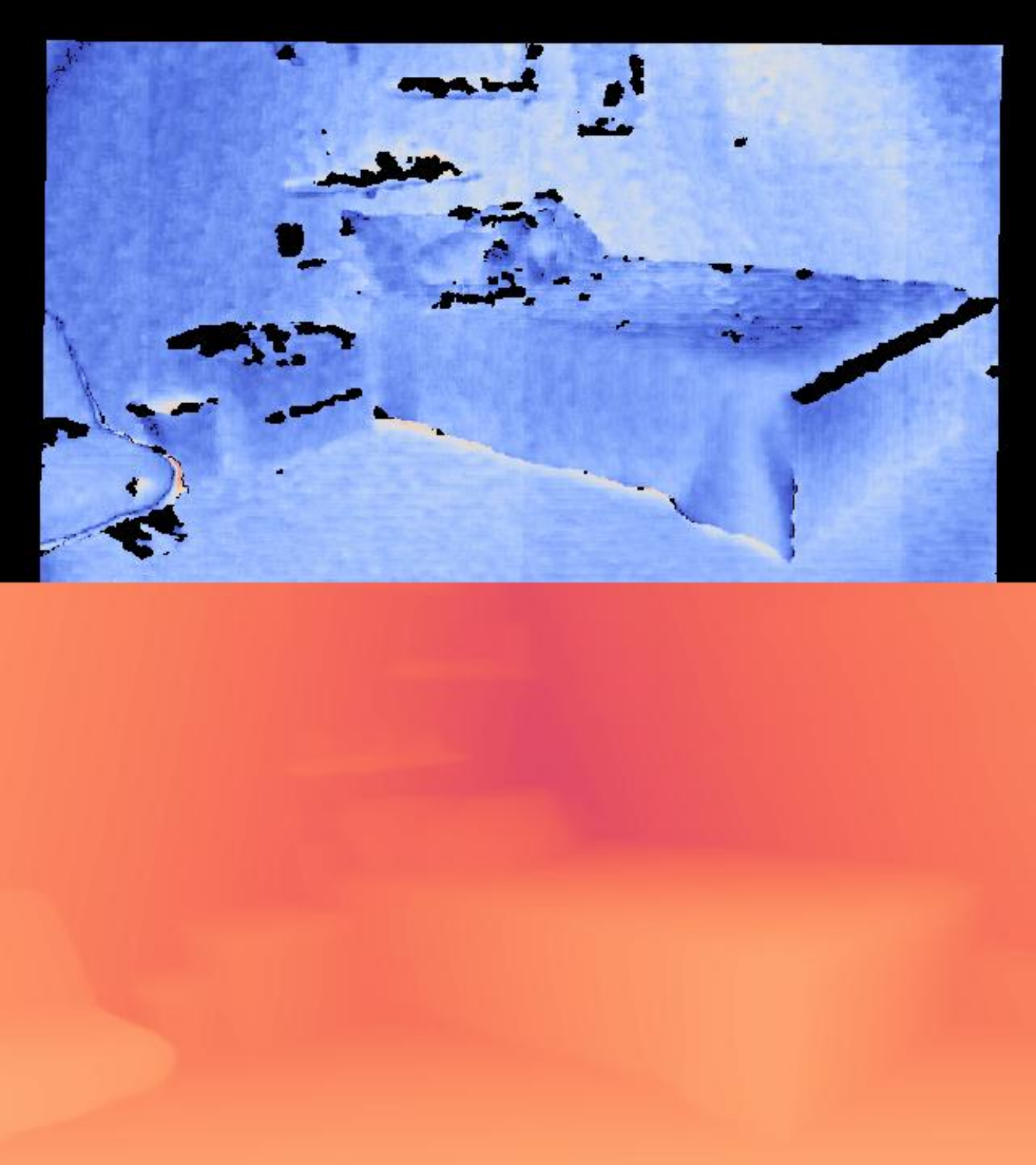}
        & \includegraphics[width=0.14\linewidth]{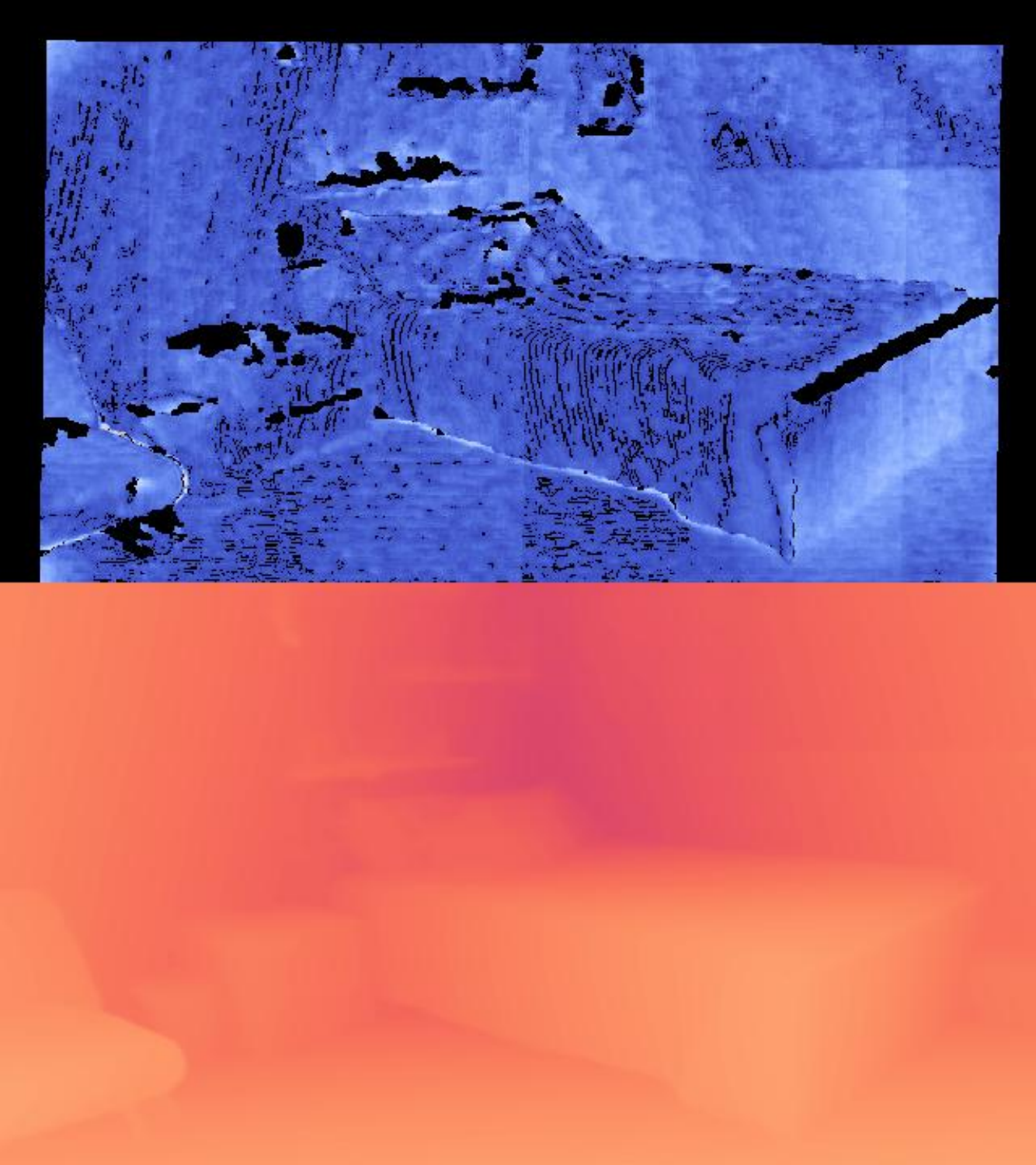}
        & \multirow{2}{*}[25pt]{\includegraphics[width=0.075\linewidth]{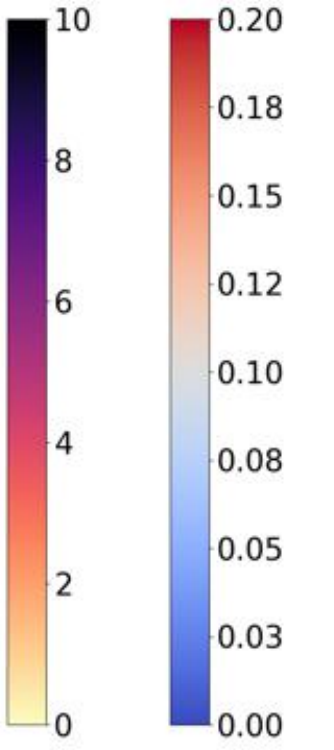}} \vspace{-8pt} \\
        & \includegraphics[width=0.14\linewidth]{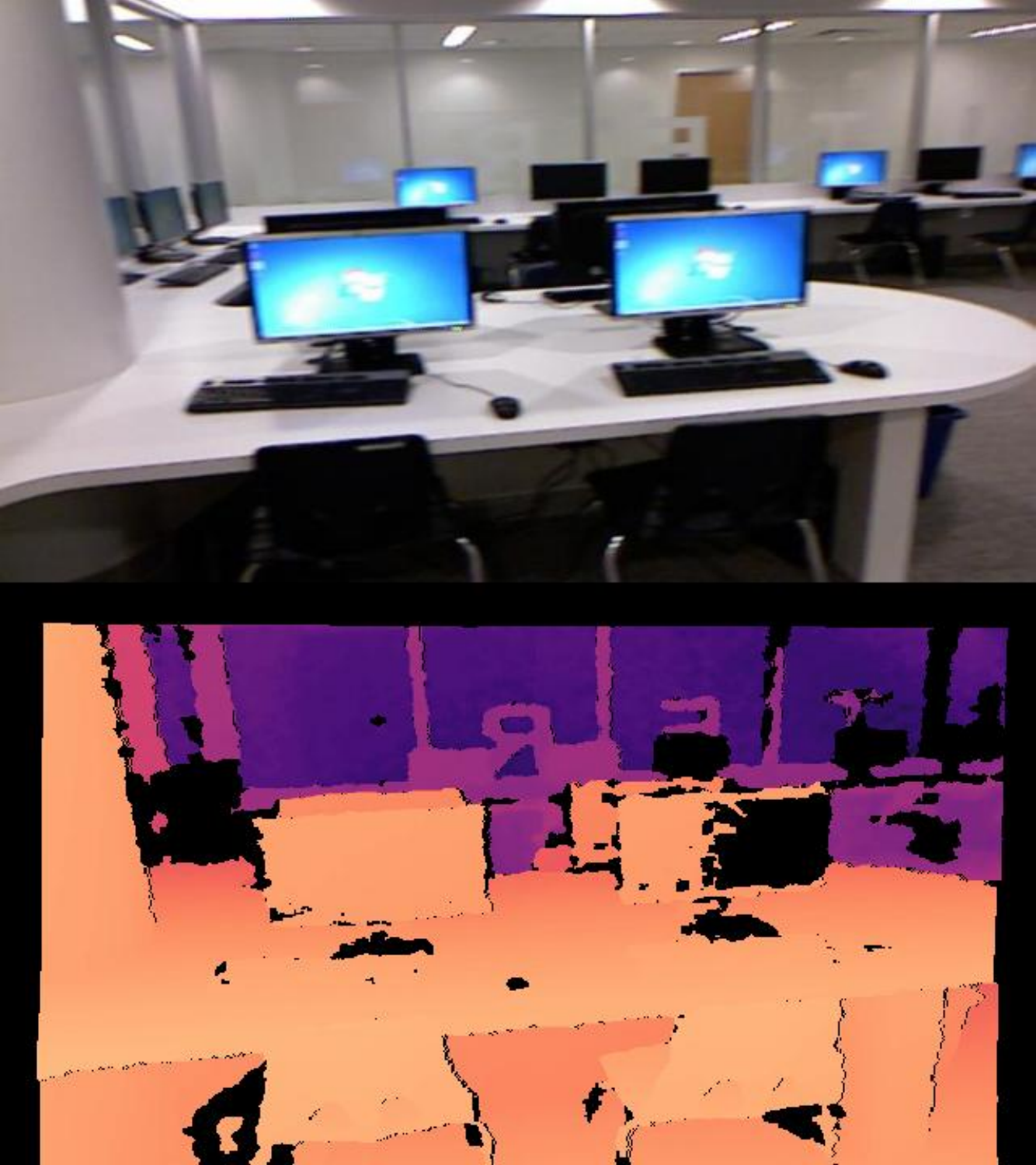}
        & \includegraphics[width=0.14\linewidth]{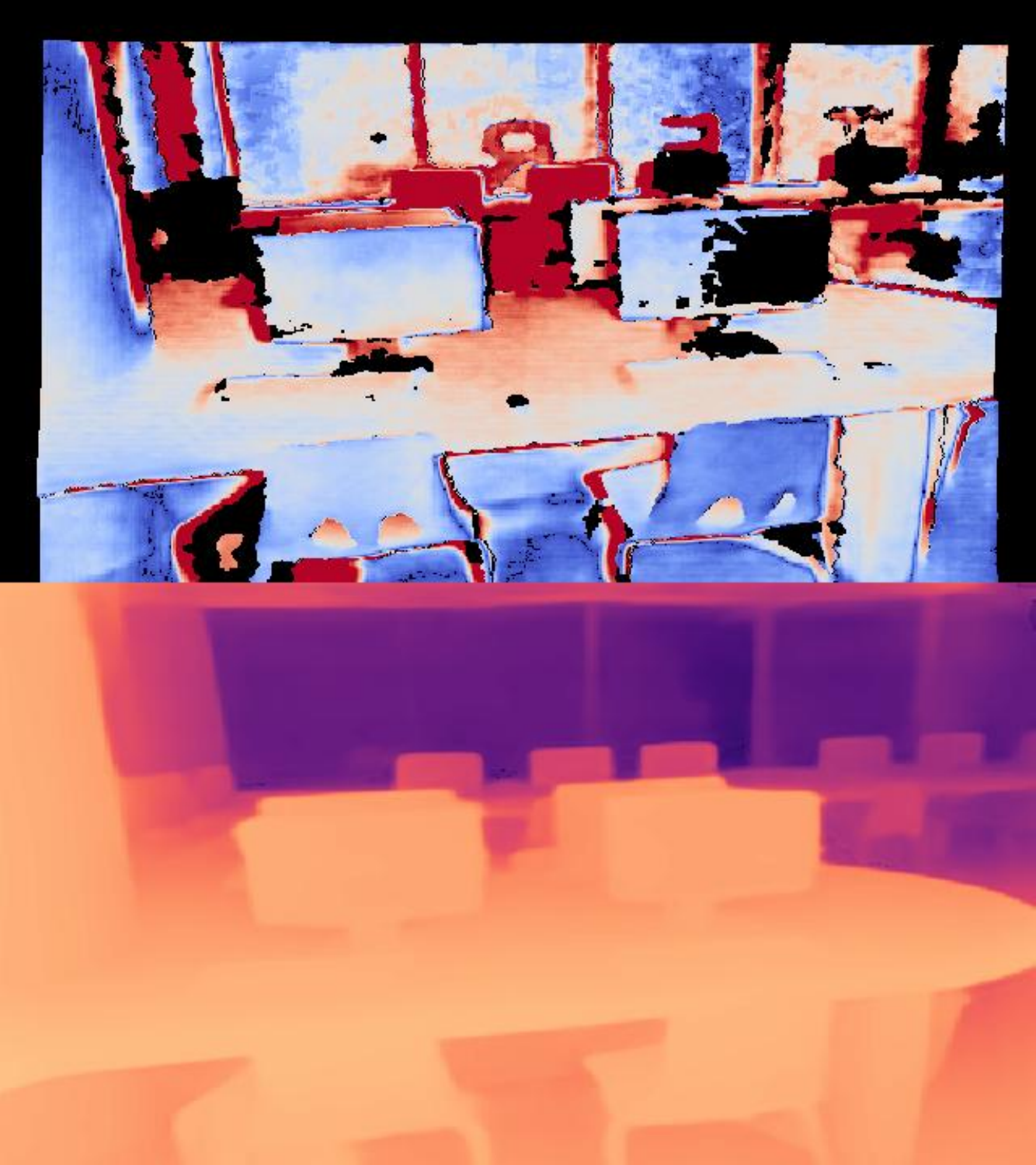}
        & \includegraphics[width=0.14\linewidth]{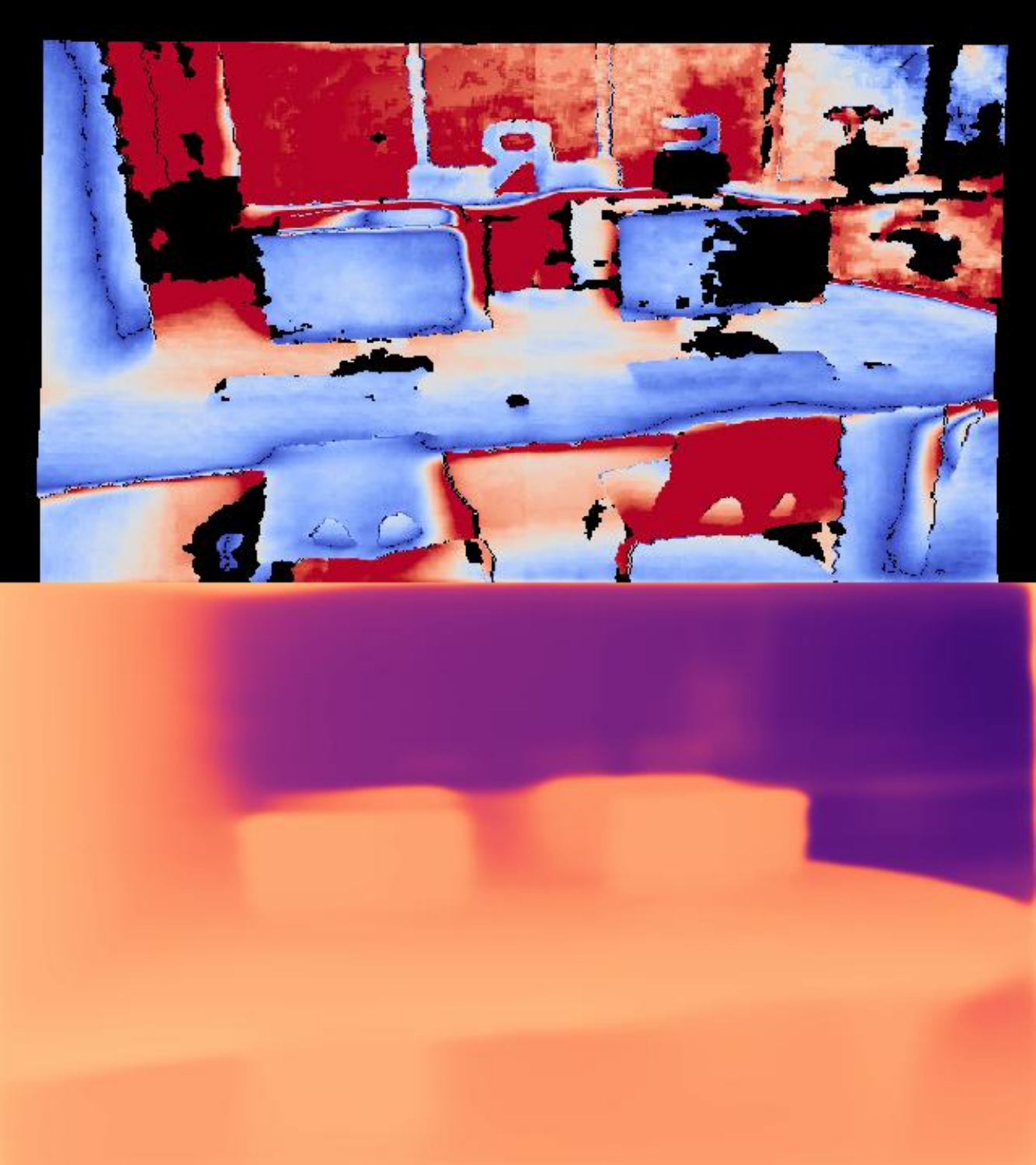}
        & \includegraphics[width=0.14\linewidth]{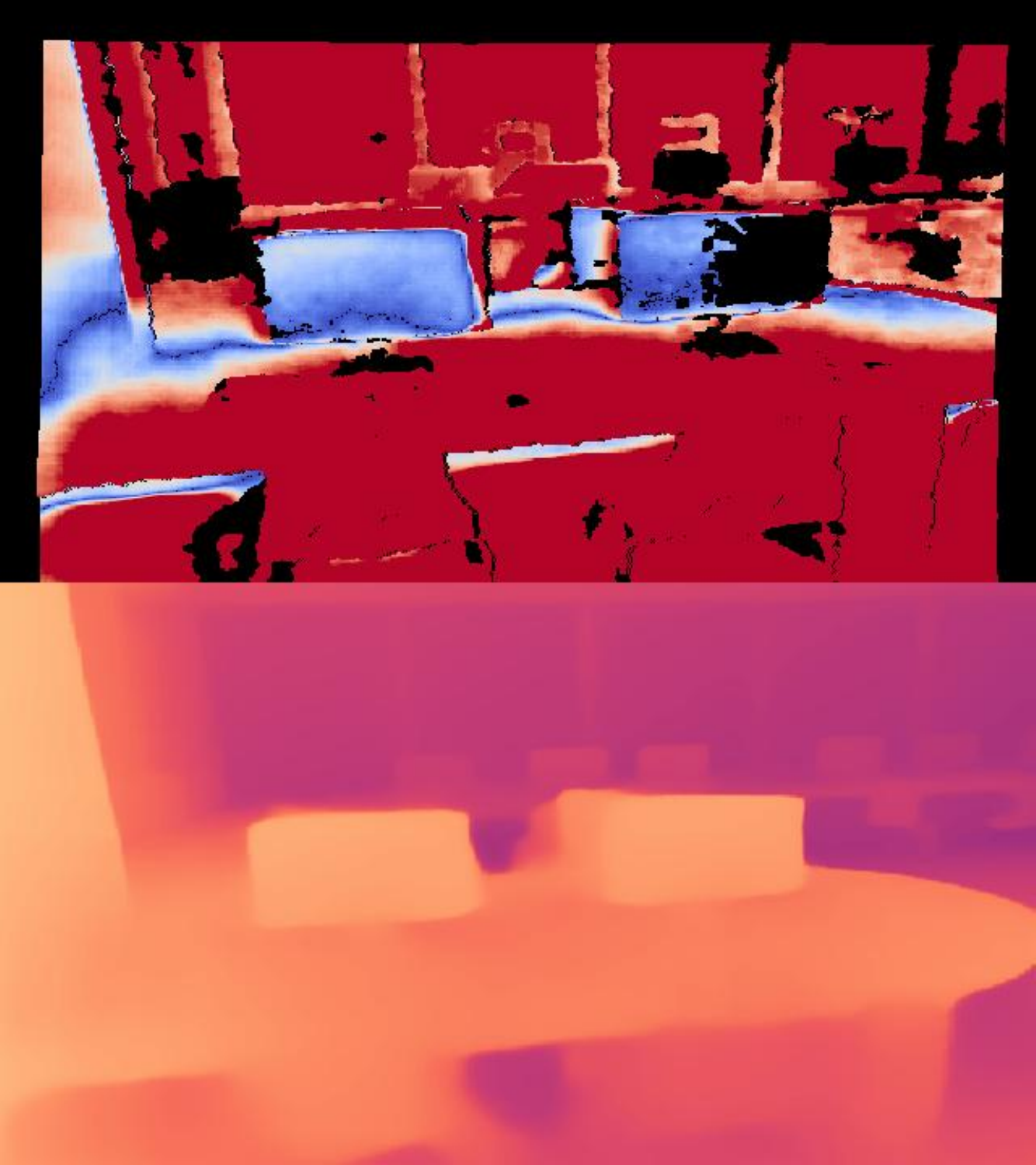}
        & \includegraphics[width=0.14\linewidth]{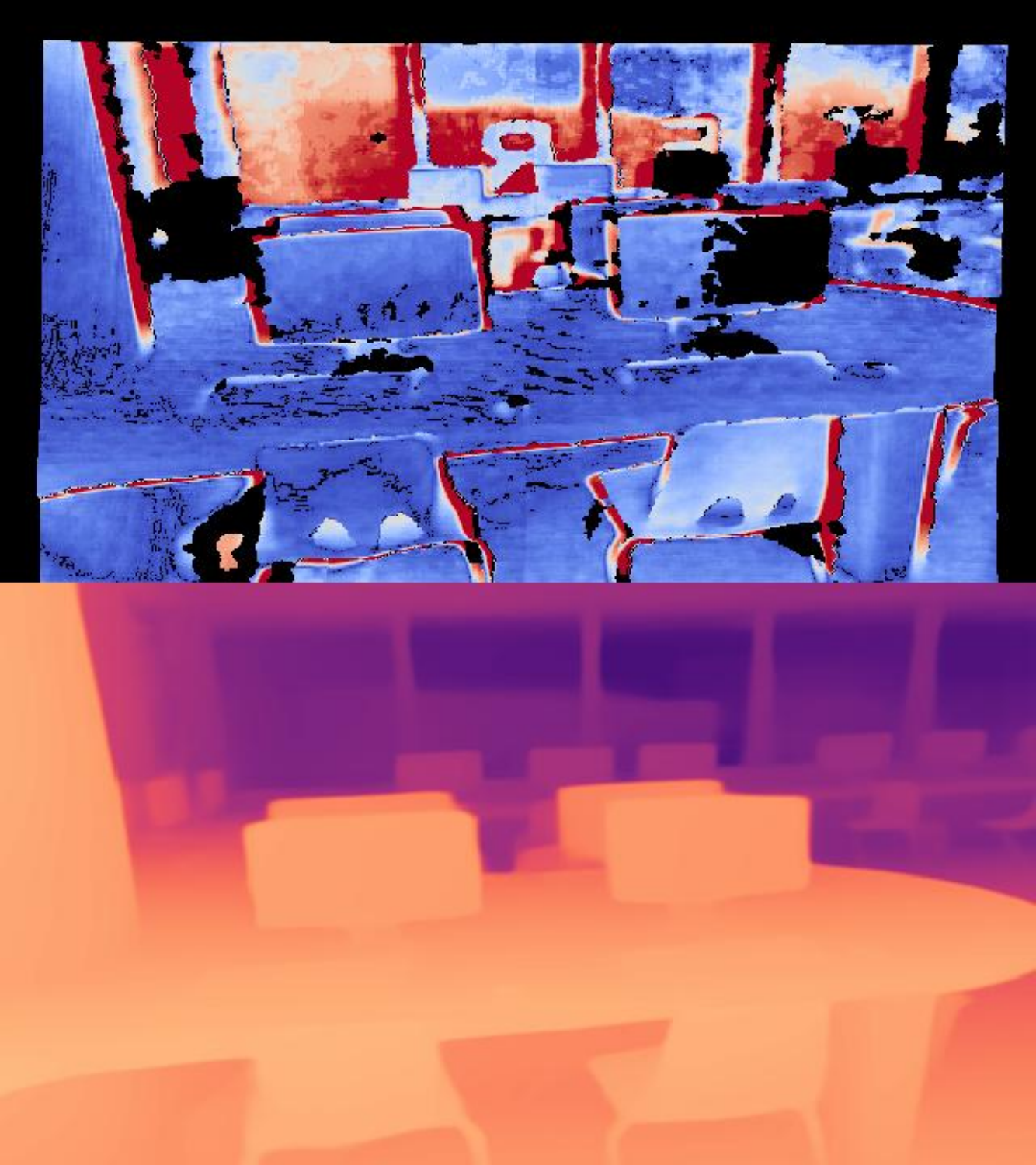}
        & \vspace{-2pt} \\

        \multirow{2}{*}[6pt]{\rotatebox[origin=c]{90}{Diode}}
        & \includegraphics[width=0.14\linewidth]{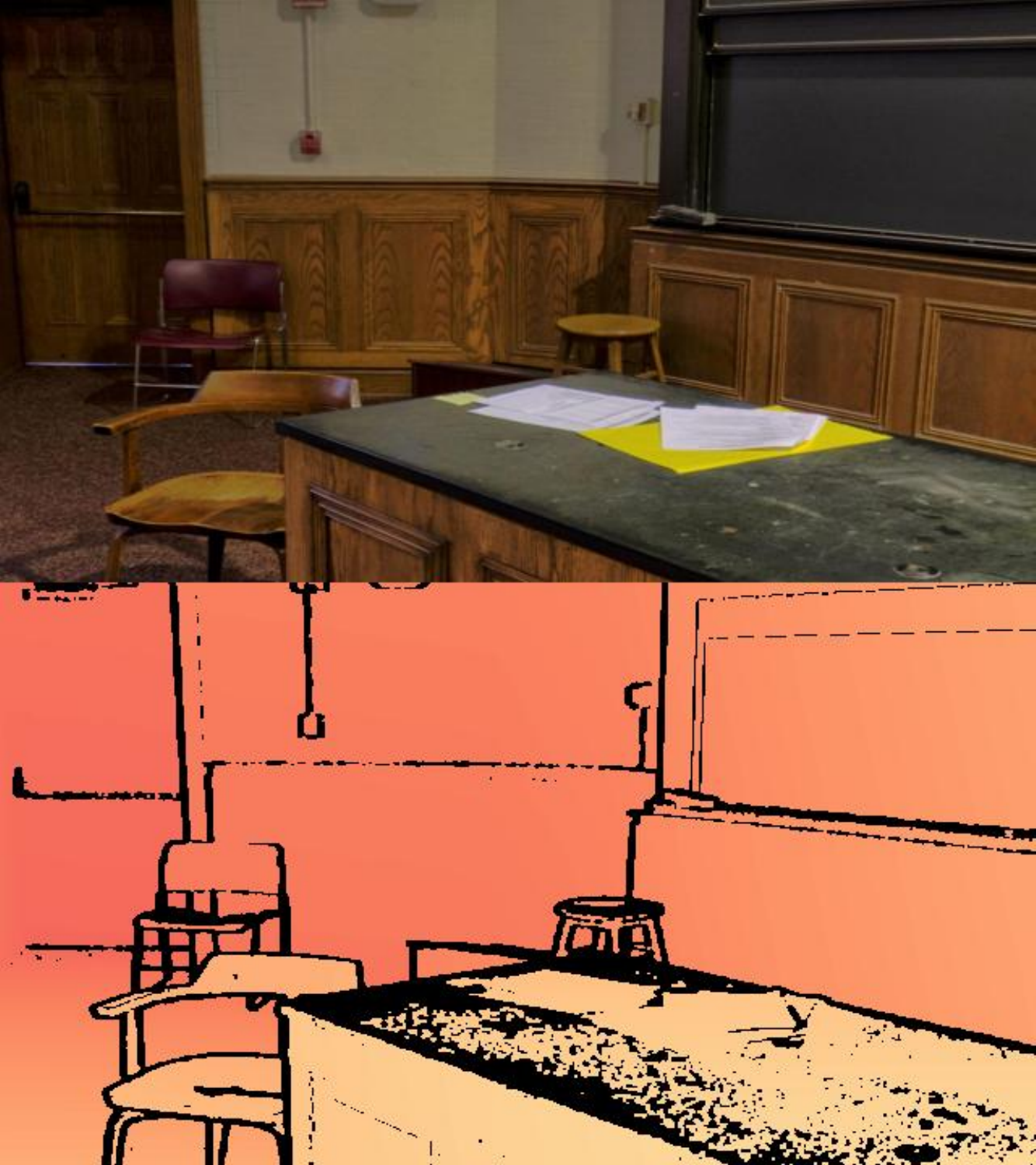}
        & \includegraphics[width=0.14\linewidth]{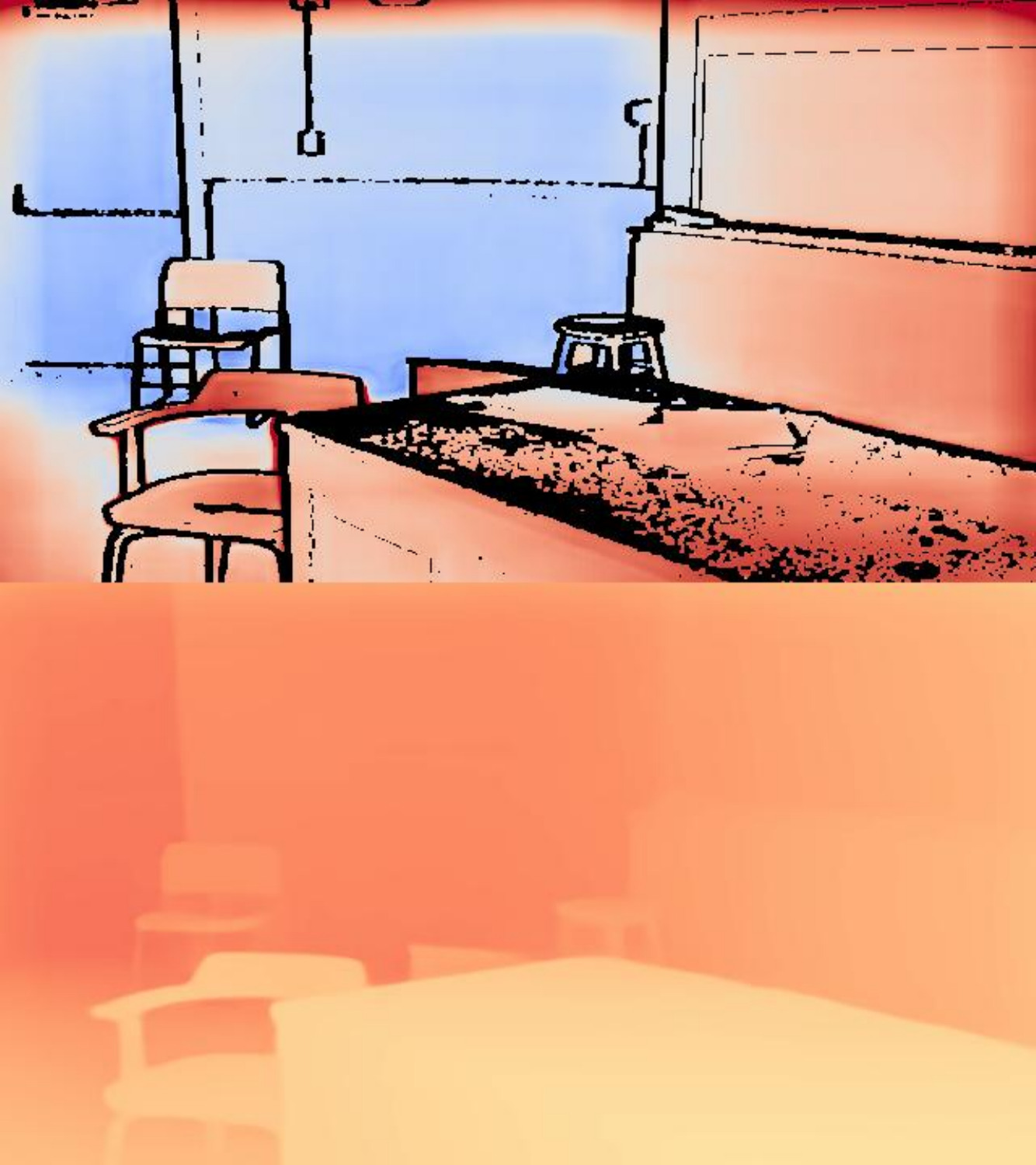}
        & \includegraphics[width=0.14\linewidth]{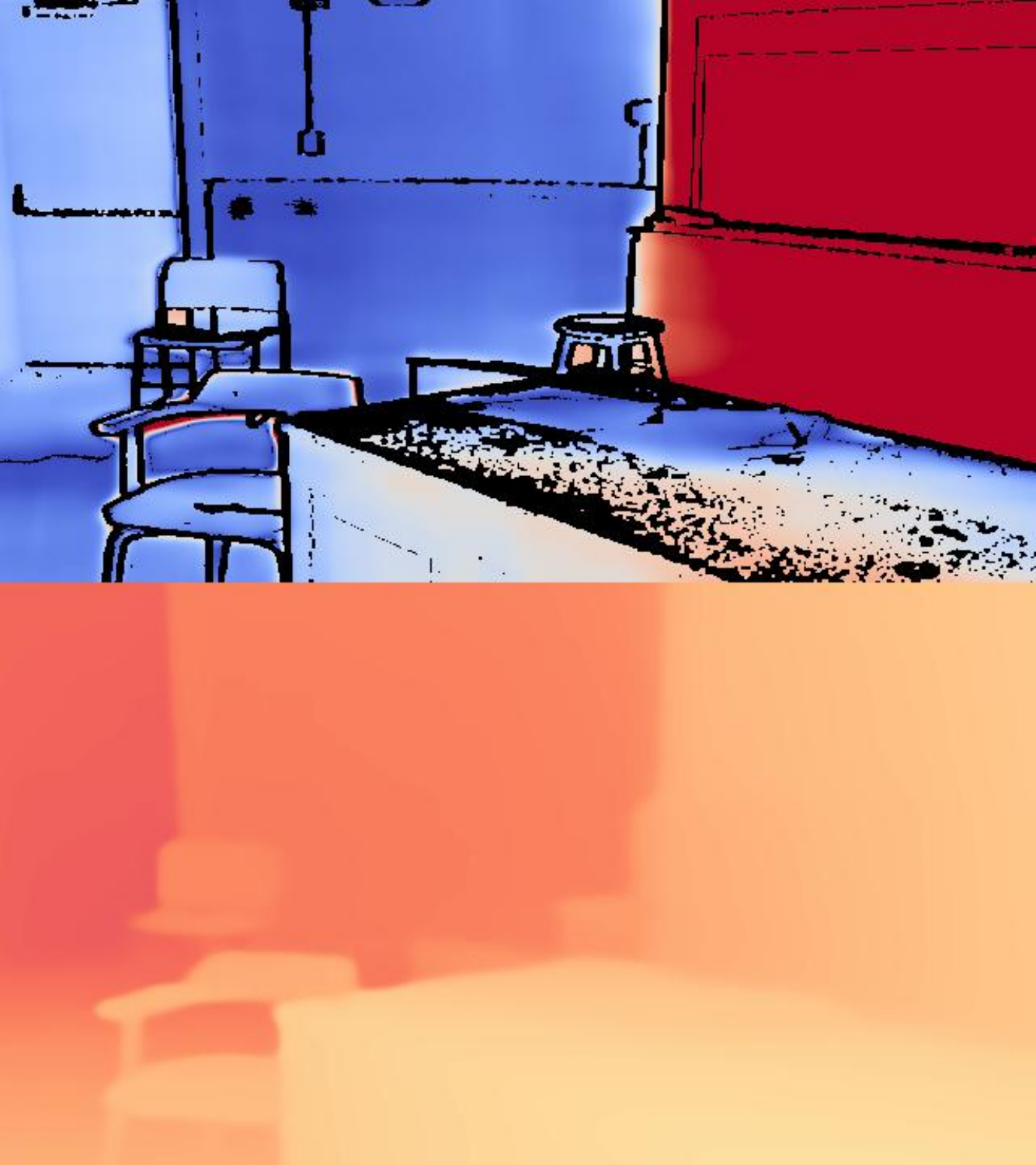}
        & \includegraphics[width=0.14\linewidth]{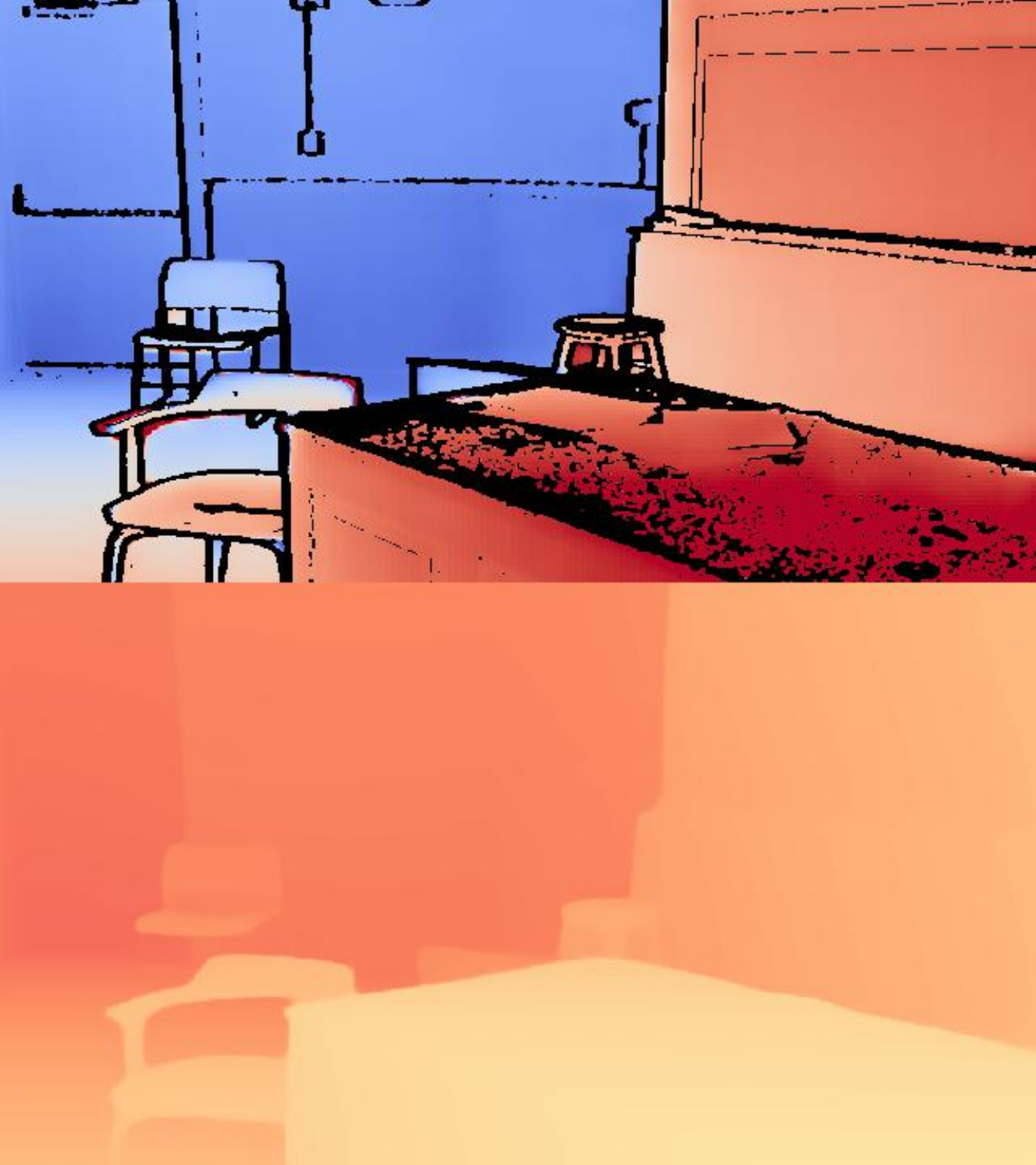}
        & \includegraphics[width=0.14\linewidth]{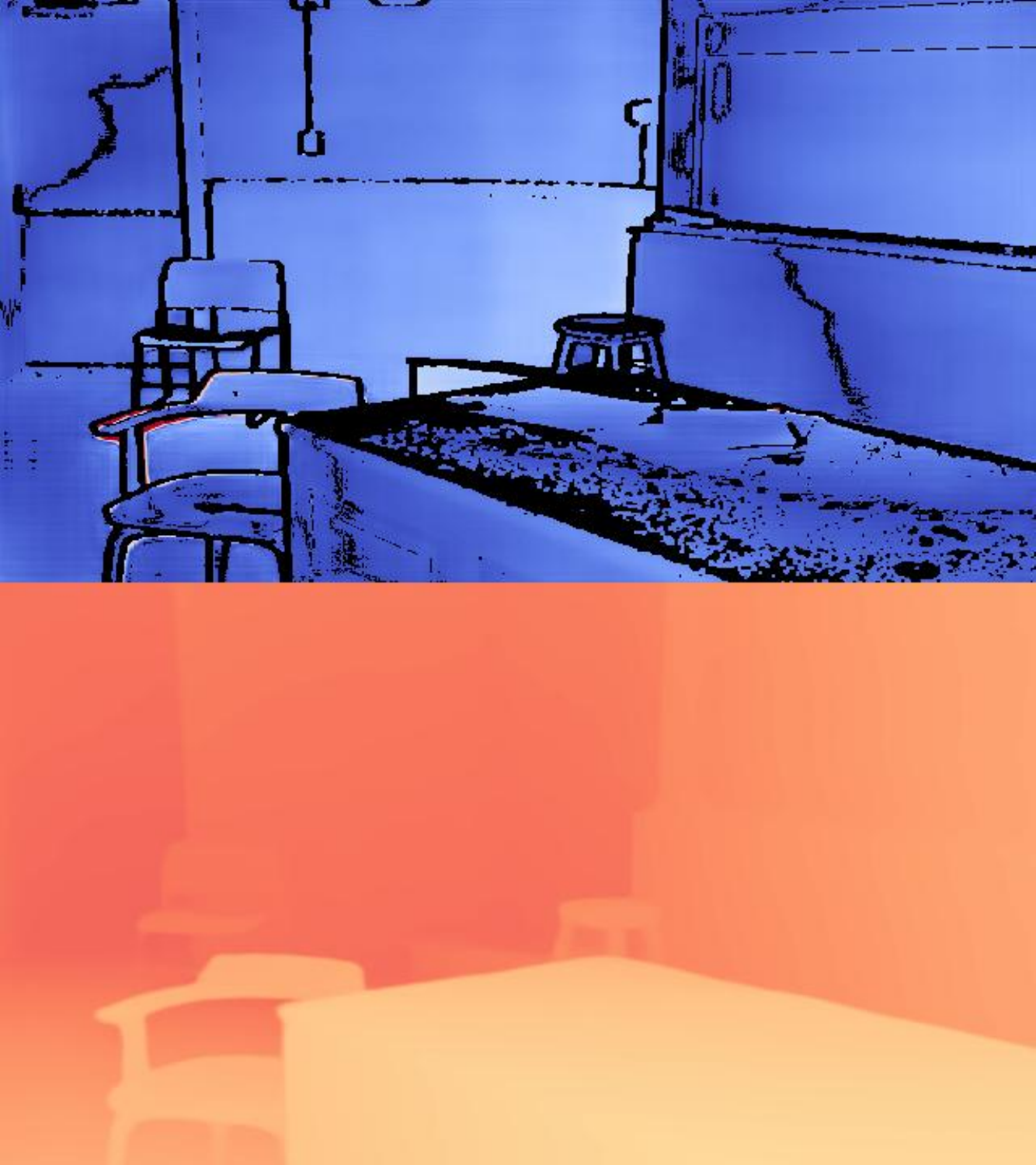}
        & \multirow{2}{*}[20pt]{\includegraphics[width=0.075\linewidth]{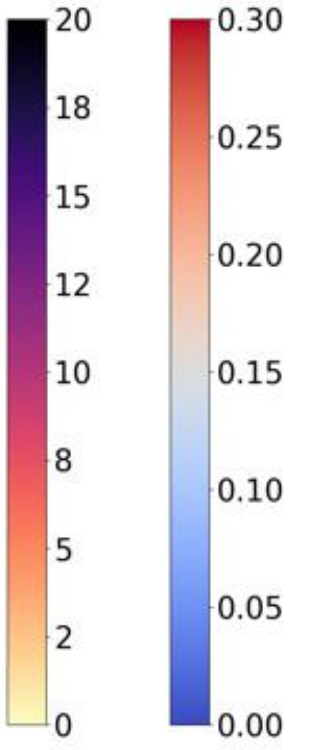}} \vspace{-8pt} \\
        & \includegraphics[width=0.14\linewidth]{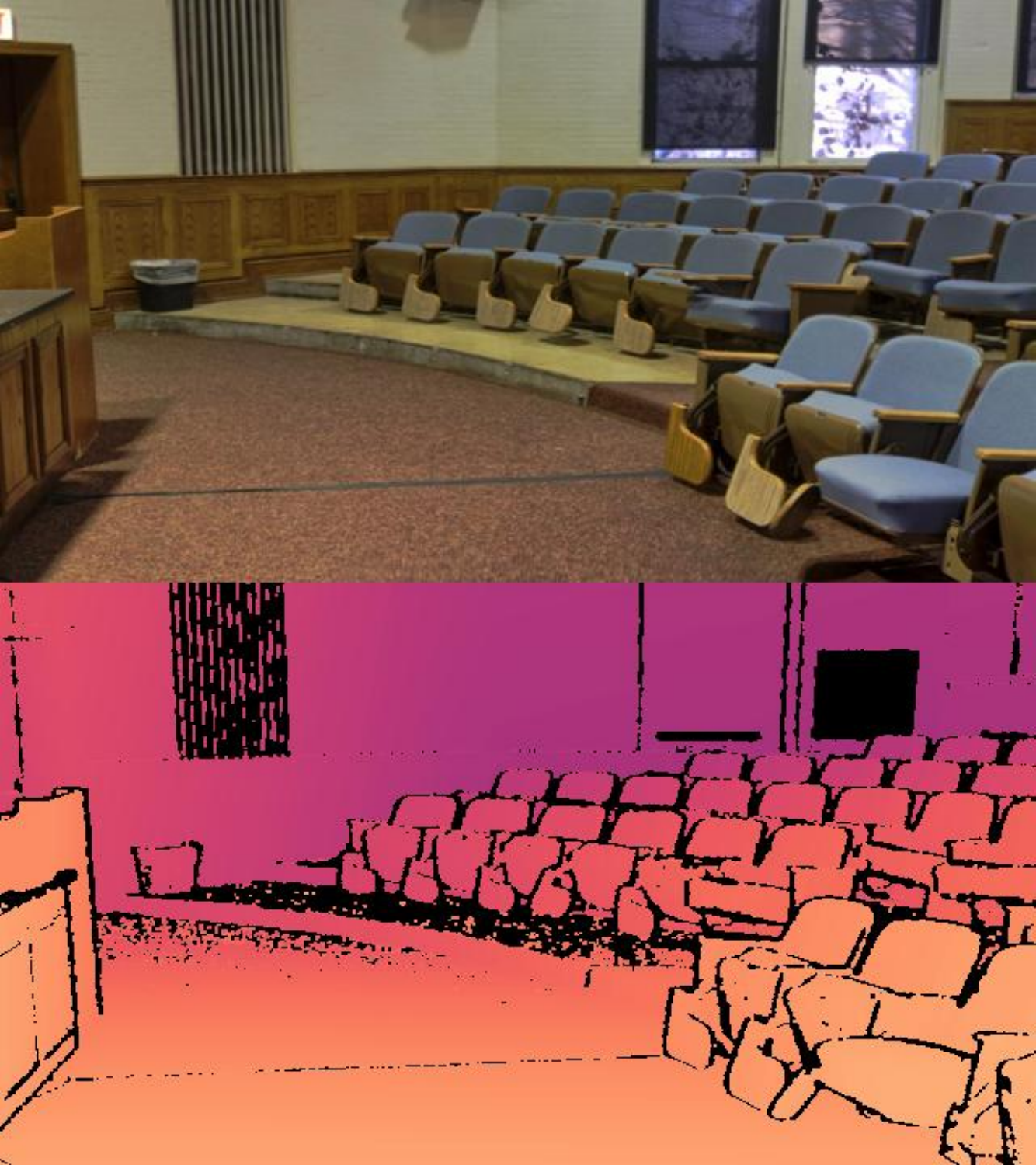}
        & \includegraphics[width=0.14\linewidth]{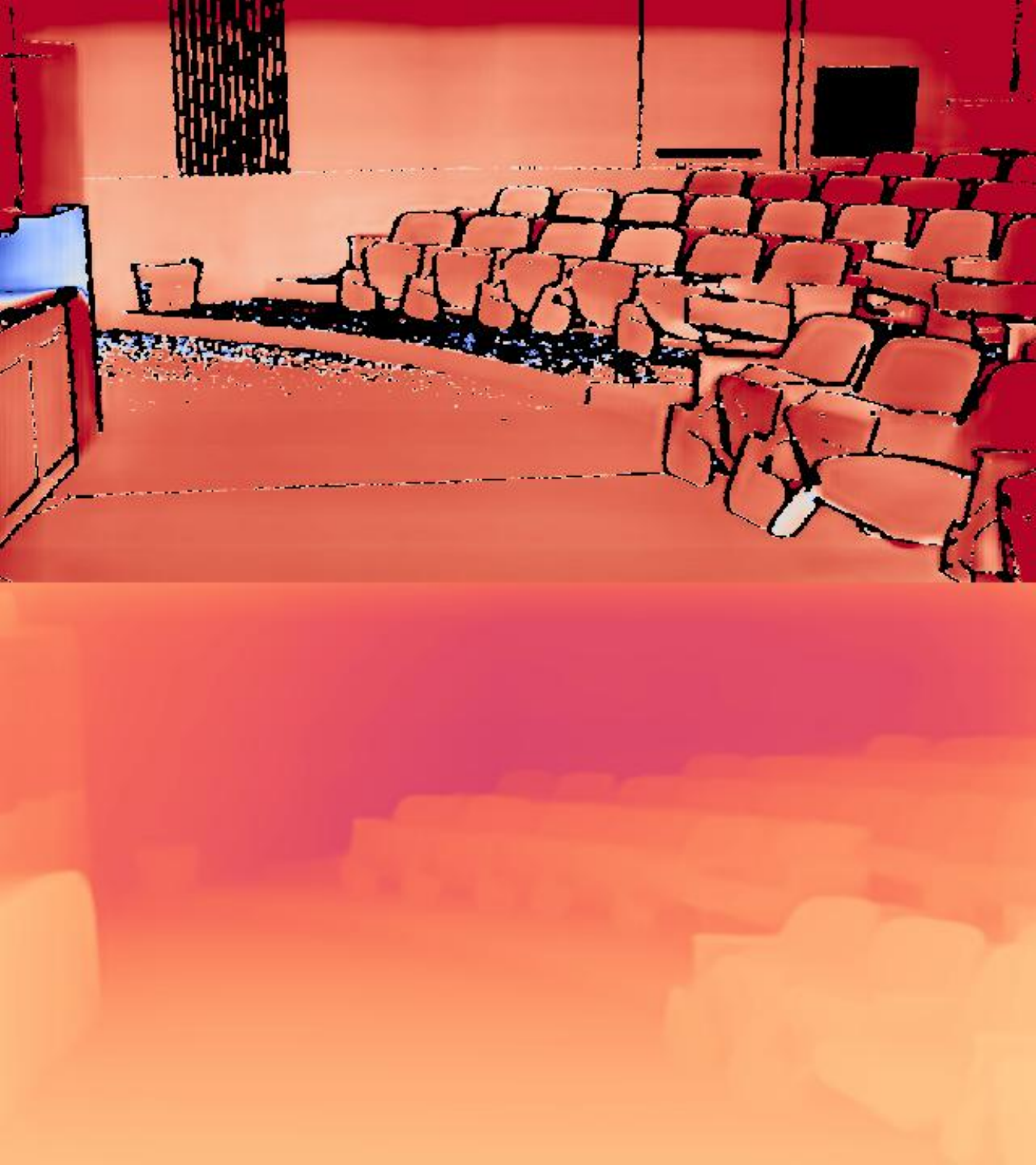}
        & \includegraphics[width=0.14\linewidth]{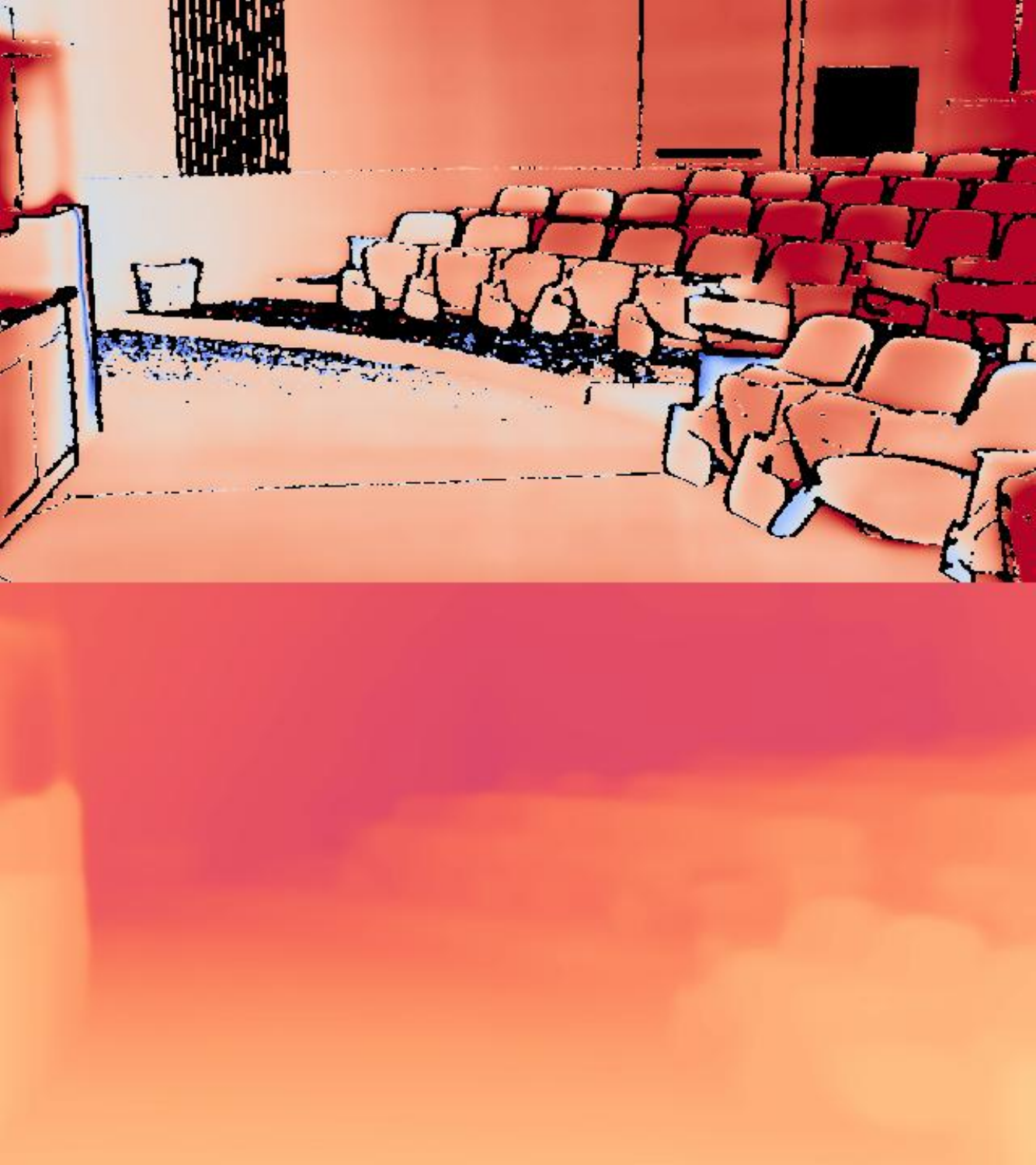}
        & \includegraphics[width=0.14\linewidth]{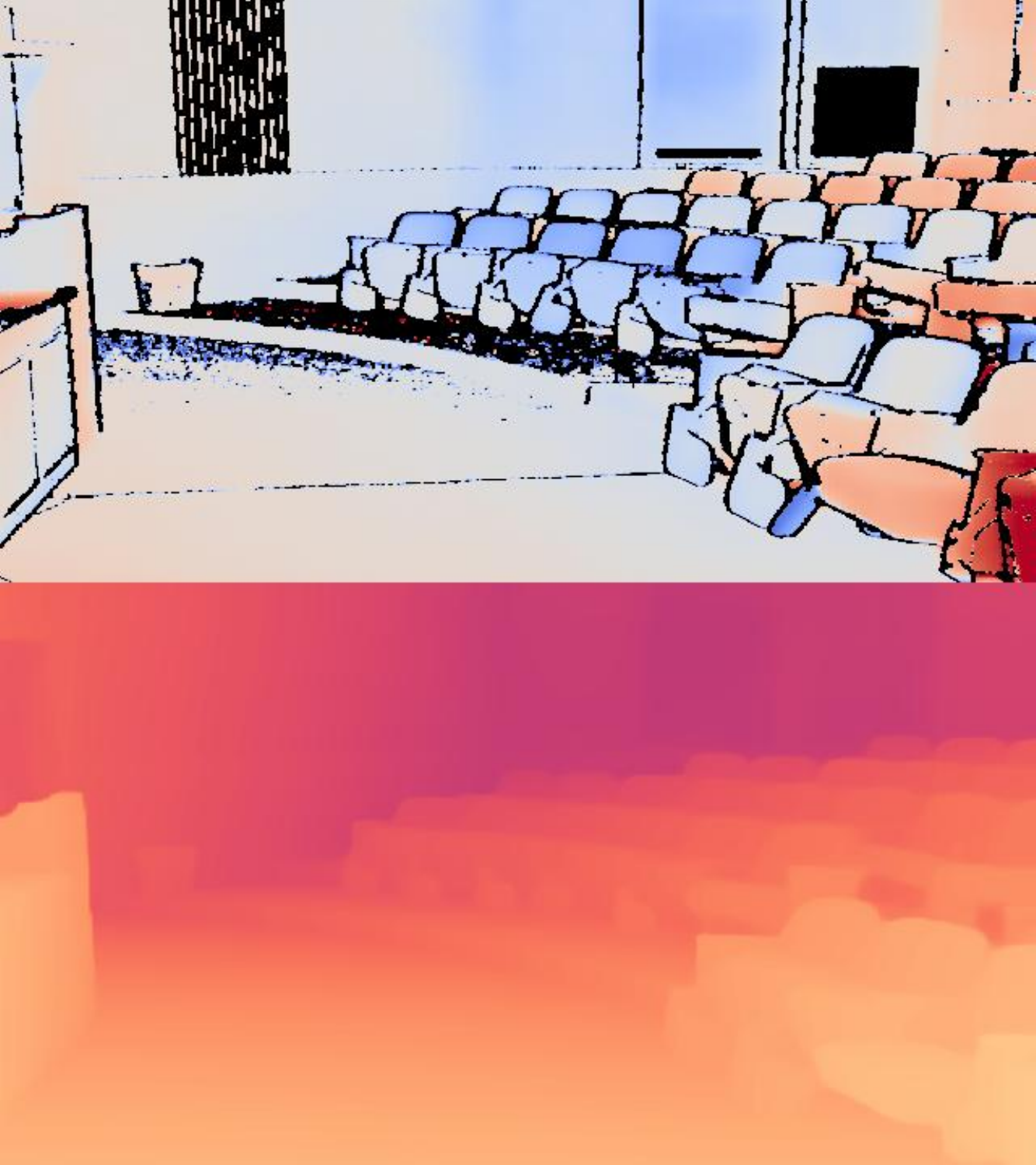}
        & \includegraphics[width=0.14\linewidth]{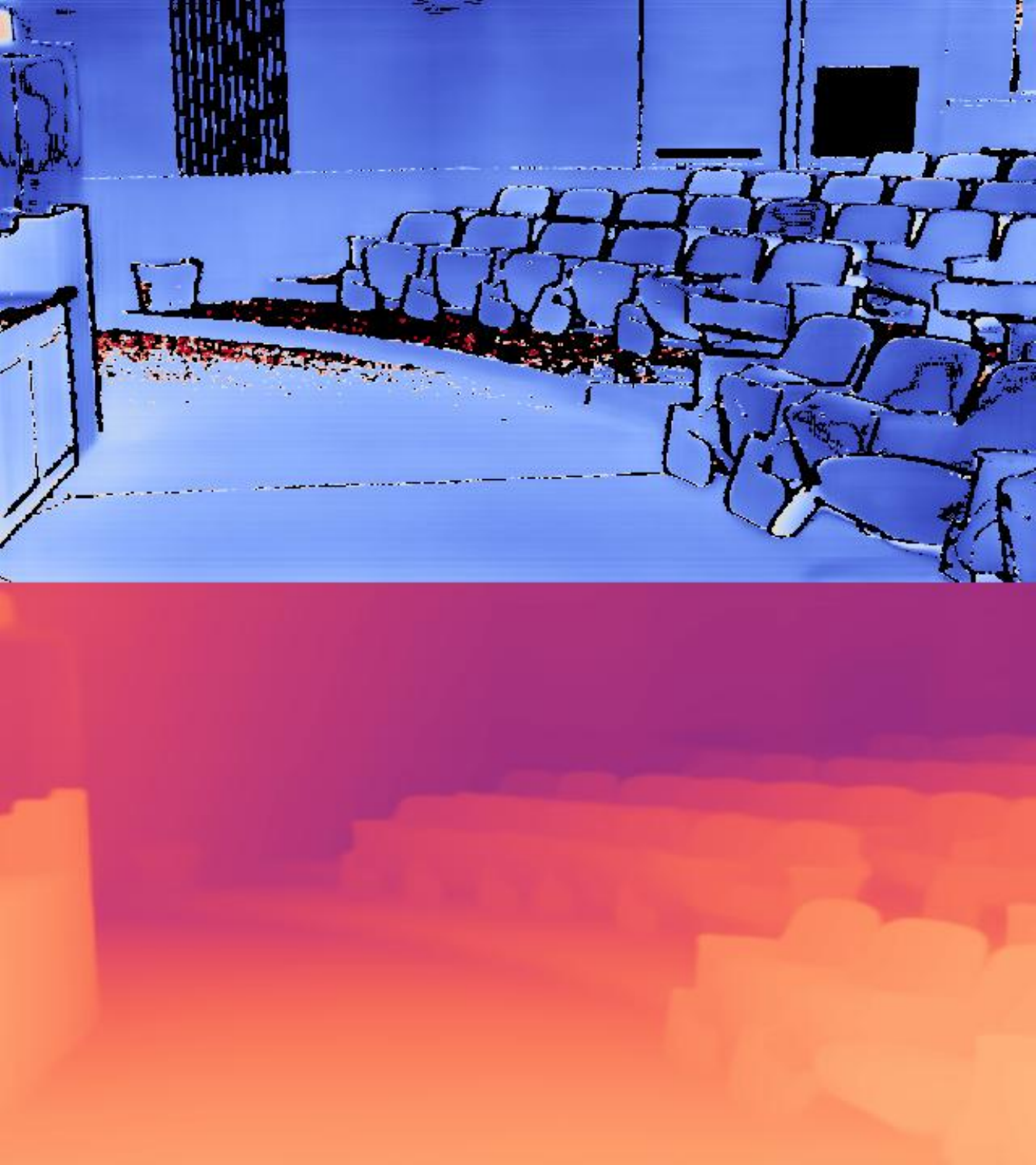}
        & \vspace{-2pt} \\

        \multirow{2}{*}[6pt]{\rotatebox[origin=c]{90}{ETH3D}}
        & \includegraphics[width=0.14\linewidth]{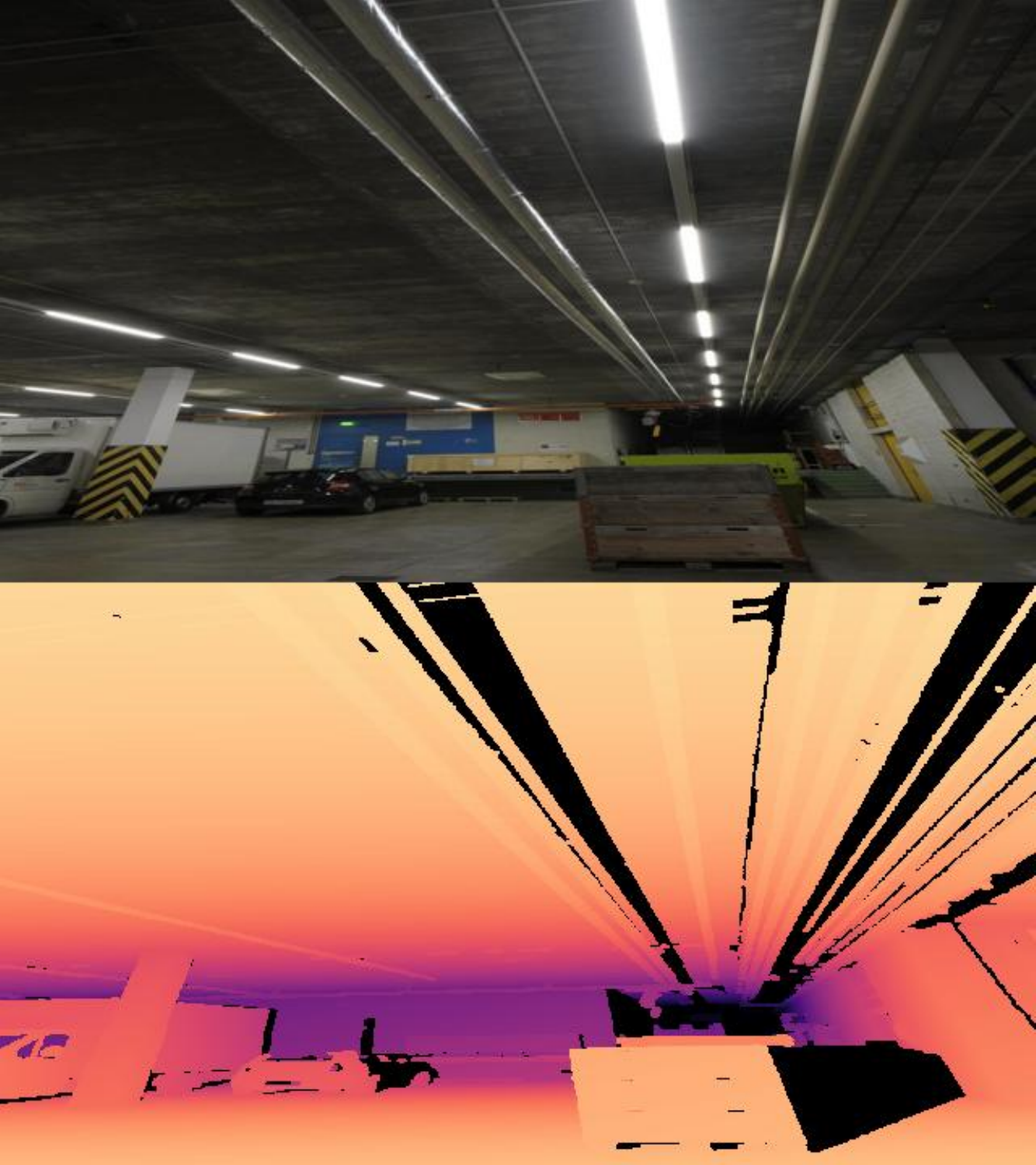}
        & \includegraphics[width=0.14\linewidth]{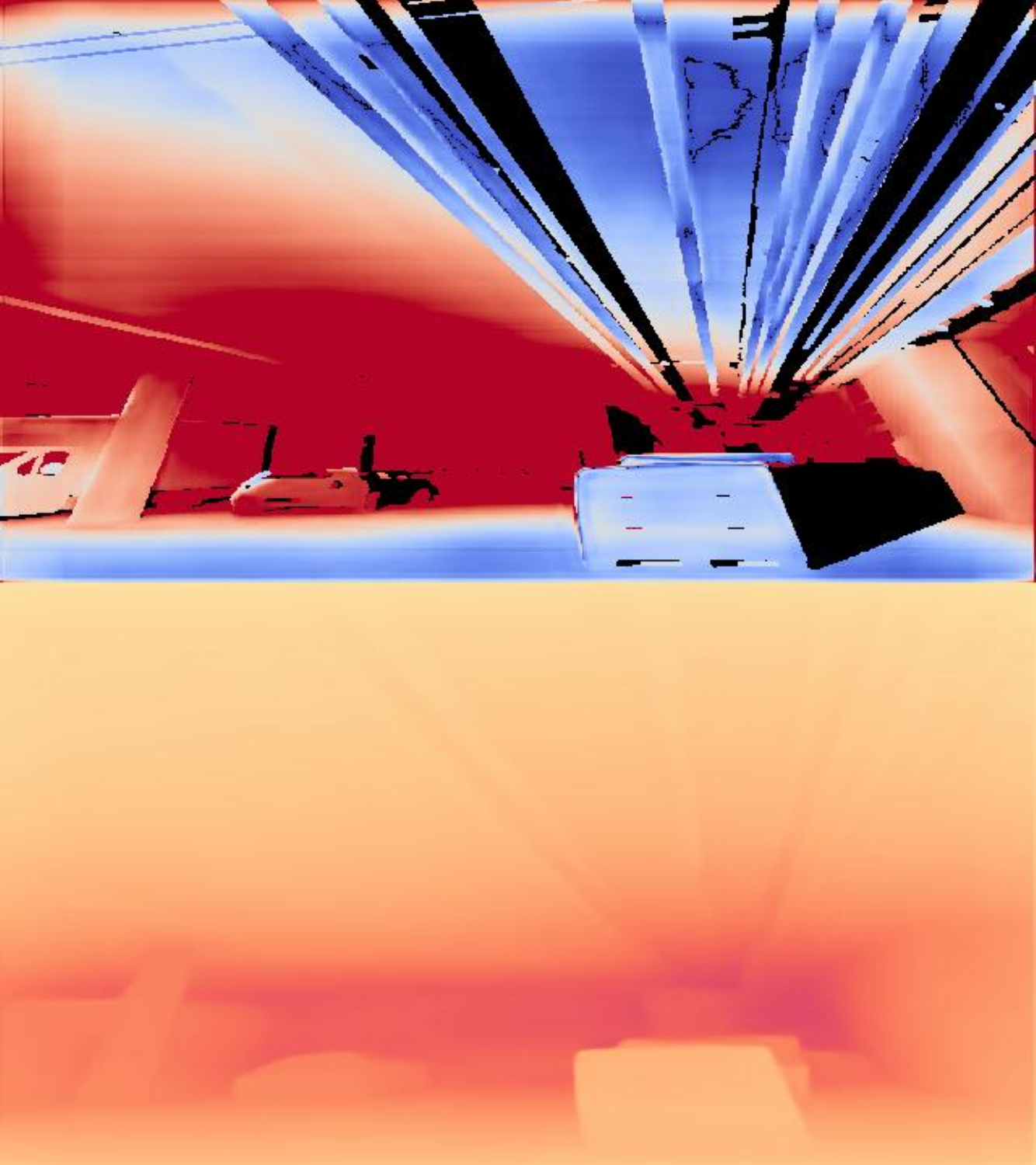}
        & \includegraphics[width=0.14\linewidth]{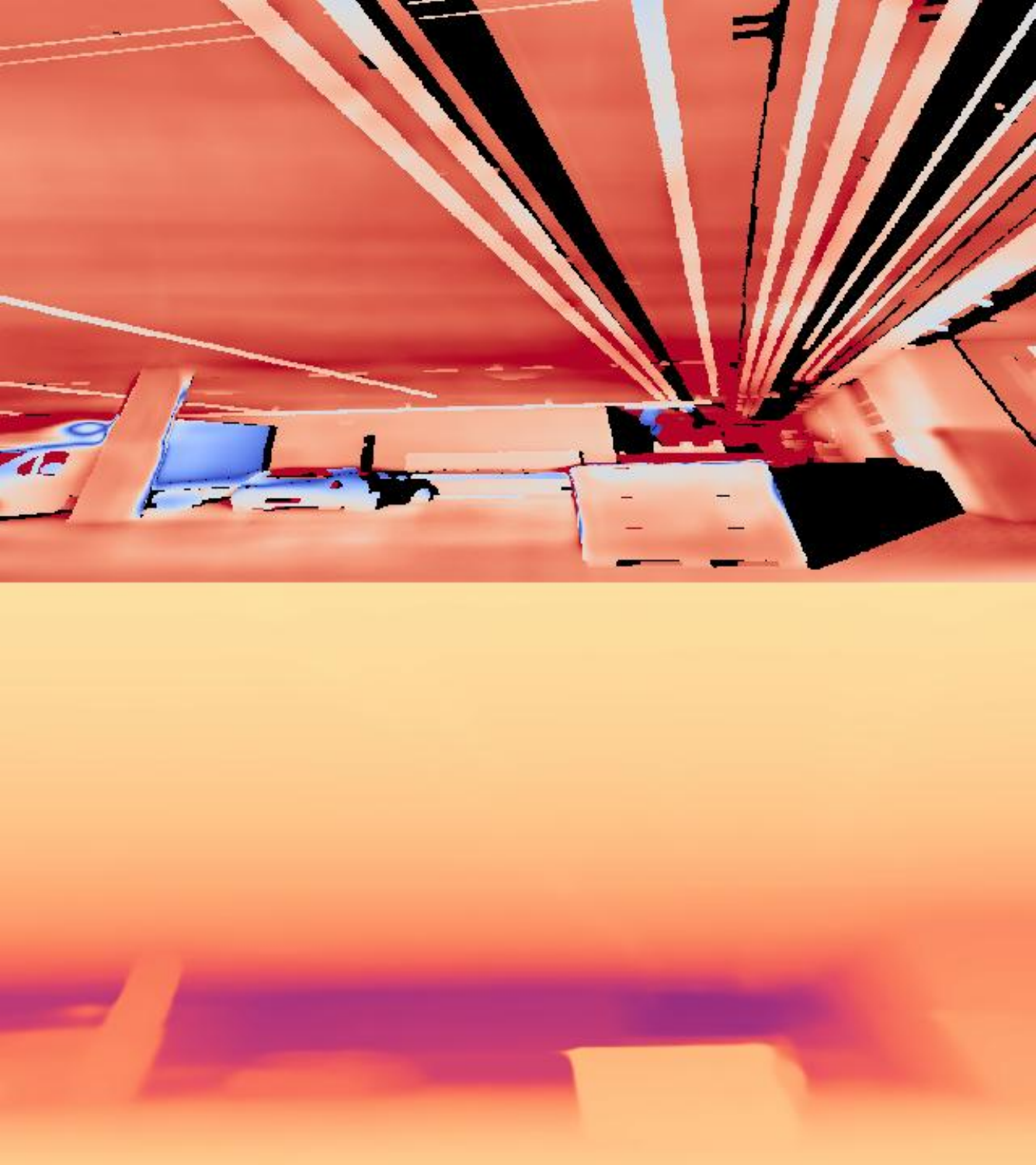}
        & \includegraphics[width=0.14\linewidth]{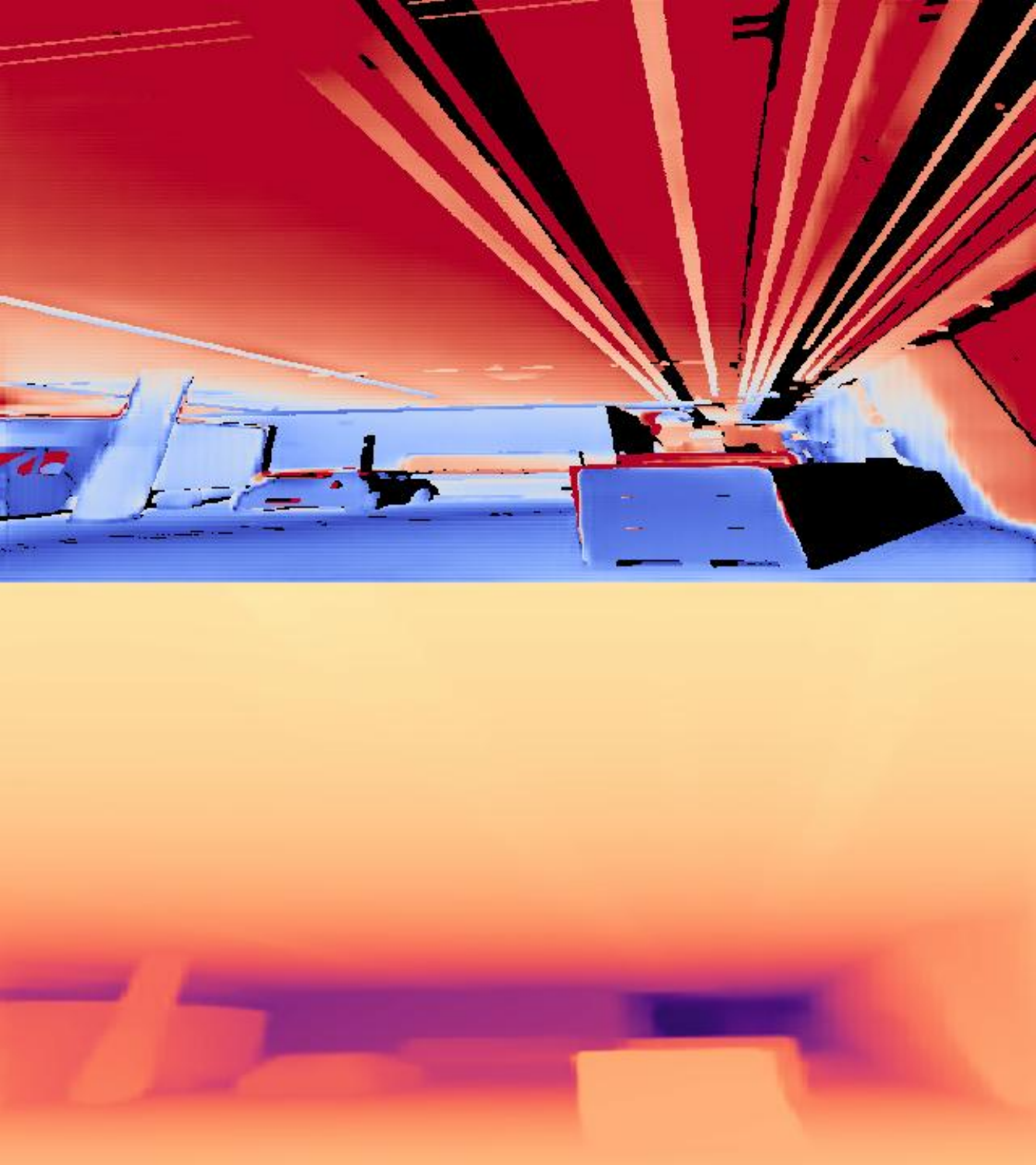}
        & \includegraphics[width=0.14\linewidth]{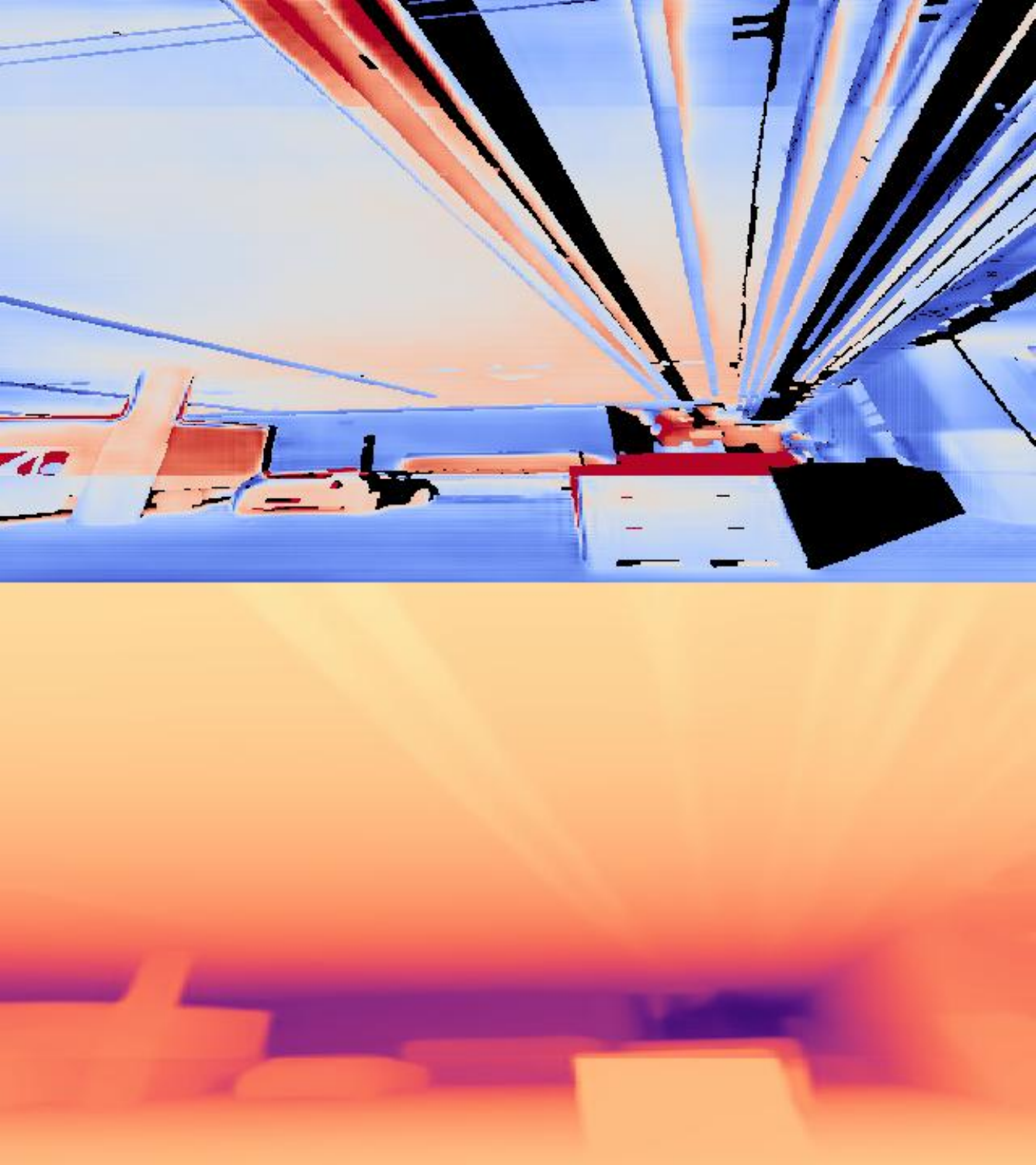}
        & \multirow{2}{*}[25pt]{\includegraphics[width=0.075\linewidth]{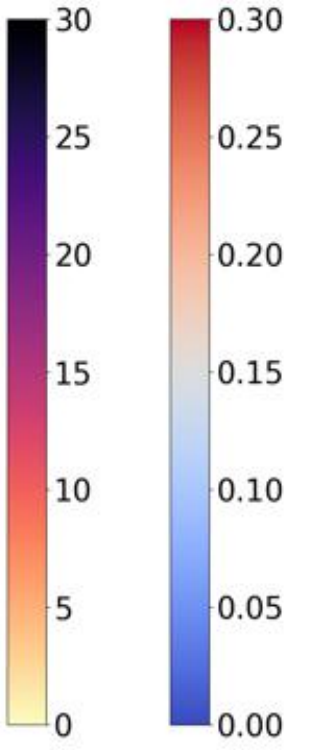}} \vspace{-8pt} \\
        & \includegraphics[width=0.14\linewidth]{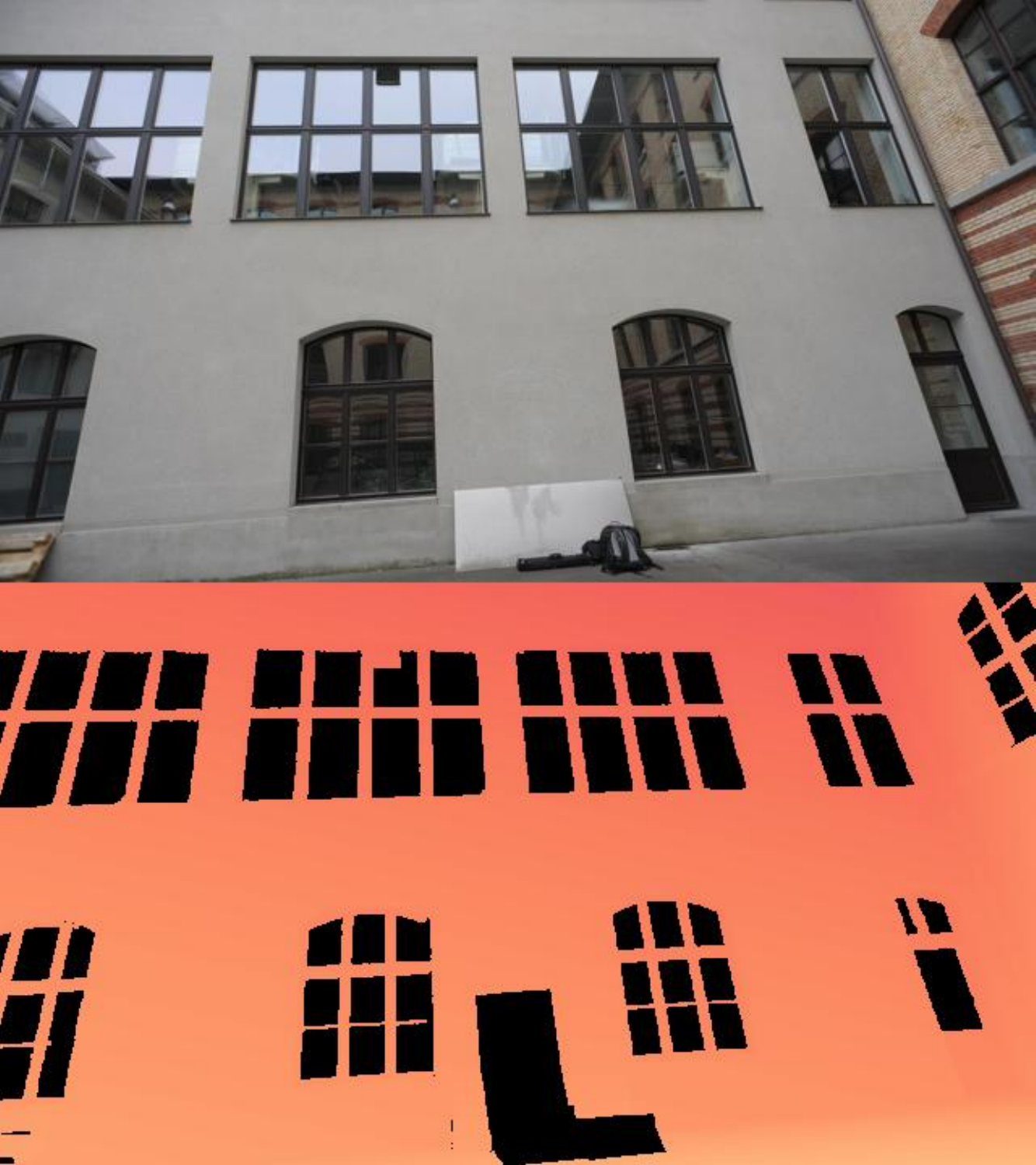}
        & \includegraphics[width=0.14\linewidth]{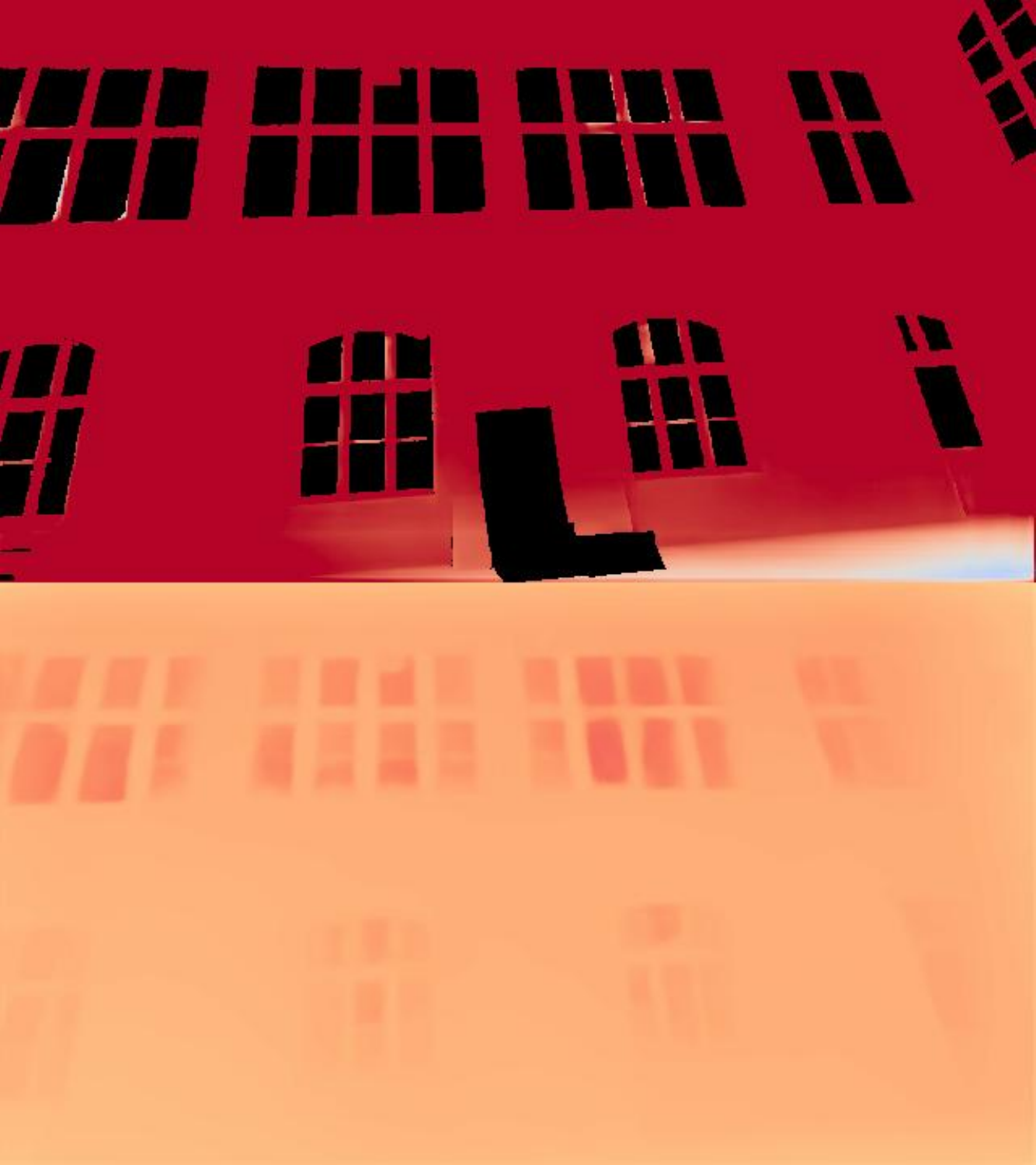}
        & \includegraphics[width=0.14\linewidth]{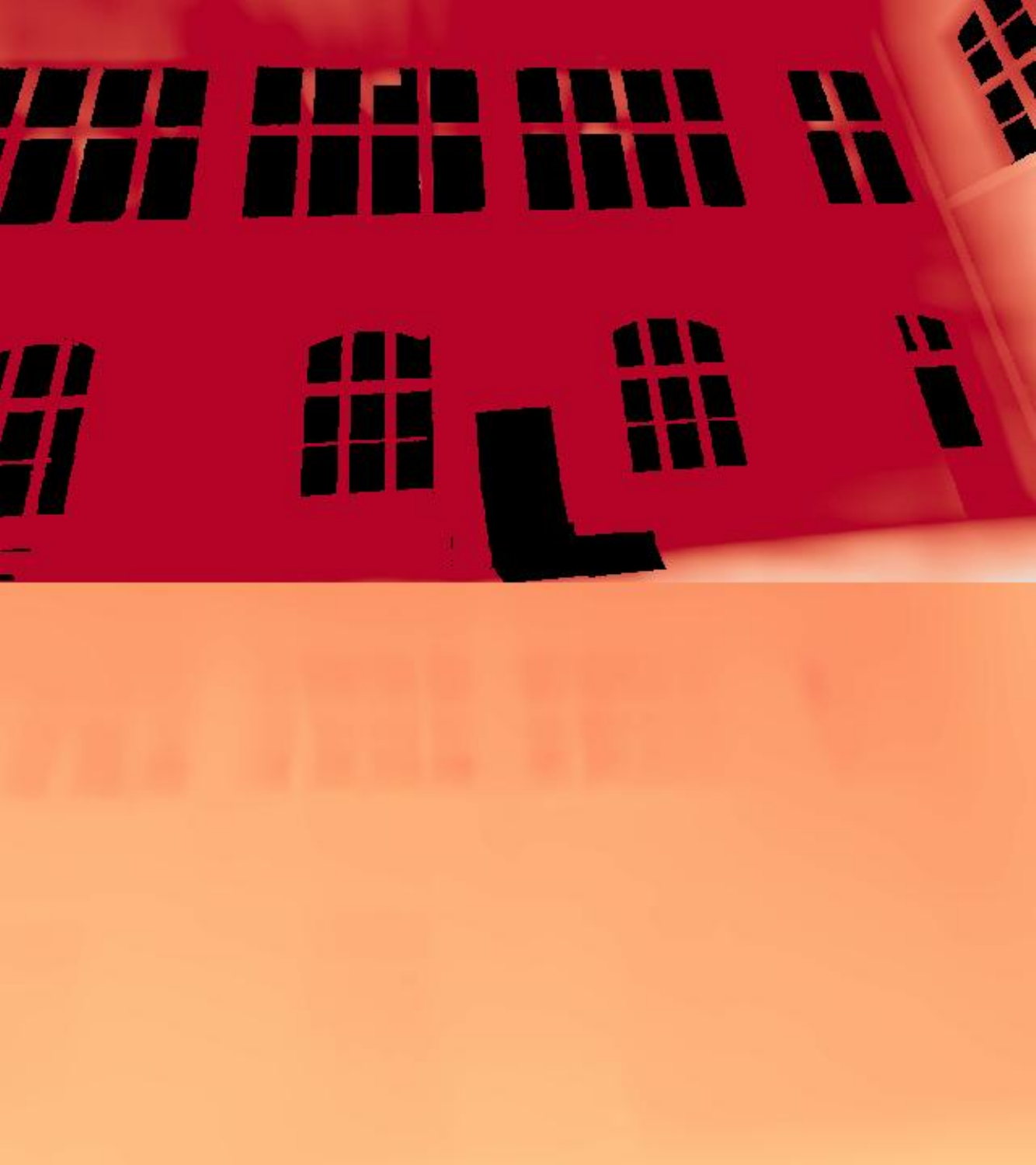}
        & \includegraphics[width=0.14\linewidth]{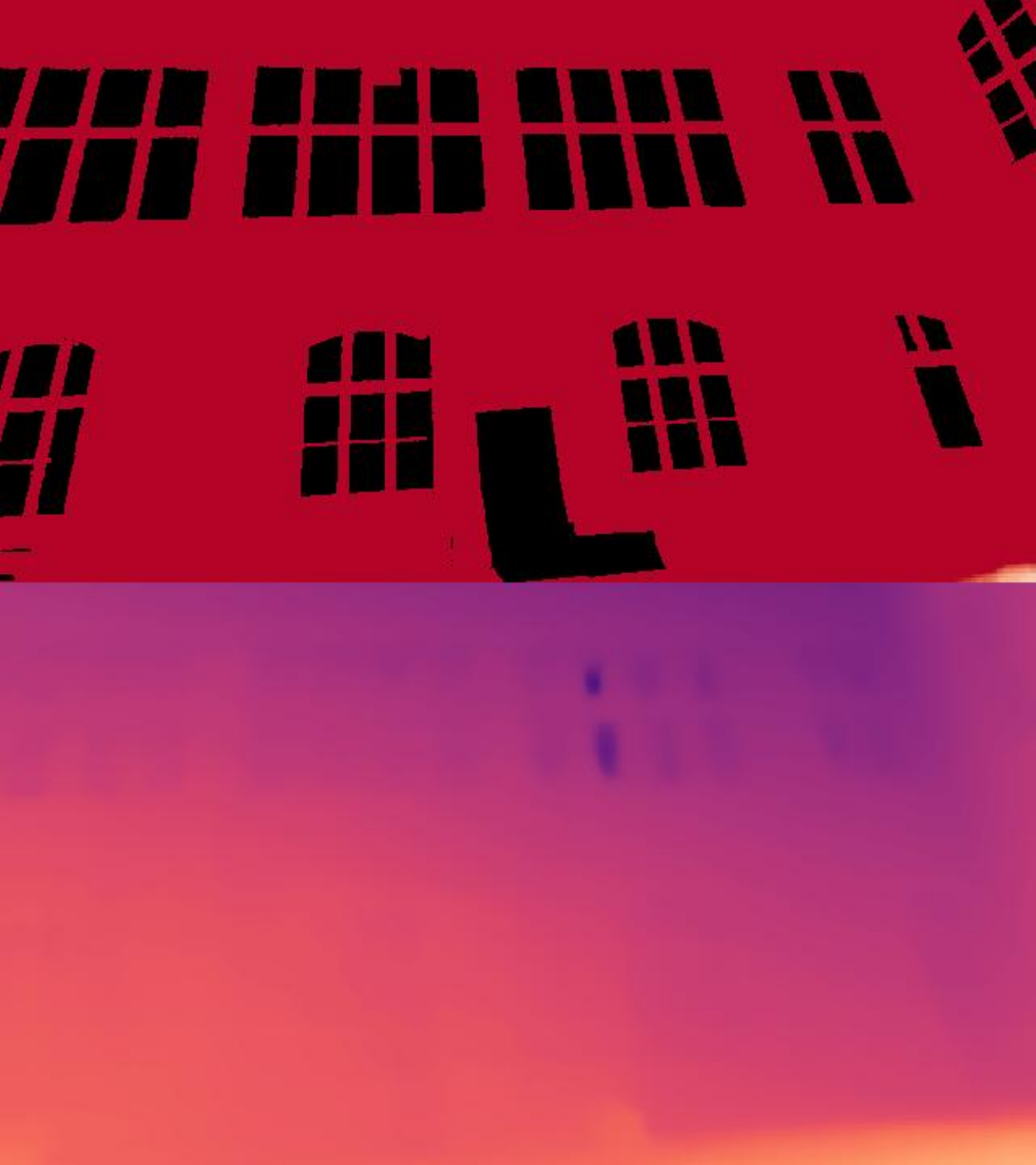}
        & \includegraphics[width=0.14\linewidth]{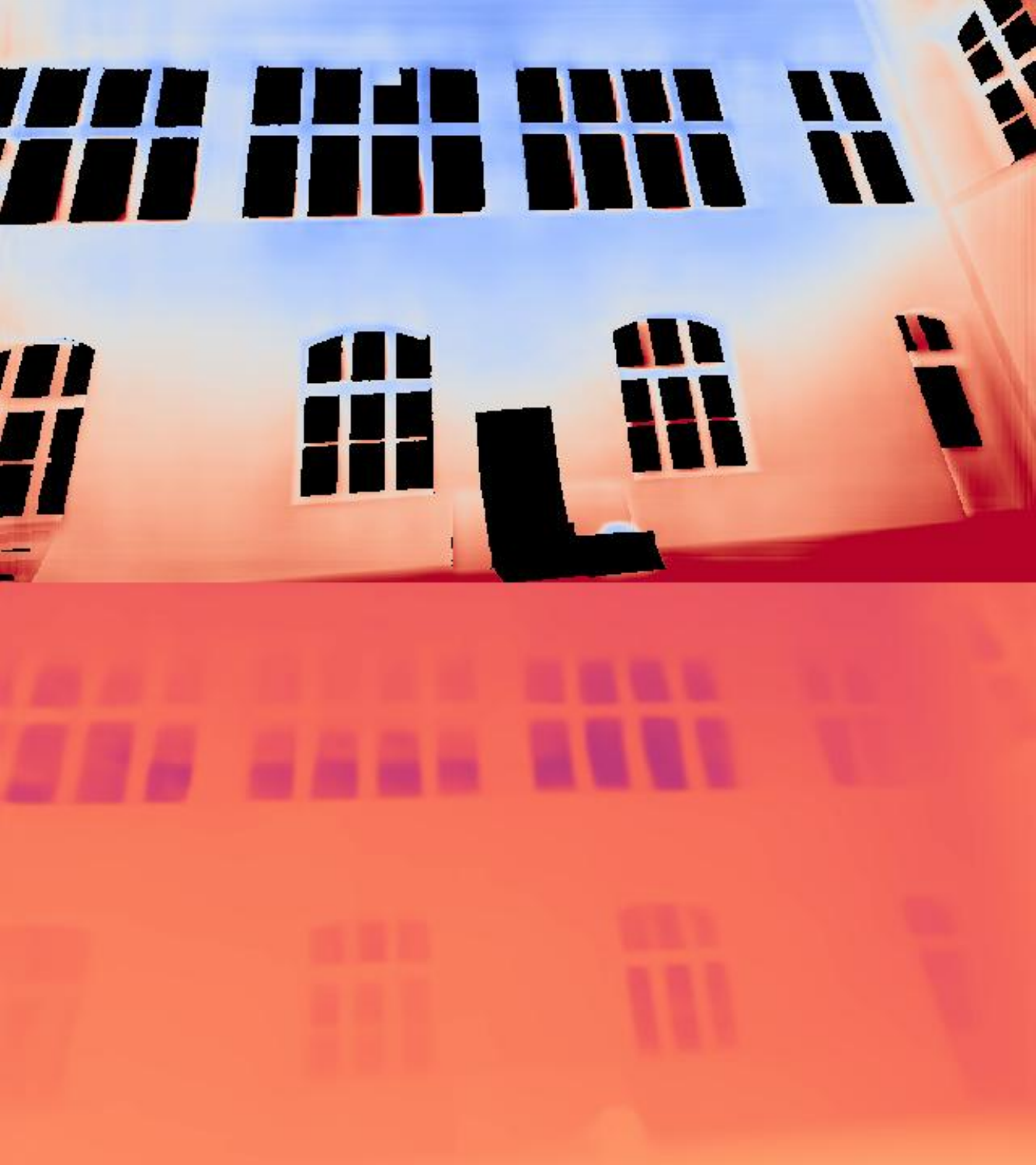}
        & \vspace{-2pt} \\

        & RGB \& GT & ZoeDepth\textsuperscript{\dag}~\cite{bhat2023zoedepth} & ZeroDepth~\cite{guizilini2023zerodepth} & Metric3D~\cite{yin2023metric3d} & \ourmodel & Meters $|$ $\mathrm{A.Rel}$ \\
    \end{tabular}
    \vspace{-8pt}
    \caption{\textbf{Zero-shot qualitative results.} Each pair of consecutive rows corresponds to one test sample. Each odd row shows the input RGB image and the absolute relative error map color-coded with \textit{coolwarm} colormap. Each even row shows GT depth and the predicted depth. The last column represents the specific colormap ranges for depth and error. (\dag): KITTI and NYU in the training set.}
    \label{fig:supp:vis1}
\end{figure*}

\begin{figure*}[t]
    \renewcommand{\arraystretch}{2}
    \centering
    \small
    \begin{tabular}{cc|cccc|c}
        \multirow{2}{*}[6pt]{\rotatebox[origin=c]{90}{DDAD}}
        & \includegraphics[width=0.14\linewidth]{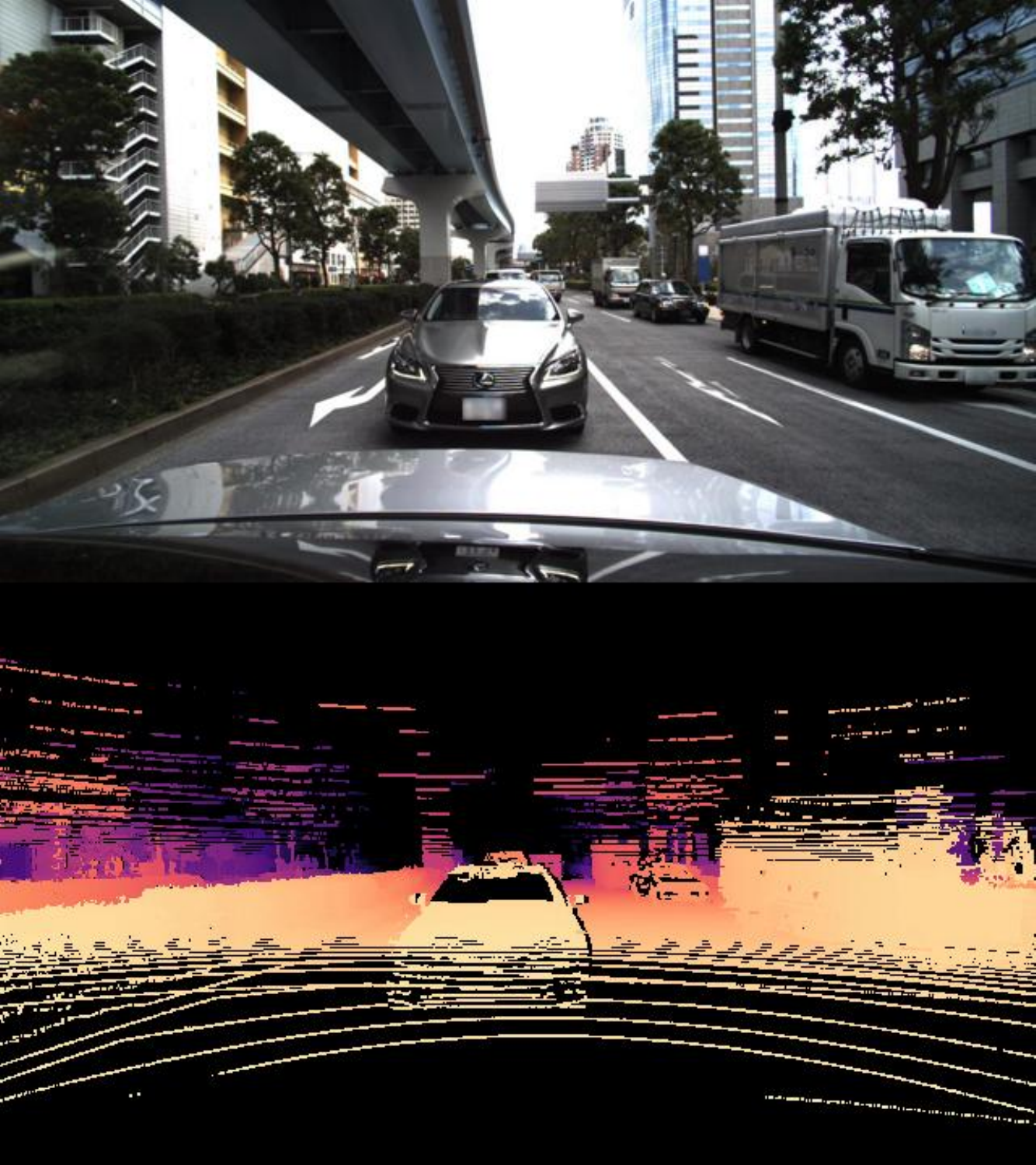}
        & \includegraphics[width=0.14\linewidth]{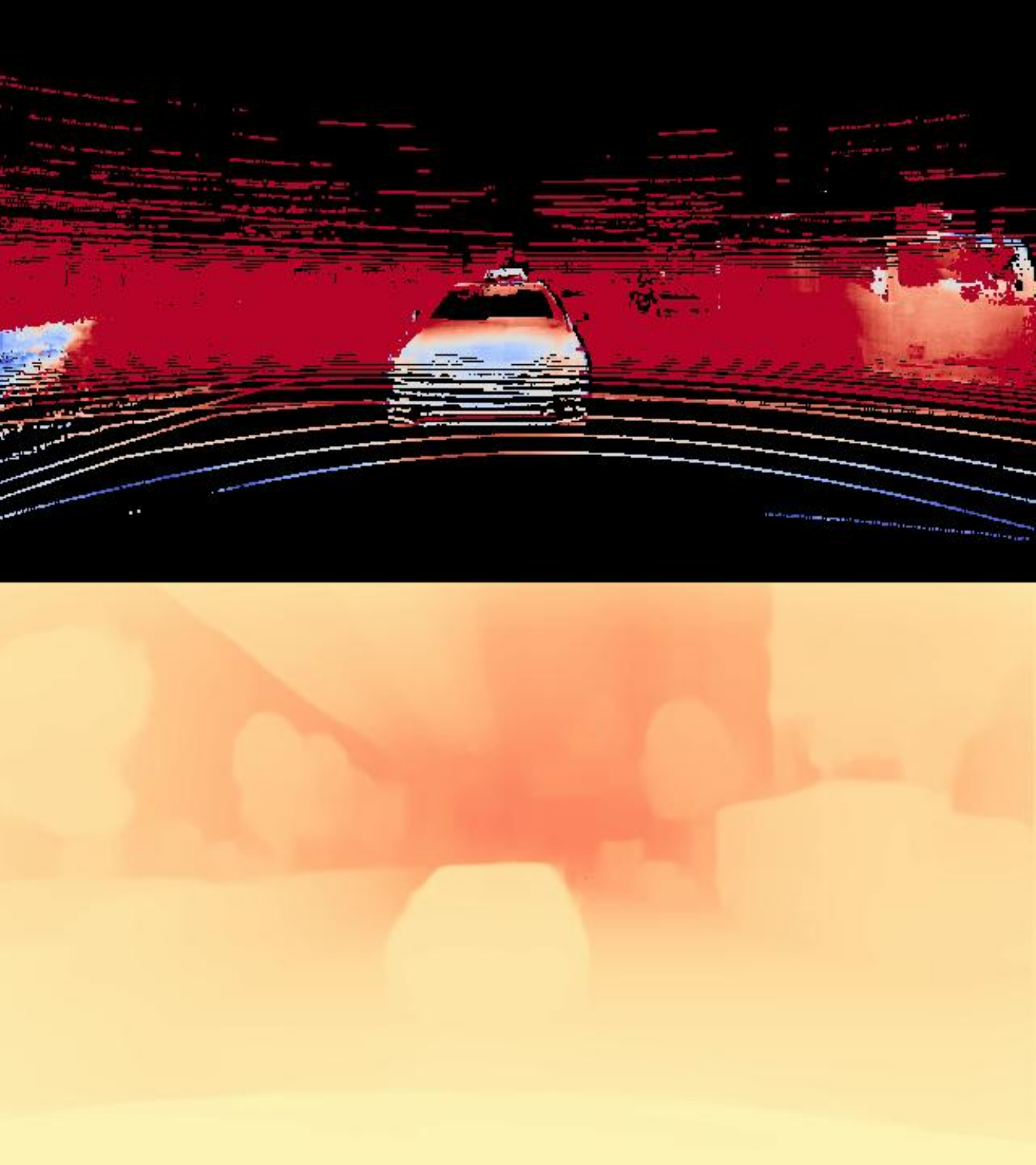}
        & \includegraphics[width=0.14\linewidth]{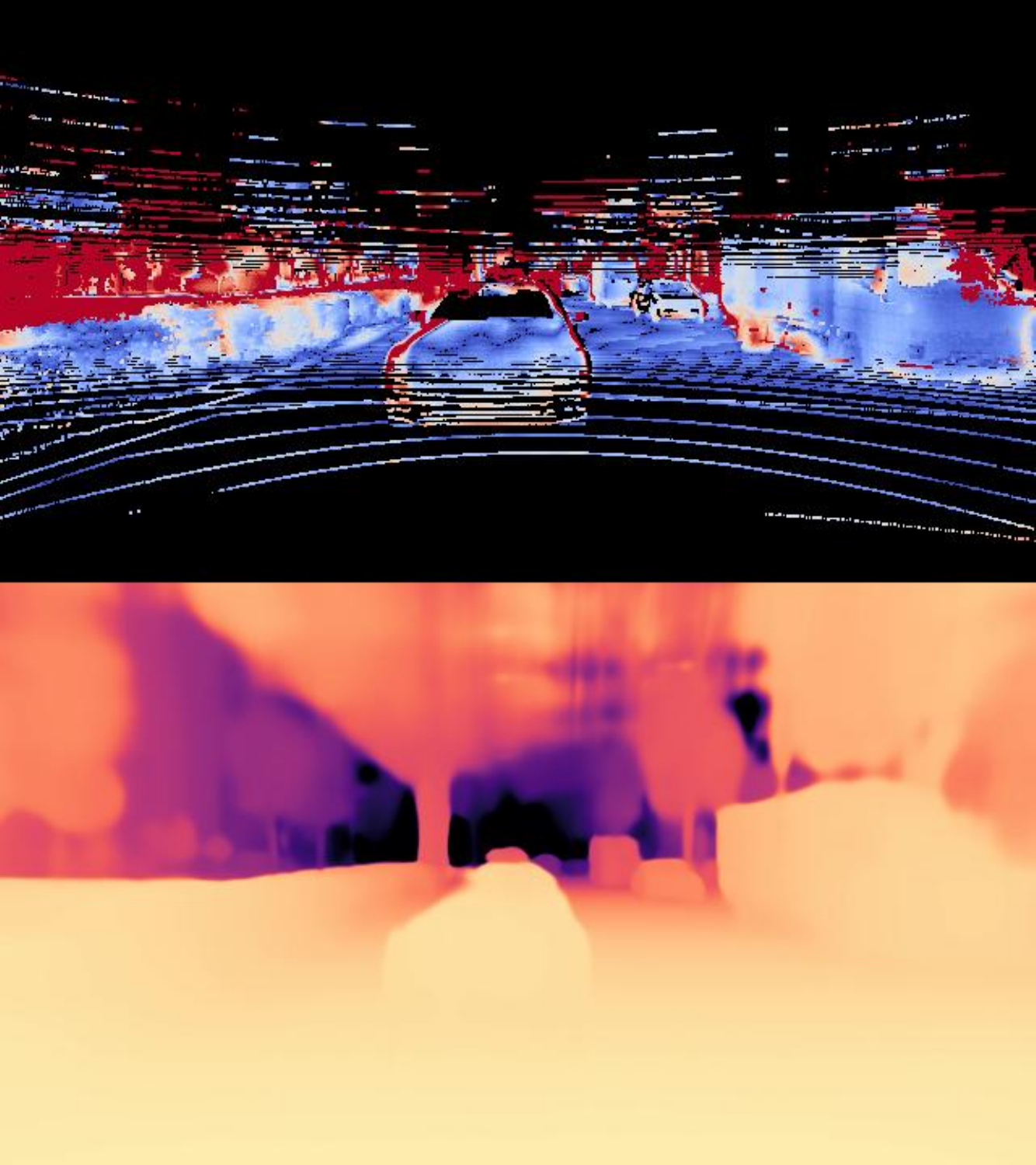}
        & \includegraphics[width=0.14\linewidth]{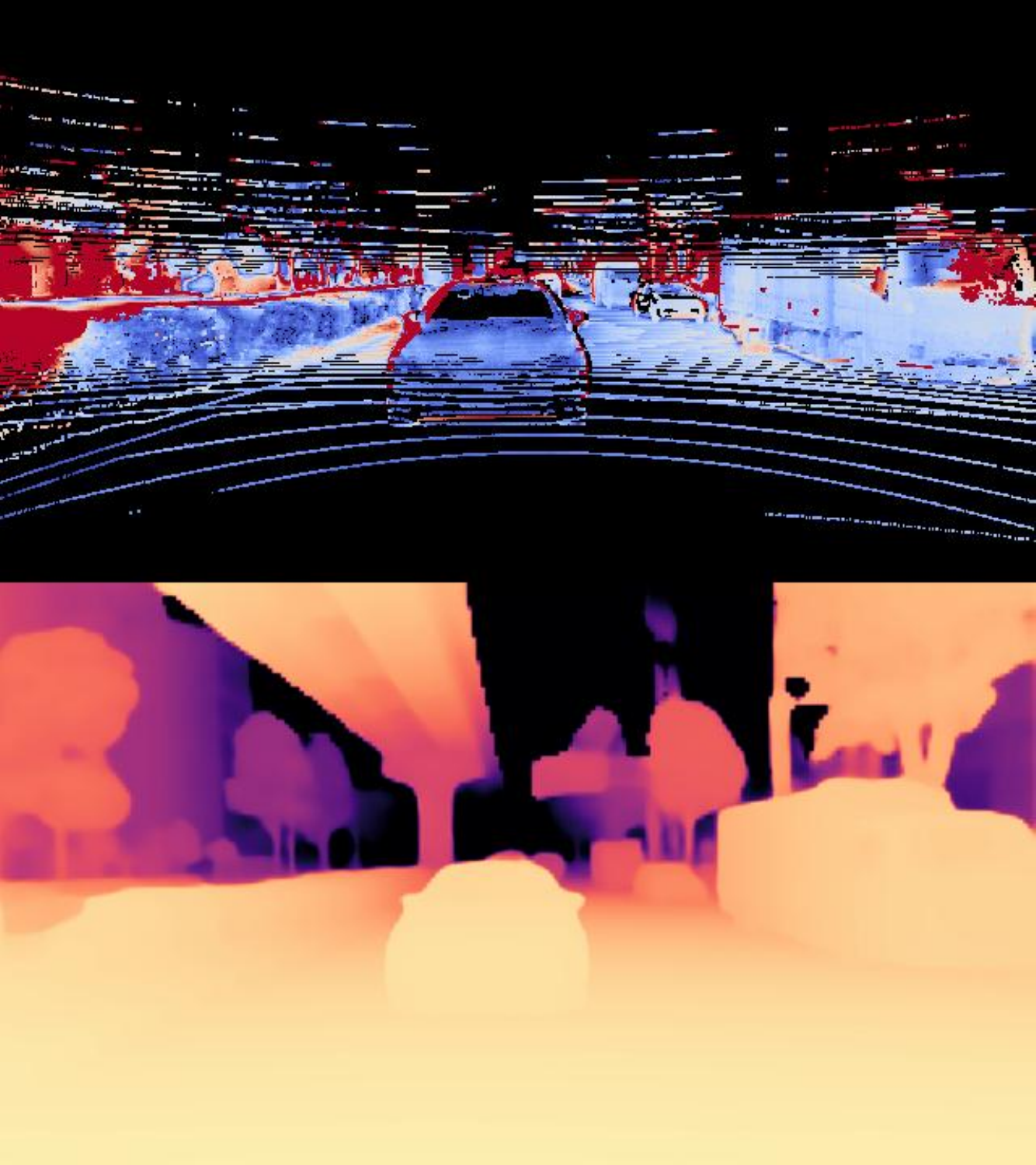}
        & \includegraphics[width=0.14\linewidth]{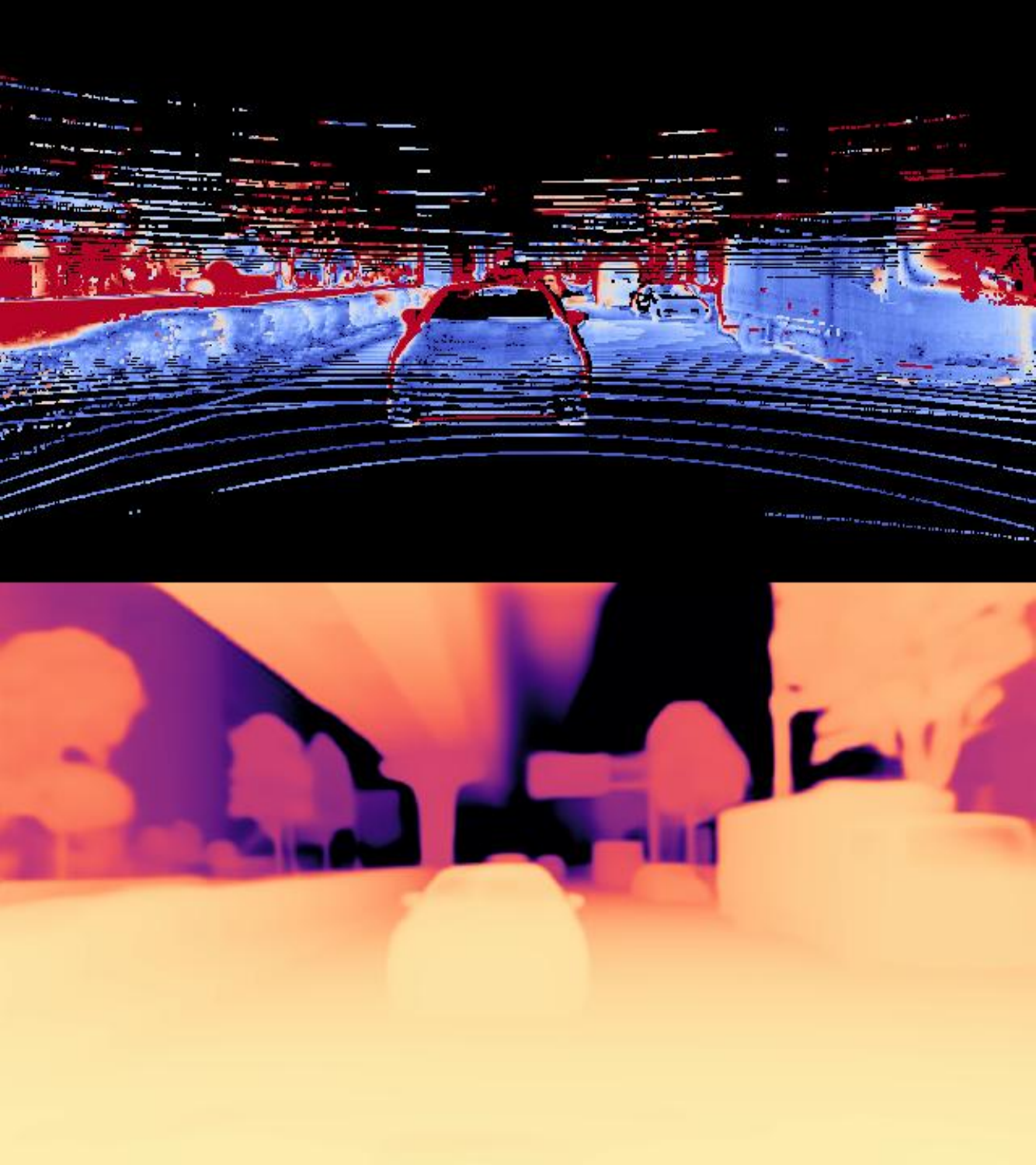}
        & \multirow{2}{*}[25pt]{\includegraphics[width=0.075\linewidth]{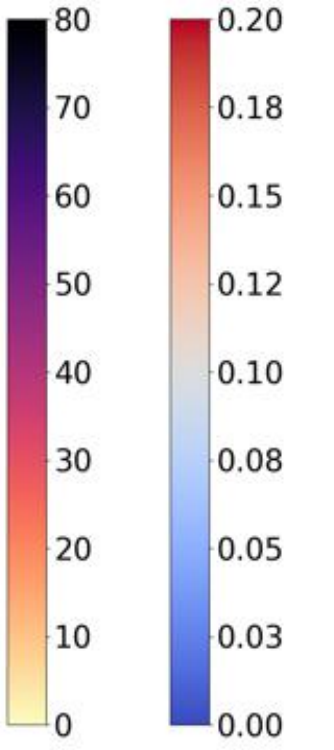}} \vspace{-8pt} \\
        & \includegraphics[width=0.14\linewidth]{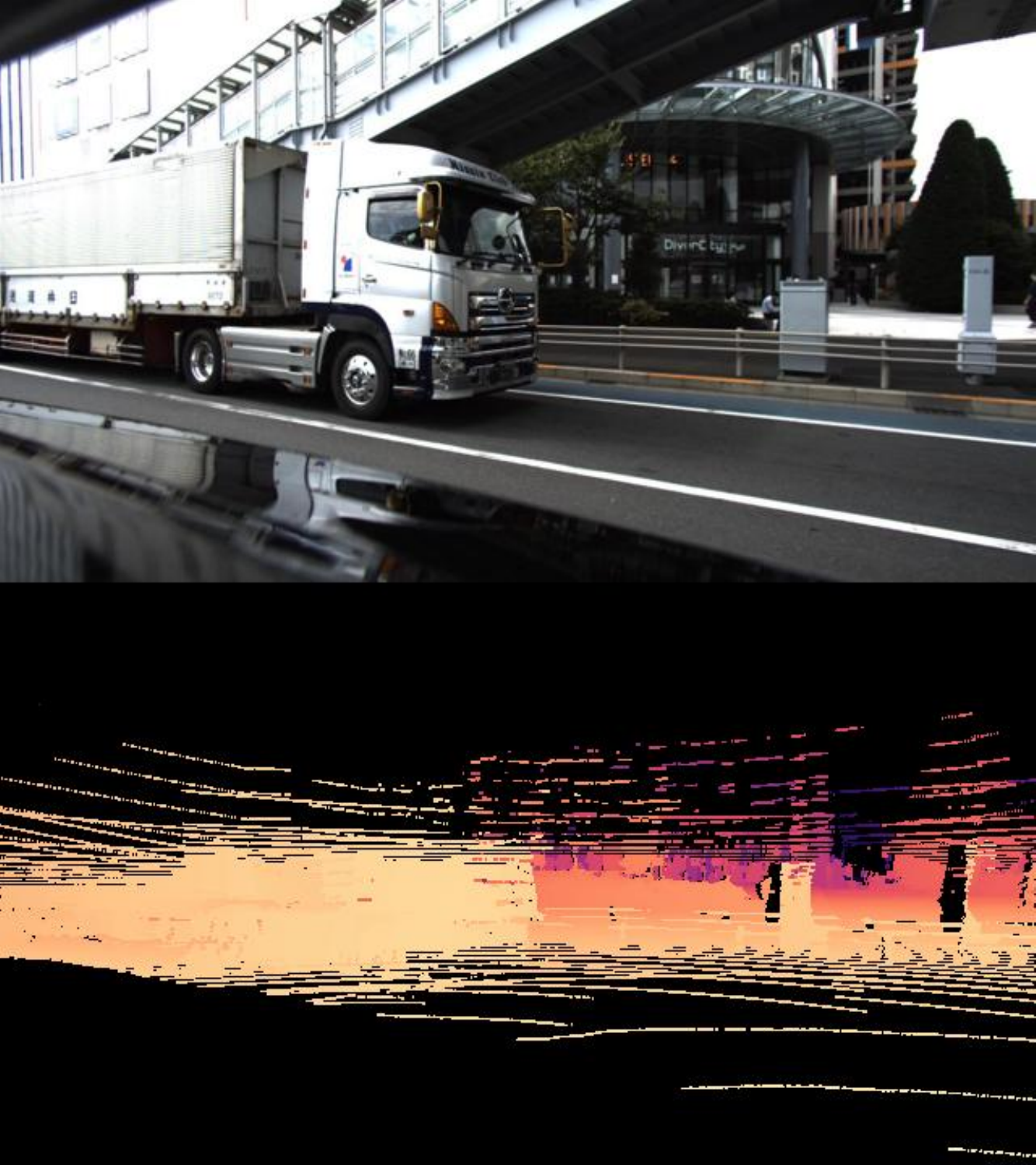}
        & \includegraphics[width=0.14\linewidth]{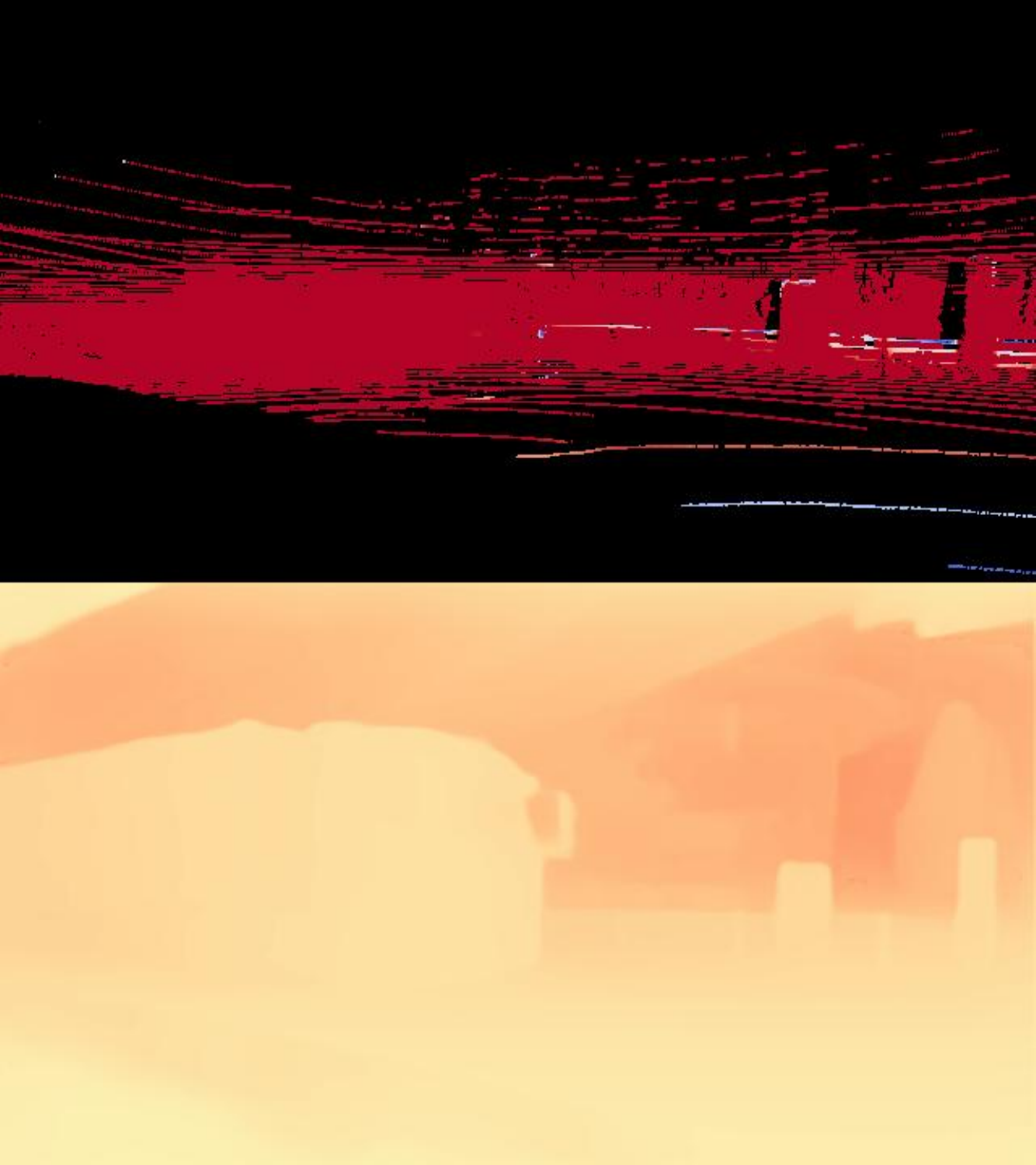}
        & \includegraphics[width=0.14\linewidth]{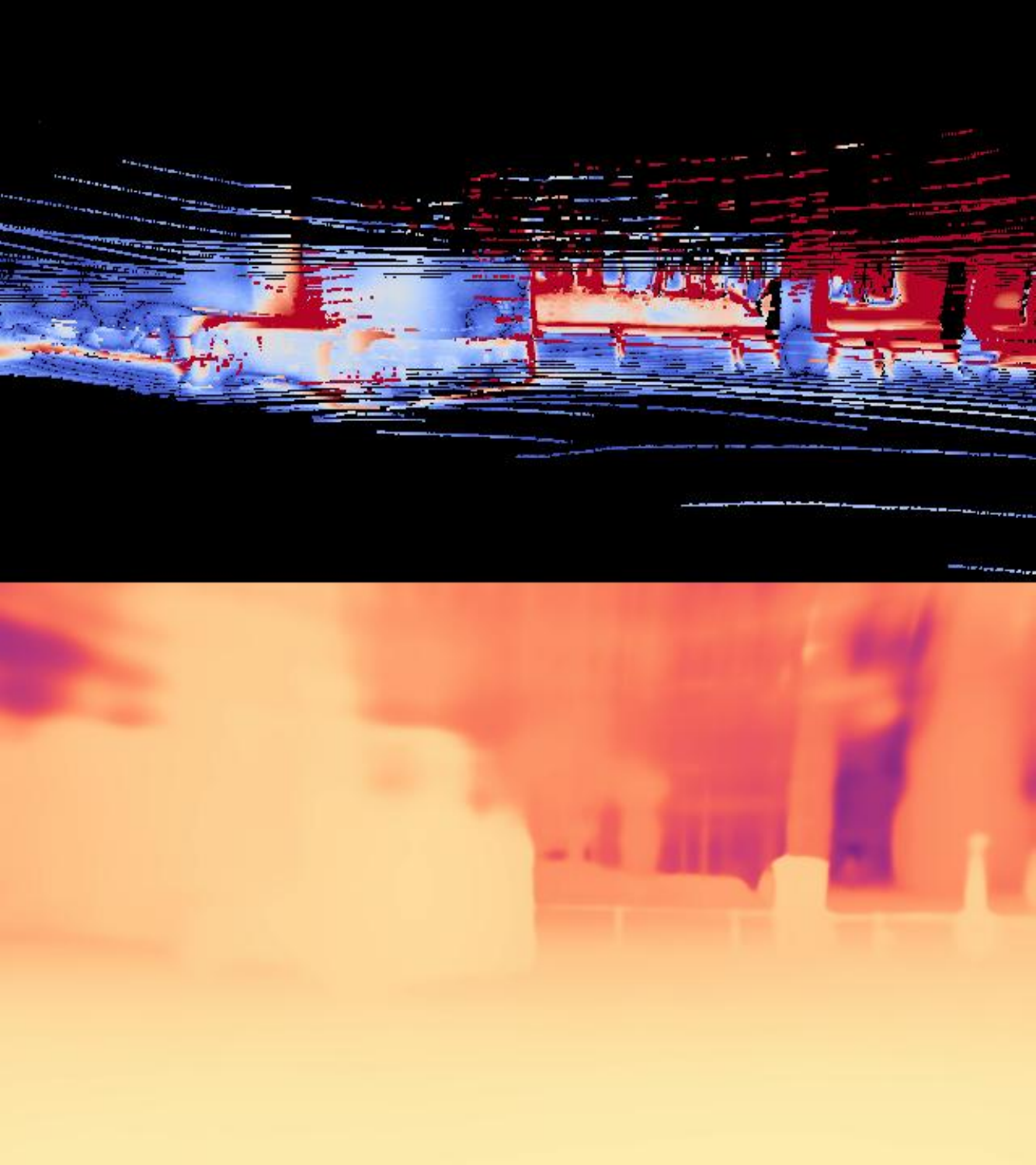}
        & \includegraphics[width=0.14\linewidth]{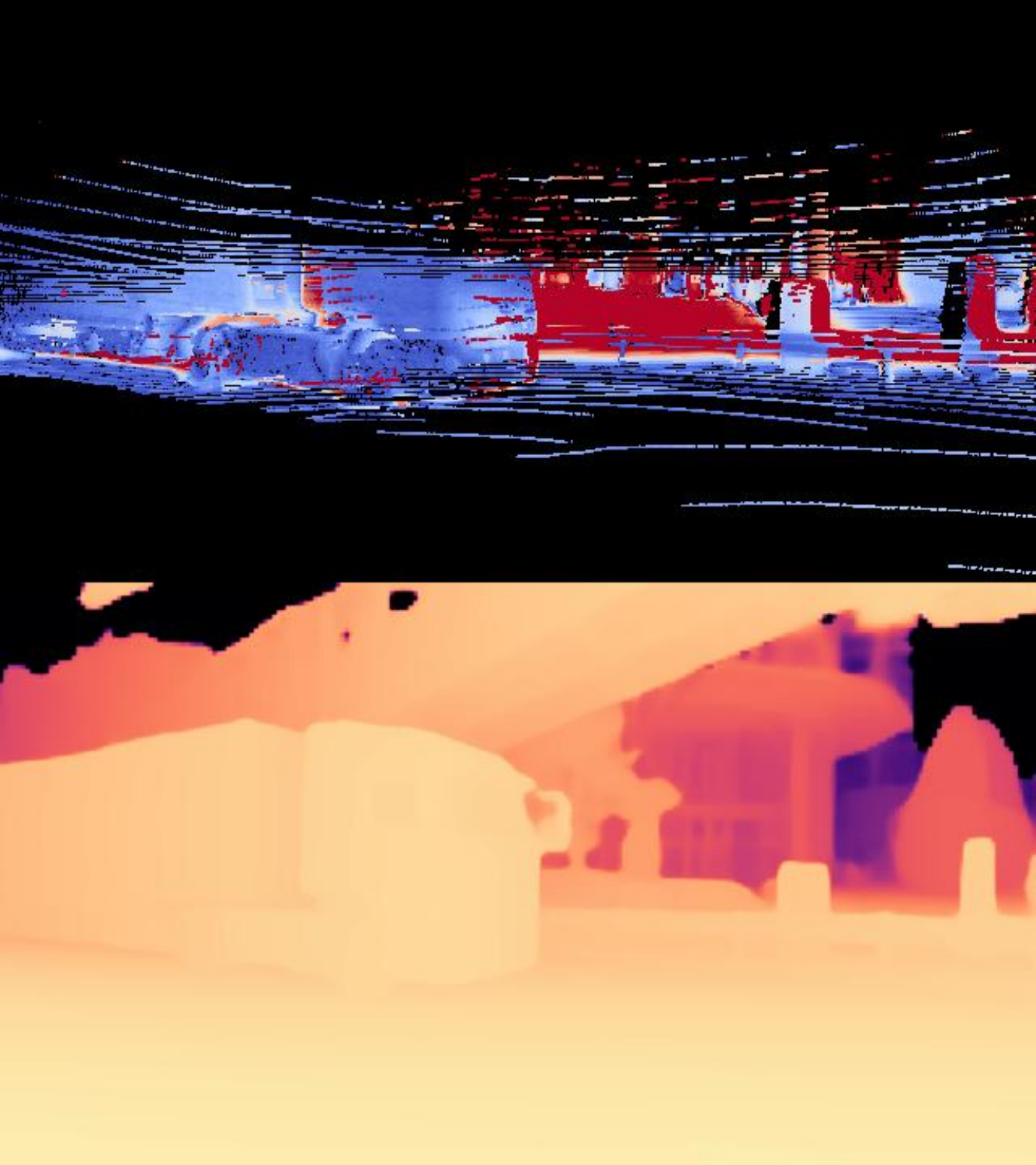}
        & \includegraphics[width=0.14\linewidth]{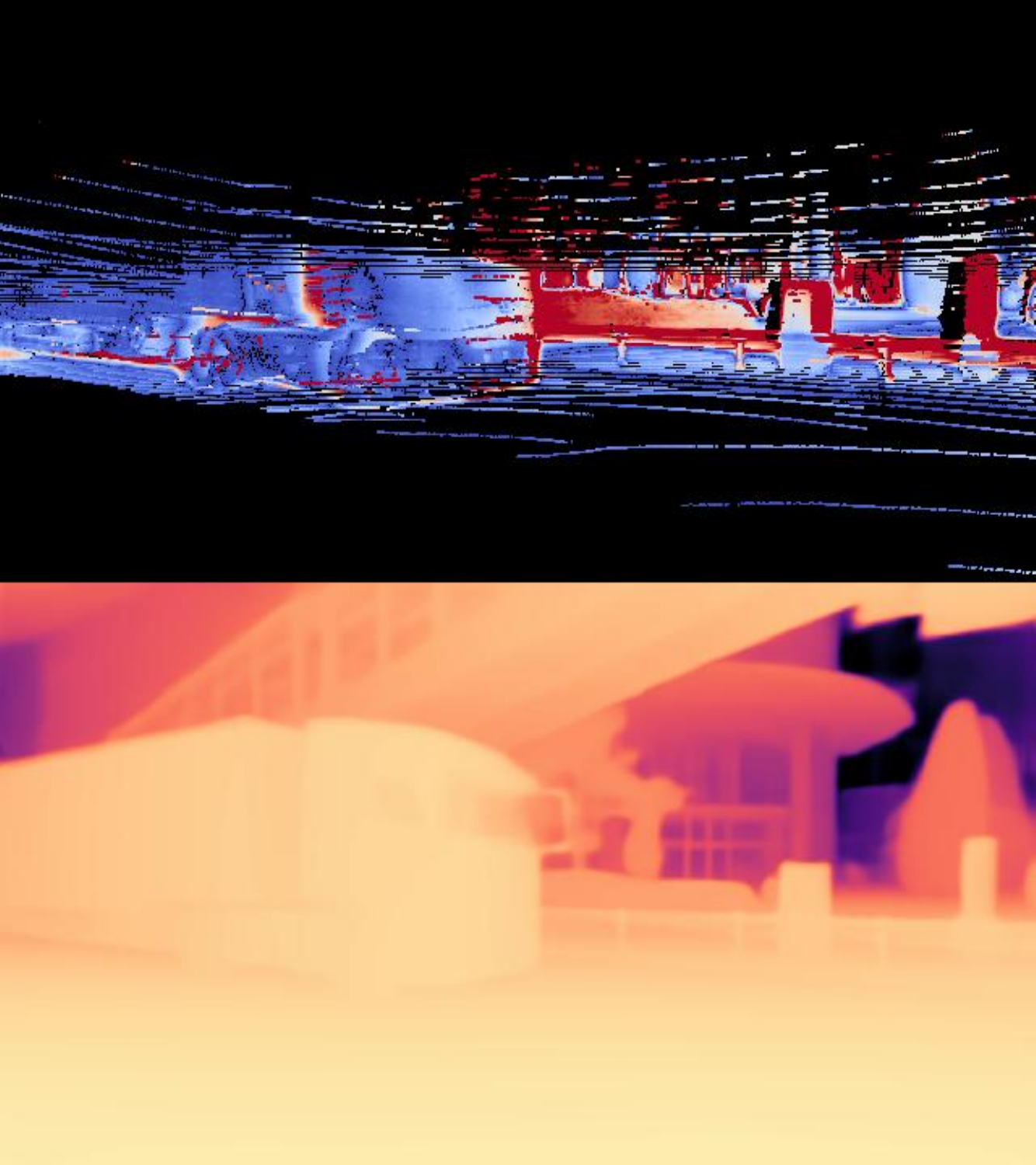}
        & \vspace{-2pt} \\

        \multirow{2}{*}[6pt]{\rotatebox[origin=c]{90}{NuScenes}}
        & \includegraphics[width=0.14\linewidth]{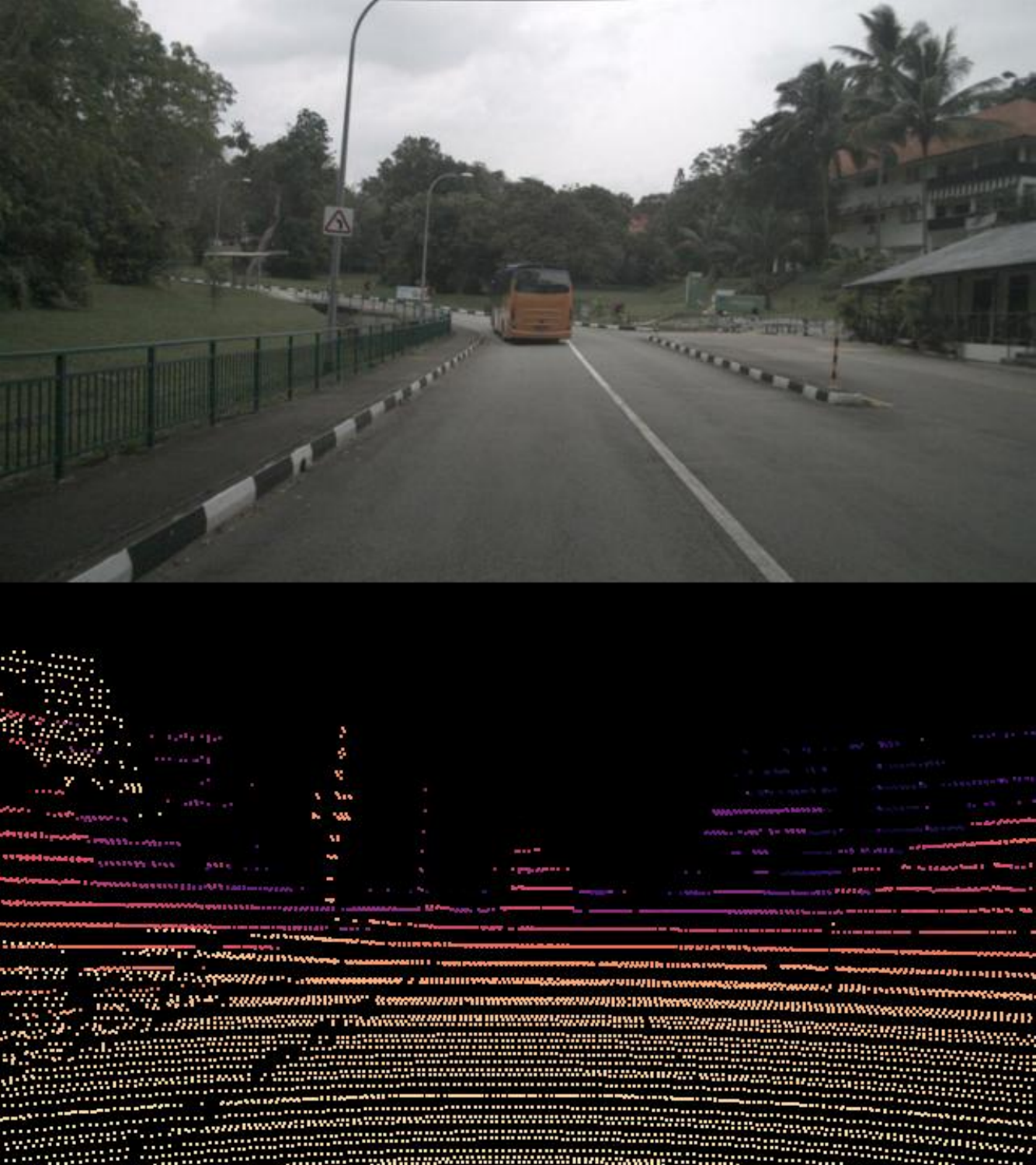}
        & \includegraphics[width=0.14\linewidth]{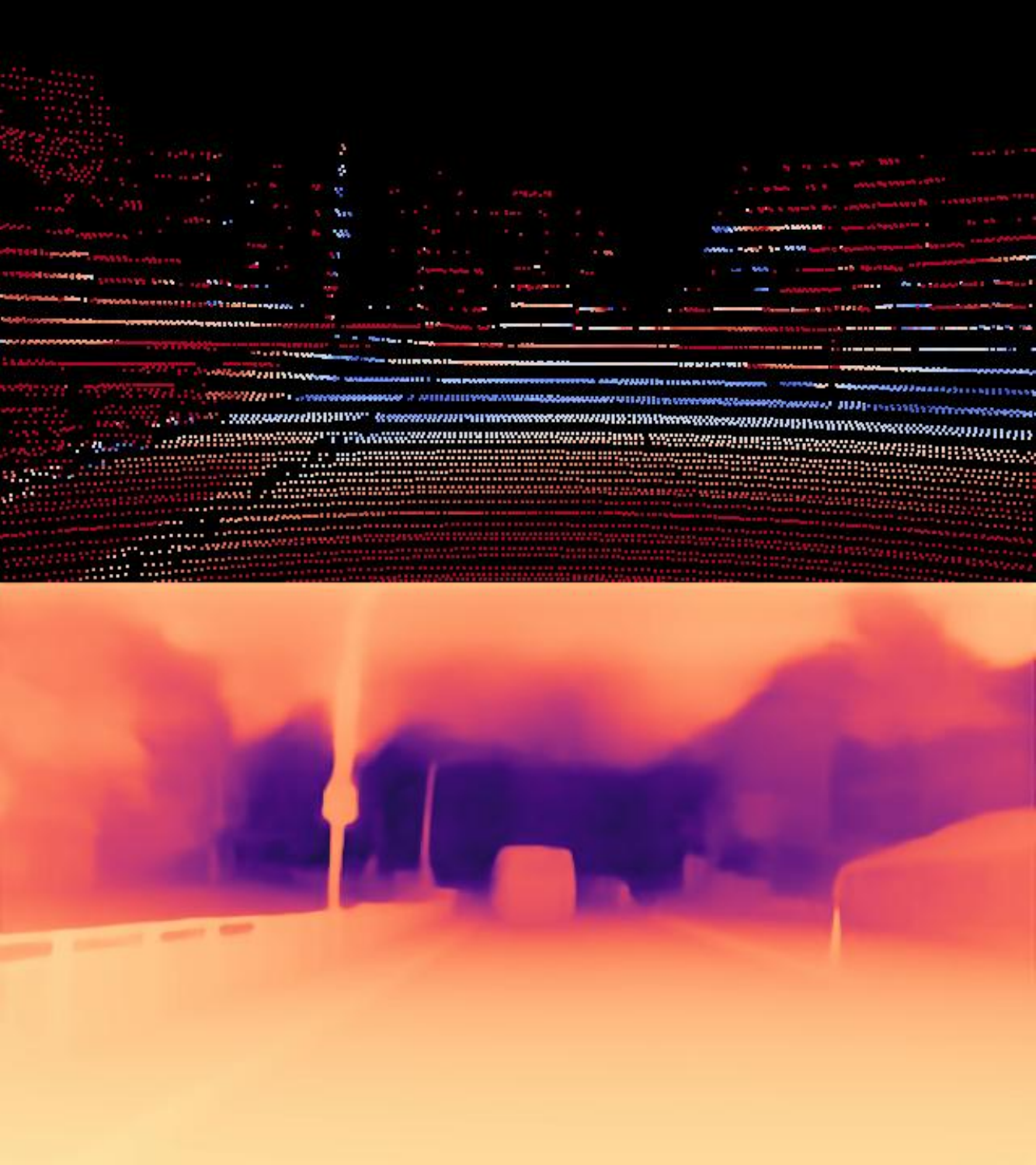}
        & \includegraphics[width=0.14\linewidth]{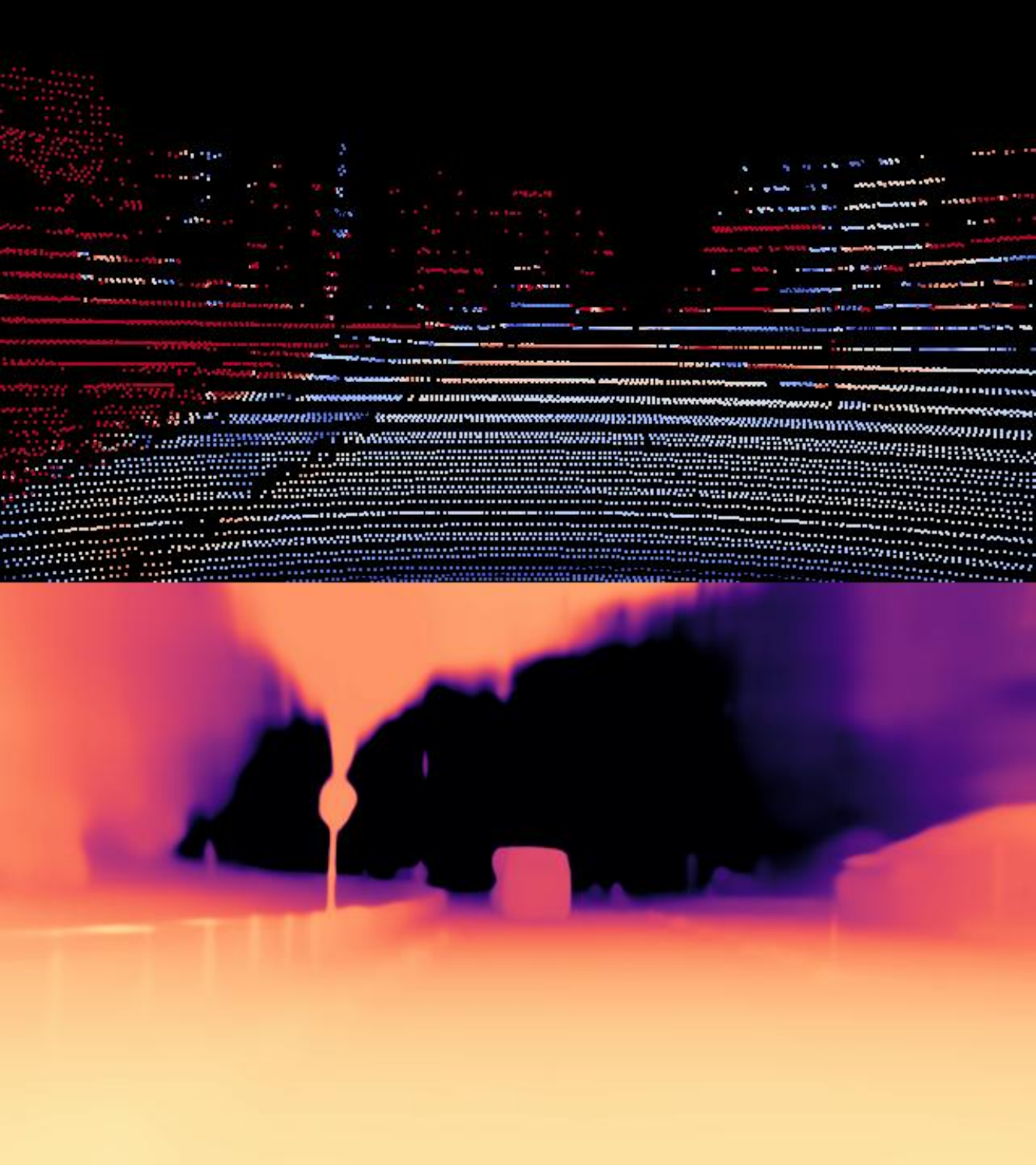}
        & \includegraphics[width=0.14\linewidth]{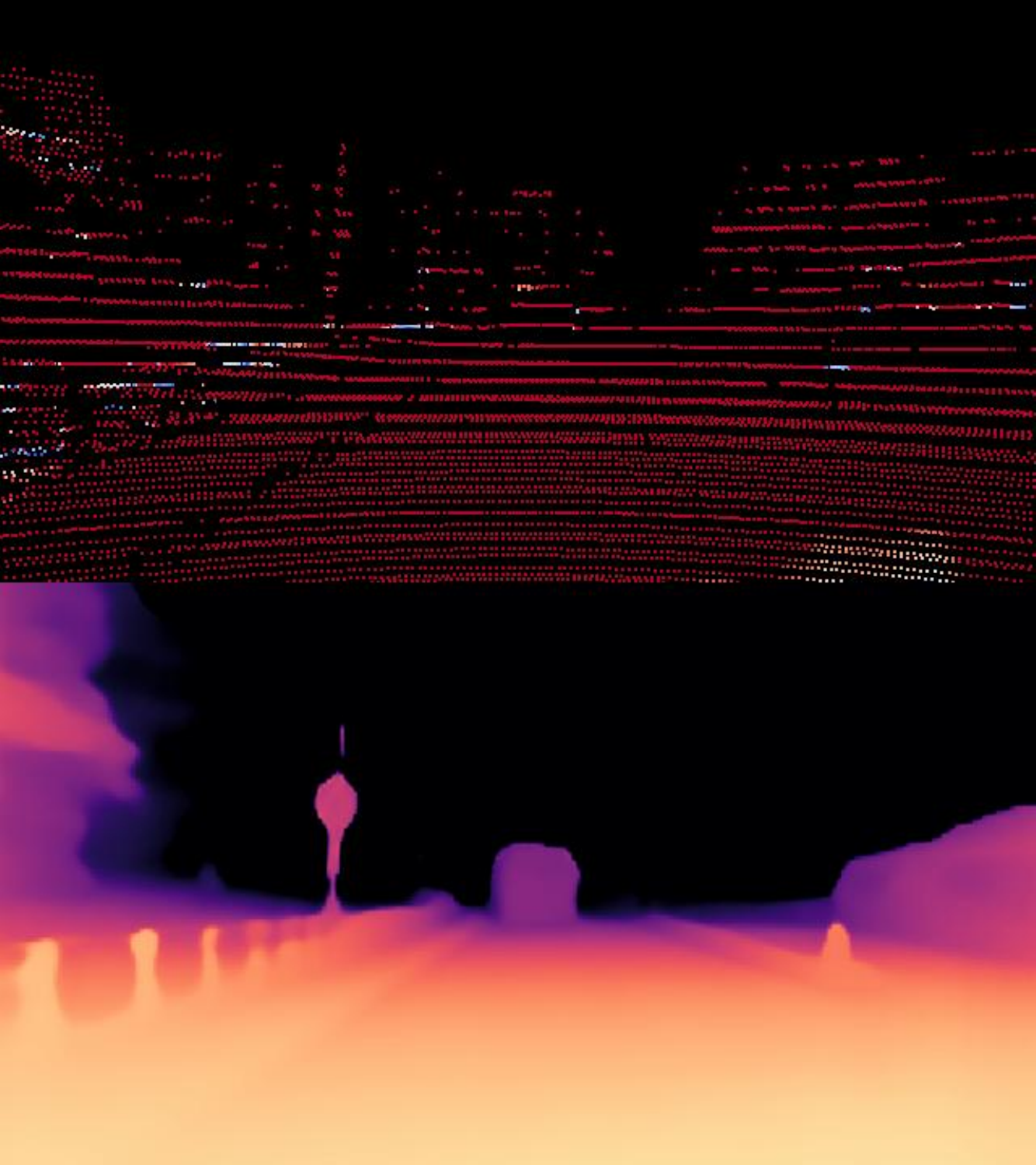}
        & \includegraphics[width=0.14\linewidth]{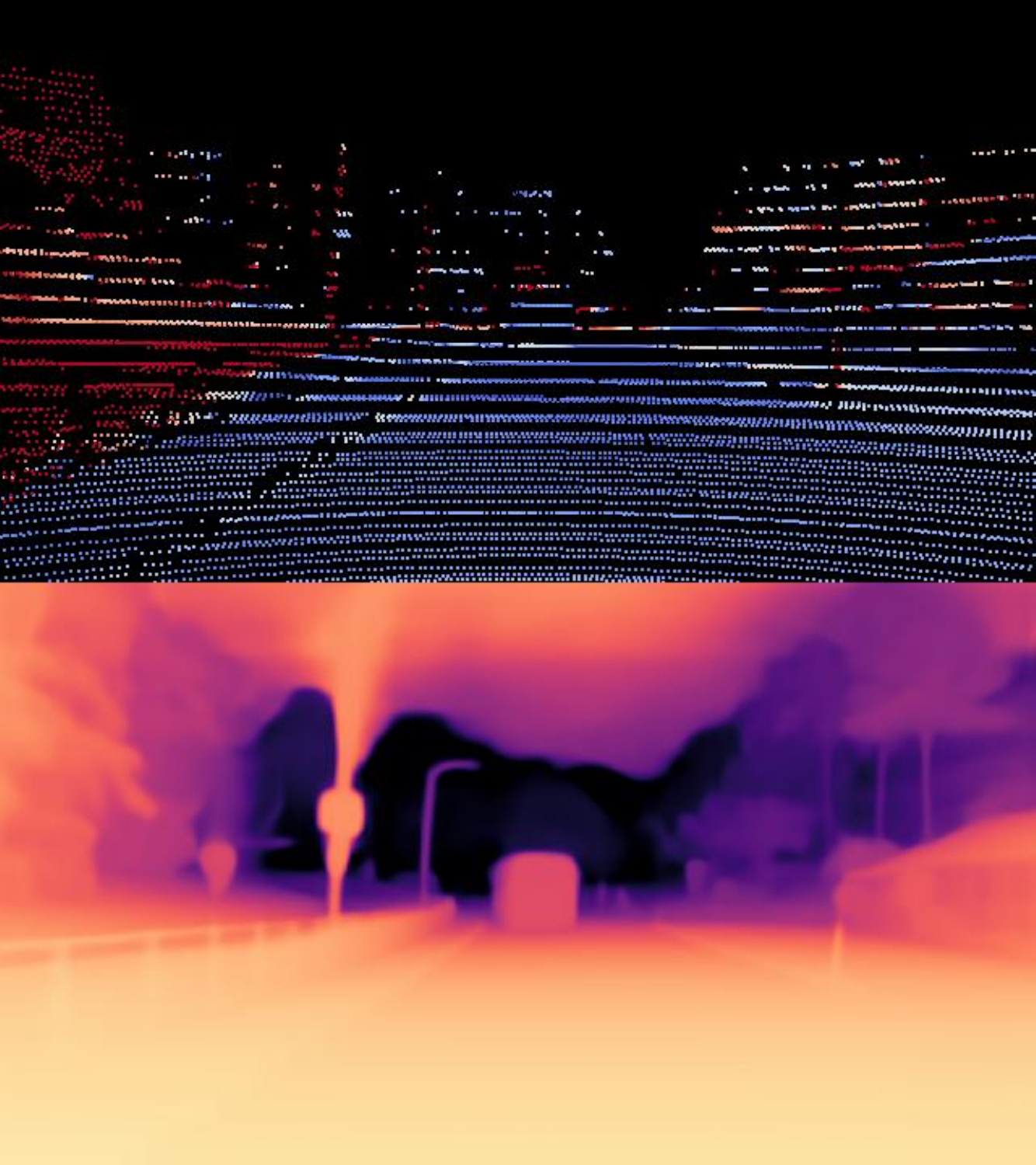}
        & \multirow{2}{*}[25pt]{\includegraphics[width=0.075\linewidth]{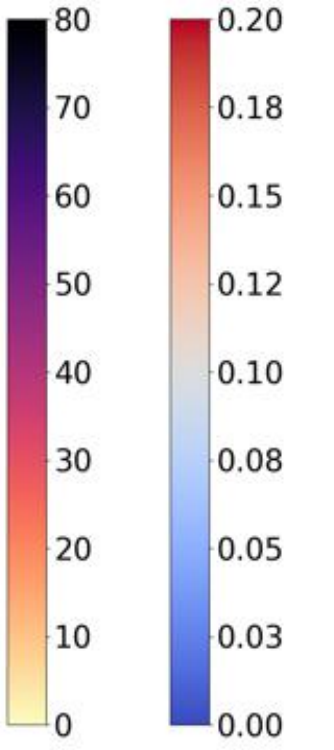}} \vspace{-8pt} \\
        & \includegraphics[width=0.14\linewidth]{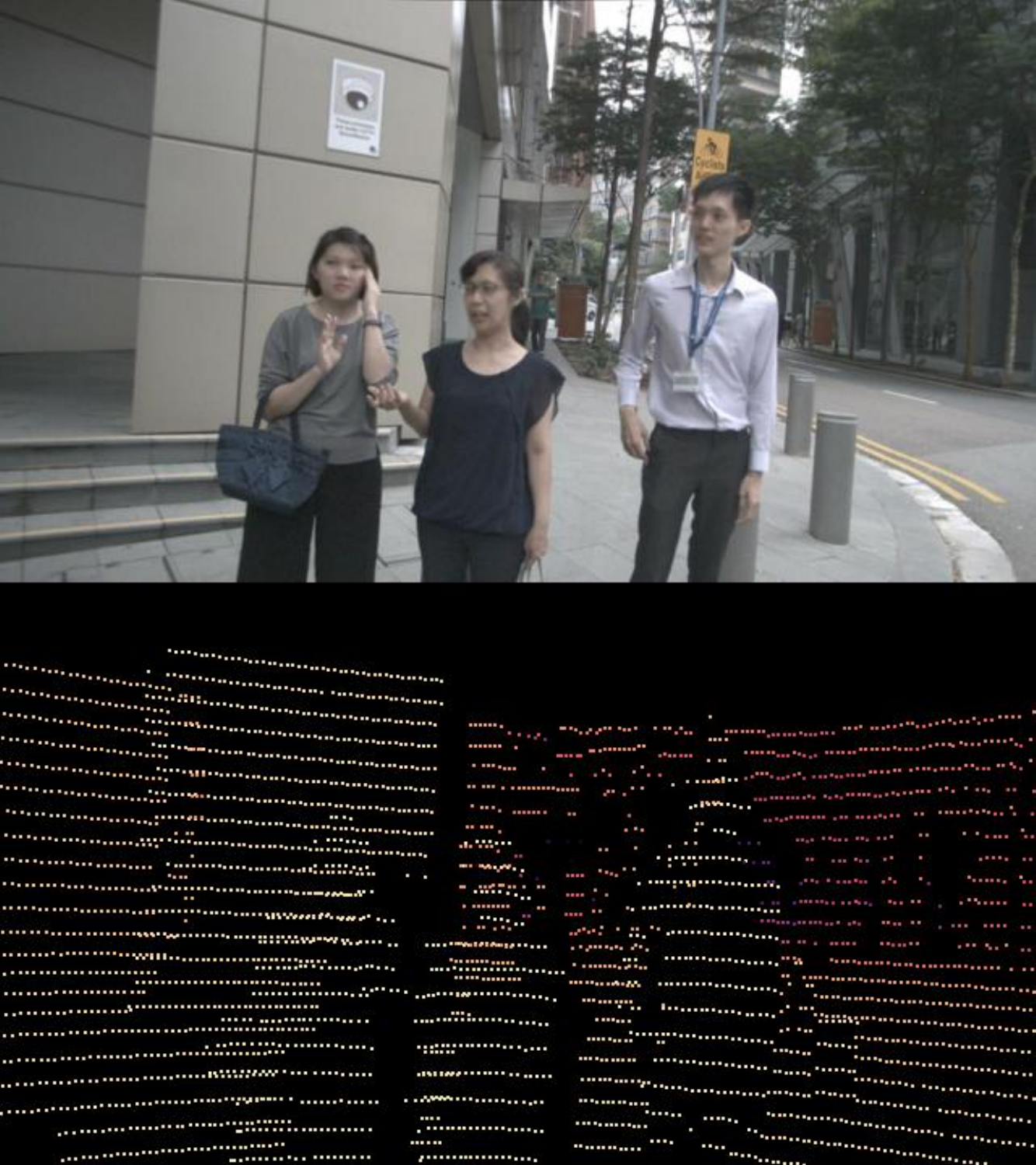}
        & \includegraphics[width=0.14\linewidth]{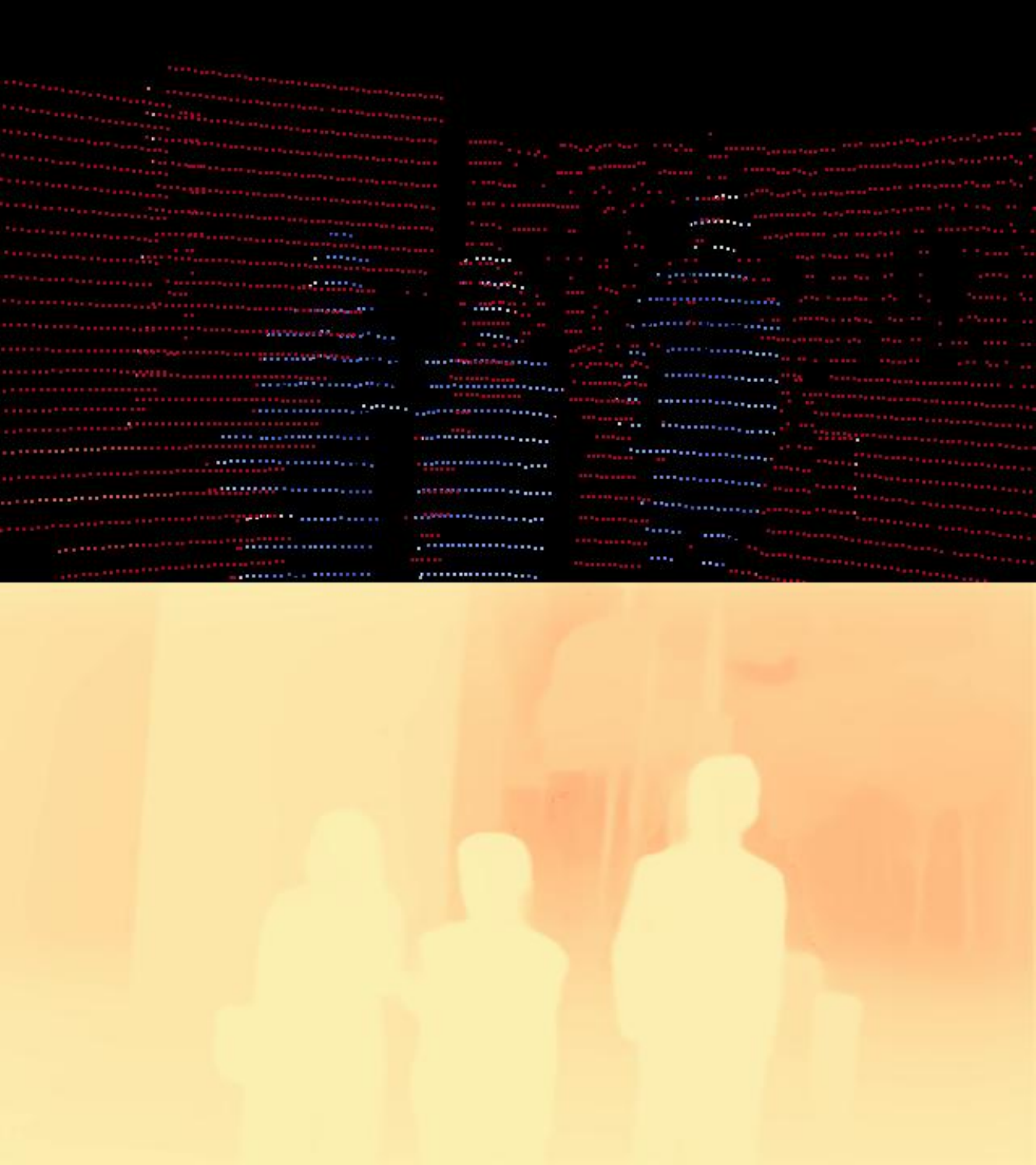}
        & \includegraphics[width=0.14\linewidth]{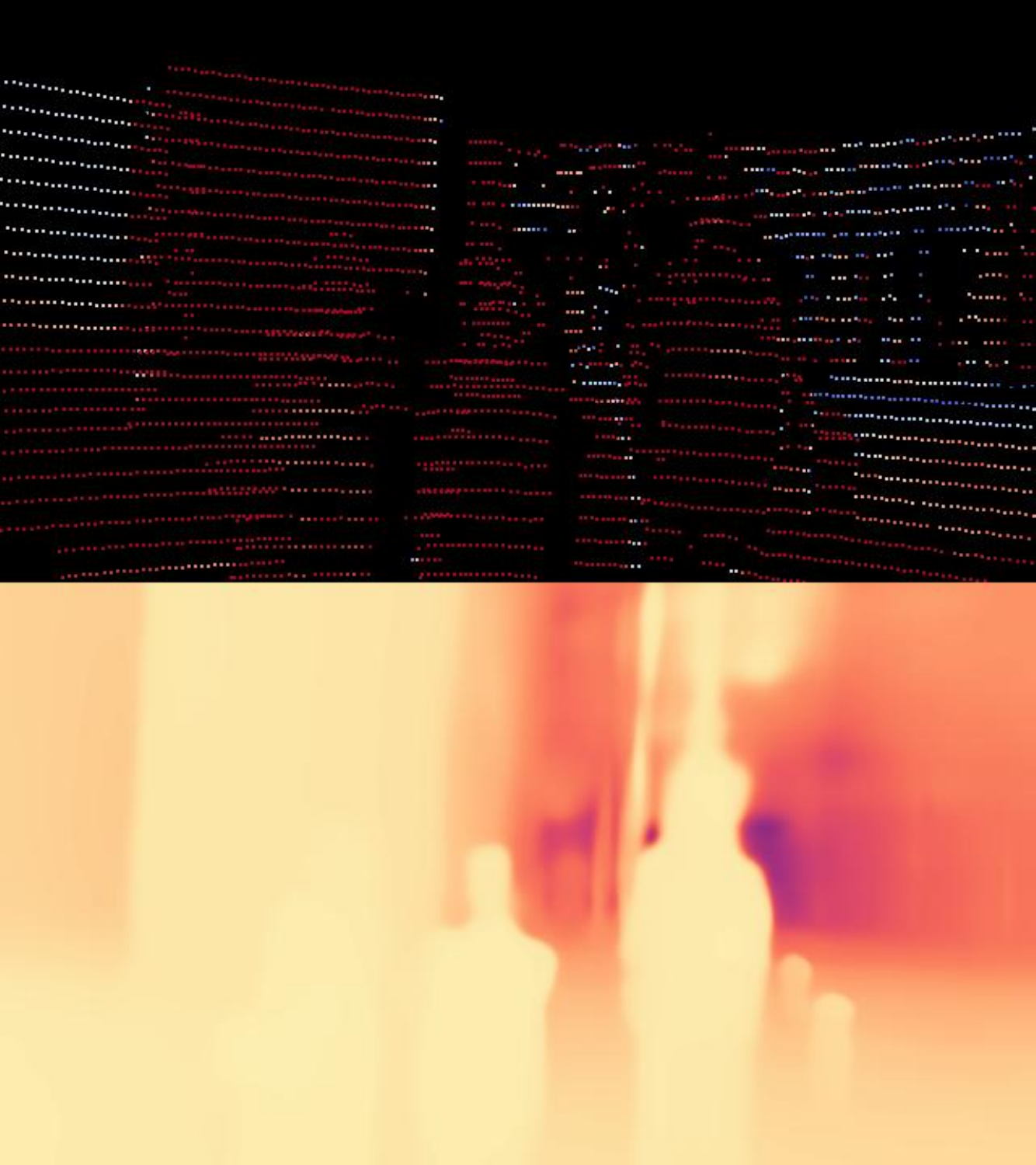}
        & \includegraphics[width=0.14\linewidth]{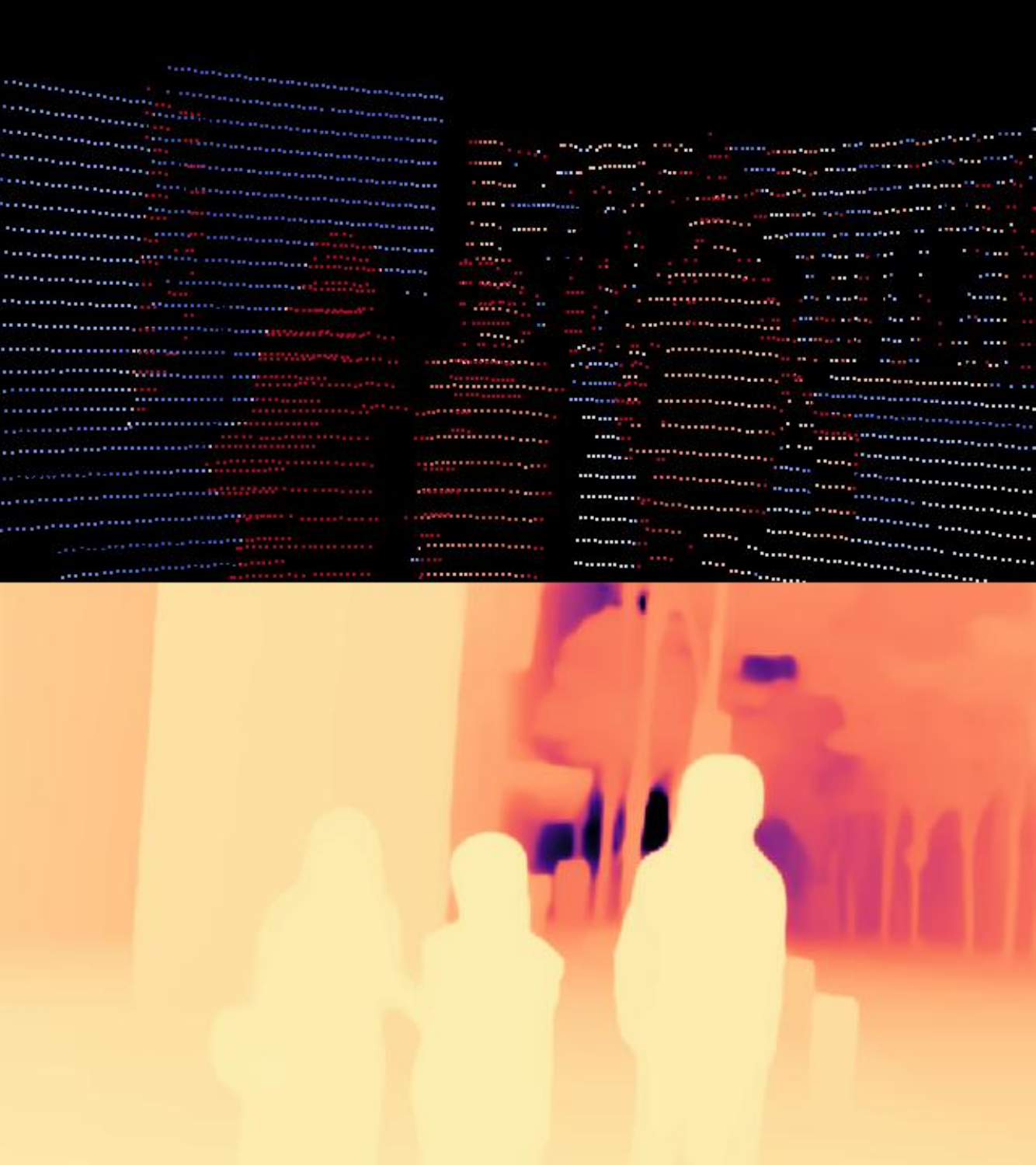}
        & \includegraphics[width=0.14\linewidth]{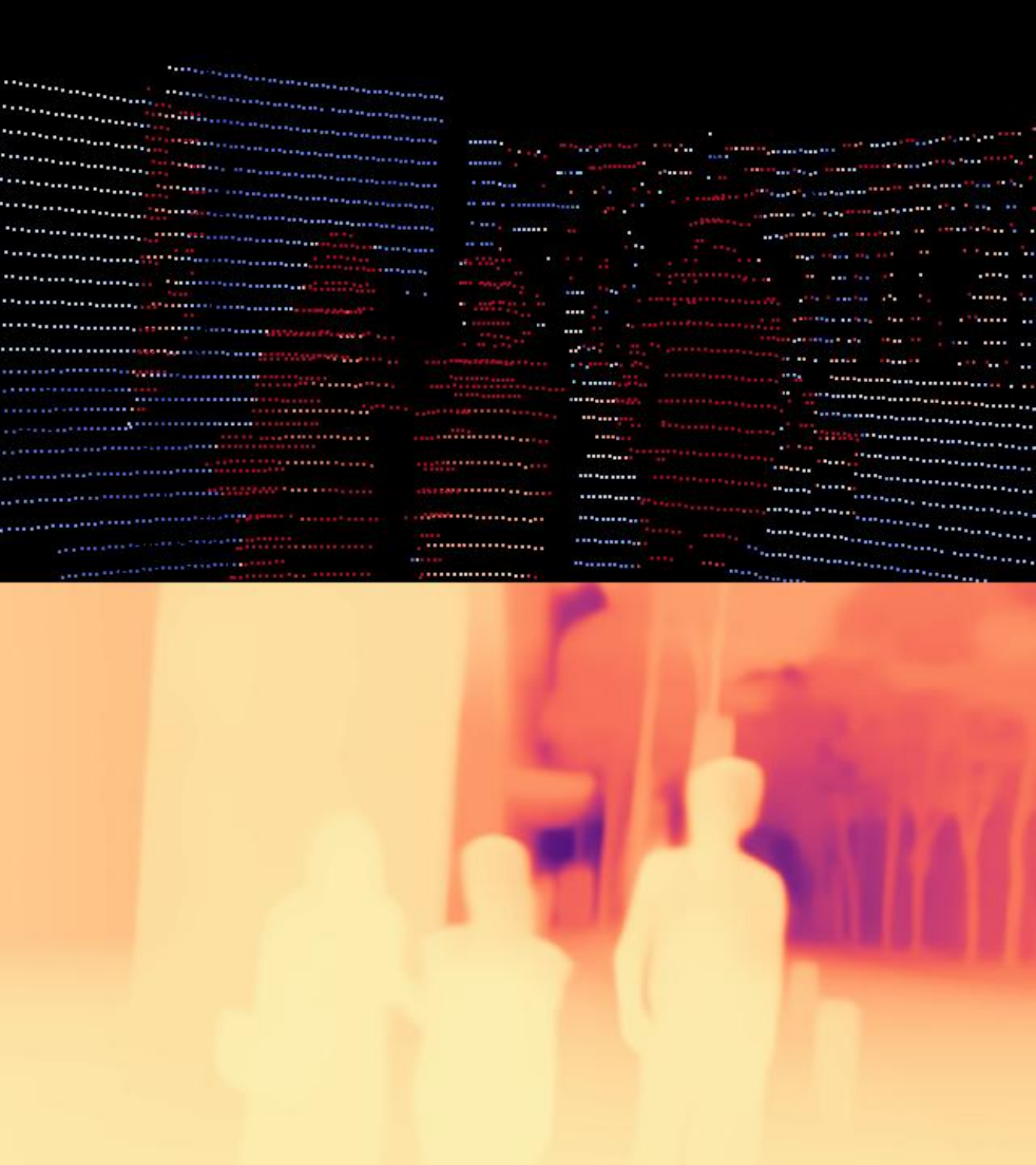}
        & \vspace{-2pt} \\

        \multirow{2}{*}[6pt]{\rotatebox[origin=c]{90}{SUN-RGBD}}
        & \includegraphics[width=0.14\linewidth]{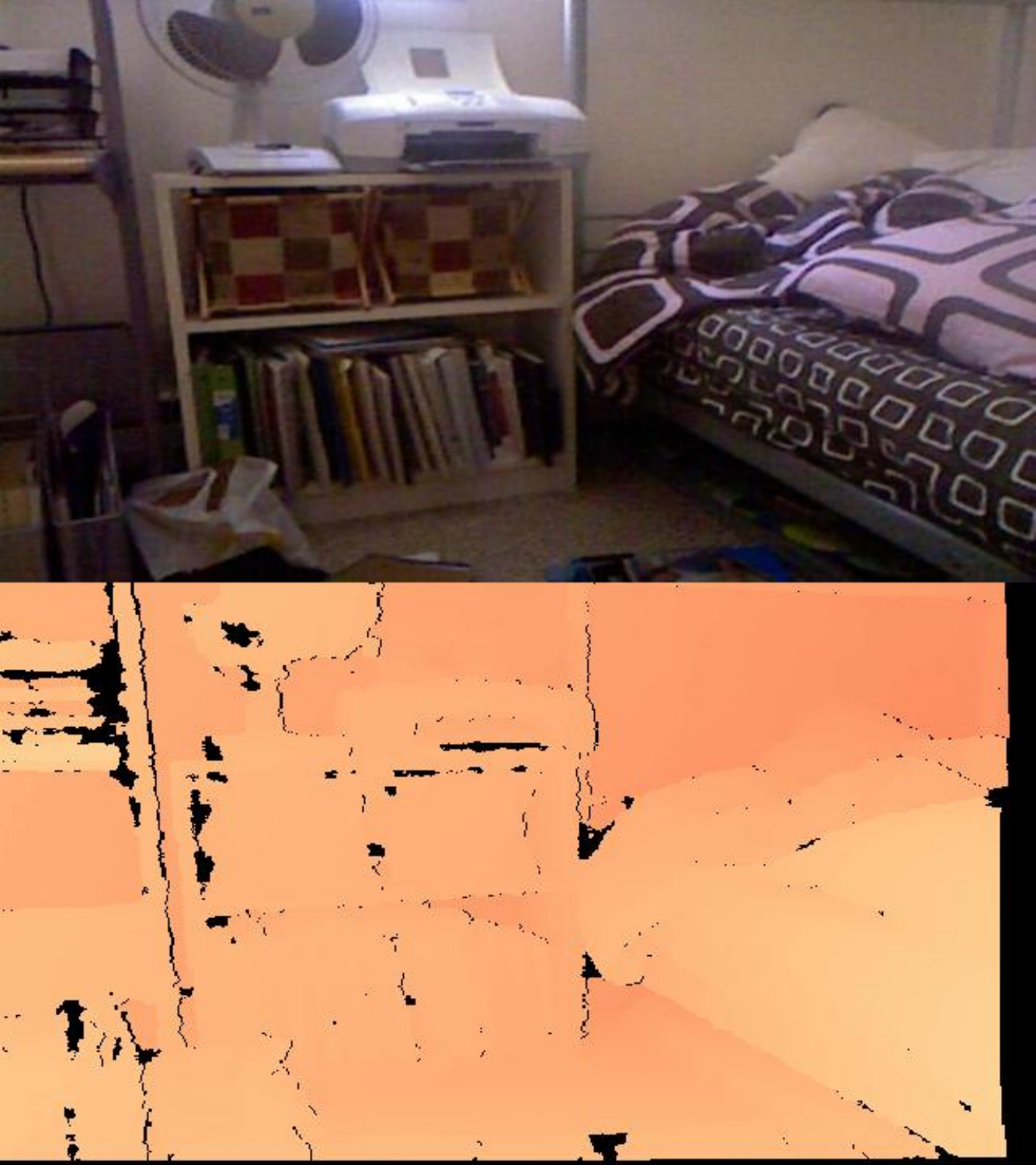}
        & \includegraphics[width=0.14\linewidth]{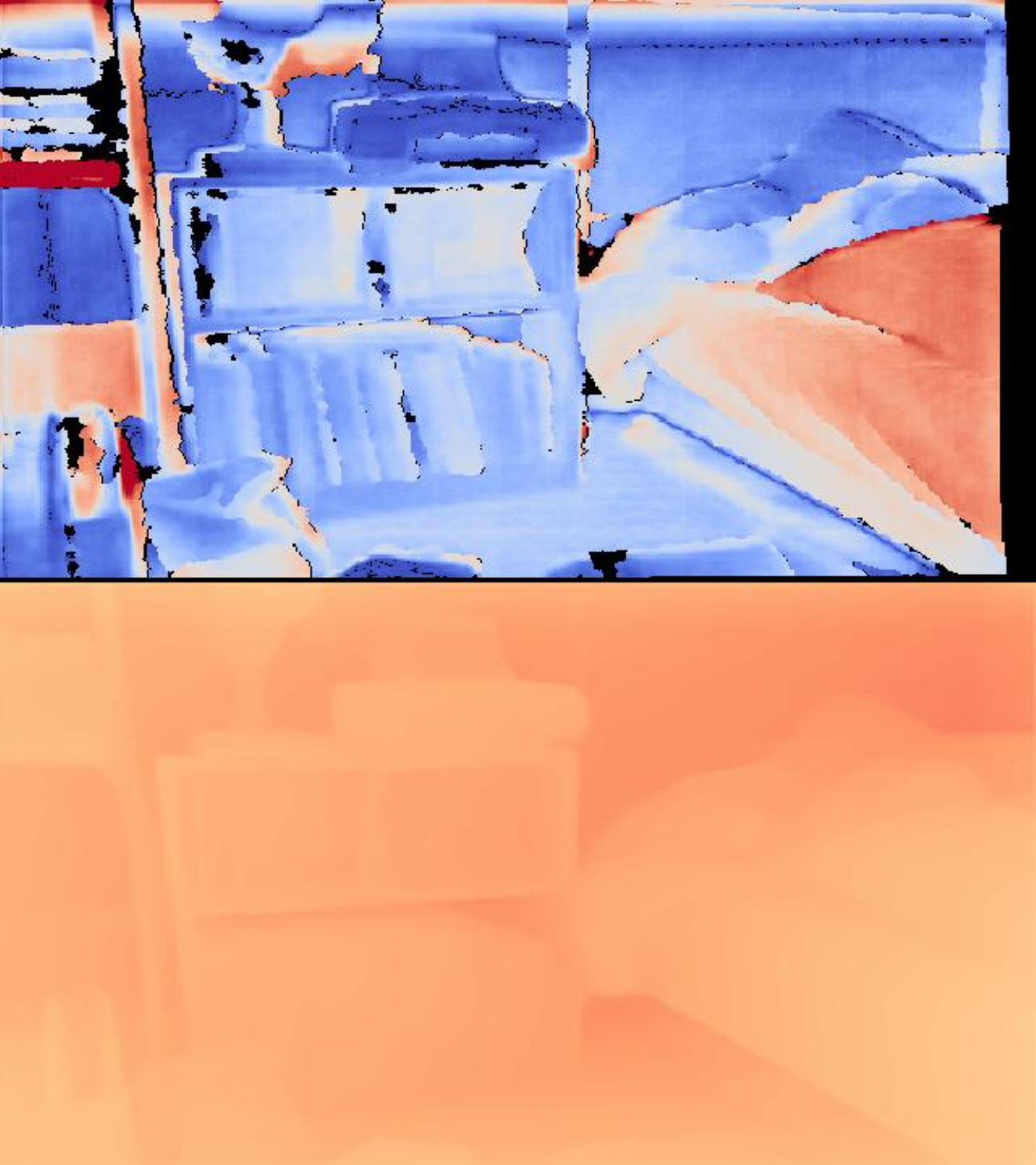}
        & \includegraphics[width=0.14\linewidth]{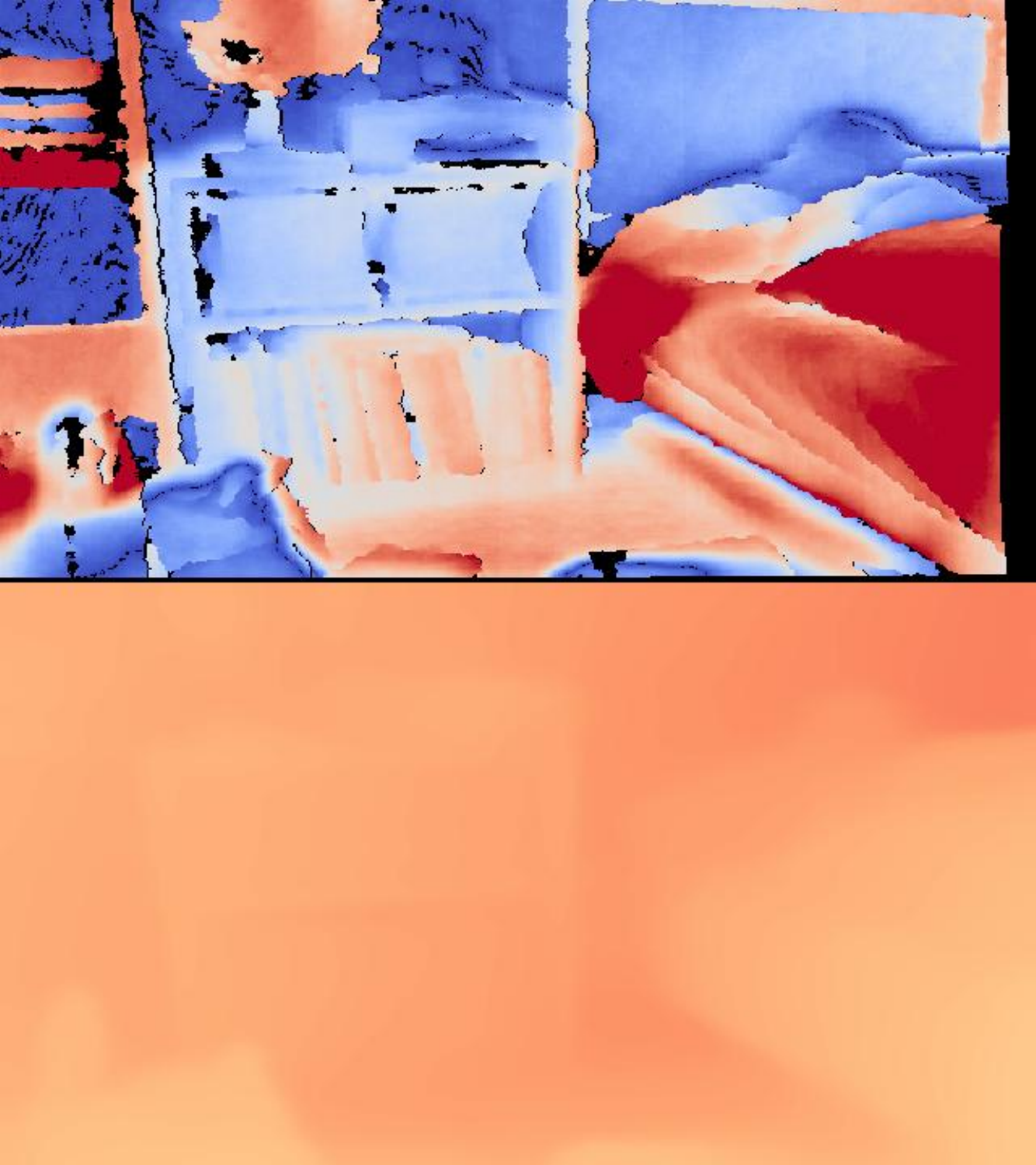}
        & \includegraphics[width=0.14\linewidth]{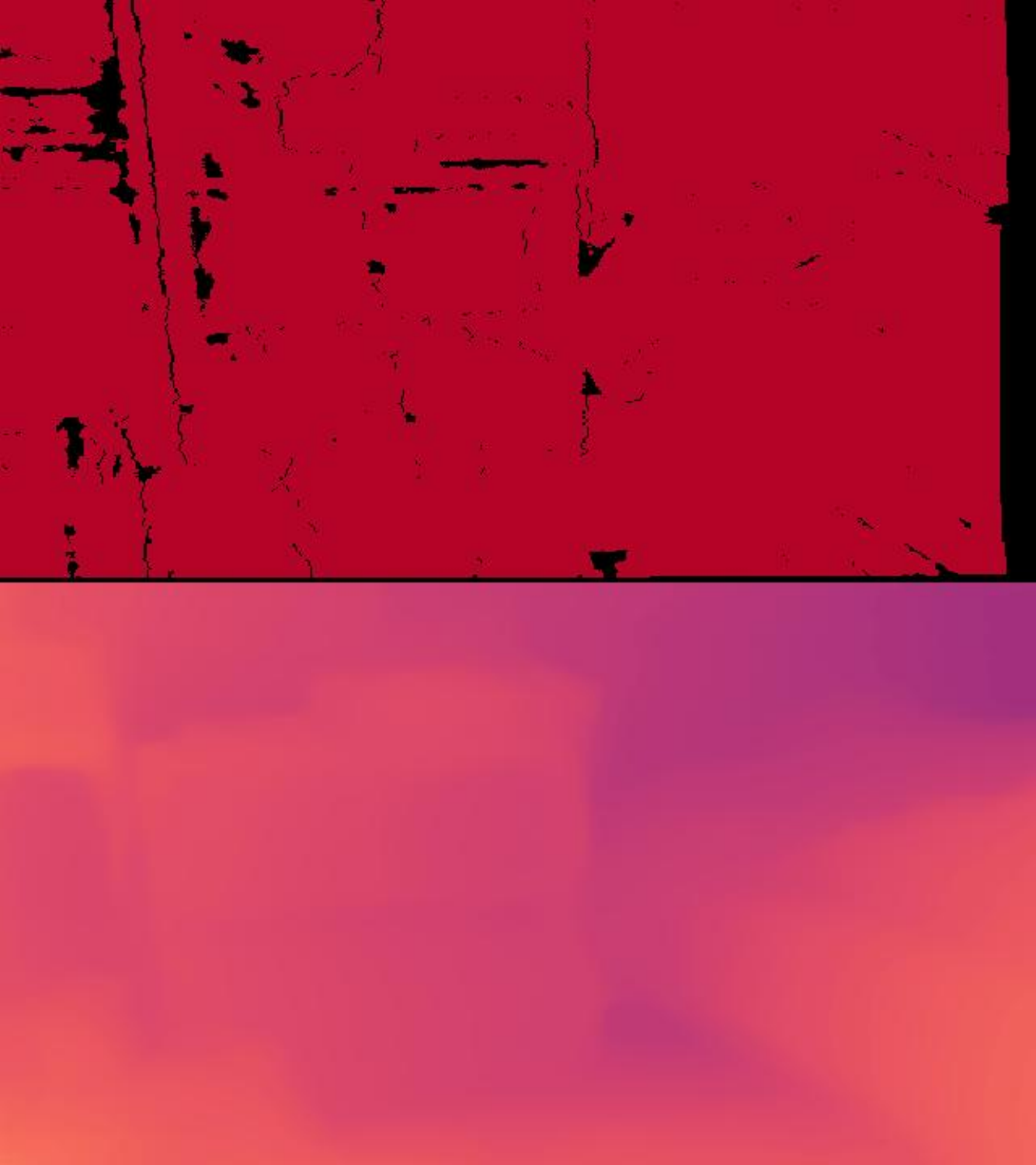}
        & \includegraphics[width=0.14\linewidth]{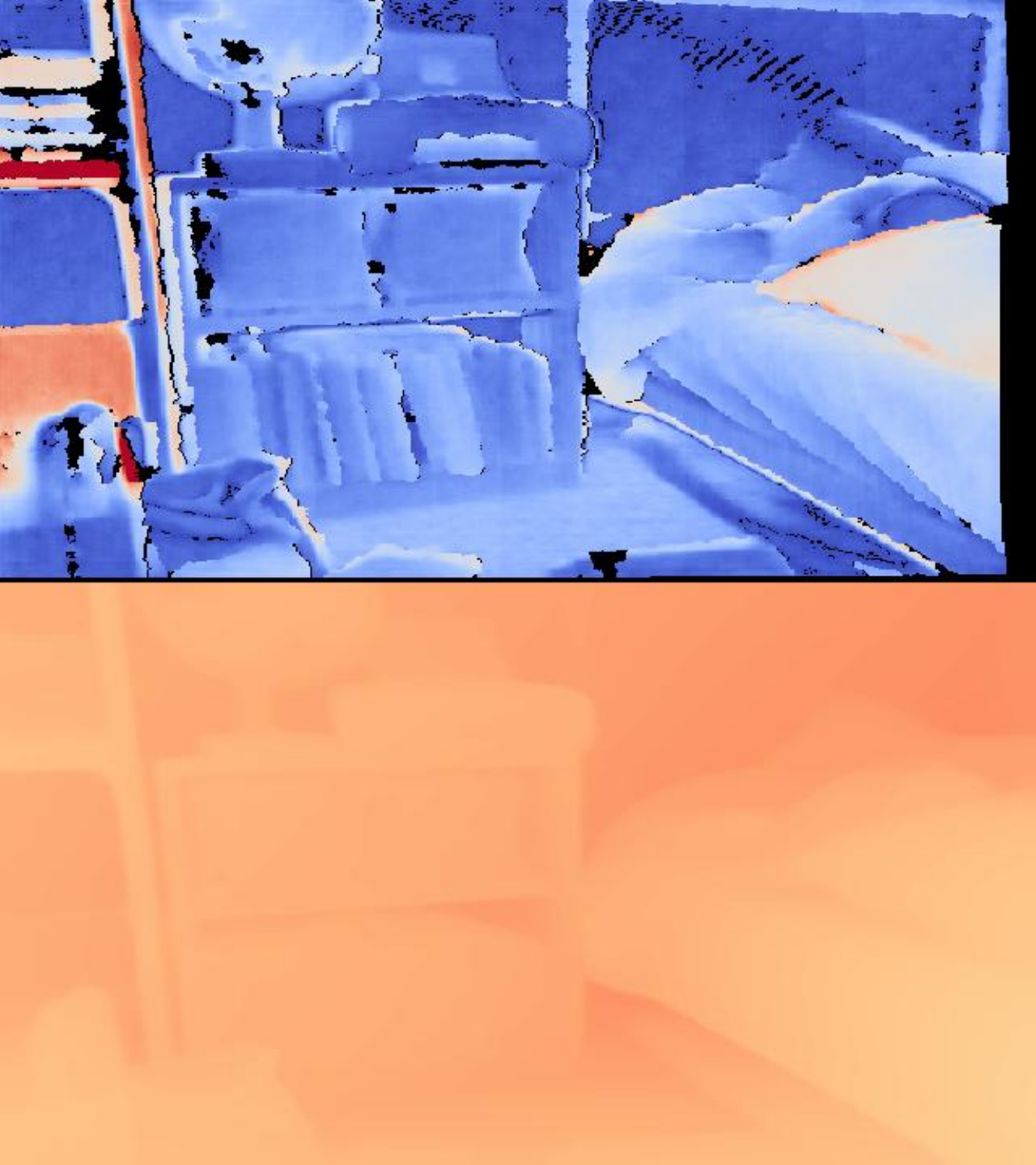}
        & \multirow{2}{*}[20pt]{\includegraphics[width=0.075\linewidth]{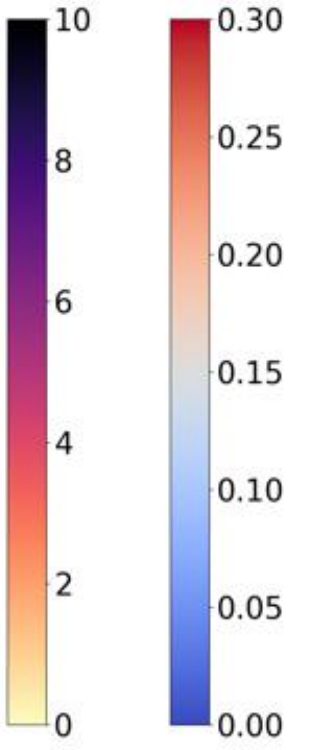}} \vspace{-8pt} \\
        & \includegraphics[width=0.14\linewidth]{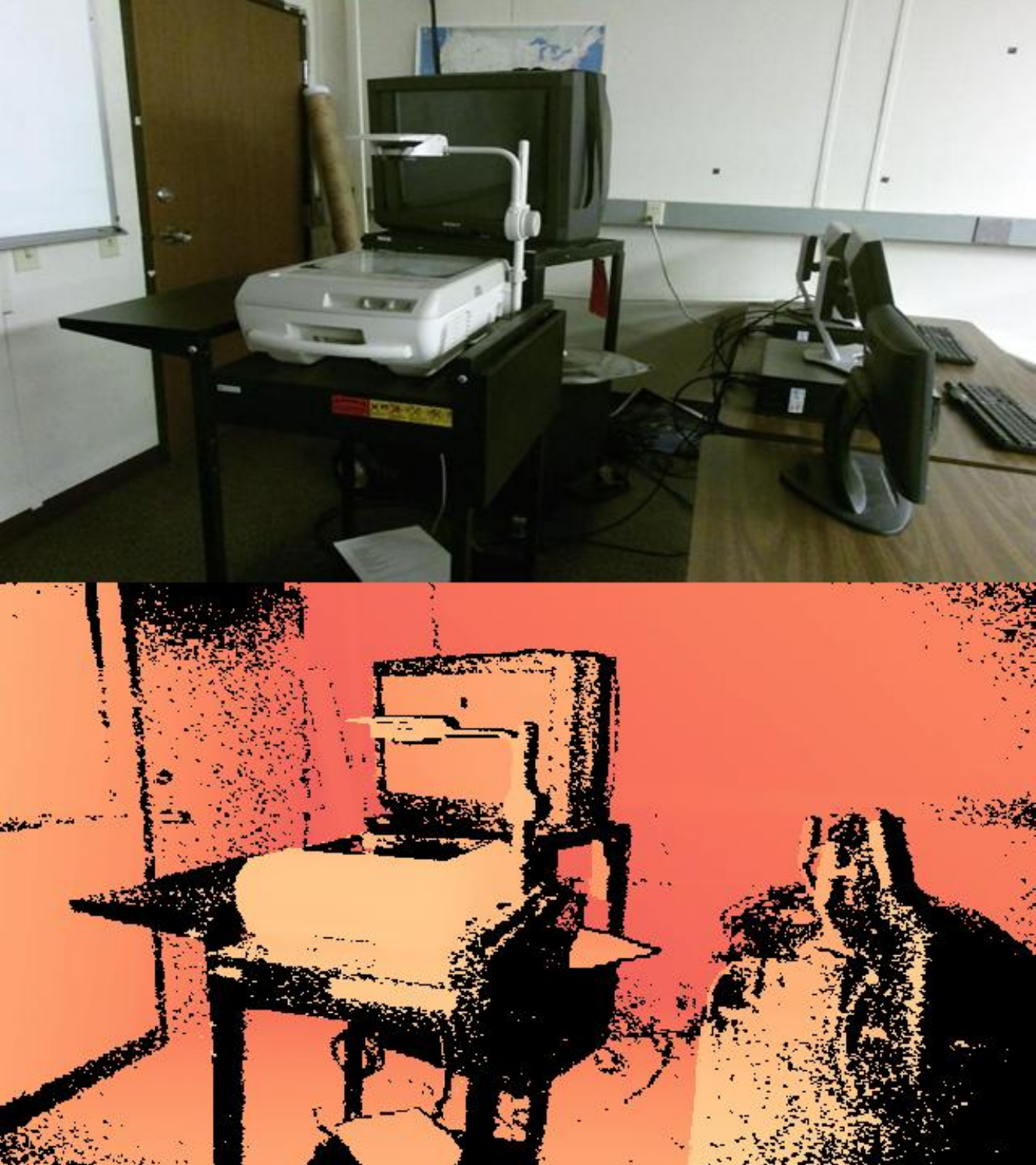}
        & \includegraphics[width=0.14\linewidth]{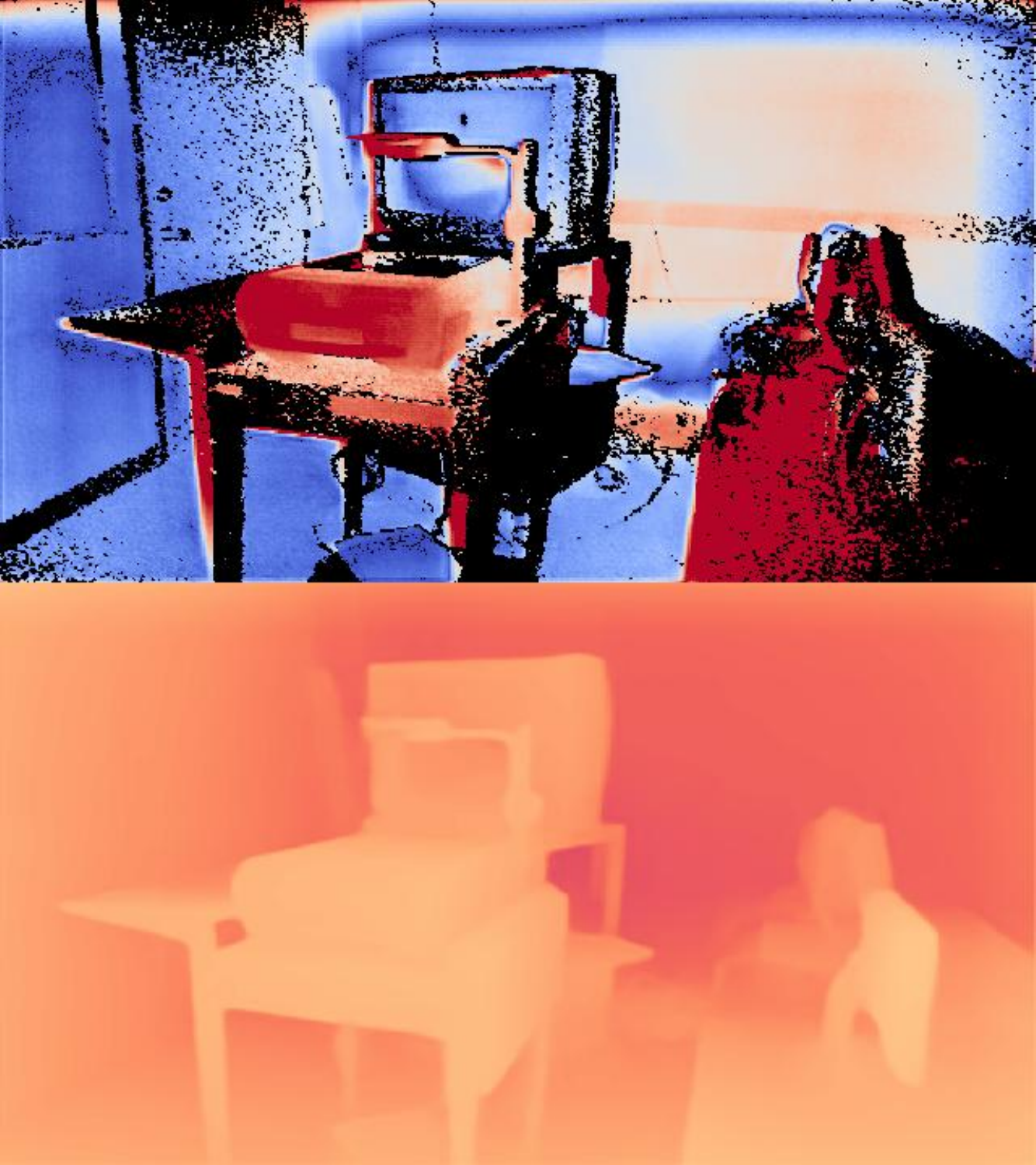}
        & \includegraphics[width=0.14\linewidth]{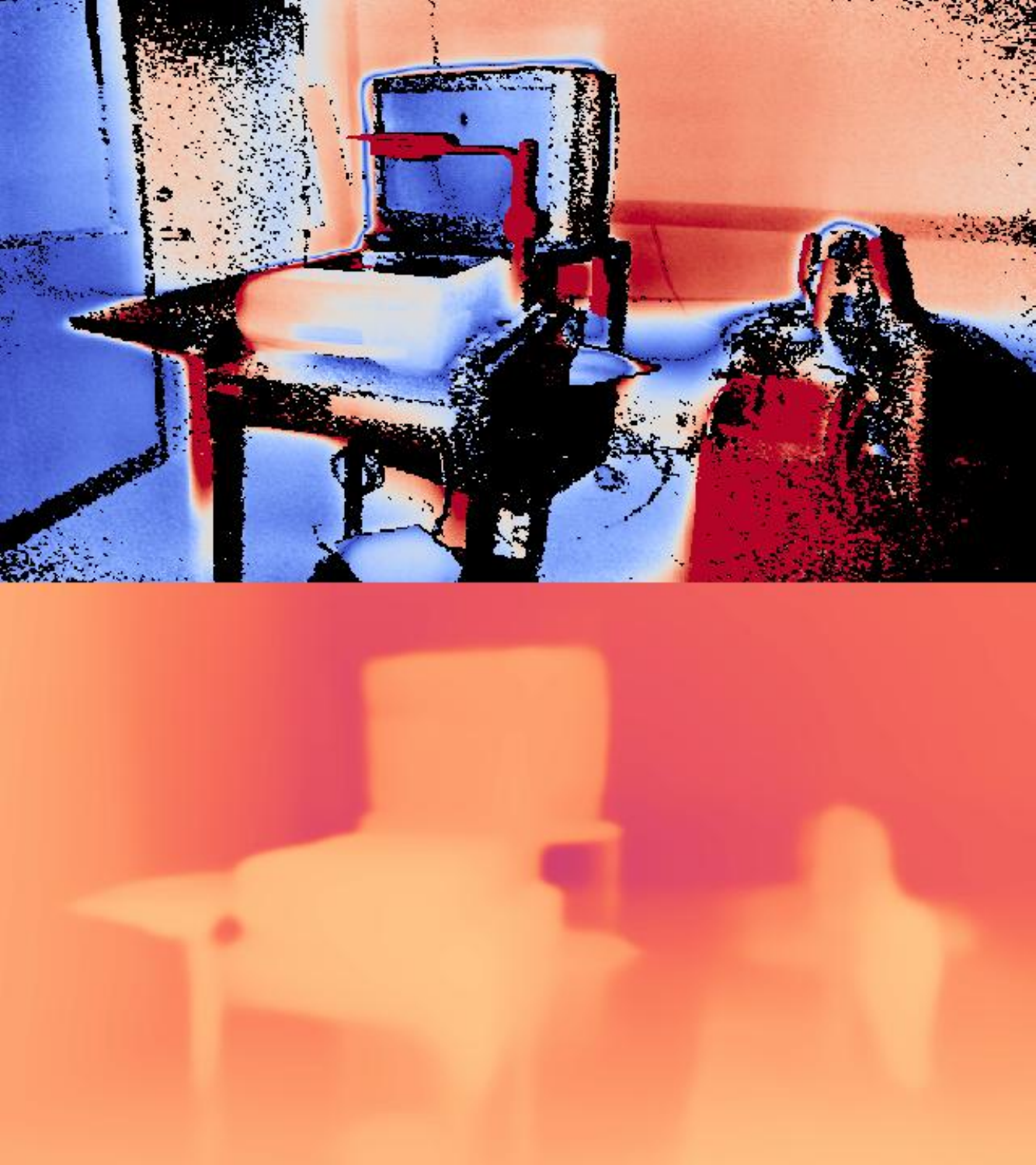}
        & \includegraphics[width=0.14\linewidth]{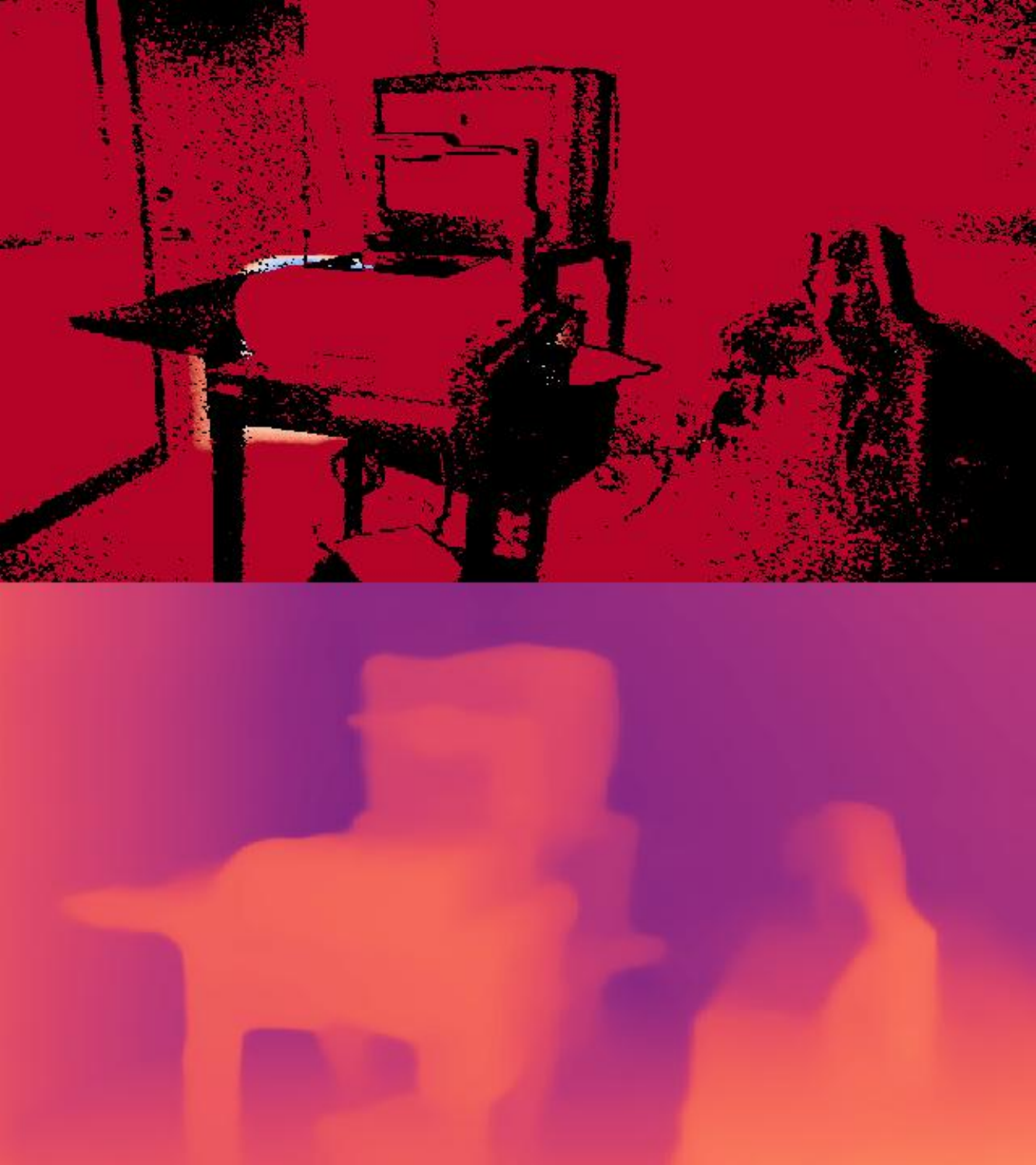}
        & \includegraphics[width=0.14\linewidth]{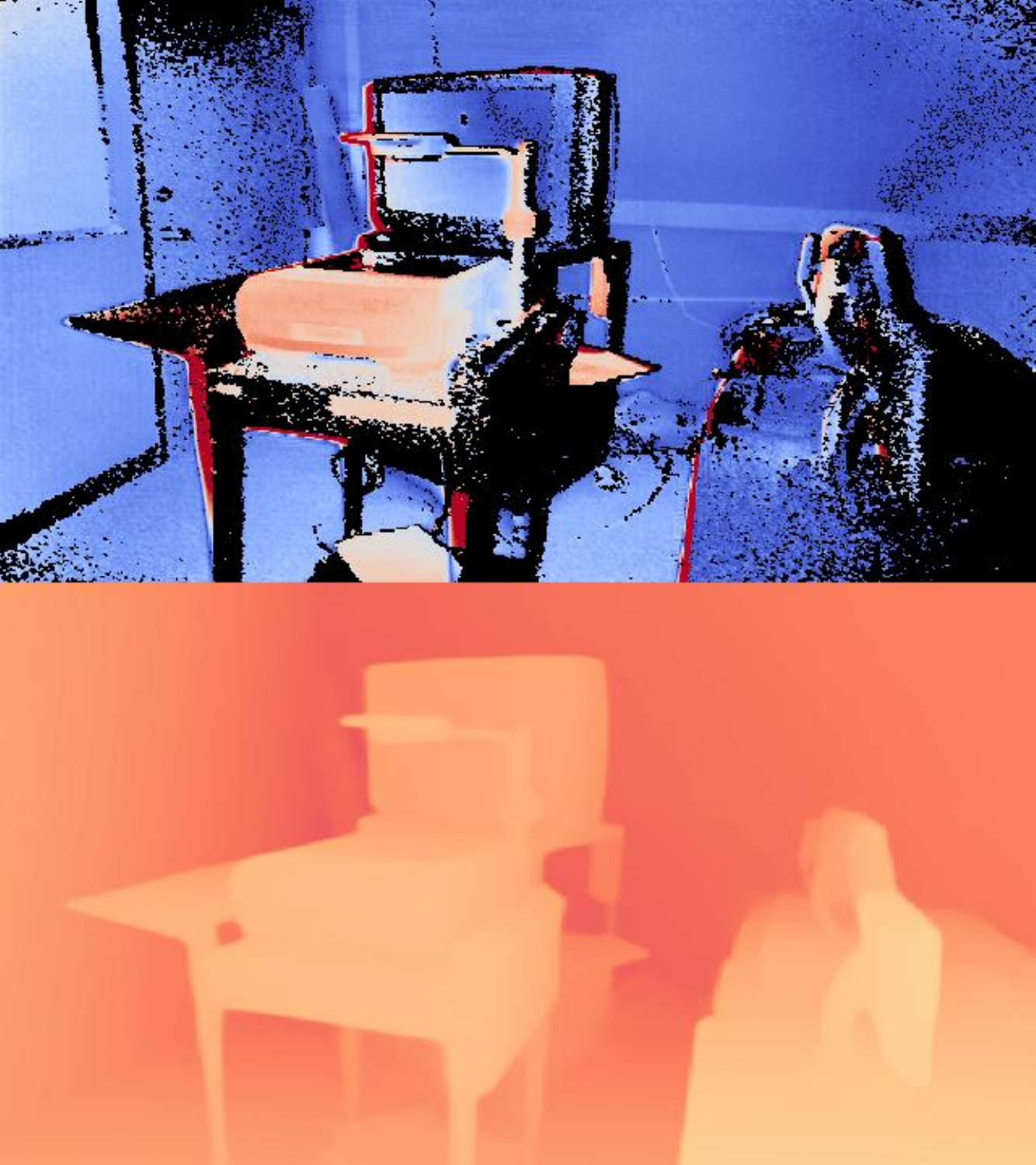}
        & \vspace{-2pt} \\

        \multirow{2}{*}[6pt]{\rotatebox[origin=c]{90}{IBims-1}}
        & \includegraphics[width=0.14\linewidth]{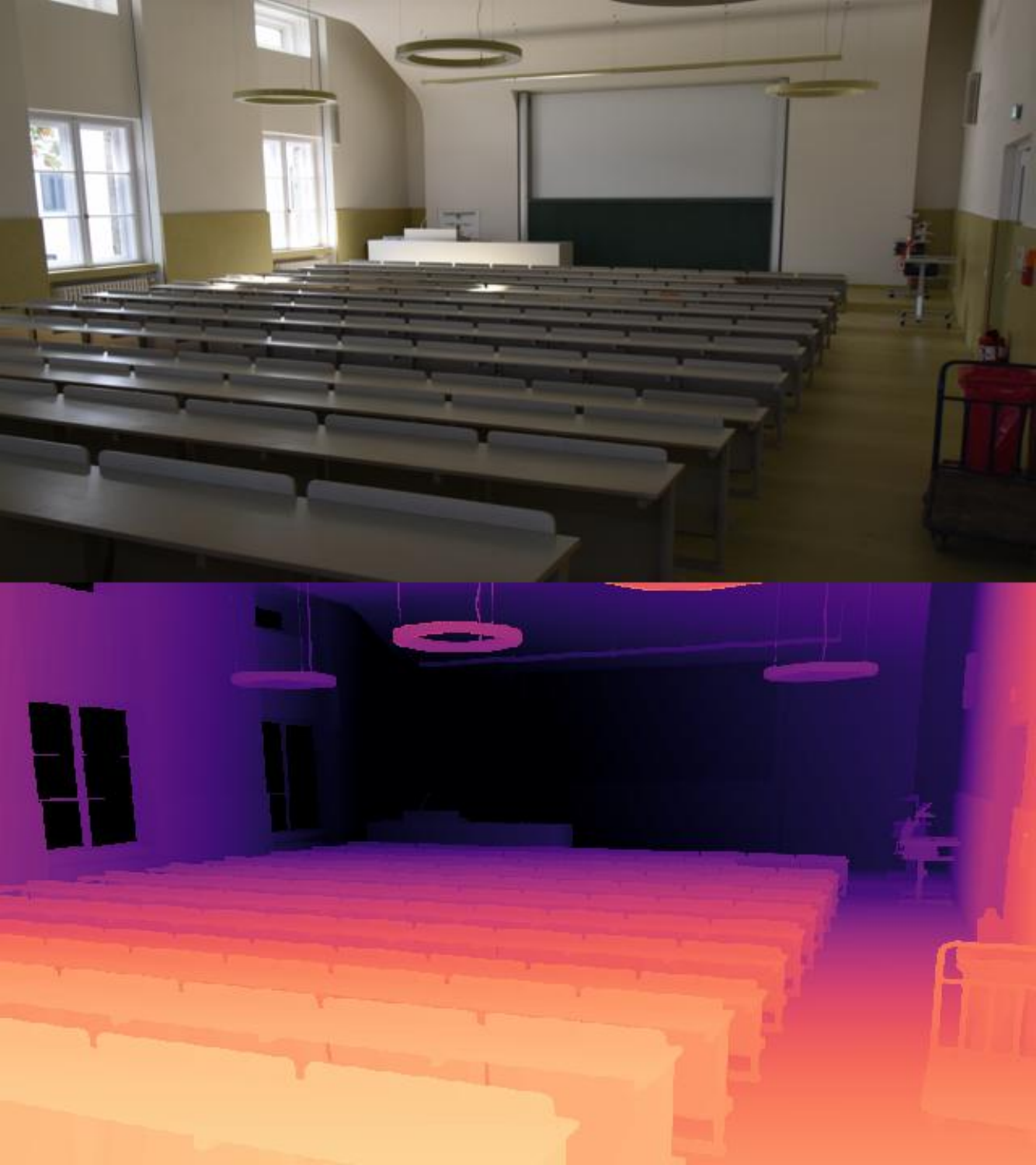}
        & \includegraphics[width=0.14\linewidth]{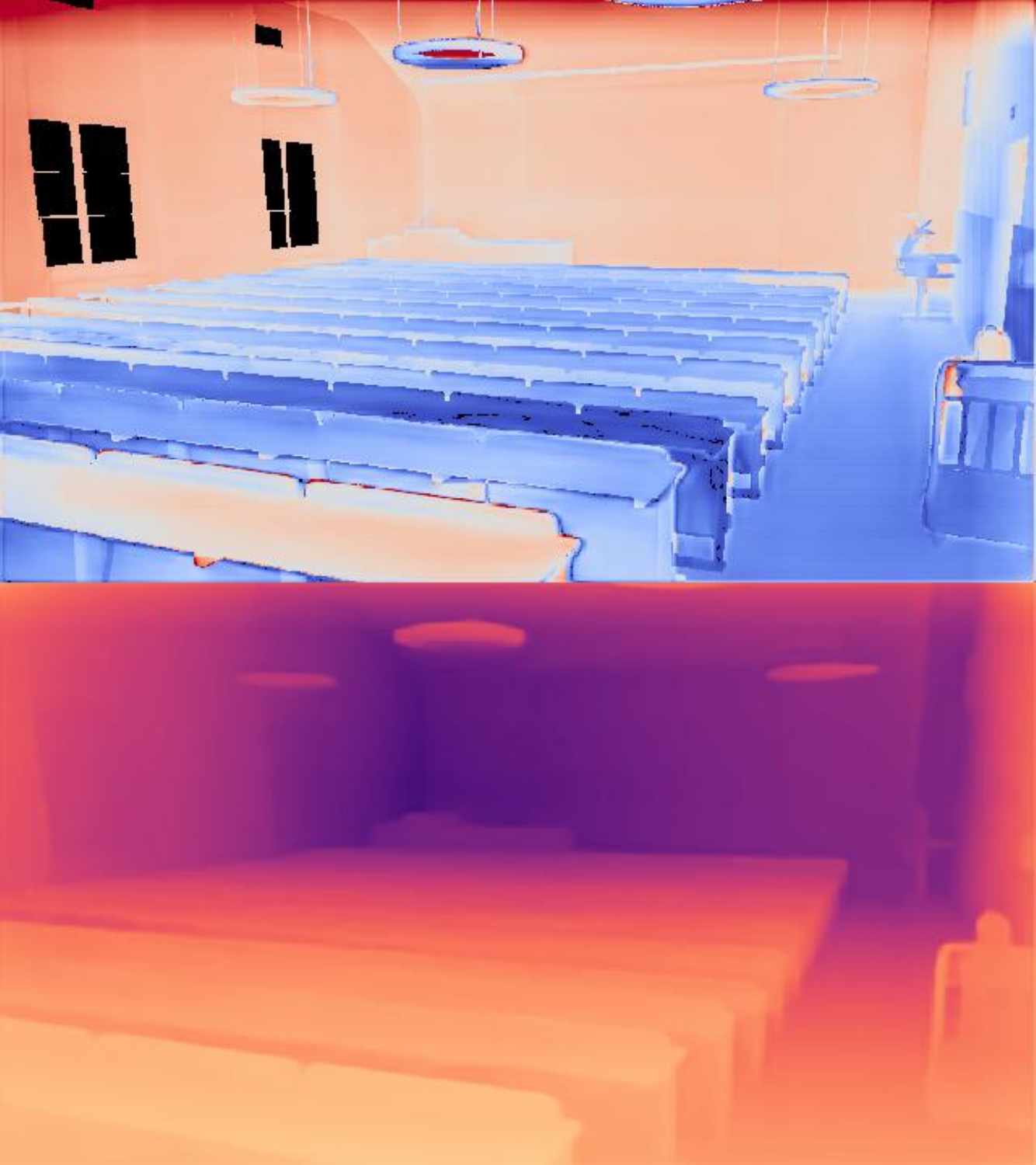}
        & \includegraphics[width=0.14\linewidth]{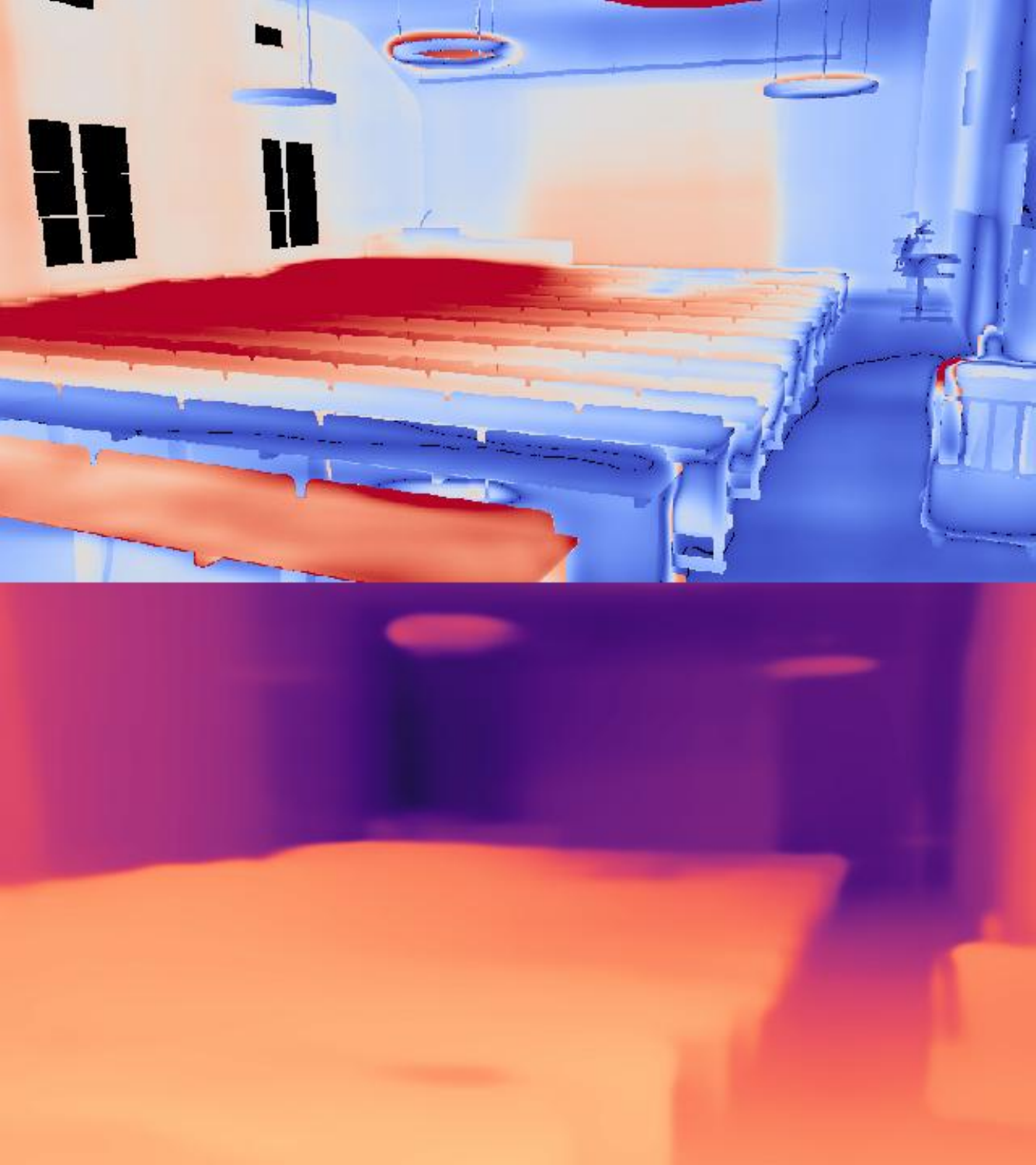}
        & \includegraphics[width=0.14\linewidth]{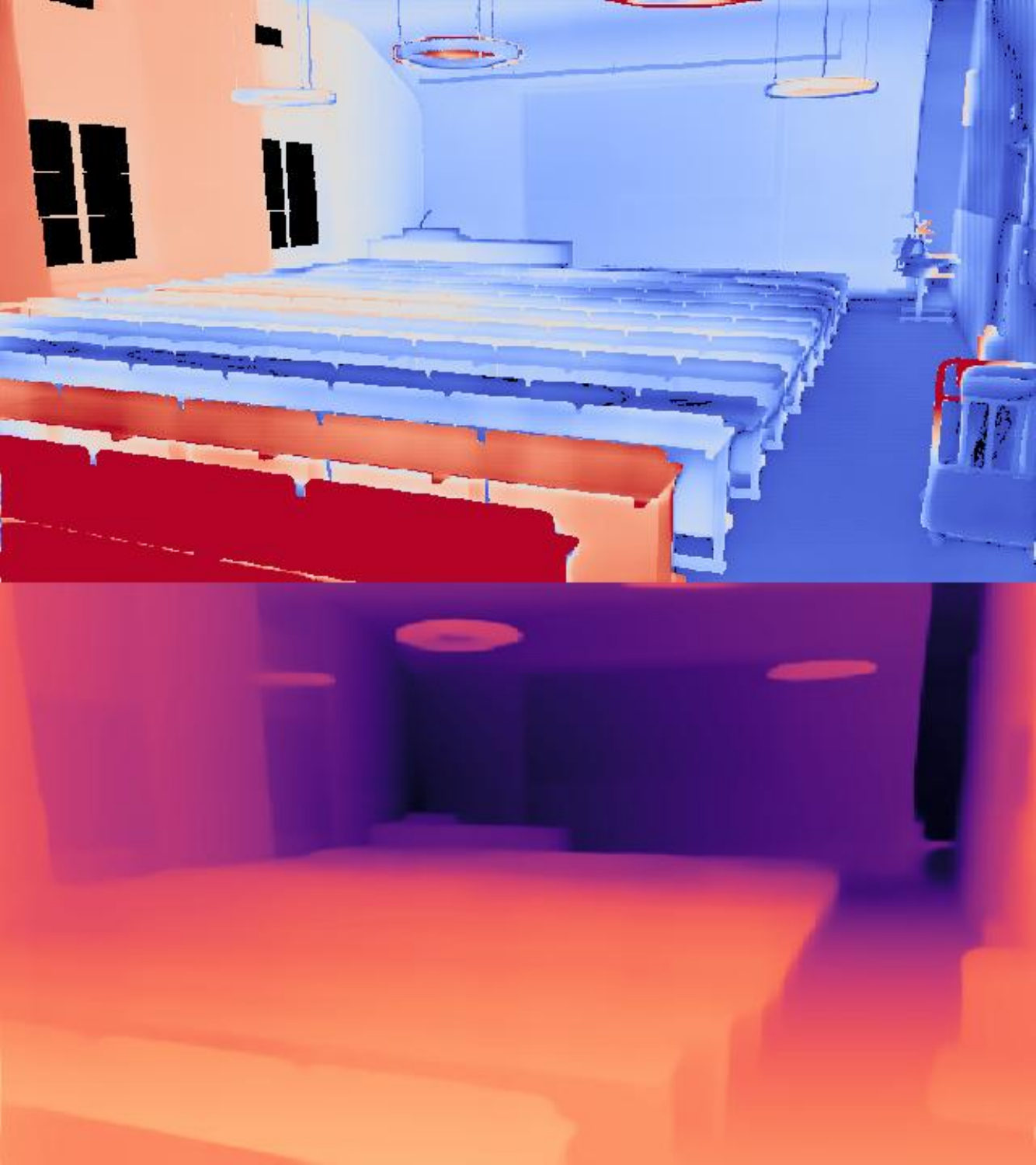}
        & \includegraphics[width=0.14\linewidth]{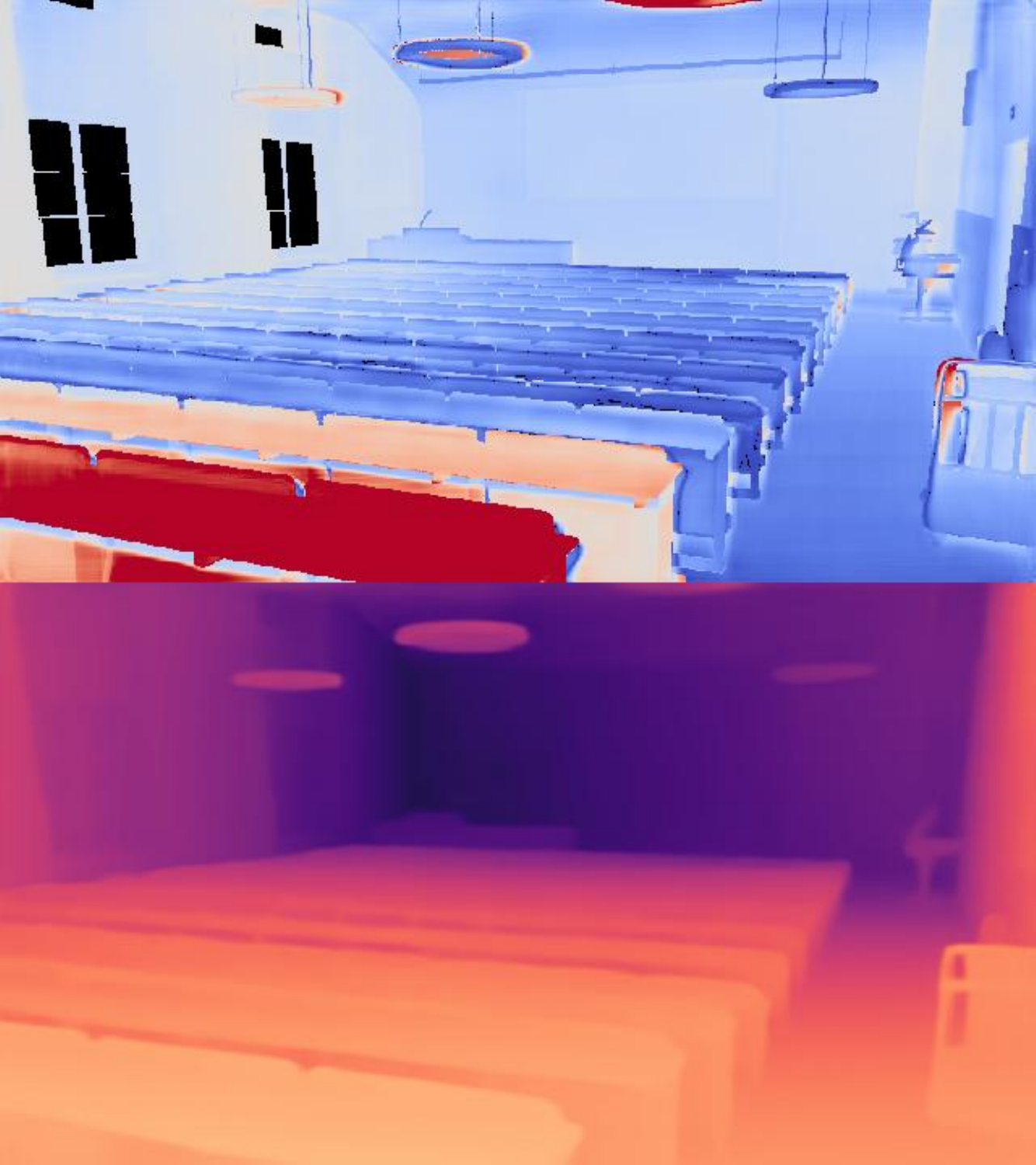}
        & \multirow{2}{*}[25pt]{\includegraphics[width=0.075\linewidth]{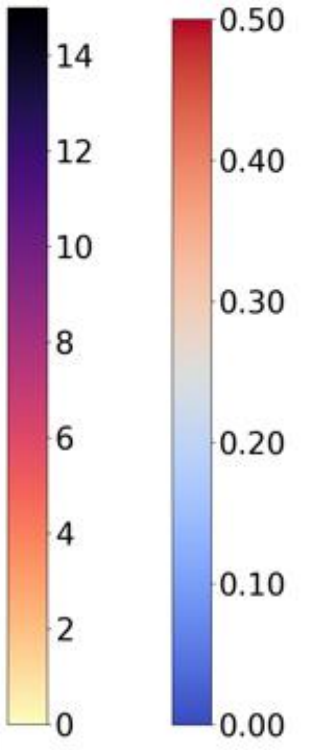}} \vspace{-8pt} \\
        & \includegraphics[width=0.14\linewidth]{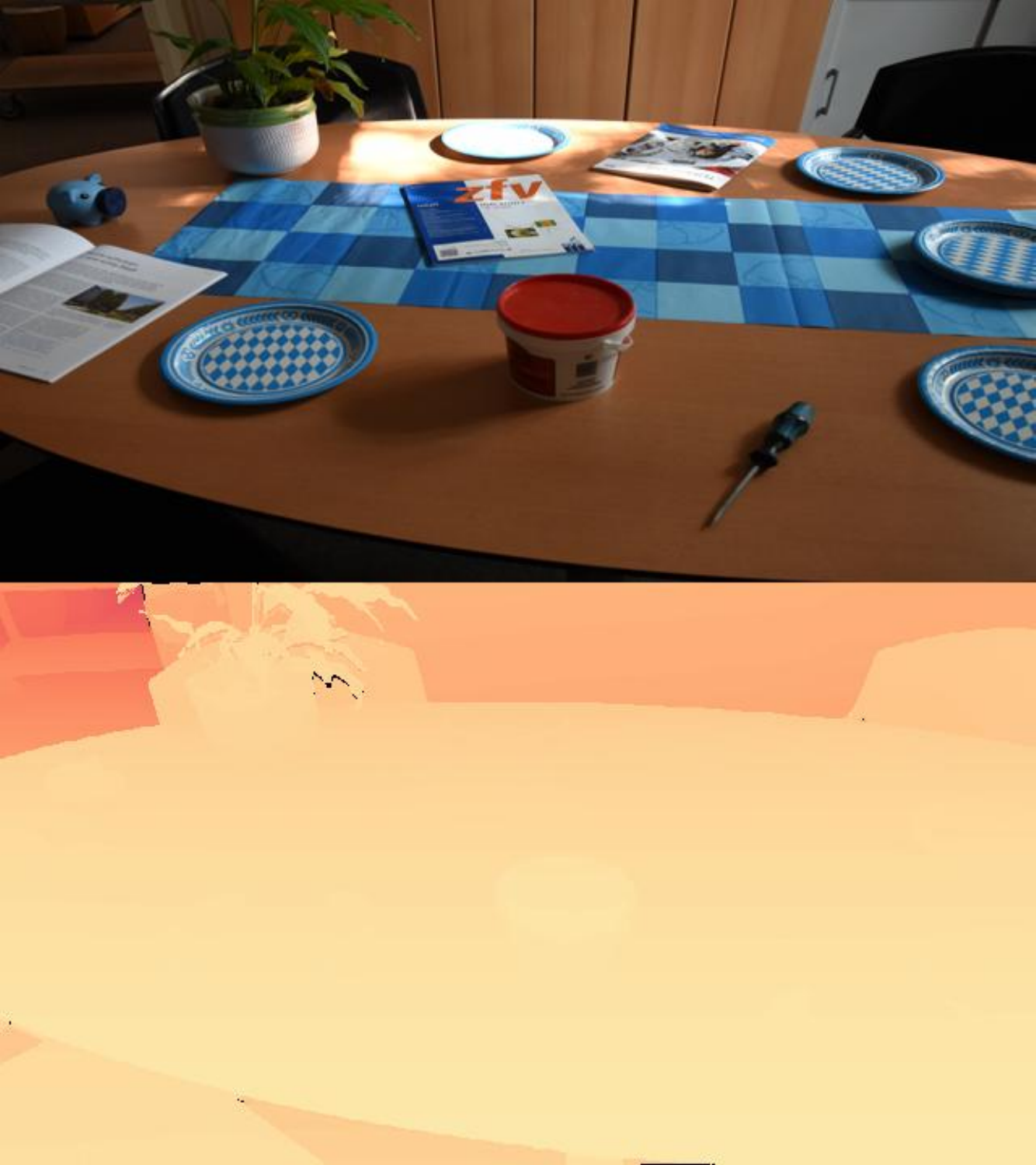}
        & \includegraphics[width=0.14\linewidth]{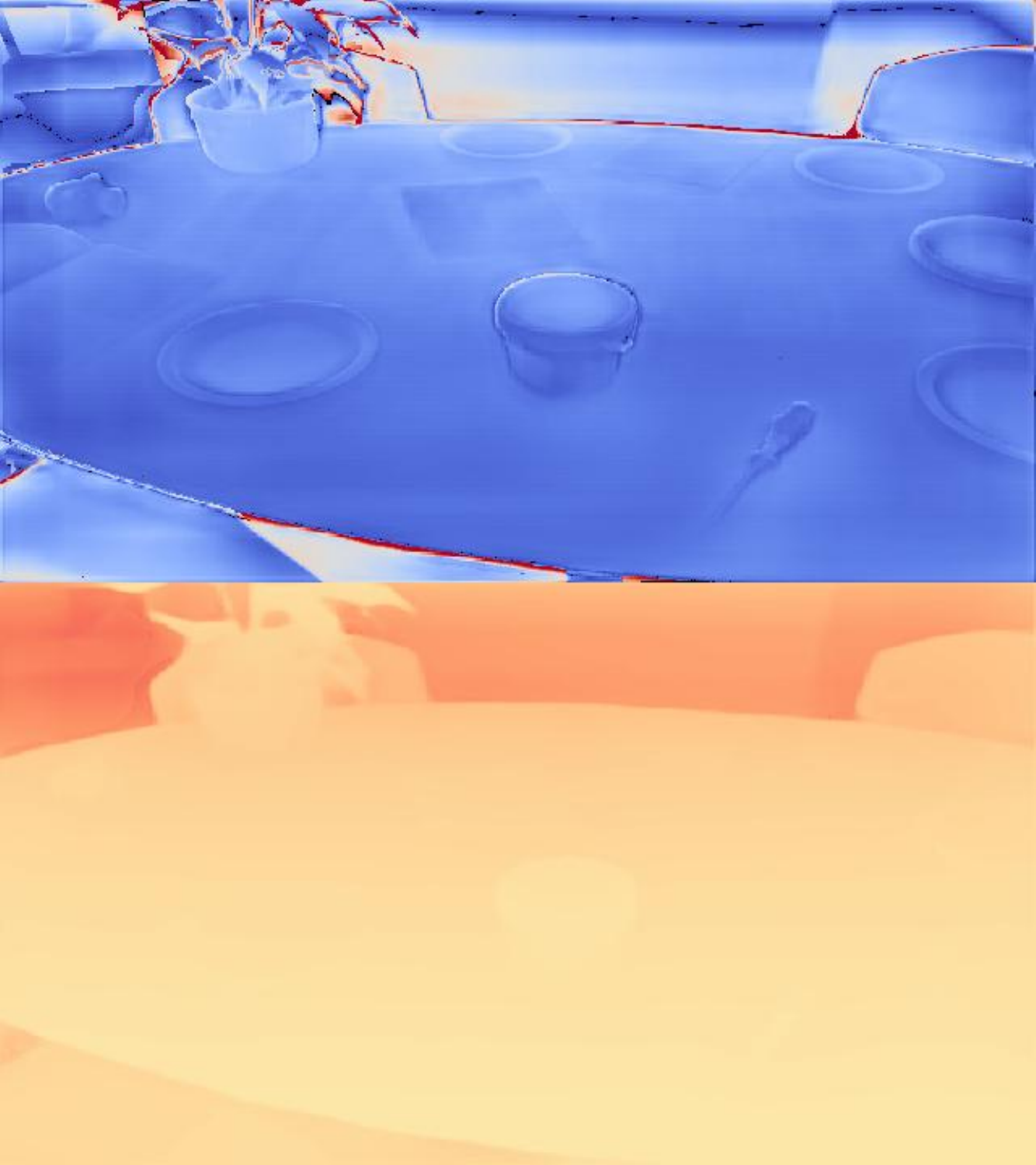}
        & \includegraphics[width=0.14\linewidth]{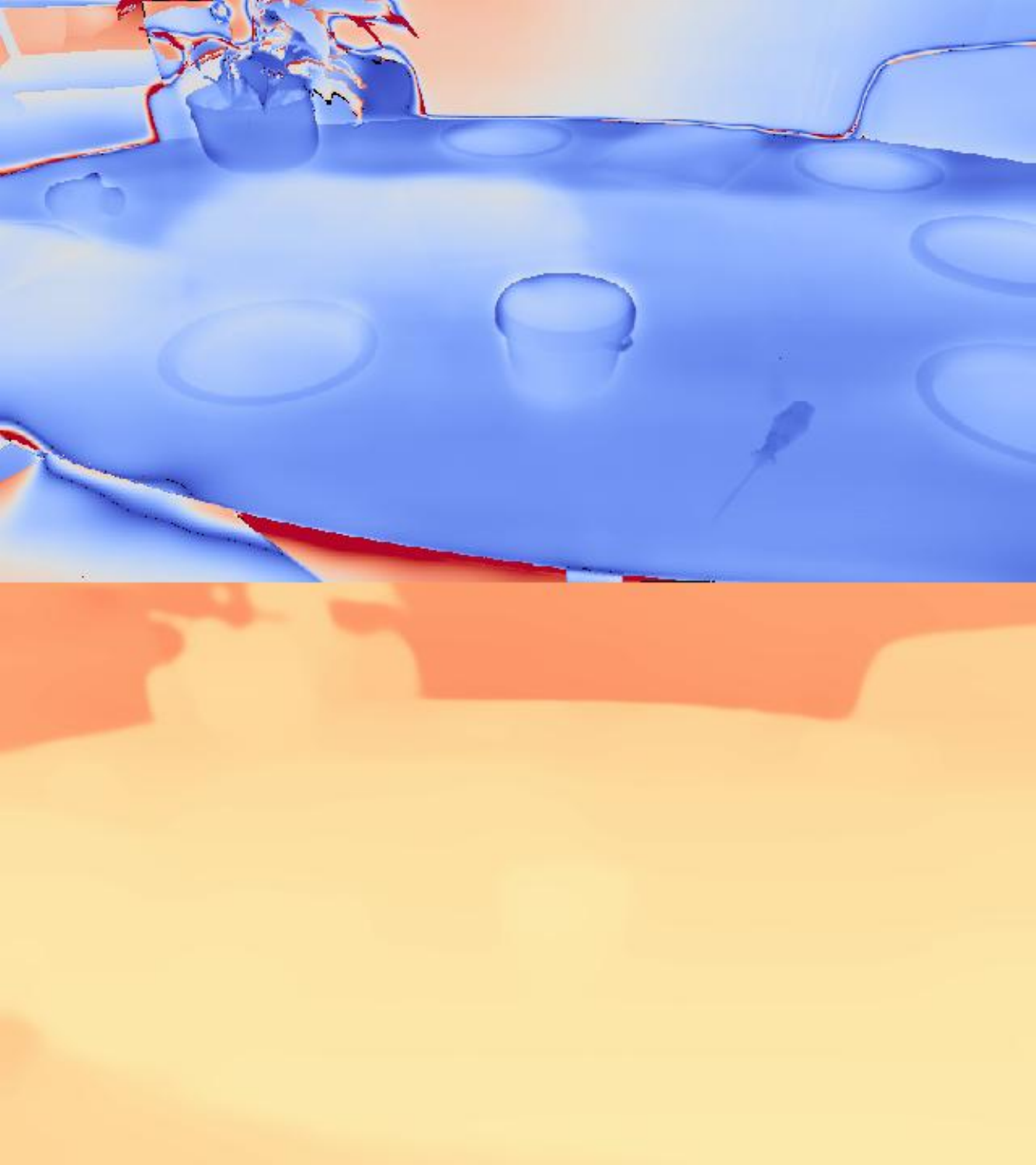}
        & \includegraphics[width=0.14\linewidth]{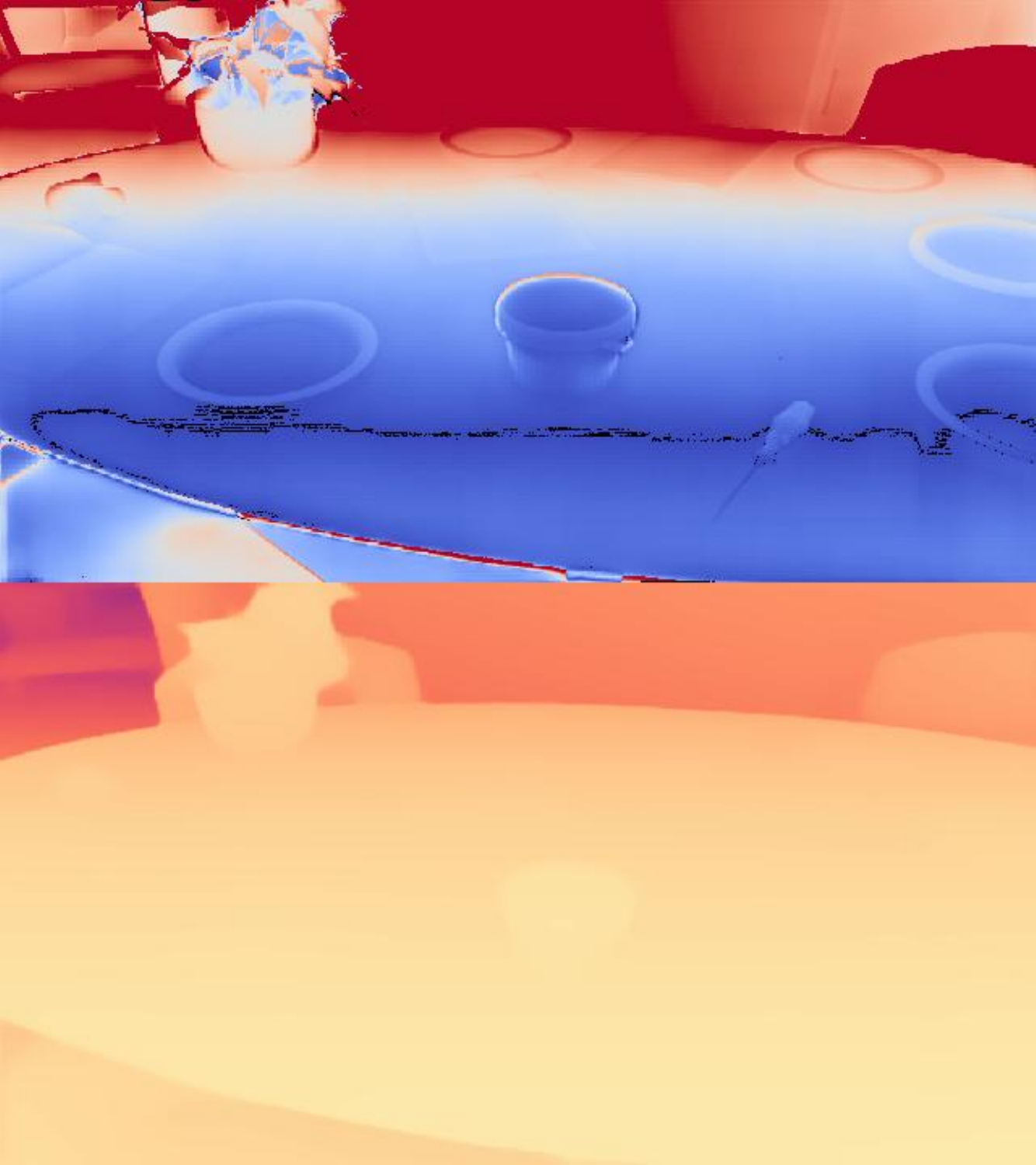}
        & \includegraphics[width=0.14\linewidth]{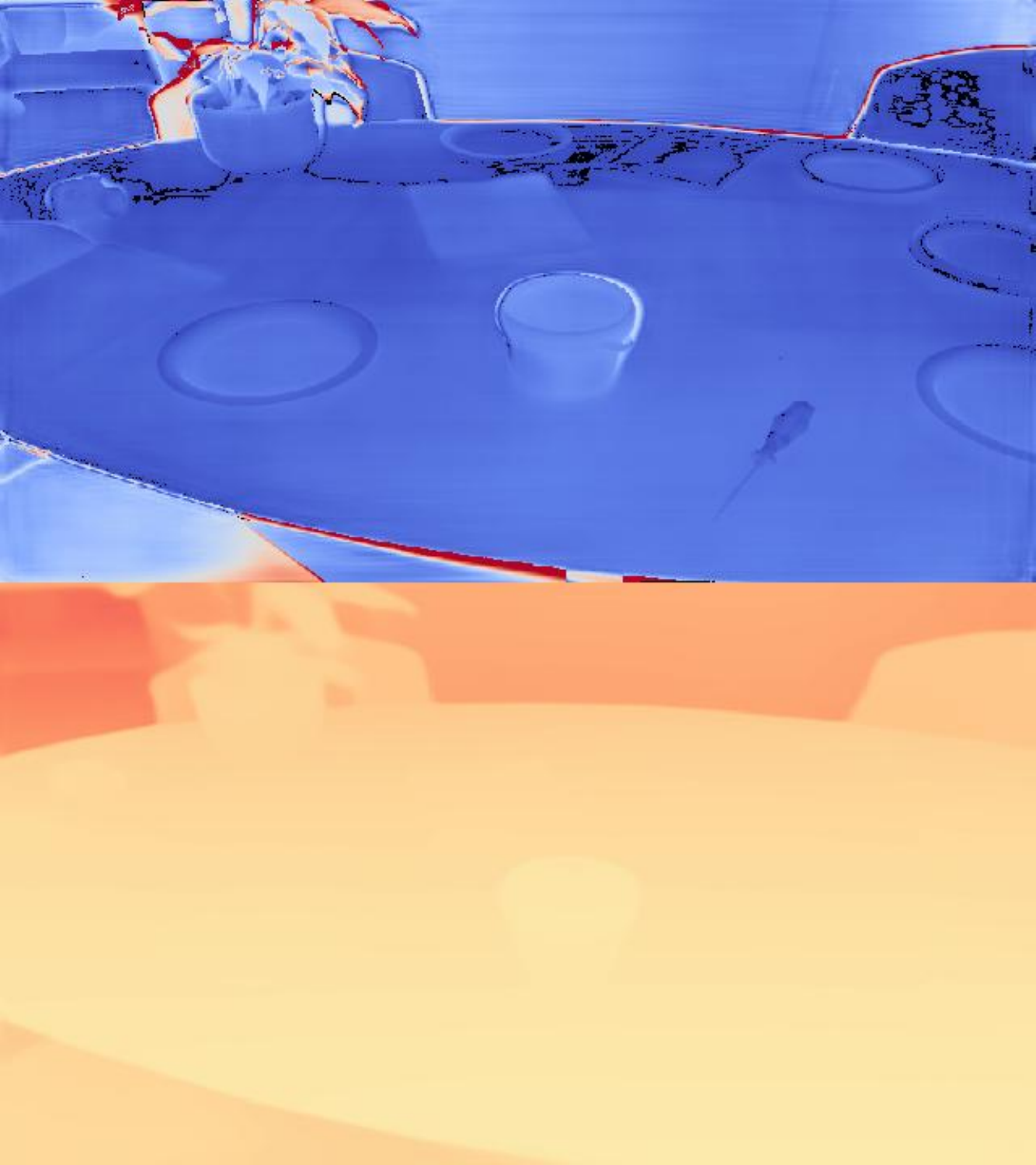}
        & \vspace{-2pt} \\

        & RGB \& GT & ZoeDepth~\cite{bhat2023zoedepth} & ZeroDepth~\cite{guizilini2023zerodepth} & Metric3D\textsuperscript{\dag}~\cite{yin2023metric3d} & \ourmodel & Meters $|$ $\mathrm{A.Rel}$ \\
    \end{tabular}
    \vspace{-8pt}
    \caption{\textbf{Zero-shot qualitative results.} Each pair of consecutive rows corresponds to one test sample. Each odd row shows the input RGB image and the absolute relative error map color-coded with \textit{coolwarm} colormap. Each even row shows GT depth and the predicted depth. The last column represents the specific colormap ranges for depth and error. (\dag): DDAD in the training set.}
    \label{fig:supp:vis2}
\end{figure*}

\begin{figure*}[t]
    \renewcommand{\arraystretch}{2}
    \centering
    \small
    \begin{tabular}{cc|cccc|c}
        \multirow{2}{*}[6pt]{\rotatebox[origin=c]{90}{VOID}}
        & \includegraphics[width=0.14\linewidth]{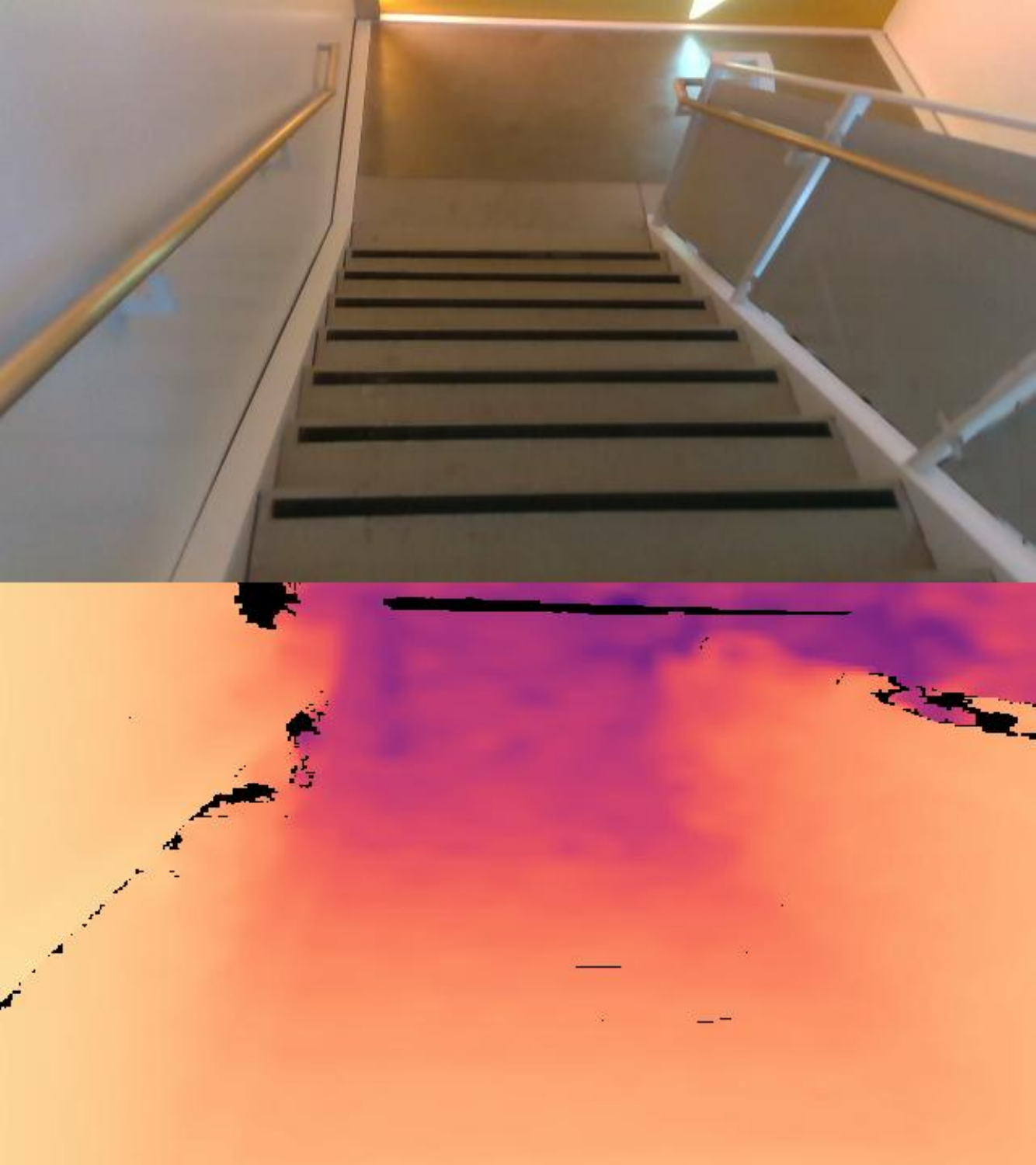}
        & \includegraphics[width=0.14\linewidth]{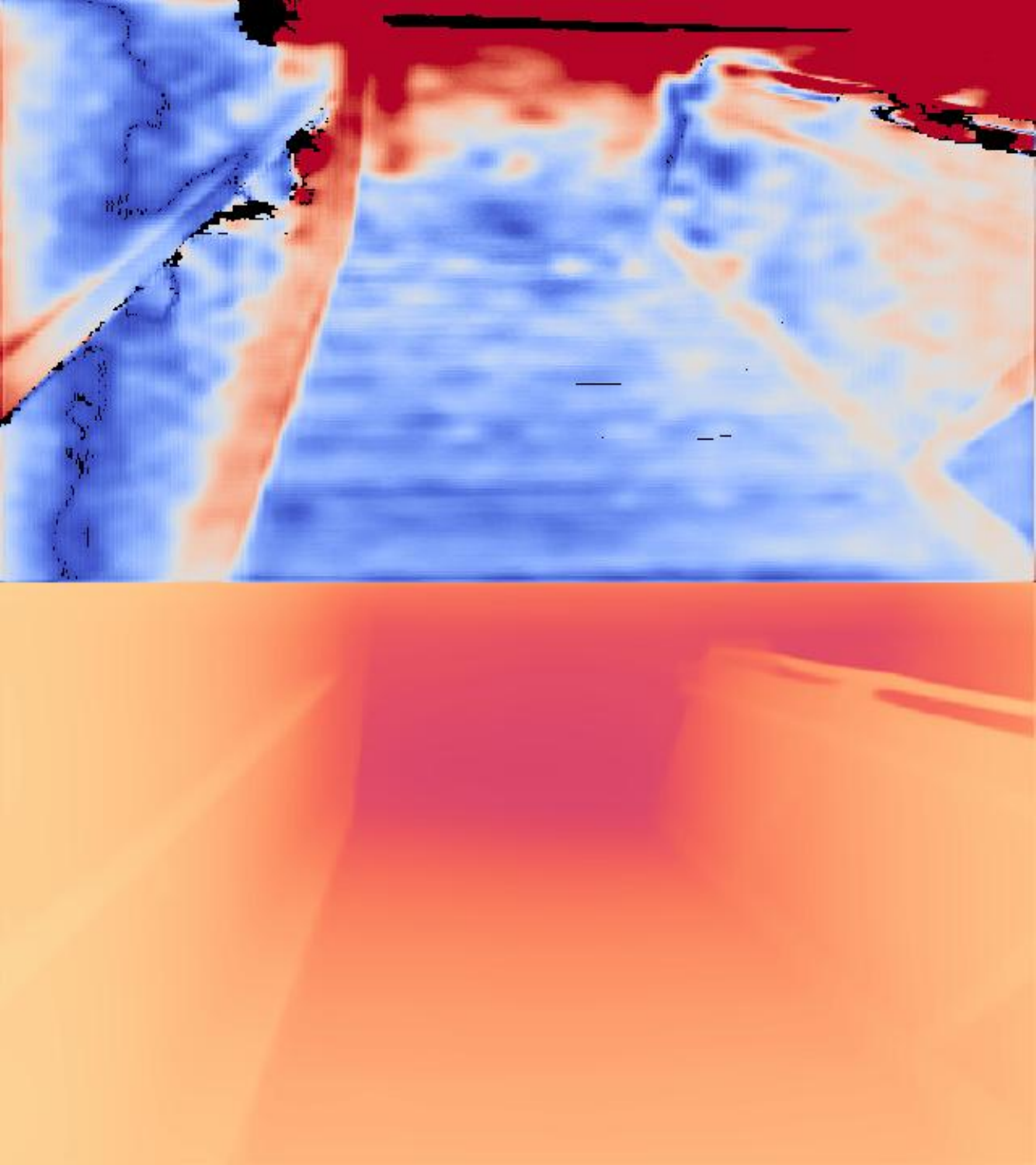}
        & \includegraphics[width=0.14\linewidth]{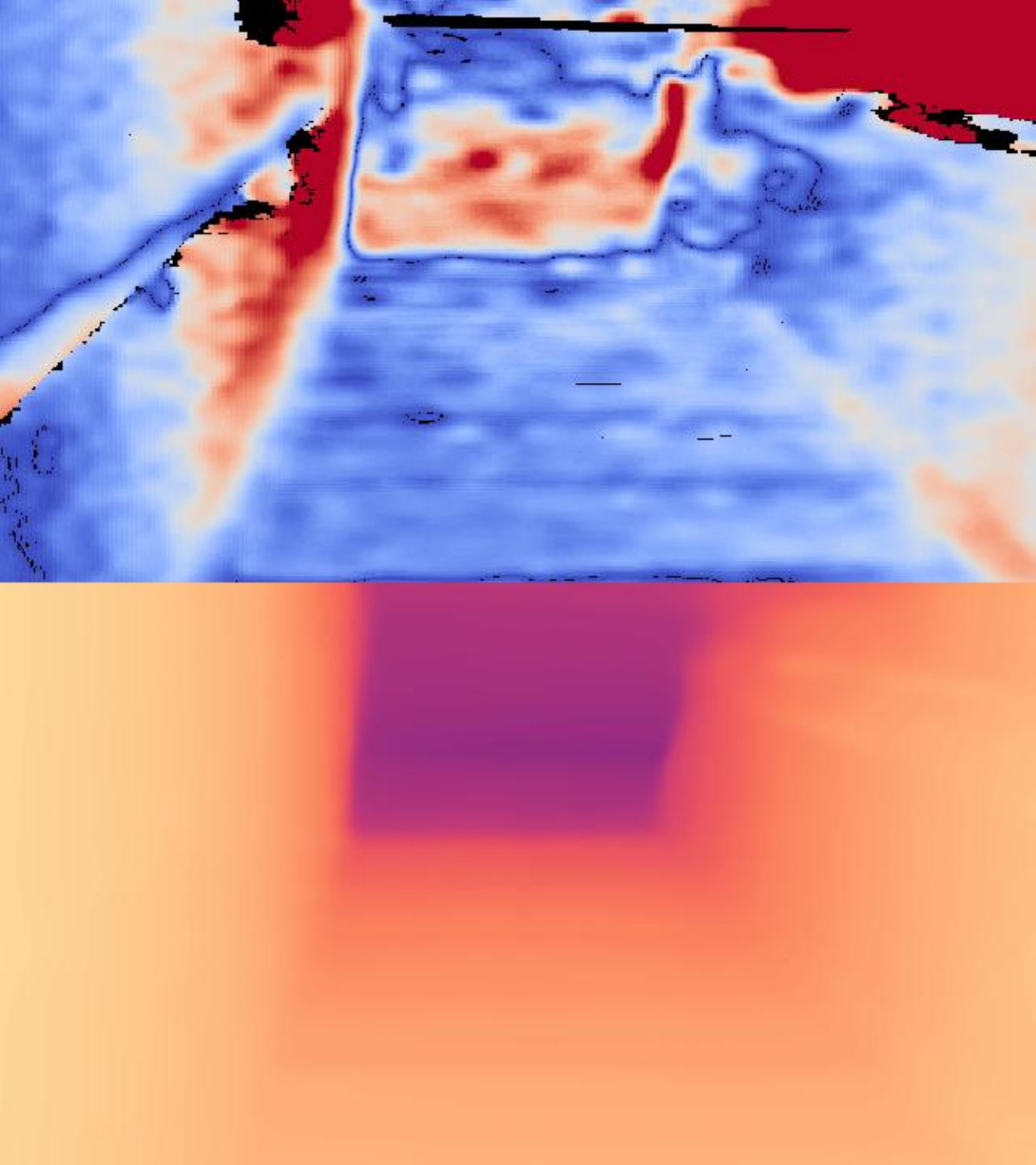}
        & \includegraphics[width=0.14\linewidth]{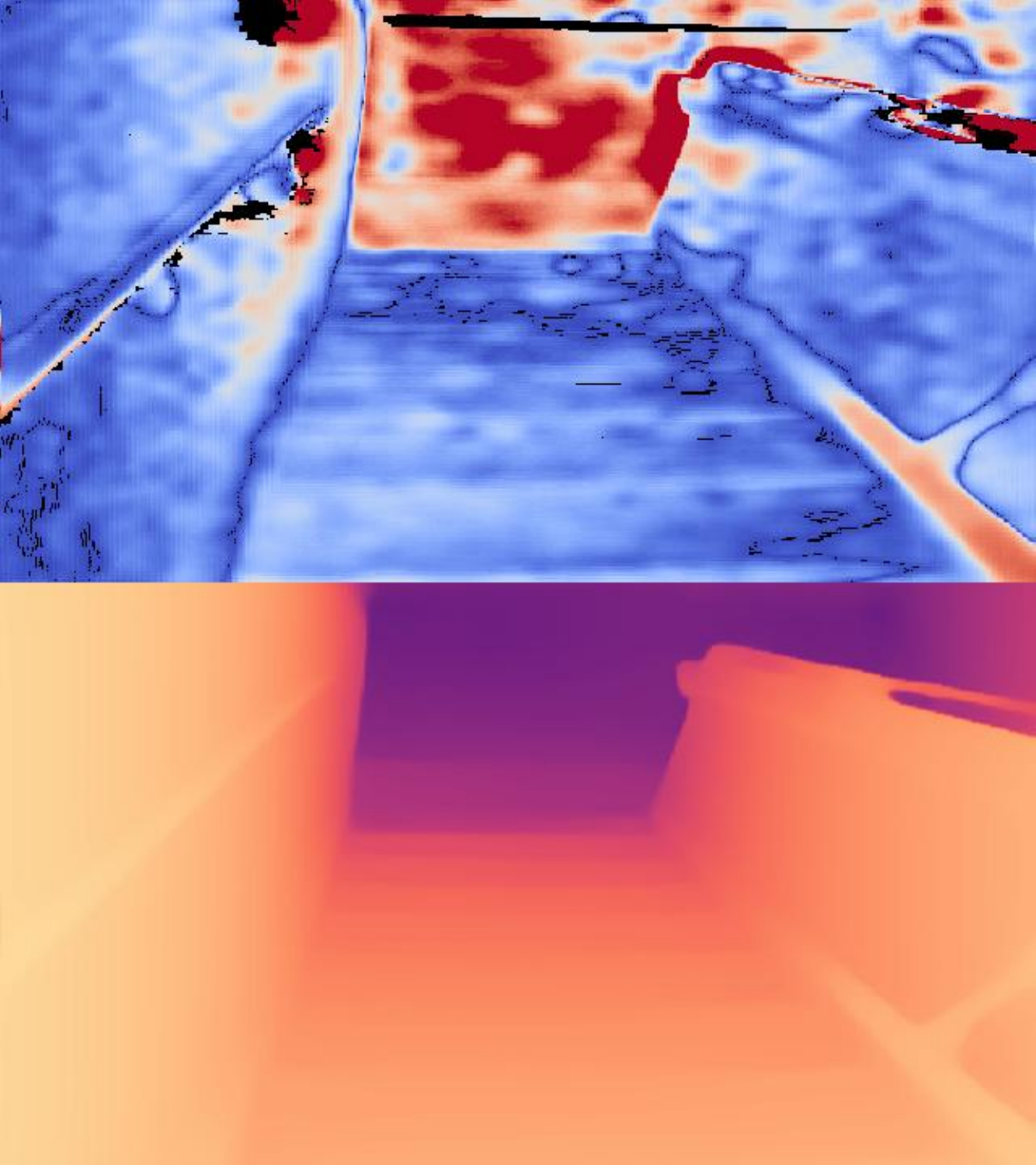}
        & \includegraphics[width=0.14\linewidth]{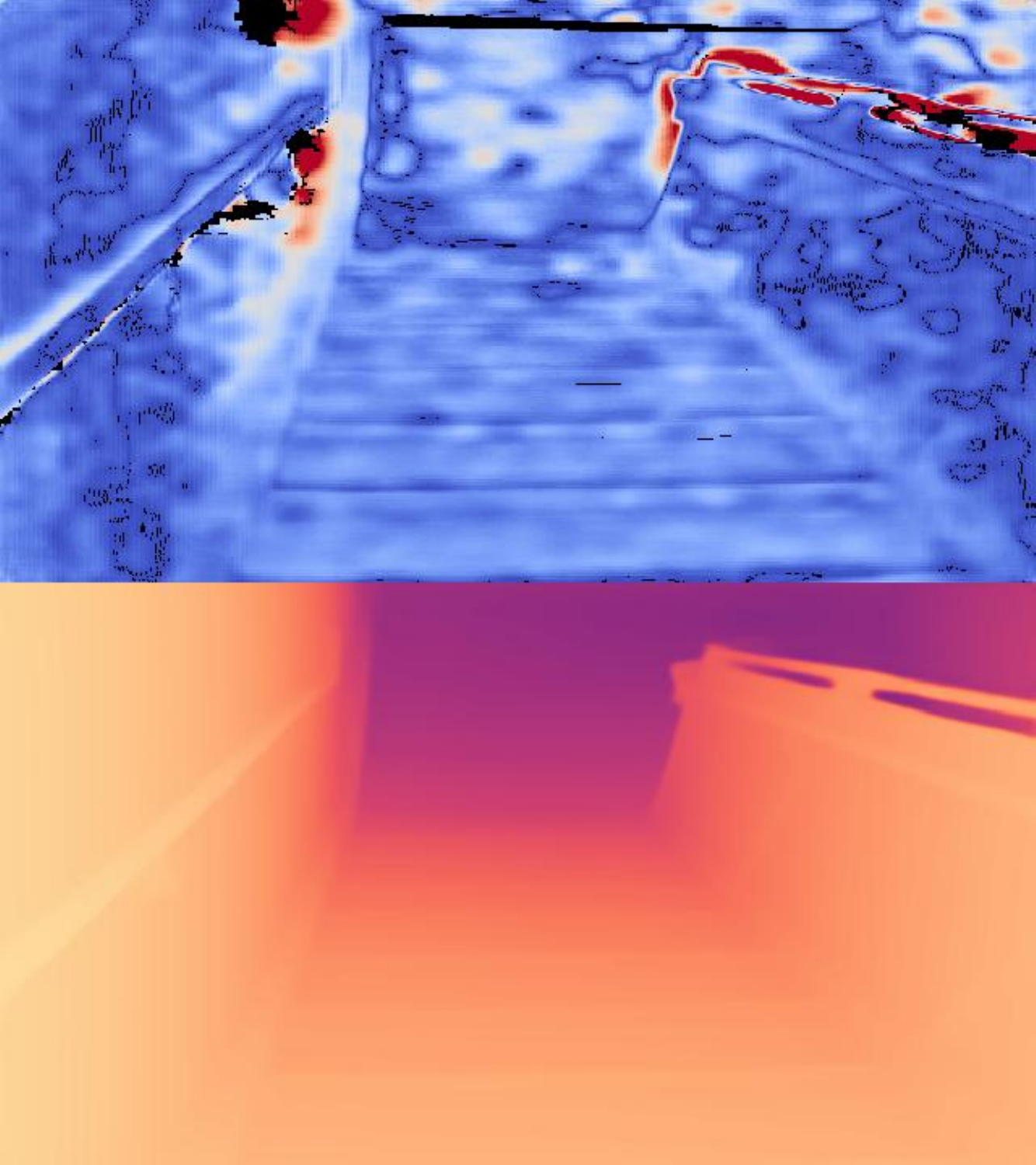}
        & \multirow{2}{*}[25pt]{\includegraphics[width=0.075\linewidth]{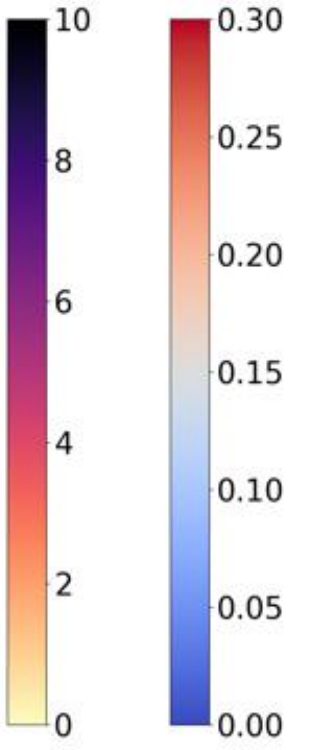}} \vspace{-8pt} \\
        & \includegraphics[width=0.14\linewidth]{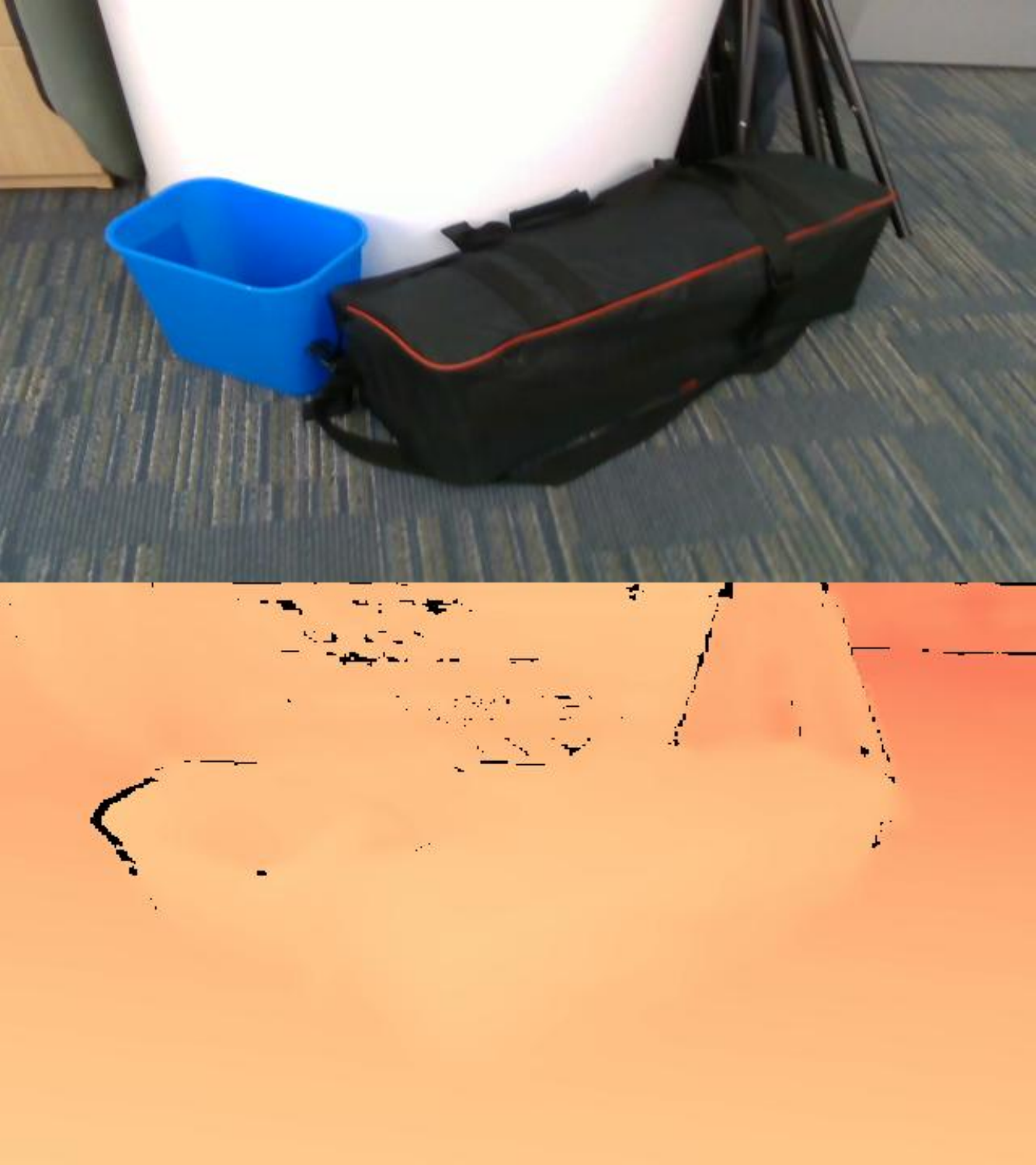}
        & \includegraphics[width=0.14\linewidth]{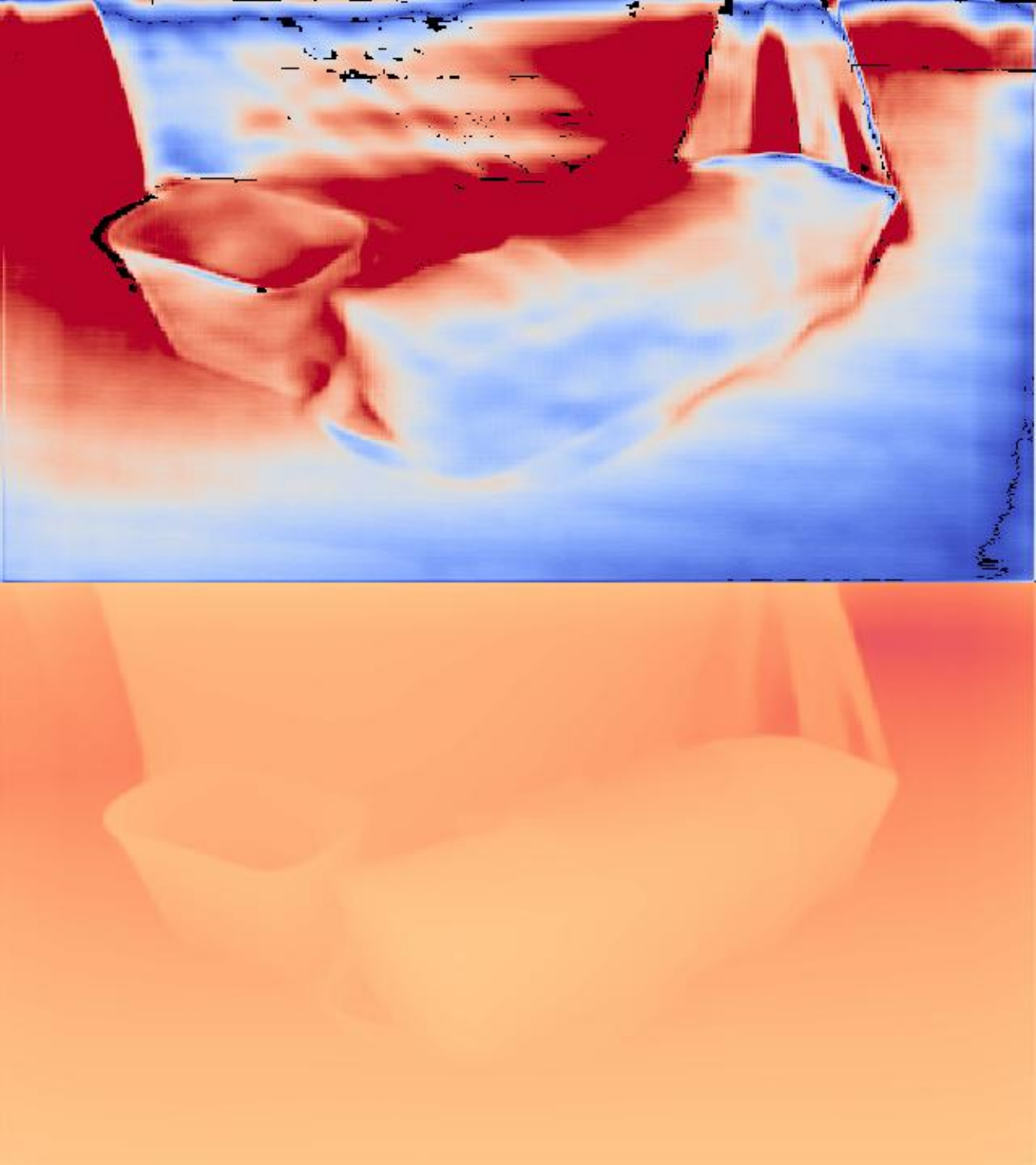}
        & \includegraphics[width=0.14\linewidth]{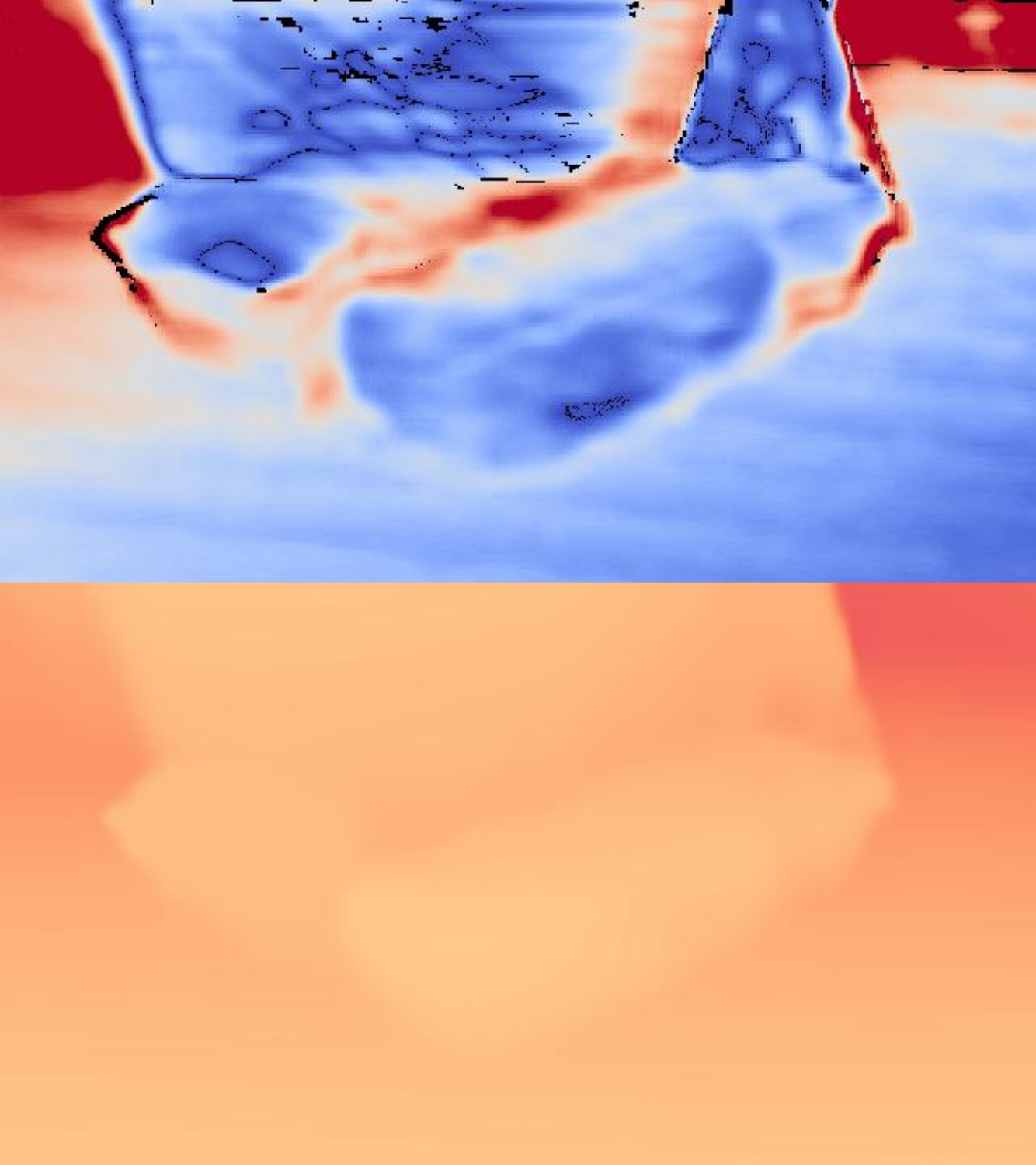}
        & \includegraphics[width=0.14\linewidth]{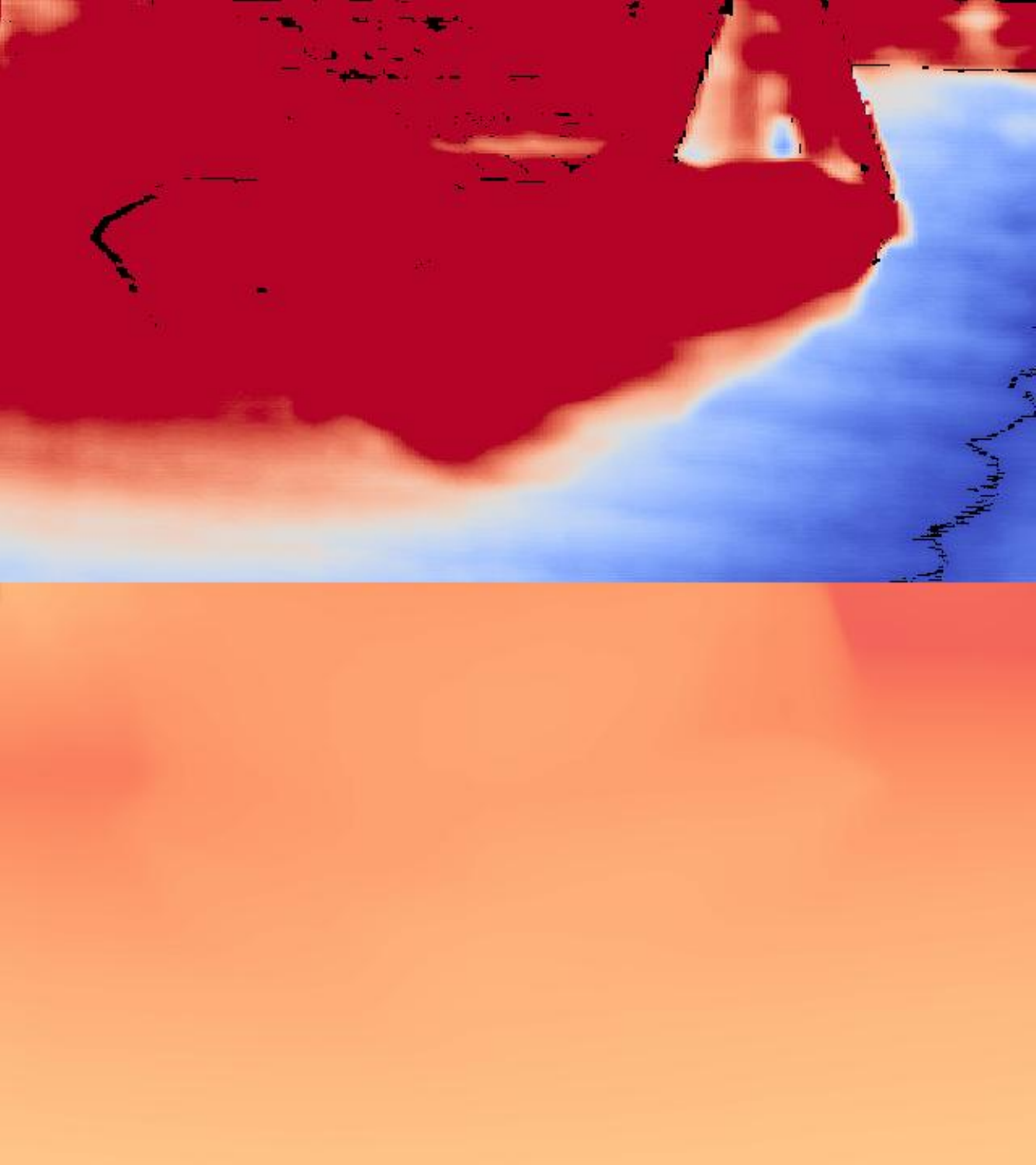}
        & \includegraphics[width=0.14\linewidth]{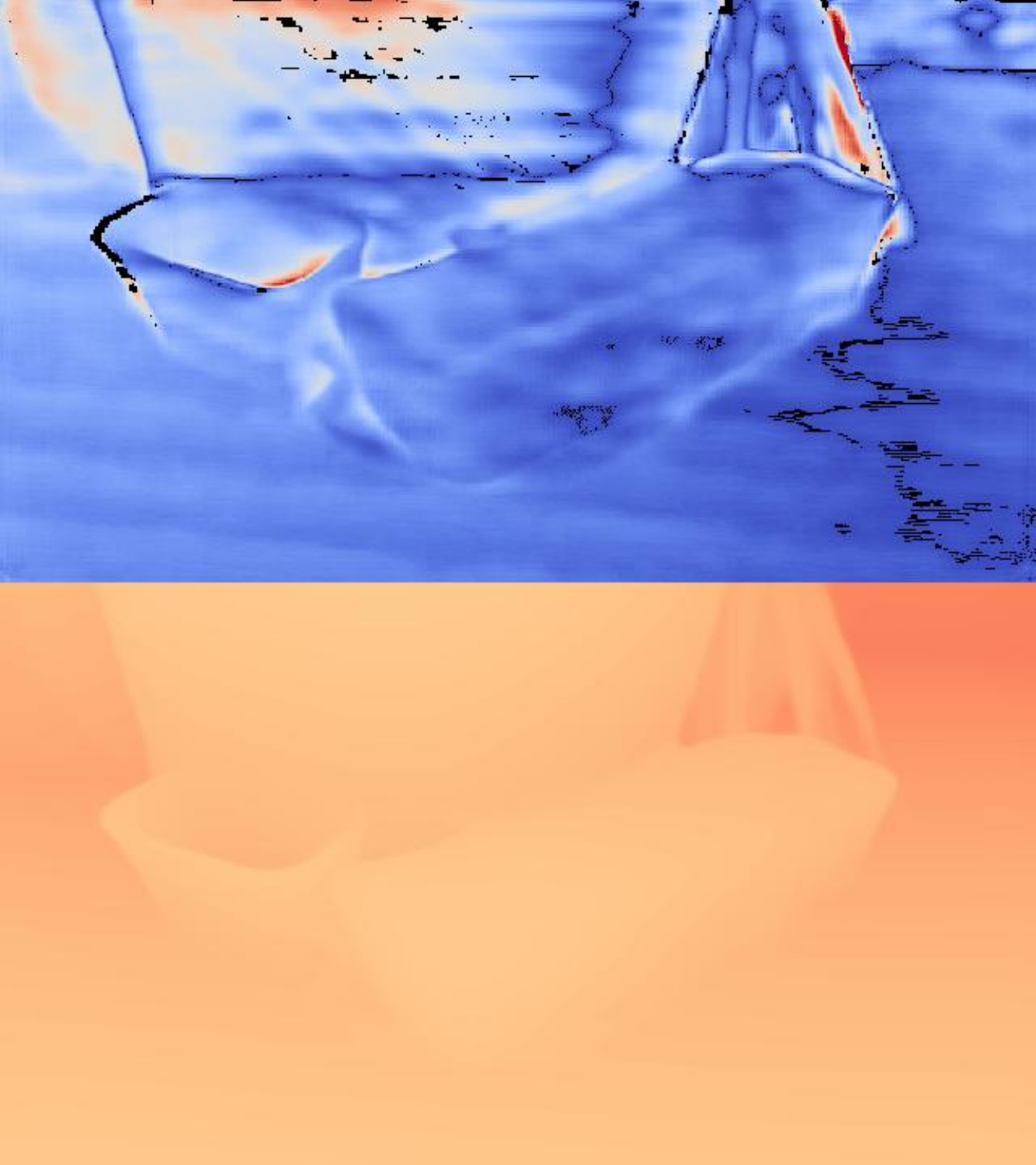}
        & \vspace{-2pt} \\

        \multirow{2}{*}[6pt]{\rotatebox[origin=c]{90}{HAMMER}}
        & \includegraphics[width=0.14\linewidth]{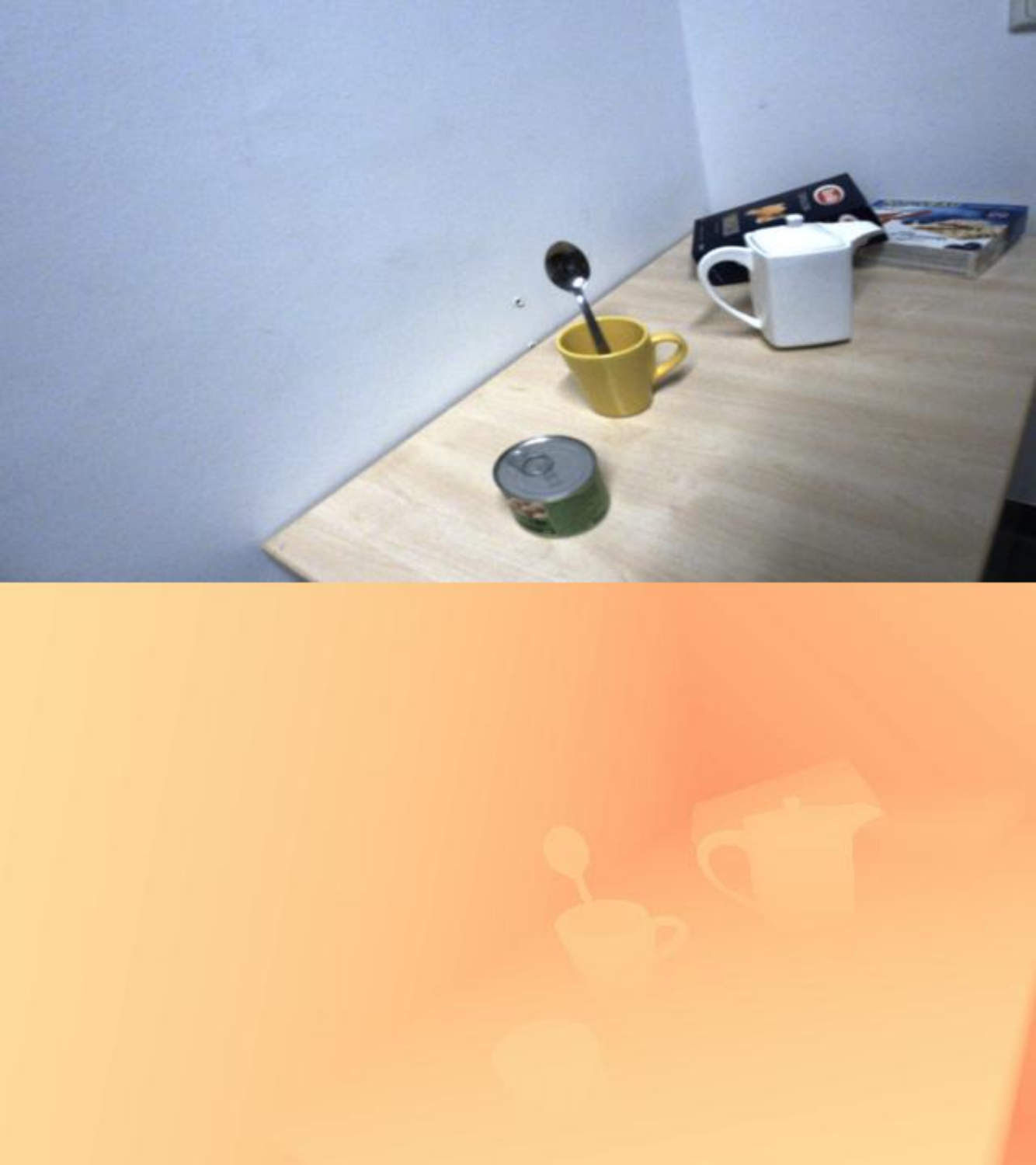}
        & \includegraphics[width=0.14\linewidth]{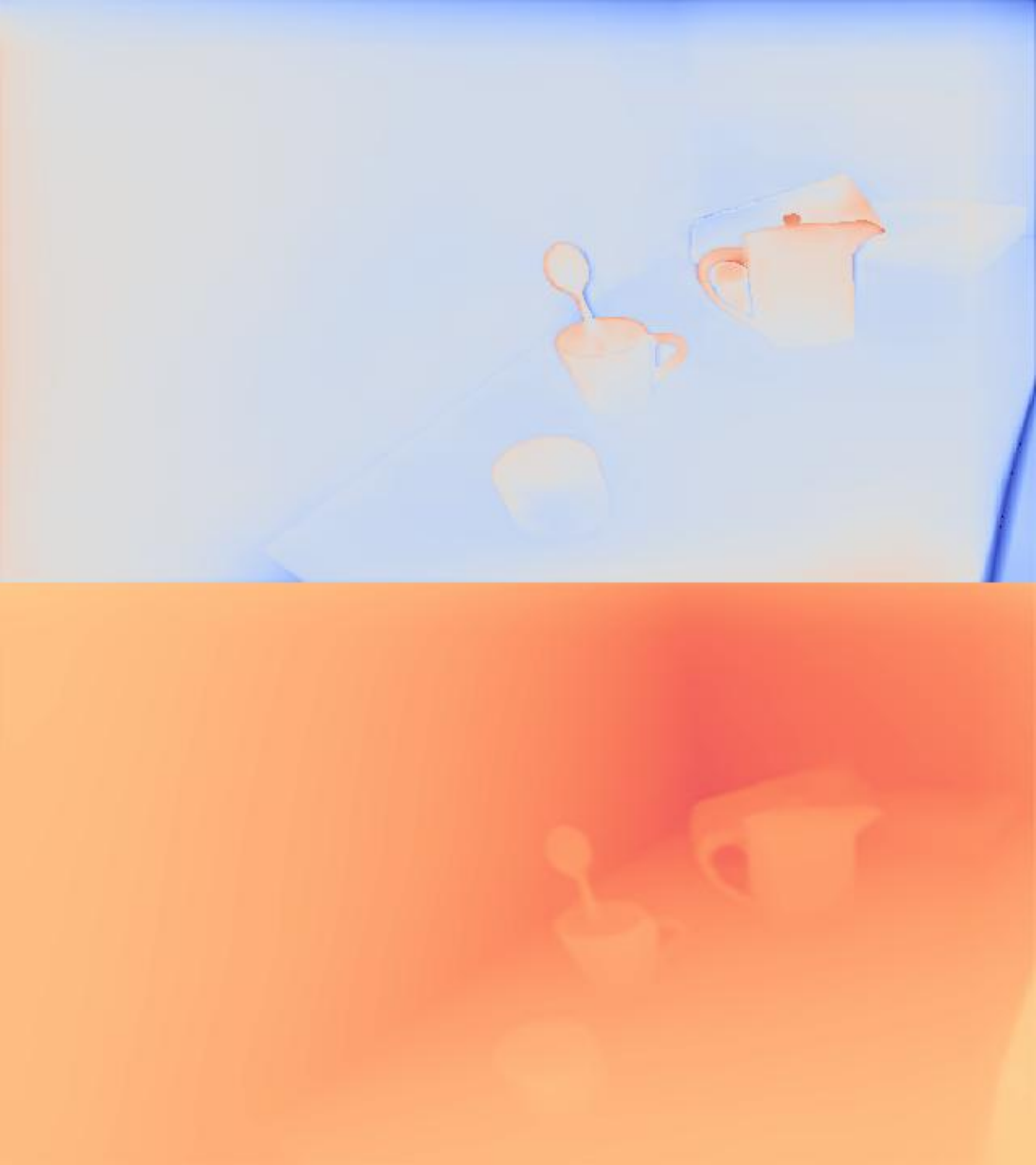}
        & \includegraphics[width=0.14\linewidth]{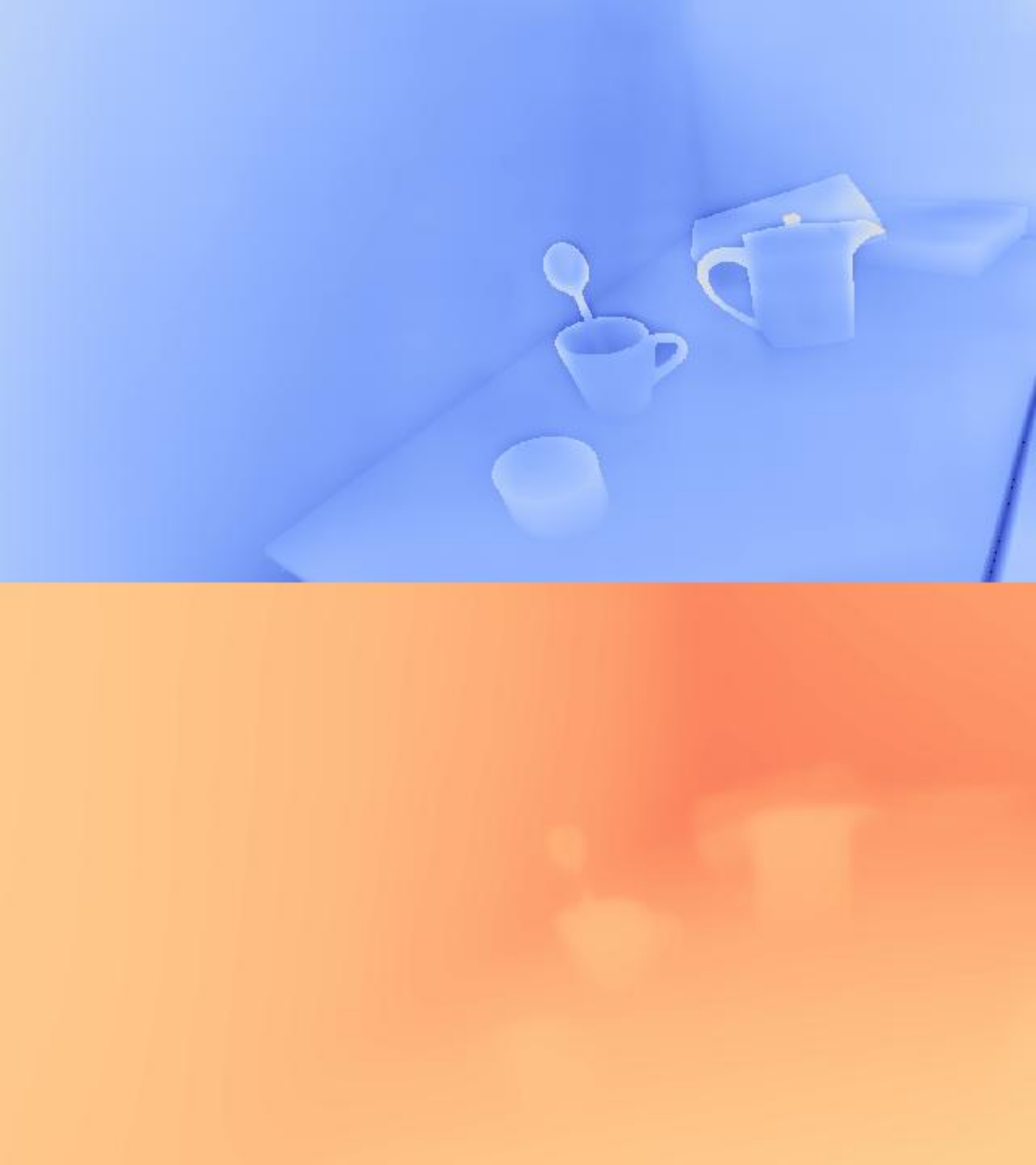}
        & \includegraphics[width=0.14\linewidth]{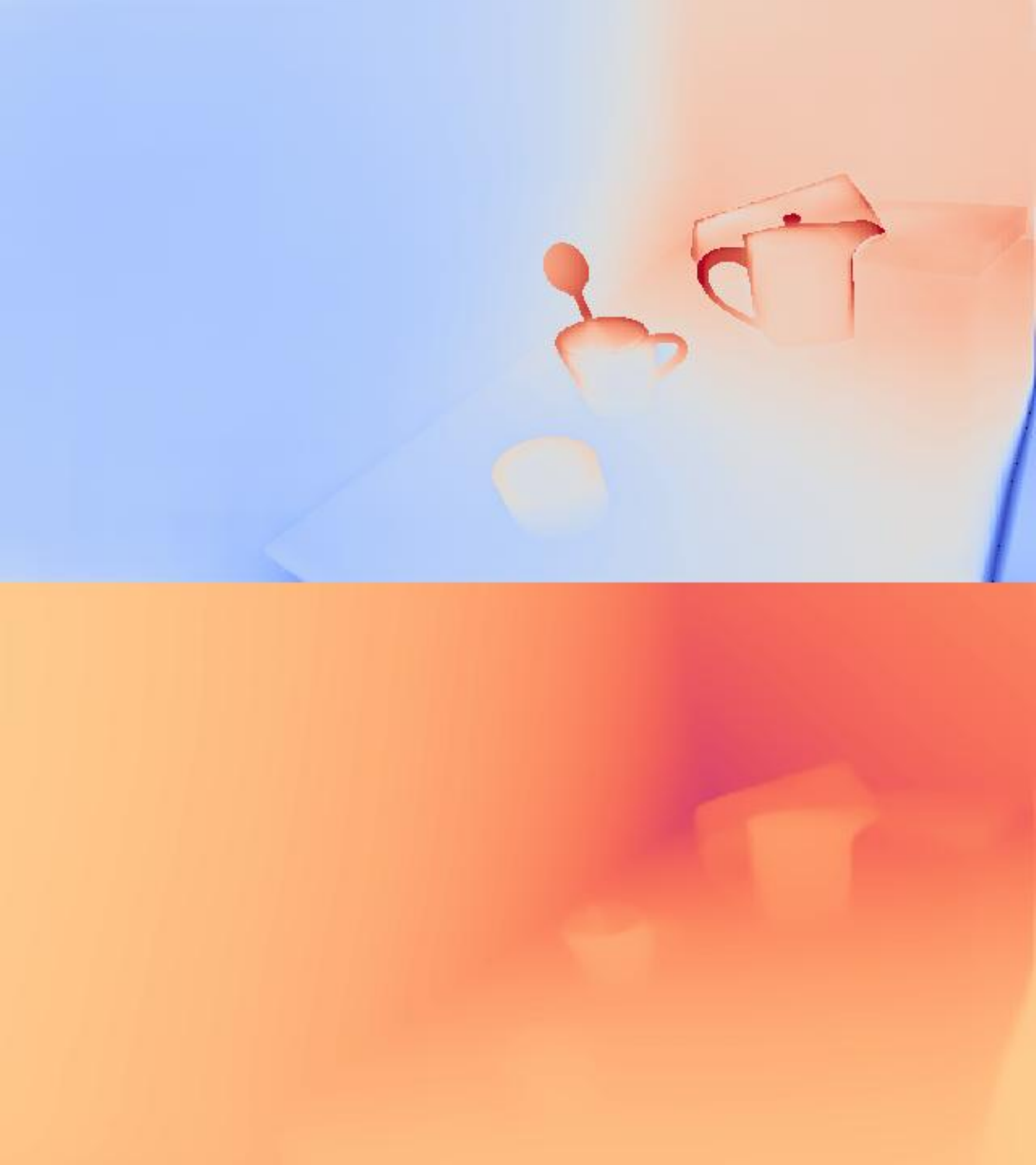}
        & \includegraphics[width=0.14\linewidth]{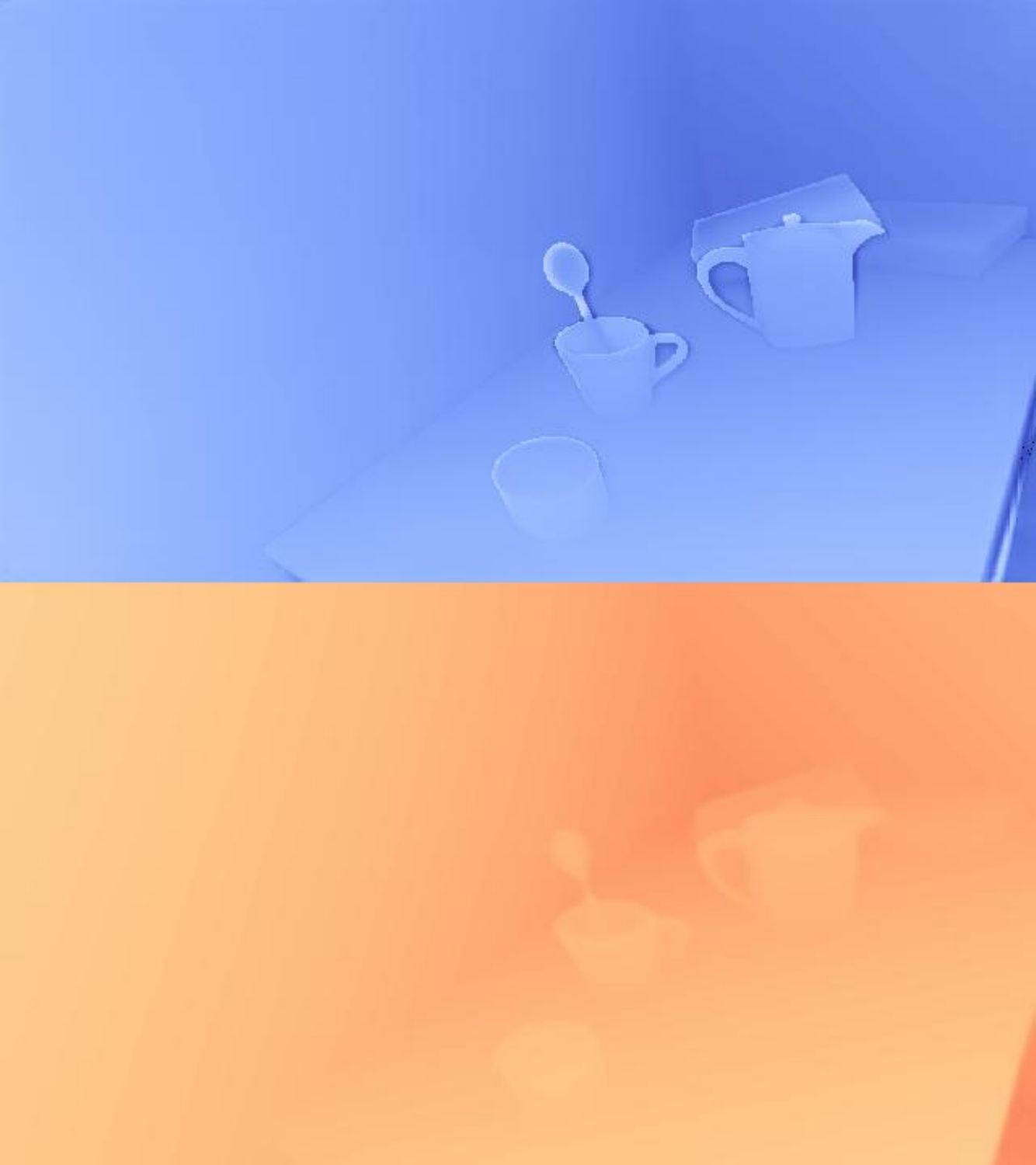}
        & \multirow{2}{*}[25pt]{\includegraphics[width=0.075\linewidth]{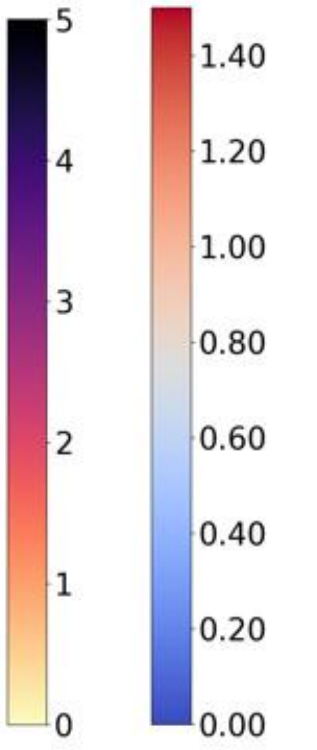}} \vspace{-8pt} \\
        & \includegraphics[width=0.14\linewidth]{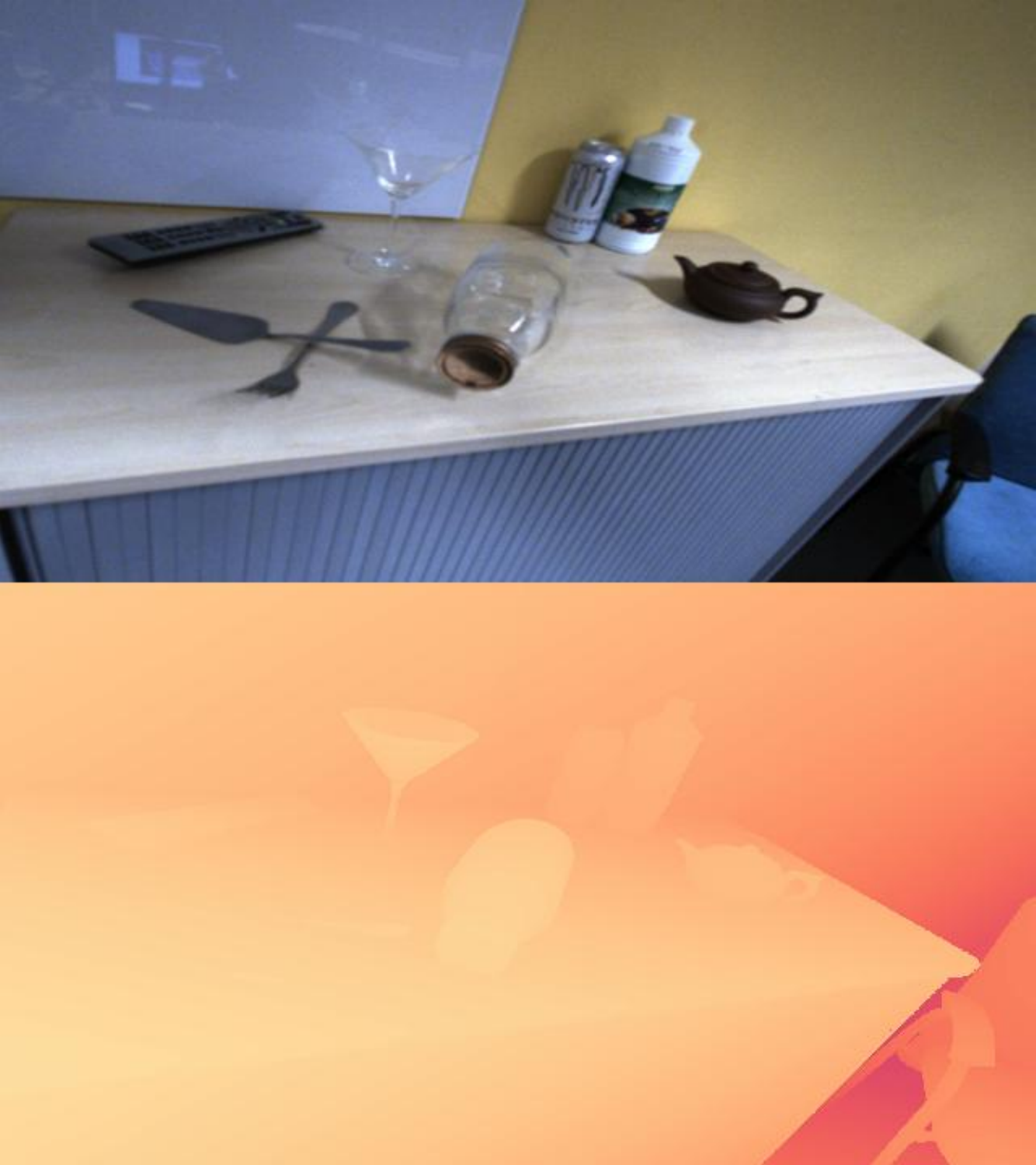}
        & \includegraphics[width=0.14\linewidth]{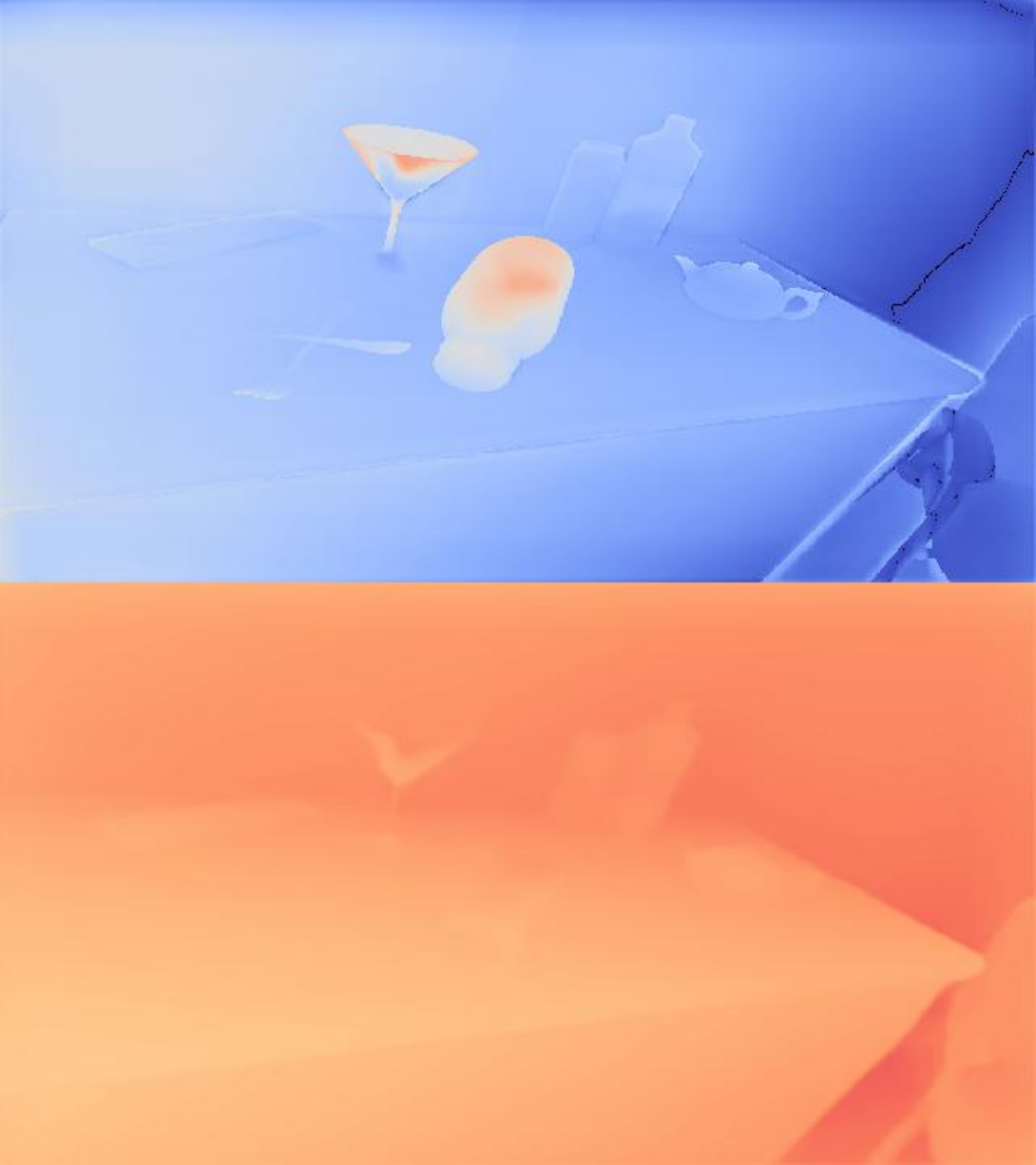}
        & \includegraphics[width=0.14\linewidth]{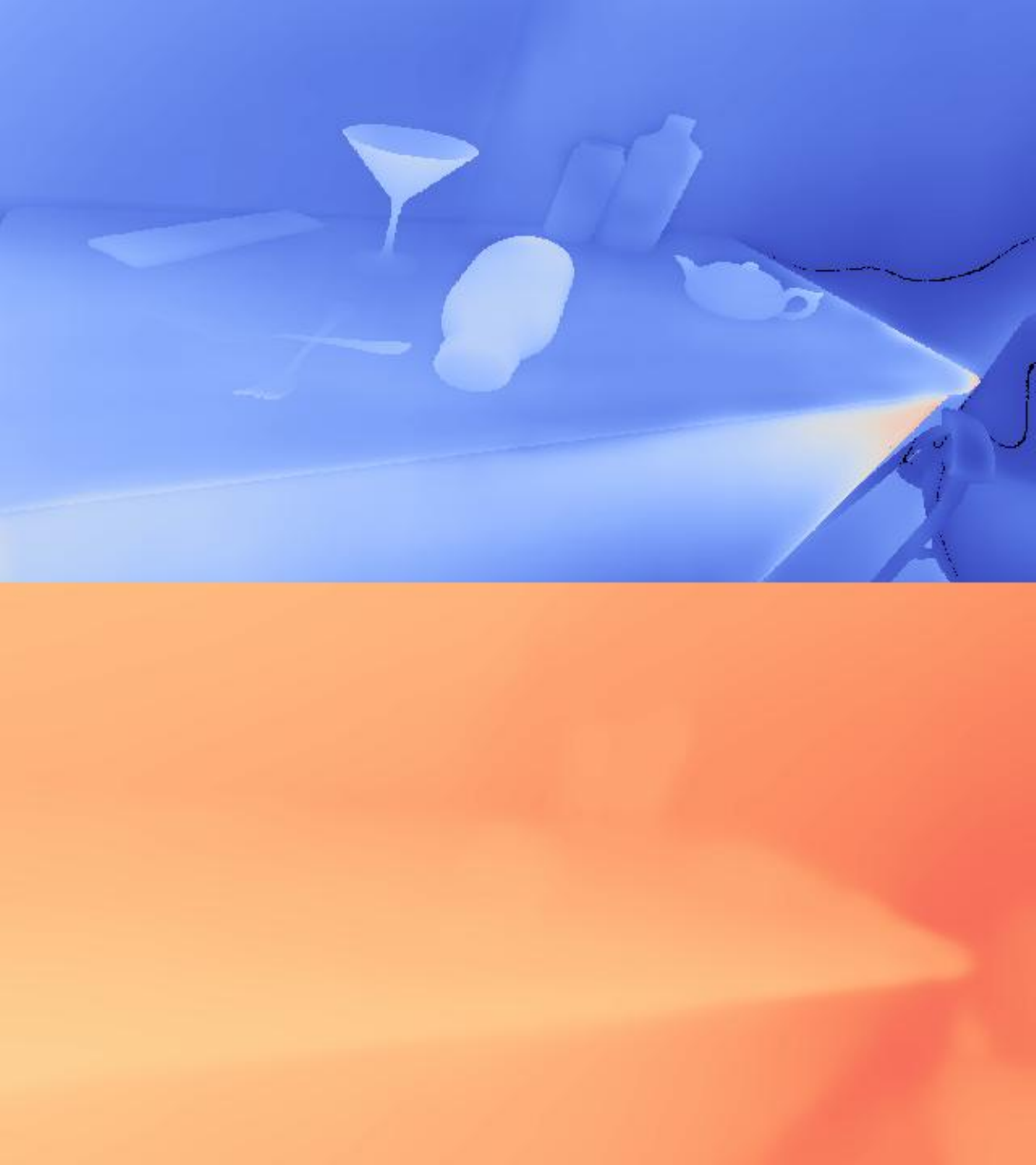}
        & \includegraphics[width=0.14\linewidth]{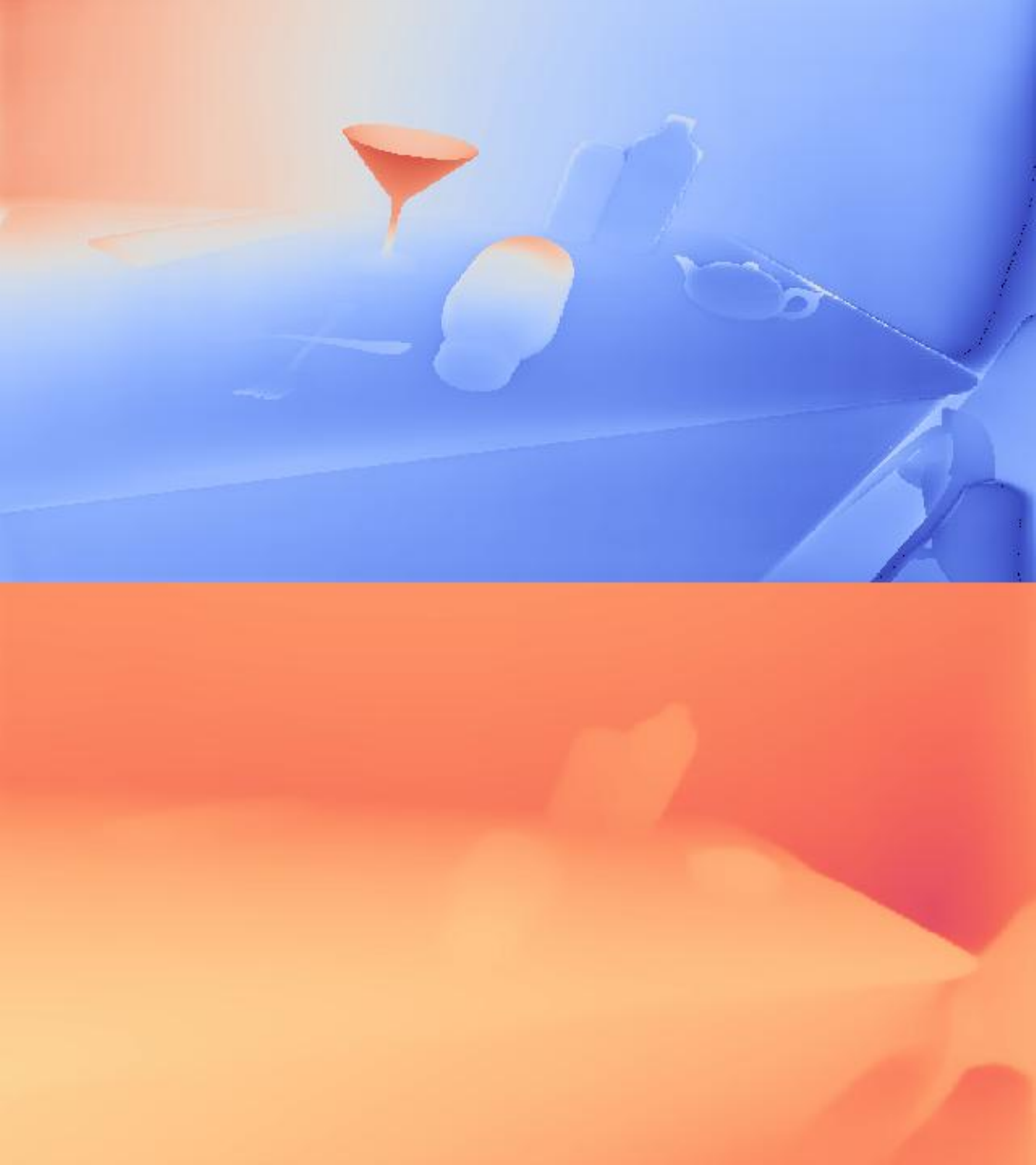}
        & \includegraphics[width=0.14\linewidth]{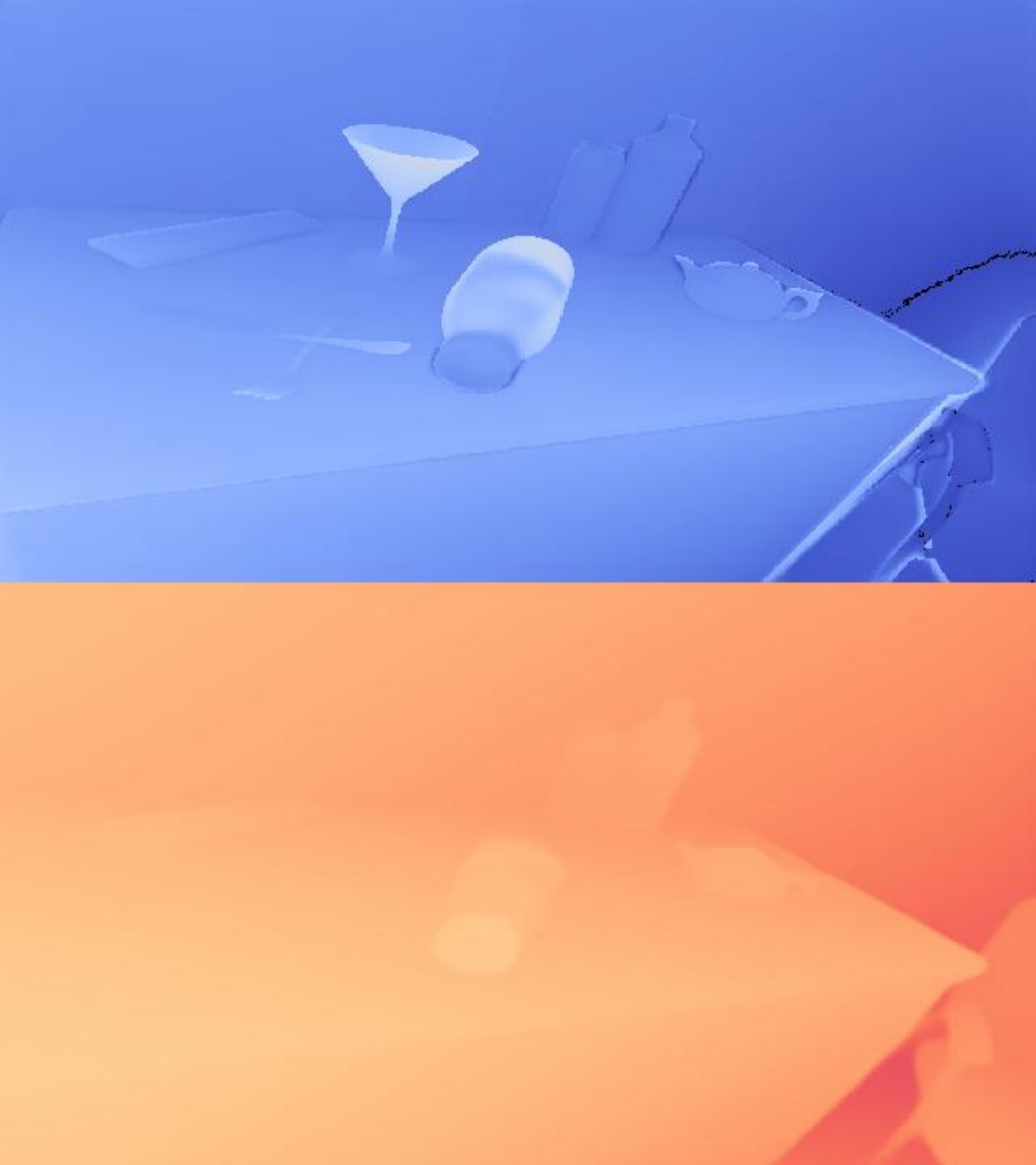}
        & \vspace{-2pt} \\

        & RGB \& GT & ZoeDepth~\cite{bhat2023zoedepth} & ZeroDepth~\cite{guizilini2023zerodepth} & Metric3D~\cite{yin2023metric3d} & \ourmodel & Meters $|$ $\mathrm{A.Rel}$ \\
    \end{tabular}
    \vspace{-8pt}
    \caption{\textbf{Zero-shot qualitative results.} Each pair of consecutive rows corresponds to one test sample. Each odd row shows the input RGB image and the absolute relative error map color-coded with \textit{coolwarm} colormap. Each even row shows GT depth and the predicted depth. The last column represents the specific colormap ranges for depth and error.}
    \label{fig:supp:vis3}
\end{figure*}

We provide here twenty more qualitative comparisons, two for each zero-shot test set: KITTI, NYU, Diode, ETH3D in \cref{fig:supp:vis1}, DDAD, NuScenes, SUN-RGBD, IBims-1 in \cref{fig:supp:vis2}, and \cref{fig:supp:vis3} displays VOID and HAMMER.
The error maps are shown after applying median-based rescaling.
The rescaling was deemed necessary to avoid some of the error maps being completely red and not informative.
Due to sparsity, DDAD and Nuscenes GT and error maps are dilated by a factor of 5, leading to visible GT depth and error maps.

\end{document}